\pdfoutput=1

\documentclass[11pt]{article}

\usepackage{acl}

\usepackage{times}
\usepackage{latexsym}
\usepackage[T1]{fontenc}

\usepackage[utf8]{inputenc}

\usepackage{microtype}

\usepackage{inconsolata}

\usepackage{amsmath,amsfonts,bm}

\def\eqref#1{equation~\ref{#1}}

\def\1{\bm{1}}

\DeclareMathAlphabet{\mathsfit}{\encodingdefault}{\sfdefault}{m}{sl}
\SetMathAlphabet{\mathsfit}{bold}{\encodingdefault}{\sfdefault}{bx}{n}

\usepackage{hyperref}       
\usepackage{url}            
\usepackage{booktabs}       
\usepackage{amsfonts}       
\usepackage{nicefrac}       
\usepackage{url}            
\usepackage{booktabs}       
\usepackage{amsfonts}       
\usepackage{color,colortbl}
\usepackage{tabularx}
\usepackage{booktabs}
\usepackage{multirow}
\usepackage{array}
\usepackage{xspace}
\usepackage{physics}
\usepackage{pifont}

\usepackage{balance}
\usepackage{makecell}
\usepackage{tcolorbox}
\usepackage{enumitem}
\usepackage{mathtools}
\usepackage{rotating}
\usepackage{tabulary}
\usepackage{subcaption}
\usepackage[export]{adjustbox}
\usepackage{ctable}
\usepackage{diagbox}
\usepackage{bbm}
\usepackage{makecell}
\usepackage[linesnumbered,ruled]{algorithm2e}
\usepackage[title]{appendix}
\usepackage{soul}
\usepackage[normalem]{ulem}
\usepackage[symbol]{footmisc}
 \useunder{\uline}{\ul}{}
\usepackage{lscape}
\graphicspath{%
	{./Pictures/}%
	{./images/}%
}

\newcommand{\cmark}{\ding{51}}%
\newcommand{\xmark}{\ding{55}}%
\newcommand{\vlbenchmark}{OLIVE}%

\newcommand{\longbenchname}{Open-world Language Instruction for Visual-language Evaluation (OLIVE)}

\newcommand{\eg}{\emph{e.g., }}
\newcommand{\citea}[1]{\citeauthor{#1} (\citeyear{#1})}

\title{What Are We Measuring When We Evaluate Large Vision-Language Models? An Analysis of Latent Factors and Biases}

\author{Anthony Meng Huat Tiong$^{1}$\footnotemark[1], Junqi Zhao$^{1}$\footnotemark[1], Boyang Li$^{1\spadesuit}$, Junnan Li, \\
	\textbf{Steven C.H. Hoi}$^{2}$, and \textbf{Caiming Xiong}$^{3}$ \\
	$^{1}$Nanyang Technological University \, $^{2}$Singapore Management University \, $^{3}$Salesforce Research\\
 {\texttt{\{anthonym001, junqi.zhao, boyang.li\}@ntu.edu.sg}} \\
  { \texttt{\{junnan4926, stevenhoi\}@gmail.com}} \qquad {\texttt{cxiong@salesforce.com}}}

\begin{document}
\maketitle
\footnotetext[1]{Equal contributions. $^\spadesuit$Corresponding Author.}
\begin{abstract}
Vision-language (VL) models, pretrained on colossal image-text datasets, have attained broad VL competence that is difficult to evaluate. A common belief is that a small number of VL skills underlie the variety of VL tests. In this paper, we perform a large-scale transfer learning experiment aimed at discovering latent VL skills from data. We reveal interesting characteristics that have important implications for test suite design. First, generation tasks suffer from a length bias, suggesting benchmarks should balance tasks with varying output lengths. Second, we demonstrate that factor analysis successfully identifies reasonable yet surprising VL skill factors, suggesting benchmarks could leverage similar analyses for task selection.
Finally, we present a new dataset, OLIVE\footnote[2]{\url{https://github.com/jq-zh/olive-dataset}}, which simulates user instructions in the wild and presents challenges dissimilar to all datasets we tested. Our findings contribute to the design of balanced and broad-coverage vision-language evaluation methods. 
\end{abstract}

\section{Introduction}
Benefiting from enormous training data, large model sizes, and pretrained large language models, the current generation of vision-language models (VLMs) (\eg \citealt{instructblip,minigpt4,llava,mplug-owl,li2023otter,awadalla2023openflamingo}) demonstrate competence in a wide range of tasks, including visual question-answering, optical character recognition, and spatial relation recognition. However, their broad competence pose a new challenge to the design of evaluation benchmarks, as many previous work focuses on only assessing one or a few capabilities, relying on data from a single distribution and annotation pipeline. This approach can result in test data that fails to fully represent all potential user inputs, leading to discrepancies between benchmark scores and real-world user experiences.

A prevailing evaluation strategy involves testing across a range of tasks and computing an average score~\cite{bitton2023visitbench,LVLM-ehub, mmbench,yu2023mmvet,li2023reformeval,fu2023mme}. This type of benchmarks usually involves a manual categorization of test tasks, as a benchmark that covers more categories is often considered more comprehensive and capable of measuring broad competence. For example, TouchStone \cite{bai2023touchstone} sorts tasks into five skills, ranging from visual recognition to visual storytelling. However, such categorizations are based on human intuition and lack support from empirical evidence. 

In this paper, we promote an alternative approach that identifies vision-language (VL) capabilities underlying various tests directly from data. Inspired by the distributional hypothesis \cite{firth-distributional-hypothesis:1957}, we characterize test tasks using neighborhood structures inferred from transfer learning. That is, transfer learning between datasets that follow similar distributions and require similar VL capabilities will likely yield high performance. By analyzing transfer performance between a large number of source and target tasks, we can observe dataset similarity, infer shared VL capabilities, and gain insights into the VL benchmarks.

Specifically, we fine-tune four popular VLMs with different strengths, BLIP-2 \cite{blip2}, Mini-GPT4 \cite{minigpt4}, LLaVA \cite{llava}, and mPLUG-Owl \cite{mplug-owl}, on 23 training (source) tasks and evaluate them on 29 test (target) tasks. Together with the original model performance before any finetuning (referred to as zero-shot performance), we obtain 2,784 performance measurements. After that, we examine the patterns and conduct Exploratory Factor Analysis, discovering six interpretable latent factors underlying the measurements.

The analyses reveal a few surprising findings. First, we find that a surface-form property, the average output length, has a surprisingly strong influence on transfer performance. This suggests current evaluation results may be affected by this length bias. Second, factor analysis is capable of discovering unexpected yet reasonable factors that explain model performance. For example, we identify factors that differentiate reading text off images from multi-hop reasoning. These findings have important implications for the design of unbiased and comprehensive VL benchmarks.

Finally, to simulate real-world user instructions, we present a new vision-language dataset, \longbenchname. \vlbenchmark{} consists of 9,450 images, 30,120 unique instructions, and 47,250 responses. Empirically, we show that \vlbenchmark{} has a transfer profile distinct from all other datasets we tested and hence provides a test complementary to existing tasks.

We summarize our contributions as follows:
\begin{itemize}
	\setlength\itemsep{0pt}
	\item We promote the approach of discovering VL skills from data and demonstrate that factor analysis is a robust and effective technique for this purpose. Our large-scale experiments lead to findings that can inform the future design of VLM test suites.
	\item We introduce a new benchmark, \vlbenchmark, to evaluate open-ended model responses to diverse instructions.
\end{itemize}

\section{Analysis Techniques}

The transfer performance from $N$ source (training) tasks to $K$ target (test) tasks on model $m$ is stored as a matrix $A^{(m)} \in \mathbb{R}^{N\times K}$. The performance numbers of different tasks cannot be compared directly due to differences in the scales of the evaluation metrics. Therefore, we first normalize the data. Subsequently, we apply Singular Value Decomposition and Factor Analysis. Both techniques can be understood as decompositions of the matrices $A^{(m)}$, albeit with different mathematical formulations.

\subsection{Normalization}
We obtain the raw performance number $b^{(m)}_{i, j}$ when we train model $m$ on the $i^\text{th}$ source task and test on the $j^\text{th}$ target task. The normalized performance $a^{(m)}_{i,j}$ is obtained as
\begin{equation}
\label{normalization}
\hspace{-0.2cm}	a^{(m)}_{i,j} = (b^{(m)}_{i,j} - b^{(m)}_{0,j})/ (\max_{j^\prime} b^{(m)}_{i,j^\prime} - b^{(m)}_{0,j}),
\end{equation} 
where $b^{(m)}_{0,j}$ denotes the performance of the pretrained model $m$ on target task $j$ without finetuning on any source task, referred to as the zero-shot performance. If finetuning on source $i$ improves over the zero-shot performance, $a^{(m)}_{i,j}$ is a positive number. Conversely, if there is negative transfer from source $i$, $a^{(m)}_{i,j}$ is negative.  The best source task, which in most cases is the in-domain i.i.d. training task, has $a^{(m)}_{i,j} = 1$. Hence, this normalization separates positive from negative transfers and shows how source tasks perform relative to the in-domain training data.The matrix $A^{(m)}$ has $a^{(m)}_{i,j}$ as its components. After separate normalizations, we concatenate the four matrices $A^{(m)}$, corresponding to the four models we fine-tune, along the source-task dimension (the rows) and obtain the aggregate performance matrix $A \in \mathbb{R}^{4N\times K}$.

\subsection{Singular Value Decomposition}
Singular Value Decomposition (SVD) is a classic technique for learning distributed representations. \citea{word2vec-matrix-factorization-2014} show that SVD produces word embeddings comparable to word2vec \cite{mikolov2013:word2vec}. The SVD of matrix $A$ can be written as
\begin{equation}
A = U\Sigma V^{\top}
\end{equation}
We perform truncated SVD using the $D$ largest singular values. After that, we use $V\Sigma^{1/2} \in \mathbb{R}^{K\times D}$ as the features of the target tasks.

\subsection{Factor Analysis}
It is a widely held belief that a small number of factors, known as cognitive abilities, underlie human performance on numerous mental activities \cite{horn2007understanding}. To uncover these latent factors, \citea{Spearman1904} developed the statistical technique of Factor Analysis. For modern treatments, we refer readers to \citea{gorsuch2014factor} and \citea{barber2012}. 

In this paper, we start with a premise similar to Spearman's, that a small number of VL capabilities are responsible for VLM performance on various test tasks. This belief is, in fact, implicitly shared, though often not explicitly stated, by most recent VLM evaluation papers~(\eg \citealt{bitton2023visitbench,LVLM-ehub, mmbench,bai2023touchstone}) that attempt to categorize VLM test scenarios based on intuitive justifications. In contrast, we apply Exploratory Factor Analysis (EFA) to uncover these factors from empirical data.

We assume that each source task imparts specific latent skills to a model. These skills, while not directly observable, manifest through the model's performance on related target tasks. Performances on target tasks that share the same skills tend to increase or decrease in a correlated manner. This pattern of correlation underpins our expectation that EFA will successfully unveil the underlying skills.

Mathematically, we treat the $i^{\text{th}}$ column of $A$, $\bm{a}_i \in \mathbb{R}^{4N}$, as the characteristic of the target task $i$, which we attempt to explain with $L$ latent factors:
\begin{equation}
    \bm{a}_i = W \bm{h}_i + \bm{\mu} + \bm{\epsilon},
\end{equation}
where $W \in \mathbb{R}^{4N \times L}$ reflects how source tasks load to the $L$ latent factors and $\bm{h}_i \in \mathbb{R}^{L}$ reflects how target task $i$ decomposes into the latent factors. $\bm{\mu}$ is the average vector across target tasks and $\bm{\epsilon}$ represents Gaussian noise. EFA differs from PCA in that it assumes the covariance of $\bm{\epsilon}$ is diagonal rather than spherical. Note that the above formulation is invariant to a rotational matrix $R$, as $W \bm{h}_i = (WR)(R^{\top}\bm{h}_i)$. We apply the Varimax rotation~\cite{kaiser1958varimax} to find $R$ so that $\bm{h}_i$ is as concentrated on as few factors as possible. 

Our preliminary analysis suggests that the captioning and VQA tasks are highly correlated and predominantly load onto a single factor, likely indicative of a general VL capability. To isolate and examine other factors, we employ linear regression to control for the influence of the dominant factor. Specifically, we first perform EFA with one factor, so that $W$ becomes a $4N$-by-$1$ vector $\bm{w}$. We then perform regression from $\bm{w}$ to $A$ by solving the following problem:
\begin{equation} 
\text{minimize} \;\; \left\| A - \bm{w} \bm{\beta}^{\top} - \bm{\gamma} \bm{1}^{\top} \right\|_F^2,
\end{equation} 
where $\bm{\beta}, \bm{\gamma} \in \mathbb{R}^{4N}$ are trainable parameters. Afterwards, we conduct EFA on the residuals, $\bar{A} = A - \bm{w} \bm{\beta}^{\top} - \bm{\gamma} \bm{1}^{\top}$, which contain information about other factors indicative of more specific VL capabilities than the first factor.

We employ both parallel analysis and Velicer's Minimum Average Partial (MAP) test to determine the optimal number of factors to extract. Parallel analysis compares the eigenvalues from our sample correlation matrix against those from random data of the same size, identifying factors that explain more variance than expected by chance. Conversely, Velicer's MAP test evaluates the average squared partial correlation for each possible number of factors, pinpointing the point at which additional factors no longer meaningfully increase the explanation of variance. Both methods converge on the decision to extract six factors.

\begin{table}[!t]
\centering 
  \small
  
  \begin{tabular}{@{}cccc@{}}
    \midrule
    \makecell{Intuitive\\Category}  & Task & Source & Target \\
    \midrule
    \multirowcell{4}{Image\\Captioning} & COCO Caption & \cmark & \cmark \\
    & Flickr30k & \cmark & \cmark \\
    & Web CapFilt & \cmark & \xmark \\
    & TextCaps & \cmark & \cmark \\ \midrule
    
    Generic VQA & VQAv2 & G & G, MC \\
    \midrule
    \multirowcell{3}{Knowledge-\\based VQA} & OK-VQA &  G & G, MC \\
    & A-OKVQA & G, MC & G, MC \\
    & ScienceQA & MC & MC \\ \midrule
    \multirowcell{2}{OCR VQA} & TextVQA & G & G, MC\\
    & OCR-VQA & G & G, MC\\ \midrule
    \multirowcell{5}{Visual\\Reasoning} & GQA & G & G, MC\\
    & VSR & MC & MC \\
    & IconQA & MC & MC \\
    & CLEVR & \xmark & G, MC \\
    & RAVEN-FAIR & \xmark & MC \\ \midrule
    \makecell{Classification} & \makecell{Hateful Memes} & MC & MC \\
    \midrule
    \multirowcell{5}{Humor \&\\Sarcasm} & \makecell{New Yorker\\Ranking} & \xmark & \cmark \\
    & \makecell{New Yorker\\Explanation} & \xmark & \cmark \\
    & \makecell{MORE} & \xmark & \cmark \\ \midrule
    \multirowcell{2}{Chart\\Reading} & OpenCQA & G & G \\
    & ChartQA & \xmark & G, MC \\ \midrule
    \multirowcell{6}{Open-ended\\Generation} & OLIVE (Ours) & \cmark & \cmark \\
    & \makecell{LLaVA Con-\\versation} & \cmark & \xmark\\
    & \makecell{LLaVA Rea-\\soning} & \cmark & \xmark\\ 
    & \makecell{LLaVA De-\\scription} & \cmark & \xmark \\ \midrule
    \multirowcell{3}{Question\\Generation\\(QG)} & VQAv2 QG & \cmark & \xmark \\
    & OK-VQA QG  & \cmark & \xmark\\
    & A-OKVQA QG  & \cmark & \xmark \\ 
  \bottomrule
\end{tabular}
 \caption{The list of source and target tasks used in experiments. G and MC indicate the generative and multiple-choice versions of the VQA tasks respectively.}
 \label{tab:tasks}
 \vspace{-2ex}
\end{table}

\section{Source and Target Tasks}
We gather 27 publicly available VL datasets and create variations, yielding 23 source tasks and 29 target tasks. We show the full list of tasks in Tab. \ref{tab:tasks} and describe them below. The performance metrics used are AUC for Hateful Memes, CIDEr~\cite{vedantam2015cider} for OpenCQA, \vlbenchmark{}, and all captioning datasets, ROUGE-L~\cite{lin2004rouge} for MORE and New Yorker Explanation, and accuracy for the remaining tasks. To focus on end-to-end performance, we do not perform any separate optical character recognition.

\vspace{0.05in}
\noindent
\textbf{Image Captioning.} Image captioning is one of the most popular image-text tasks and is commonly used as a pretraining task for VLMs~\cite{chen2021visualgpt, pnpvqa}. Here we select two classic datasets: COCO Caption~\cite{coco} and Flickr30k~\cite{flickr30k}. In addition, we include TextCaps \cite{textcaps}, which involves the description of textual content in images. We also include as a source task Web CapFilt, a set of synthetic image captions covering a wide variety of web images. Web CapFilt was generated by BLIP \cite{li2022blip} for self-training. We hypothesize that its diversity could be beneficial in transfer learning.

\vspace{0.05in}
\noindent
\textbf{Visual Question-answering (VQA).} VQA is another very popular image-text task due to the versatility of the question-answering format. VQAv2~\cite{goyal2017making} is probably the most prominent VQA benchmark, featuring more than 200,000 COCO images and 1 million questions. Other variations include knowledge-grounded VQA, OCR VQA, Chart VQA, and so forth, which we discuss below. 

Performance measurement in VQA can be challenging, as there are often many correct answers to the same question. As a remedy, we create two target tasks for every VQA dataset. The first is the generative (G) version, which considers an answer to be correct only when it matches exactly one of the ground-truth answers. The second is the multiple-choice (MC) version, where the model chooses one from five options. To convert a generative VQA dataset to the MC version, we create five options for every question, including at most two correct answers to account for their linguistic variations. After that, we add incorrect choices by sampling answers from other questions and selecting those with top-k probabilities according to InstructBLIP \cite{instructblip}. During inference, we feed all options to the model and choose the one with the highest average word probability as the model prediction. 

\begin{figure*}[!t]
	\begin{center}
	\includegraphics[width=\textwidth]{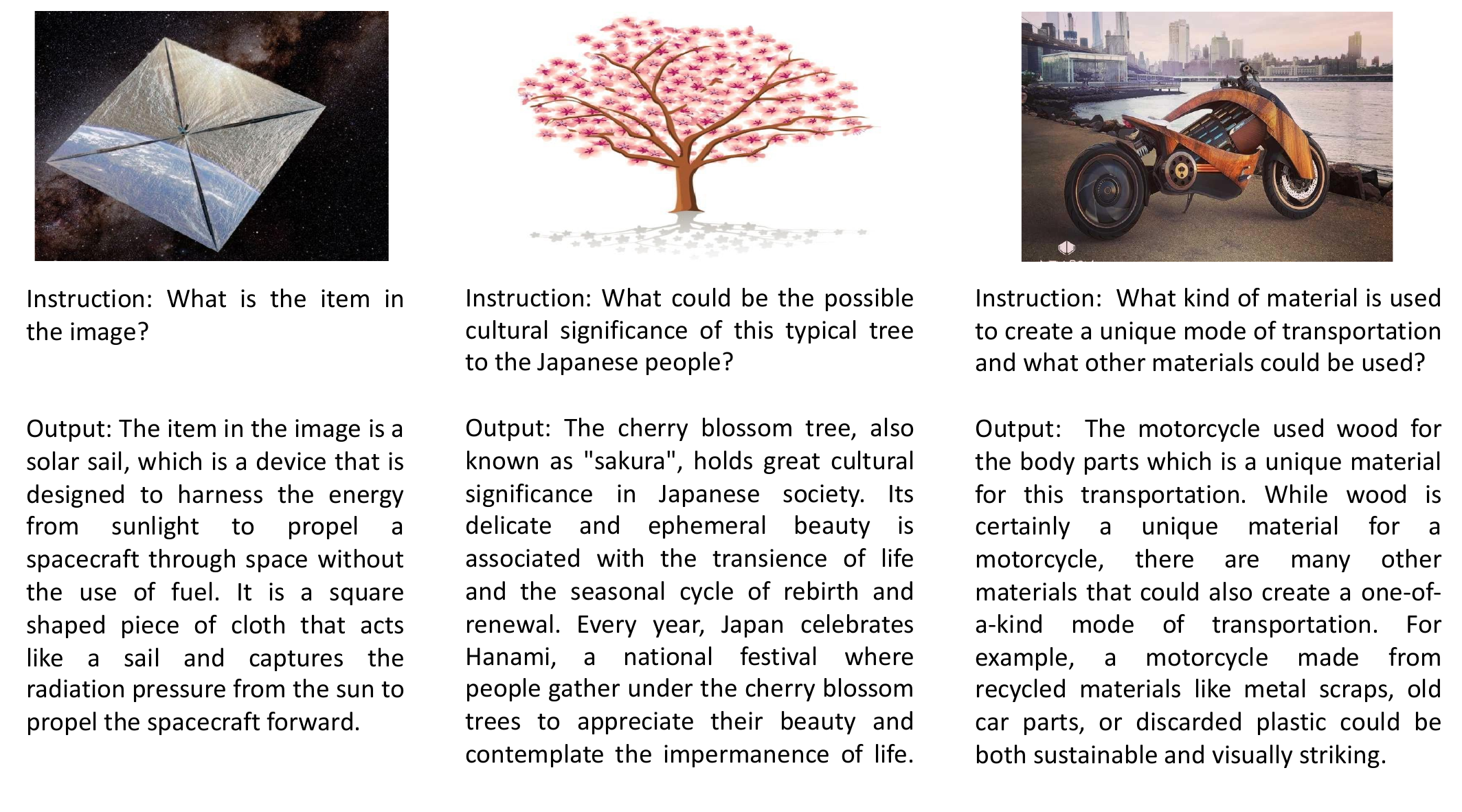}
	\vspace{-6ex}
	\end{center}
	\caption
	{
    Examples of the \vlbenchmark{} benchmark for different categories. From left to right: visual recognition, knowledge-based QA, and creative writing.
	}
	\label{fig:eval_olive_example}
	  \vspace{-2ex}
\end{figure*}

\vspace{0.05in}
\noindent
\textbf{Knowledge-grounded VQA.} These tasks require the model to apply world knowledge not present in the input to answer questions. ScienceQA~\cite{scienceqa} focuses on the contents of science textbooks. OK-VQA~\cite{marino2019ok} is primarily about visual recognition and knowledge recall, whereas A-OKVQA~\cite{aokvqa} often requires an additional step of reasoning.

\vspace{0.05in}
\noindent
\textbf{OCR VQA.} TextVQA~\cite{textvqa} and OCR-VQA~\cite{ocrvqa} are two VQA datasets that require recognition of text in images. OCR-VQA focuses on reading text from book covers, whereas TextVQA often requires locating an object before reading the text on it.

\vspace{0.05in}
\noindent
\textbf{Chart Reading.} OpenCQA~\cite{opencqa} and ChartQA~\cite{chartqa} feature questions about the content of diagrams and charts. OpenCQA expects descriptive, long-form answers, whereas ChartQA is mainly concerned with data extraction and comparison using short answers. 

\vspace{0.05in}
\noindent
\textbf{Visual Reasoning.} The term "reasoning" is used broadly in the VLM literature. It sometimes refers to shallow tasks like counting (\emph{e.g., how many apples are in the image?}), spatial relations and grounding (\emph{e.g., what is adjacent to the cylinder?}), but it can also include logical or algebraic operations. In this category, we include five datasets, GQA~\cite{hudson2019gqa}, VSR~\cite{vsr},  CLEVR~\cite{clevr}, IconQA~\cite{iconqa}, RAVEN-FAIR~\cite{ravenfair}. GQA and VSR mainly contain natural images, whereas IconQA features cartoons, and RAVEN-FAIR comprises abstract diagrams. CLEVR consists of synthetically rendered images of 3D objects. Among these, RAVEN-FAIR is the most challenging, as it is derived from the Raven's test \cite{raven1938}, an intelligence test designed for humans that requires complex reasoning.

\vspace{0.05in}
\noindent
\textbf{Image Classification.} Hateful Memes~\cite{hatefulmemes} is a binary classification task that distinguishes hateful memes from other meme images. 

\vspace{0.05in}
\noindent
\textbf{Humor and Sarcasm Understanding.} \citea{newyorker} show that VLMs perform poorly at understanding humor. From their paper, we adopt New Yorker Ranking, which involves selecting the best humorous caption for a cartoon from the New Yorker magazine, and New Yorker Explanation, which asks the model to explain why the cartoon and its caption invoke humor. The MORE dataset~\cite{more} involves explaining why a textual statement associated with a natural image is sarcastic.

\vspace{0.05in}
\noindent
\textbf{Question and Open-ended Generation.} We adapt three datasets, VQAv2, OK-VQA, A-OKVQA, for the task of question generation from an image and an answer. Further, we take the three subsets of LLaVA-Instruct-150K~\cite{llava}, which respectively focus on free-form conversation, detailed description, and reasoning. We use these as source tasks but not target tasks, since these artificial data may not be representative of real use cases. 
	
\vspace{0.05in}
\noindent
\textbf{\vlbenchmark.}
Additionally, we include a new dataset, \vlbenchmark{}, a highly diverse, human-corrected multimodal dataset, which we create to simulate in-the-wild user queries to VLMs. Once a VLM is publicly released, it tends to receive substantially more diverse and idiosyncratic inputs than the available academic datasets. \vlbenchmark{} is our attempt to simulate such user queries, so that we may train and evaluate the VLMs under conditions similar to their ultimate use case. 

We briefly describe the data curation process here and refer readers to the Appendices~\ref{Appendix_olive_data_collection} and ~\ref{Appendix_olive_prompt} for more details. First, we randomly sample 9,450 images from LAION-Aesthetics~\cite{schuhmann2022laion}, consisting of diverse web images. We take the original LAION caption from the dataset and a few generated image captions from BLIP-2 as the complete image description. Next, we feed customized prompts to ChatGPT~\cite{chatgpt} to generate instructions conditioned on the image description and five responses to each instruction. Thereafter, a team of data annotators is recruited to verify and refine the data in order to remove errors, shortcut biases, and harmful content. This procedure yields 30,120 unique instructions and 47,250 responses. We use 6,750 instruction-response pairs for training and another 6,750 for validation, and leave the rest for test. Fig.~\ref{fig:eval_olive_example} shows three examples that require both detailed visual understanding and extensive commonsense knowledge.

\section{Experiments}
\subsection{Setup}
Considering that different VLMs may exhibit different training behaviors, our analysis uses four popular VLMs: BLIP-2 \cite{blip2}, MiniGPT-4 \cite{minigpt4}, LLaVA \cite{llava}, and mPLUG-Owl \cite{mplug-owl}, which have mostly not been exposed to the datasets in focus. 
As minor exceptions, BLIP-2 and MiniGPT-4 were pretrained on COCO Caption and Web CapFilt. mPLUG-Owl has been exposed to COCO Caption, and LLaVA was pretrained on the three LLaVA datasets. We avoid models that have been finetuned on many VQA datasets, such as InstructBLIP \cite{instructblip}, LLaVA 1.5 \cite{liu2023improvedllava}, and Qwen-VL \cite{Qwen-VL}.

For each model, we fine-tune the parameters that are trainable during their respective vision-language pretraining. On each source task, we train for 10K steps with a batch size of 192 for BLIP-2 and 128 for MiniGPT-4, mPLUG-Owl, and LLaVA. Other model details and hyperparameters are in Appendices~\ref{Appendix_model_details} and ~\ref{Appendix_hyperparameters}. 

\subsection{Results}
We defer the transfer performance tables to Appendix \ref{Appendix_model_perf} due to space constraints.  
With these results, we first examine the transfer learning power of the source tasks. Within every model, we rank the source tasks by the sum of their normalized performance across all target tasks. After that, for each source task, we compute the harmonic mean of its rankings across all models. The results are in Tab. \ref{tab:source_task_mean_ranking}. A-OKVQA (MC), VQAv2, ScienceQA, A-OKVQA and OCR-VQA hold the top-5 positions.

In addition, we examine the effects of output lengths. We partition the source and target tasks into three mutually exclusive and collectively exhaustive groups according to their average output lengths: 1-3 words, 6-12 words, and more than 40 words. We show the average normalized transfer performance across these groups and the top-5 best source tasks for each group in Tab. \ref{tab:source_target_length_relationship}.

\begin{figure}[!t]
	\centering
	\includegraphics[width=\columnwidth]{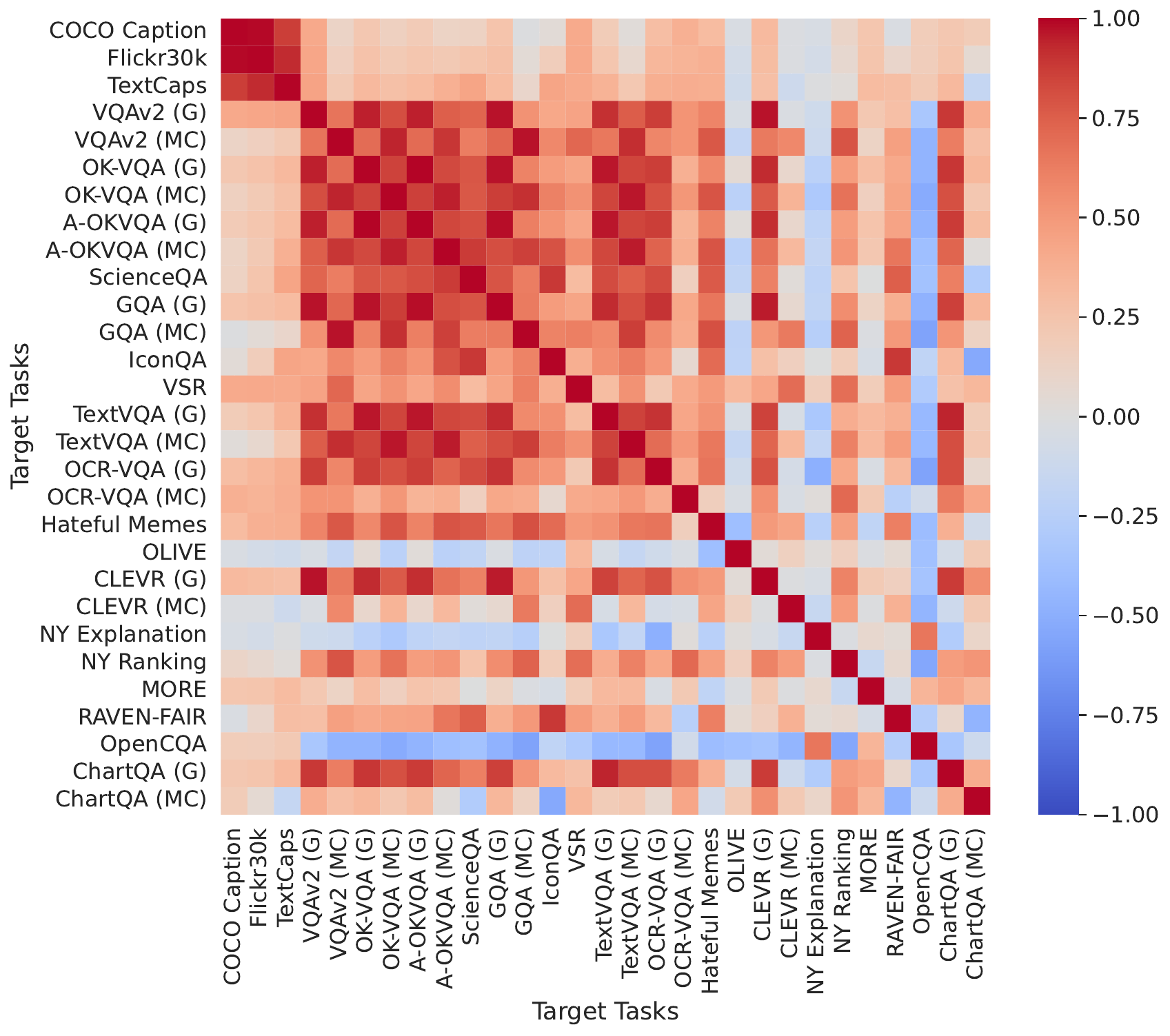}
\vspace{-3ex}
  \caption
  	{
  	Cosine similarity between target tasks computed using SVD features.
	  } 
  \label{fig:svd_similarity}
  \vspace{-1ex}
 \end{figure}

Next, we investigate the similarity of target tasks. We perform truncated SVD on $A$ with the first $D=8$ singular values. After that, we compute the cosine similarity between target tasks and visualize the results in Fig.  \ref{fig:svd_similarity}. With a mean similarity of $-0.06$, \vlbenchmark{} is the third least similar to other target tasks (see details in Appendix~\ref{Appendix_target_task_cos_sim_svd}).

\begin{figure}[!t]
	\centering
	\includegraphics[width=\columnwidth]{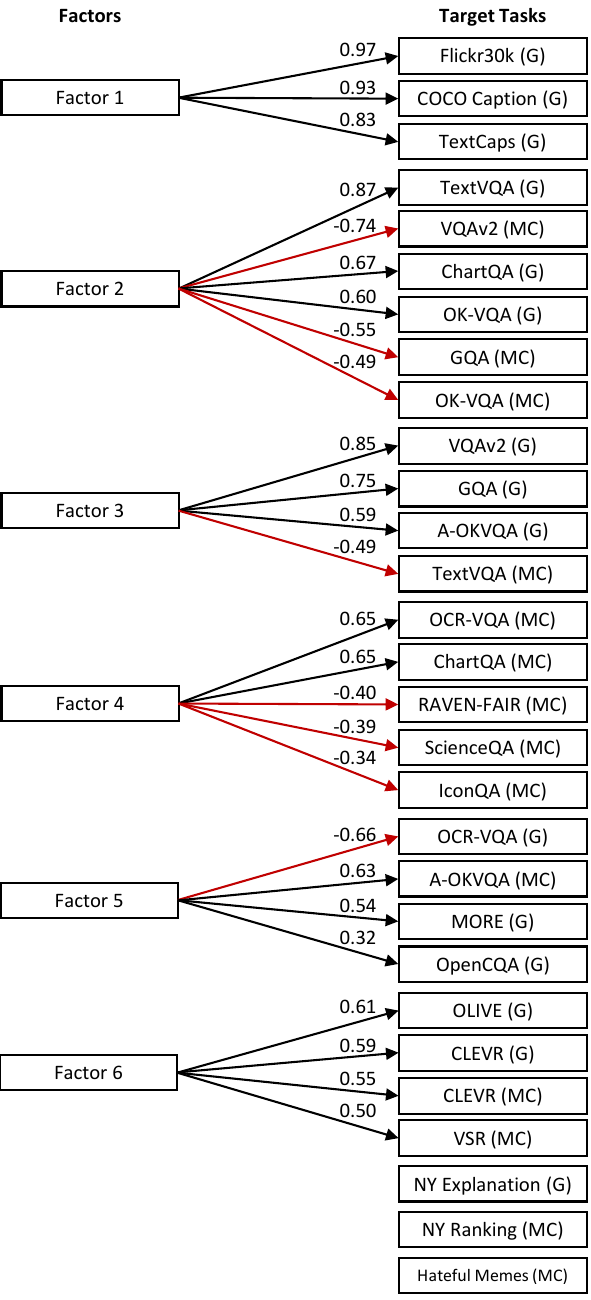}
  \caption
  	{
  	Results of EFA on the residuals $\bar{A}$. Black arrows indicate positive loadings; red arrows indicate negative loadings. Cut-off for factor loadings = 0.3.
	  } 
  \label{fig:fa_residuals}
  \vspace{-2ex}
 \end{figure}

Finally, we run EFA on the residual matrix, $\bar{A}$, and present the outcomes in Fig. \ref{fig:fa_residuals}. We plot the most significant factor loading for each target task, retaining only those that exceed an absolute value threshold of 0.3. Loadings close to 1 or -1 signify strong influences of a factor on a target task. Three tasks, New Yorker Explanation and Ranking, and Hateful Memes, do not have loadings more than 0.3 on any discovered factor, suggesting they do not share with the other tasks some VL capabilities that can be discovered by EFA. The full results are available in Appendix~\ref{Appendix_factor_analysis_details}. 

\subsection{Discussion}
In this section, we highlight important findings from our experiments. 

\begin{table}[!tbp]
\centering
\resizebox{\columnwidth}{!}
{
\begin{tabular}{@{}c|ccc@{}}
\toprule

\multirowcell{2}{Source Task \\ Output Length} & \multicolumn{3}{c}{Target Task Output Length}   \\\cmidrule{2-4}
                            & 1-3 & 6-12         & >40 \\
\midrule
1-3        & \textbf{-0.03} / \textbf{1.00} & -0.78 / 0.79 &      -0.85 / 0.44         \\
6-12     &  -0.49 / 0.64   &    -0.43 / \textbf{0.75}      & \textbf{-0.43} / 0.48    \\
>40       &    -0.90 / 0.43      &     -0.87 / 0.28      & \textbf{-0.26} / \textbf{0.55}  \\
\midrule
\multirowcell{5}{Top-5 Source \\ Tasks (Mean \\ Length)} & VQAv2 (1) & Web CapFilt (12) & \makecell{LLaVA Conver-\\sation (48)} \\
& A-OKVQA (MC) (1) & COCO Caption (10) & OpenCQA (56) \\
& A-OKVQA (1) & OCR-VQA (3) & TextCaps (12) \\
& OK-VQA (1) & Flickr30K (12) & Flickr30K (12) \\
& TextVQA (1) & ScienceQA (3) & A-OKVQA (MC) (1) \\

\bottomrule

\end{tabular}
}

\caption
{
Mean normalized performance by the mean output lengths of source and target tasks. In-domain sources are excluded. In the top section, we report the mean normalized performance across all (left) and top-5 (right) source tasks.  
}
\label{tab:source_target_length_relationship}

\end{table}
\begin{table}[!t]
\centering
\small
\setlength\tabcolsep{5pt}
\begin{tabular}{@{}lc@{}}
\toprule
Source Task        & Harmonic Mean Score \\
\midrule
A-OKVQA   (MC)     & 1.3  \\
VQAv2 (G)              & 1.3  \\
ScienceQA (MC)         & 3.8  \\
A-OKVQA (G)            & 4.6  \\
OCR-VQA (G)           & 6.0  \\
GQA (G)               & 6.2  \\
Flickr30k (G)         & 7.2  \\
OK-VQA (G)             & 7.8  \\
Web CapFilt (G)        & 7.9  \\
IconQA (MC)             & 8.4  \\
OpenCQA (G)            & 9.5  \\
TextVQA (G)           & 9.5  \\
VSR (MC)               & 10.0 \\
Hateful Memes (MC)                & 11.7 \\
COCO Caption (G)      & 13.3 \\
TextCaps (G)          & 15.0 \\
VQAv2 QG (G)        & 15.9 \\
LLaVA Conversation (G) & 17.6 \\
OLIVE (G)             & 17.8 \\
A-OKVQA QG (G)      & 17.9 \\
OK-VQA QG (G)        & 19.7 \\
LLaVA Reasoning (G)    & 21.1 \\
LLaVA Description (G) & 22.5 \\
\bottomrule
\end{tabular}%
\caption
{
Harmonic mean ranking score for source tasks. 
}
\label{tab:source_task_mean_ranking}
\vspace{-1ex}
\end{table}

\vspace{0.05in}
\noindent \textbf{The Output Length Bias.}
Tab.~\ref{tab:source_target_length_relationship} demonstrates that transfer performance is strongly influenced by the output lengths. In the top section, mismatch between the output lengths results in significant performance drops. In the bottom section, the best source tasks almost always have similar output lengths to the target tasks. This surprising finding shows that output length may be a shortcut feature for VLMs, suggesting that future test suites need a balance among tasks with different output lengths.

\vspace{0.05in}
\noindent \textbf{EFA Overview.} 
The six EFA factors resemble hierarchical clusters from SVD features (Appendix~\ref{Appendix_hierarchical_clustering}). For instance, both techniques identify a factor (Factor 1) or a cluster around the three captioning tasks, COCO, Flickr30k, and TextCaps. 
However, EFA offers insights into both the positive and negative ends of the same factor, rendering the factors more interpretable. EFA also picks up more VL skills than SVD, such as text reading versus reasoning. 

It is notable that EFA is influenced by the length bias. For example, it fails to find a humor factor shared by the two New Yorker tasks. This omission occurs because the similarity between the two tasks is masked by their differences in output lengths and MC-vs-G evaluation, leading to markedly different transfer profiles in Fig.~\ref{fig:svd_similarity}. Nonetheless, EFA successfully identifies several reasonable factors, and we discuss Factors 2-6 below.

\vspace{0.05in}
\noindent \textbf{Factors 2 and 3: Generative vs. MC Evaluation.}
Factors 2 and 3 capture the contrast between generative and MC evaluation of VQA. Generative VQAs exhibit positive loadings on both factors while MC VQAs exhibit negative loadings. Furthermore, specialized VQAs requiring OCR skills, such as TextVQA and ChartQA, load positively on Factor 2 and negatively on Factor 3, whereas generic VQAs, such as VQAv2 and GQA, exhibit the opposite pattern.

We identify two reasons for the differences between generative and MC evaluations. First, generative evaluation requires an exact match with at least one ground-truth answer, which can result in false negatives for valid answers phrased differently. In contrast, MC evaluation compares average word probabilities and does not mandate strict word matching. Second, the exact match requirement in generative evaluation makes it more sensitive to the output lengths of source tasks.

\begin{figure}[!t]
	\centering
	\includegraphics[width=\columnwidth]{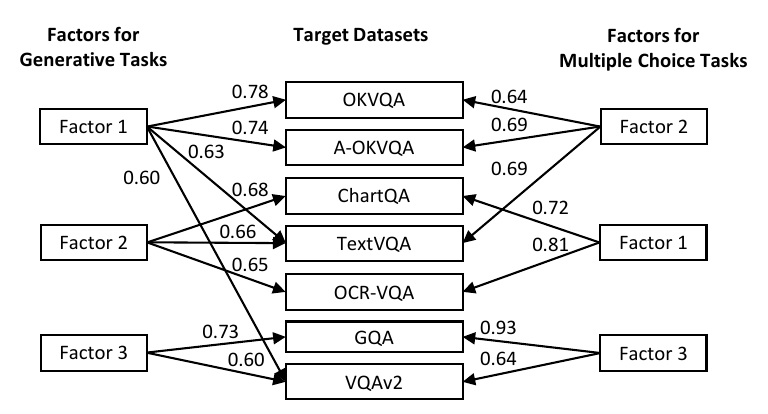}
  \caption
  	{EFA results when we extract 3 factors from the 7 generative VQA tasks and the 7 MC VQA tasks separately. We merge the results for display. Cut-off for factor loadings = 0.6.
	  } 
  \label{fig:fa_gen_mc_tasks}
    \vspace{-1ex}
\end{figure}

Nevertheless, when we analyze factors internal to the generative and MC tasks (Fig. \ref{fig:fa_gen_mc_tasks}), we observe very similar structures. In both groups, we observe a knowledge-based QA factor, which includes OKVQA and A-OKVQA, a OCR-related factor, which includes OCR-VQA and ChartQA, and a generic or spatial relation factor, which includes GQA and VQAv2. This underscores the robustness of EFA in identifying underlying structures when provided with appropriate input data.

\vspace{0.05in}
\noindent 
\textbf{Factors 4 \& 5: Text Reading vs Reasoning.}
Factors 4 and 5 distinguish between tasks that involve simple text extraction from those that require deeper and multi-hop reasoning. OCR-VQA and ChartQA, which primarily focus on reading text and numbers from images, exhibit loadings opposite to RAVEN-FAIR, ScienceQA, and IconQA, which demand strong logical reasoning skills, and opposite to A-OKVQA, MORE, and OpenCQA, which necessitate the use of external knowledge and contextual understanding. The fact that EFA can find these reasonable skills illustrates its power. 

\vspace{0.05in}
\noindent \textbf{Factor 6: Spatial Reasoning.} 
Factor 6 is characterized by spatial reasoning, as CLEVR and VSR are both designed for this purpose. Notably, while \vlbenchmark{} shows the highest loading on Factor 6, its communality (overall variance explained) is only 0.4. The remaining variance in \vlbenchmark{} is not explained by the factors identified in our analysis. This implies that although \vlbenchmark{} requires spatial reasoning skills, these skills only account for a small portion of skills required by \vlbenchmark. 

\begin{table}[!tbp]
\centering
\resizebox{\columnwidth}{!}{%
\begin{tabular}{ccc}
\toprule

VLMs in $A_{train}$ & VLM in $A_{test}$ &  L2-norm Error \\
\midrule
MiniGPT-4, mPLUG-Owl, LLaVA  & BLIP-2    & 85.6 \\
BLIP-2, mPLUG-Owl, LLaVA     & MiniGPT-4 & 14.2 \\
BLIP-2, LLaVA, MiniGPT-4     & mPLUG-Owl & 16.3 \\
BLIP-2, MiniGPT-4, mPLUG-Owl & LLaVA     & 63.0 \\
MiniGPT-4, mPLUG-Owl         & LLaVA     & 25.2 \\
\bottomrule
\end{tabular}%
 }
  \caption
  	{
Robustness analysis of EFA. The generalization among mPLUG-Owl, MiniGPT-4, and LLaVA indicates that EFA is robust.
	  } 
  \label{tbl:efa_robustness}
  \vspace{-2ex}
\end{table}

\vspace{0.05in}
\noindent 
\textbf{A-OKVQA and VQAv2 are effective source tasks.} These two are among the sources with the highest transfer performance (Fig. \ref{tab:source_task_mean_ranking}). They not only transfer well to VQA tasks but also to other complex tasks, such as New Yorker Ranking. 
We hypothesize that the large and diverse data in VQAv2 contributes to its strong transferability.
Interestingly, even though A-OKVQA is 24 times smaller than VQAv2, it still transfers well. We hypothesize that the main skill that A-OKVQA requires, knowledge-enabled reasoning, is an important skill for VL competence. In contrast, OK-VQA, which is designed for knowledge recall, is not as beneficial to the target tasks. 

\vspace{0.05in}
\noindent 
\textbf{Humor, sarcasm, and abstract reasoning remain difficult.}
All models we tested struggle to understand humor and sarcasm, as captured by the New Yorker datasets and MORE. The models also perform barely above chance level on RAVEN-FAIR, an abstract reasoning task. Surprisingly, EFA is able to correctly place RAVEN-FAIR under the reasoning factor (negative Factor 4), despite the tiny variance caused by overall poor performance. 

\vspace{0.05in}
\noindent 
\textbf{Robustness and Inter-model Similarity.}
We empirically verify the robustness of EFA using a leave-one-out analysis. We partition the aggregate performance matrix $A$ into a training set of data from three models, $A_{train} \in \mathbb{R}^{3N \times K}$, and a test set of data from the remaining model, $A_{test} \in \mathbb{R}^{N \times K}$.
Similar to before, we first perform a one-factor EFA on $A_{train}$ and calculate the residuals $\bar{A}_{train}$ and $\bar{A}_{test}$. Then, we perform a six-factor EFA, which decomposes $\bar{A}_{train}$  to the factor weights $W_{train}$, the factors $H_{train}$, and a mean vector $\mu_{train}$, 
\begin{equation}
\label{efa_train}
\bar{A}_{train} = W_{train} H_{train}^{\top} + \mu_{train} + \epsilon_{train}.
\end{equation}
After that, we use the identified $H_{train}$ and $\mu_{train}$ to reconstruct $\bar{A}_{test}$ and calculate the L2-norm of the reconstruction error $\epsilon_{test}$, which measures the generalization of the factor models, 
\begin{equation}
\label{efa_test}
\bar{A}_{test} = W_{test} H_{train}^{\top} + \mu_{train} + \epsilon_{test}.
\end{equation}
We repeat the experiment five times, with different combinations of VLMs in $A_{train}$ and $A_{test}$. The results are presented in Tab.~\ref{tbl:efa_robustness}.

We observe that the factors can generalize to mPLUG-Owl and MiniGPT-4 fairly well, but not so well to BLIP-2 and LLaVA. However, if we remove BLIP-2 from the training set, as shown by the last row of Tab.~\ref{tbl:efa_robustness}, the factors would generalize to LLaVA. We attribute this observation to the fact that BLIP-2 is more thoroughly trained than the other three models. In particular, its average zero-shot performance across all target tasks is 47.6. LLaVA, mPLUG-Owl, and MiniGPT-4 score 29.5, 23.4, and 22.6, respectively, suggesting these three networks are much more similar to each other.
Hence, we argue that the observed generalization among mPLUG-Owl, MiniGPT-4, and LLaVA indicates that EFA is robust. 

\vspace{0.05in}
\noindent 
\textbf{Implications of Factor Analysis.} Our findings have the following implications for the design of VL benchmarks. Firstly, to prevent shortcut learning and giving unfair advantages to any source training tasks, VL benchmarks should contain tasks of different output lengths and include both generative and MC evaluation. Secondly, instead of intuition-based categories, VL benchmarks may group tasks based on statistically discovered VL factors, and score VLMs accordingly. 

\begin{table}[!tbp]
\centering
\resizebox{\columnwidth}{!}{%
\begin{tabular}{ccccc}
\toprule
Linguistic Diversity           & \vlbenchmark{}  & VQAv2  & GQA    & LLaVA  \\
\midrule
Number of words       & 4 M   & 7 M & 9 M  & 28 M  \\
Word distribution entropy   & \textbf{9.62}  & 8.90  & 6.74  & 9.46 \\
\bottomrule
\end{tabular}%
}
  \caption
  	{
Linguistic diversity. \vlbenchmark{} has higher linguistic diversity than VQAv2, GQA and LLaVA.
	  } 
  \label{tbl:olive_lin_diversity}
\end{table}
\begin{table}[!tbp]
\centering
\resizebox{\columnwidth}{!}{%
\begin{tabular}{cccccc}
\toprule
\multicolumn{2}{c}{Image Diversity}           & \vlbenchmark{}  & VQAv2  & GQA    & LLaVA  \\
\midrule
\multicolumn{2}{c}{Number of images}       & 9 K   & 123 K & 73 K  & 81 K  \\
\midrule
\multirow{2}{*}{Generalized variance} & k = 8  & \textbf{348.40} & 205.61 & 198.06 & 207.50 \\
       & k = 16 & \textbf{84.28}  & 66.85  & 68.41  & 70.07 \\
\bottomrule
\end{tabular}%
}
  \caption
  	{
Image diversity. \vlbenchmark{} has higher generalized variance than the other datasets.
	  } 
  \label{tbl:olive_img_diversity}
\end{table}

\subsection{Analysis of OLIVE}
\noindent 
\textbf{Uniqueness.} 
Intended to simulate real-world user instructions, \vlbenchmark{} features a unique transfer profile and has very low cosine similarity with other tests (Fig.~\ref{fig:svd_similarity}). Additionally, EFA only explains 0.4 of the total variance in \vlbenchmark{}, indicating that the identified factors cannot fully explain model behaviors on \vlbenchmark{}. These facts support our thesis that \vlbenchmark{} measures an aspect of VL capabilities that few existing datasets do. 

\vspace{0.05in}
\noindent 
\textbf{Diversity.} 
We evaluate the linguistic and image and diversity of \vlbenchmark{}, and compare them against those of existing datasets, such as VQAv2, GQA, and LLaVA (conversation, description and reasoning). Notably, both VQAv2 and GQA are large datasets with 658K and 956K data points respectively. LLaVA, similar to \vlbenchmark{}, is generated using a GPT model, albeit without any human correction. 

For linguistic diversity, we show the entropy of the multinomial word distributions in Tab.~\ref{tbl:olive_lin_diversity}. \vlbenchmark{} exhibits higher entropy, indicating a greater level of diversity compared to the other datasets. 

For image diversity, we calculate the generalized variance for CLIP image features, which is the product of the top-k eigenvalues of the covariance matrix. This is akin to calculating the variance explained by the top-k principal components of PCA. We consider k values of 8 and 16. Tab.~\ref{tbl:olive_img_diversity} shows that \vlbenchmark{} achieves higher generalized variance than the other datasets, implying greater diversity.

\section{Related Work}

\textbf{Relationships among Tasks.} Relationships between machine learning tasks has been studied from the perspective of transfer learning and multi-task learning. A number of works focus on identifying transfer relationships from empirical data \cite{taskonomy, achille2021information,
dwivedi2019representation, achille2019task2vec,
xi2023connectivity}. A typical strategy is to train a base network on multiple source tasks and test the resulting networks on target tasks. In multi-task learning, the research focuses on identifying optimal grouping of tasks that should be learned together to maximize synergy~\cite{standley2020tasks, fifty2021efficiently_task_grouping_mt, 
ben2003exploiting, 
kumar2012learning,
song2022efficient}. Different from the above, our work focuses on vision-language tasks and identifying latent factors and potential biases responsible for the observed performance.

\vspace{0.05in}
\noindent 
\textbf{Broad-coverage Multimodal Test Suites.} As VLMs begin to excel on an ever growing set of tasks, the test suites have also grown in size. Early benchmarks contain only 
a few tasks. For example, \citea{zhou2022vlue} employs 4 tasks and \citea{lin2021gem} employs 8 tasks in 4 groups. More recent benchmarks 
\cite{bugliarello2022iglue,bitton2023visitbench,bai2023touchstone, yu2023mmvet, LVLM-ehub, li2023reformeval}
uses increasingly more tasks in order to achieve broader coverage of VL capabilities, and group tasks into VL capabilities based on human intuition. For instance,
\citea{seed-bench} categorizes tasks into 12 aspects focusing on spatial and temporal understanding.
To the best of our knowledge, our work is the first data-driven approach to identify the VL capabilities.

\section{Conclusions}
In this work, we aim to empirically uncover latent factors and biases that contribute to the performance of various VLMs on VL tasks. Using EFA, we identify six highly interpretable factors and biases that affect performance. We also contribute a new dataset, \vlbenchmark. We hope this research will lead to the creation of VL benchmarks with more balanced and complete coverage of VL capabilities.

\section{Acknowledgments}
Anthony Meng Huat Tiong is supported by Salesforce and Singapore Economic Development Board under the Industrial Postgraduate Programme. 
Junqi Zhao and Boyang Li are supported by the Nanyang Associate Professorship and the National Research Foundation Fellowship (NRF-NRFF13-2021-0006), Singapore. We thank Flitto\footnote[3]{\url{https://datalab.flitto.com/en}} for their support and expertise in \vlbenchmark{} annotation.

\section{Ethical Considerations}
The potential risks of large language models have been discussed in literature, including ~\citet{flan}, ~\citet{llama}, and ~\cite{vicuna2023}. We neither propose nor release new models in this work. For easy replication, we use open-source visions-language models to analyze publicly available, academic datasets. 

We consider issues of privacy, toxicity, and fair compensation in the production of \vlbenchmark. We utilize images from LAION-aesthetics, whose privacy policy is in compliance with GDPR. For a given image, we generate an instruction and outputs using ChatGPT. To mitigate the potential issues of hallucination, toxicity, and harmful content in ChatGPT generated content, we hire an annotation company, Flitto, to thoroughly review and correct any errors presented in the data. We establish specific guidelines for annotations, prioritizing content that is free from harmful information. The identity of the annotators are anonymized to safeguard their privacy. We pay Flitto \$3 USD per image to ensure fair compensation.

\section{Limitations}
We focus our study of transfer performance using only one source task instead of multi-task learning setting due to the computational constraint. Thus, we do not investigate the interaction of multiple source tasks on the target tasks. To assess the model transfer performance, our work requires a substantial amount of computation and scaling our approach to diverse models and datasets is inefficient.

\bibliography{bib}
\newpage
\appendix

\section{Data Collection Protocol for \vlbenchmark}
\label{Appendix_olive_data_collection}
\vlbenchmark{} comprises 9,450 images, 30,120 unique instructions and 47,250 reponses. The examples can be broadly categorized into 4 groups: visual recognition, creative writing, knowledge-based, and elaborated description. Some examples are shown in Figure~\ref{fig:eval_olive_example}.

In our benchmark \vlbenchmark{}, we use the text-only version of ChatGPT to generate instructions and outputs for each image. Specifically, we sample images from LAION-Aesthetics~\cite{schuhmann2022laion} and use the BLIP-2~\cite{blip2} captioning model to encode the visual information in each image into captions.  These generated captions, along with the original LAION captions - which may contain entity-specific knowledge useful for generating varied instruction-output data - are then fed into ChatGPT.  Additionally, for each of the aforementioned categories, we manually annotate a few seed examples, and use these as in-context examples to guide ChatGPT.

The instructions and outputs generated by ChatGPT could contain incorrect information due to model hallucination, which undermines their reliability for use as an evaluation benchmark.
Recognizing this, we hire an annotation company, Flitto which recruits human annotators to thoroughly inspect and correct any erroneous data. They are task to: 1) ensure that the instructions contain minimal shortcut information, which could enable the model to produce correct outputs without having to understand the image, 2) verify the accuracy of the output and confirm that it is free from harmful content, and 3) fact-check knowledge-based information. This comprehensive review process helps to enhance the overall quality and reliability of the data.

\section{ChatGPT Prompts for \vlbenchmark{}}
\label{Appendix_olive_prompt} 
Following~\cite{llava} and ~\cite{alpaca}, we construct prompts for ChatGPT~\cite{chatgpt} to generate instructions and outputs for different categories: visual recognition, elaborated description, knowledge-based and creative writing. 
For elaborated description, we randomly sample from a list of instructions which inquire about image description.

\begin{tcolorbox}[title=Prompt for generating creative writing instructions]
	You are given several image captions, each describing the same image you are observing. Using your creativity and imagination, think of a new instruction that can be induced from the image captions.\\
	
	Here are the requirements:\\
	1. Try not to repeat the verb for each instruction to maximize diversity.\\
	2. The language used for the instruction also should be diverse. Either an imperative sentence or a question is permitted.\\
	3. The type of instruction should be diverse.\\
	4. The instruction must not involve counting.\\
	5. Make the instruction challenging by not including the visual content details in the instruction so that one must use the captions to understand the instruction.\\
	6. Replace the name of the object entity with a generic term or category, for example replace bus as this vehicle, dress as this clothing, etc.\\
	7. The format of the instruction should follow the examples shown below. Make sure it is numbered and end with '\#\#\#'.
\end{tcolorbox}

\begin{tcolorbox}[title=Prompt for generating knowledge-based instructions]
	You are given several image captions, each describing the same image you are observing. Using your creativity and imagination, think of a new instruction that can be induced from the image captions.\\
	
	Here are the requirements:\\
	1. Try not to repeat the verb for each instruction to maximize diversity.\\
	2. The language used for the instruction also should be diverse. Either an imperative sentence or a question is permitted.\\
	3. The instruction should be diverse and ask a question that requires reasoning, not just simple visual recognition.\\
	4. Given the instruction, one should require first understanding the visual content, then based on the background knowledge or reasoning, either explain why the things are happening that way, or provide guides and help to user's request.\\
	5. Make the instruction challenging by not including the visual content details in the instruction so that the user must use the captions to understand the instruction.\\
	6. Replace the name of the object entity with a generic term or category, for example replace bus as this vehicle, dress as this clothing, etc.\\
	7. The instruction must not involve counting.\\
	8. The format of the instruction should follow the examples shown below. Make sure it is numbered and end with '\#\#\#'.
\end{tcolorbox}

\begin{tcolorbox}[title=Prompt for generating visual recognition instructions]
	You are given several image captions, each describing the same image you are observing. Using your creativity and imagination, think of a new instruction that can be induced from the image captions.\\
	
	Here are the requirements:\\
	1. Try not to repeat the verb for each instruction to maximize diversity.\\
	2. The language used for the instruction also should be diverse. Either an imperative sentence or a question is permitted.\\
	3. The instruction should ask about the visual content of the image, including the object types, object actions, object locations, etc. Only include instruction that has definite answers founded in the captions.\\
	4. Include complex instruction that is relevant to the content in the image, for example, asking about background knowledge of the objects in the image, asking to discuss about events happening in the image, etc. Again, do not ask about uncertain details.\\
	5. Make the instruction challenging by not including the visual content details in the instruction so that the one must use the captions to understand the instruction.\\
	6. Replace the name of the object entity with a generic term or category, for example replace bus as this vehicle, dress as this clothing, etc.\\
	7. The instruction must not involve counting.\\
	8. The format of the instruction should follow the examples shown below. Make sure it is numbered and end with '\#\#\#'.
\end{tcolorbox}

\begin{tcolorbox}[title=List of instructions for elaborated description (part 1)]
	\begin{itemize} 
		\itemsep0em
		\item Provide a vivid description of the image.
		\item What is a suitable paragraph that describes this image?
		\item Compose a passage that depicts this image.
		\item What is this image about?
		\item What's happening in the scene?
		\item Can you describe the main features of this image for me?
		\item What are the key details in this picture?
		\item Can you elaborate on the elements of the picture provided?
		\item What do you think is going on in this photo?
		\item Can you provide a comprehensive description of the image?
		\item Describe the following image in detail.
		\item Provide a detailed portrayal of what's captured in this image.
		\item Offer an intricate description of the image you see.
		\item Please share a thorough run down of the image that has been presented.
		\item Could you elaborate on the contents of the displayed image with thoroughness?
		\item Clarify the contents of the displayed image with elaborate detail.
	\end{itemize}
\end{tcolorbox}

\begin{tcolorbox}[title=List of instructions for elaborated description (part 2)]
	\begin{itemize} 
		\itemsep0em
		\item Can you offer a comprehensive portrayal of the image?
		\item Could you highlight and elaborate on the details of the image?
		\item Portray the image with a vivid comprehensive narrative.
		\item Analyze the image in a descriptive manner.
		\item Write an well-detailed depiction of the given image.
		\item How would you describe this photo in great detail?
		\item Can you give a detailed account of what you see in this image?
		\item Describe this image using your own words.
		\item Please describe what you see in the image with as much detail as possible.
		\item I need you to depict the image with utmost detail.
		\item Can you describe the image below in exhaustive detail?
		\item Please provide a complete description of what is shown in the picture.
		\item I would like you to give a detailed clarification of the contents of the displayed image.
		\item Could you provide a detailed and comprehensive representation of the image?
		\item Provide a comprehensive illustration of the image.
		\item Illustrate the image using a well-detailed description.
		\item Write a rich narrative for this image.
		\item Give a thorough description for the given image.
		\item Write a vivid account of the moment captured in this image.
		\item Create a narrative that is rich and vivid based on the image presented.
	\end{itemize}
\end{tcolorbox}

\begin{tcolorbox}[title={Prompt for generating visual recognition, knowledge-based and creative writing outputs}]
	You are given an instruction and several image captions, each caption describing the same image you are observing. Generate an output resulting from following the instruction.\\
	
	Here are the requirements:\\
	1. The output is the response to the instruction and the caption.\\
	2. The output must utilize the information in the caption and must not contradict the caption.\\
	3. If the output is unknown without further context, generate "unknown" as the output.\\
	4. When using the information from the caption, directly explain the scene, do not mention that the information source is the caption. Always answer as if you are directly looking at the image.\\
	5. Provide detailed output when answering complex instruction. For example, give detailed examples or reasoning steps to make the content more convincing and well-organized.\\
	6. The format of the output should follow the examples shown below. Make sure it is numbered and end with '\#\#\#'.
\end{tcolorbox}

\begin{tcolorbox}[title={Prompt for generating elaborated description outputs}]
You are given several image captions, each caption describing the same image you are observing.\\

Here are the requirements:\\
1. Generate an output that describes the image in detail.\\
2. The output must utilize the information in the caption and must not contradict the caption. Do not include description of objects that is not presented in the caption.\\
3. When using the information from the caption, directly explain the scene, do not mention that the information source is the caption. Always answer as if you are directly looking at the image.\\
4. The format of the output should follow the examples shown below. Make sure it is numbered and end with '\#\#\#'.
\end{tcolorbox}

\section{Model Details}
\label{Appendix_model_details}
We experiment with 4 different VLMs as follows:
\begin{itemize}
    \item BLIP-2 utilizes ViT-G/14~\cite{fang2023eva} as the image encoder and FlanT5\textsubscript{XL}~\cite{flan} as the LLM. We initialize BLIP-2 from the pretrained checkpoint and only fine-tune the Q-former parameters. Both the image encoder and the LLM are frozen. The total and trainable parameters are 4B and 187M respectively.
    \item MiniGPT-4 adopts ViT-G/14~\cite{fang2023eva} as the image encoder and Vicuna\textsubscript{7B}~\cite{vicuna2023} as the LLM. It consists of the BLIP-2 Q-former and a linear layer connecting the image encoder and the LLM. The Q-former is initialized from BLIP-2. All parameters are frozen except the linear layer. The total and trainable parameters are 8B and 3M respectively.
    \item mPLUG-Owl adopts ViT-L/14~\cite{radford2021learning} as the image encoder and LLaMA\textsubscript{7B}~\cite{llama} as the LLM. It consists of a visual abstractor module between the image encoder and the LLM. All parameters are frozen except LoRA~\cite{hu2022lora} parameters on the LLM. The total and trainable parameters are 7B and 4M respectively.
    \item  LLaVA adopts ViT-L/14~\cite{radford2021learning} as the image encoder and LLaMA\textsubscript{7B}~\cite{llama} as the LLM.
    It consists of a linear layer in between the image encoder and the LLM. All parameters are frozen except the linear layer and LoRA~\cite{hu2022lora} parameters on the LLM. The total and trainable parameters are 7B and 164M respectively.
\end{itemize}

\section{Additional Hyperparameters}
\label{Appendix_hyperparameters}
We individually finetune models for each task using datasets in the instruction format. Only one instruction template is used per task, as preliminary experiments show using multiple templates per task degrades performance.

For all experiments using the same model architecture, we keep the hyperparameters constant. We set the training iteration to 10K steps. 
The batch size for BLIP-2 is 192 and 128 for the other three models.  
For BLIP-2, MiniGPT-4 and mPLUG-Owl, 
we train the model using AdamW optimizer with a weight decay of 0.05. The learning rate is linearly increased from $1\mathrm{e}{-8}$ to $1\mathrm{e}{-5}$ in the first 200 steps and then cosine decayed to 0. 
For LLaVA, 
we use a weight decay of 0. The learning rate is linearly rises from 0 to $2\mathrm{e}{-5}$ across the initial 200 steps and then cosine decay to 0.

We output model performances at intervals of 1,000 iterations and select the best checkpoint using the validation set for evaluation. 

All experiments are performed on a machine with 8 or 16 Nvidia A100 GPUs. 
On average, each experiment involves around 2 hours of training and another 2 hours of evaluation. We utilize LAVIS~\cite{lavis} library for training of BLIP-2, MiniGPT-4 and mPLUG-Owl. For LLaVA, we utilize LLaVA original author's codebase for training. All evaluations are performed on LAVIS.

\section{Complete Results}
\label{Appendix_model_perf} 
In this section, we show all the experimental results from all four models. Tables \ref{eval:affinity_matrix_raw_blip_landscape}-\ref{eval:affinity_matrix_raw_mplugowl_landscape} show the raw transfer learning performance, where rows denote the source tasks and columns are target tasks. Tables \ref{eval:affinity_matrix_norm_blip_landscape}-\ref{eval:affinity_matrix_norm_mplugowl_landscape} show the normalized performance, where the rows (source tasks) are sorted in a descending order of average performance. 

Among the source tasks, LLaVA Conversation shows strong transfer to \vlbenchmark{} for BLIP-2 and LLaVA. The relatively good transferability could attribute to the fact that LLaVA Conversation and \vlbenchmark{} share some similarities in data distribution since the instruction-response pairs are generated using OpenAI GPT models~\cite{chatgpt, gpt4}. However, the key difference is that \vlbenchmark{} is inspected by human annotators to rectify erroneous data, while LLaVA Conversation does not undergo this correction process.

\begin{landscape}
\begin{table}[!tp]
	\resizebox{\columnwidth}{!}{%
		\begin{tabular}{c|c|ccccccccccccccccccccccccccccc}
			\toprule
			Source   Task &
			\setlength\extrarowheight{0pt}\begin{tabular}[c]{@{}c@{}}Dataset\\ Size \end{tabular} &
			\multicolumn{29}{c}{Target Task} \\
			&
			&
			\setlength\extrarowheight{0pt}\begin{tabular}[c]{@{}c@{}}COCO\\ Caption \end{tabular}  &
			\setlength\extrarowheight{0pt}\begin{tabular}[c]{@{}c@{}}Flickr\\ 30k\end{tabular}  &
			\setlength\extrarowheight{0pt}\begin{tabular}[c]{@{}c@{}}Text\\ Caps\end{tabular}  &
			\multicolumn{2}{c}{VQAv2} &
			\multicolumn{2}{c}{OK-VQA} &
			\multicolumn{2}{c}{A-OKVQA} &
			\setlength\extrarowheight{0pt}\begin{tabular}[c]{@{}c@{}}Science\\ QA\end{tabular} &
			\multicolumn{2}{c}{GQA} &
			\setlength\extrarowheight{0pt}\begin{tabular}[c]{@{}c@{}}Icon\\ QA \end{tabular} &
			VSR &
			\multicolumn{2}{c}{CLEVR} &
			\setlength\extrarowheight{0pt}\begin{tabular}[c]{@{}c@{}}RAVEN-\\ FAIR \end{tabular} &
			\multicolumn{2}{c}{\setlength\extrarowheight{0pt}\begin{tabular}[c]{@{}c@{}}Text\\ VQA\end{tabular}} &
			\multicolumn{2}{c}{\setlength\extrarowheight{0pt}\begin{tabular}[c]{@{}c@{}}OCR-\\ VQA\end{tabular}} &
			\setlength\extrarowheight{0pt}\begin{tabular}[c]{@{}c@{}}Open\\ CQA \end{tabular} &
			\multicolumn{2}{c}{\setlength\extrarowheight{0pt}\begin{tabular}[c]{@{}c@{}}Chart\\ QA\end{tabular}} &
			HM &
			\setlength\extrarowheight{0pt}\begin{tabular}[c]{@{}c@{}}NY\\ Explain \end{tabular} &
			\setlength\extrarowheight{0pt}\begin{tabular}[c]{@{}c@{}}NY\\ Rank \end{tabular} &
			MORE &
			\vlbenchmark \\
 &
&
&
&
&
G &
MC &
G &
MC &
G &
MC &
&
G &
MC &
&
&
G &
MC &
&
G &
MC &
G &
MC &
&
G &
MC &
&
&
&
&
\\
\midrule
Zero-shot &
- &
\cellcolor[HTML]{80C77D}128.8 &
\cellcolor[HTML]{A0D07F}79.2 &
\cellcolor[HTML]{CADC81}71.4 &
\cellcolor[HTML]{9ECF7F}63.0 &
\cellcolor[HTML]{E3E383}64.7 &
\cellcolor[HTML]{BAD780}40.9 &
\cellcolor[HTML]{CBDC81}59.2 &
\cellcolor[HTML]{C5DB81}43.6 &
\cellcolor[HTML]{DDE182}70.2 &
\cellcolor[HTML]{F3E884}69.6 &
\cellcolor[HTML]{C6DB81}43.9 &
\cellcolor[HTML]{F7E984}46.8 &
\cellcolor[HTML]{FCEA84}48.0 &
\cellcolor[HTML]{B8D780}65.7 &
\cellcolor[HTML]{9ACE7F}33.9 &
\cellcolor[HTML]{F8E984}36.9 &
\cellcolor[HTML]{FCC37C}12.4 &
\cellcolor[HTML]{B3D580}26.6 &
\cellcolor[HTML]{D6E082}62.7 &
\cellcolor[HTML]{ECE683}36.9 &
\cellcolor[HTML]{93CC7E}72.6 &
\cellcolor[HTML]{F8E984}6.9 &
\cellcolor[HTML]{98CE7F}9.1 &
\cellcolor[HTML]{82C77D}51.1 &
\cellcolor[HTML]{FEEB84}55.0 &
\cellcolor[HTML]{F0E784}8.4 &
\cellcolor[HTML]{DEE283}53.5 &
\cellcolor[HTML]{78C47D}14.4 &
\cellcolor[HTML]{FEDF81}5.2 \\
COCO Caption &
567 K &
\cellcolor[HTML]{63BE7B}140.9 &
\cellcolor[HTML]{94CD7E}83.0 &
\cellcolor[HTML]{BFD981}75.1 &
\cellcolor[HTML]{FCC27C}32.5 &
\cellcolor[HTML]{FEEB84}63.3 &
\cellcolor[HTML]{FDC87D}23.8 &
\cellcolor[HTML]{FEE883}56.1 &
\cellcolor[HTML]{FDC97D}27.2 &
\cellcolor[HTML]{D4DF82}70.7 &
\cellcolor[HTML]{F8E984}68.8 &
\cellcolor[HTML]{FEDC81}33.1 &
\cellcolor[HTML]{FEE983}45.9 &
\cellcolor[HTML]{E6E483}51.9 &
\cellcolor[HTML]{ACD380}66.2 &
\cellcolor[HTML]{CEDD82}31.0 &
\cellcolor[HTML]{FED980}34.9 &
\cellcolor[HTML]{FCC37C}12.4 &
\cellcolor[HTML]{F9EA84}18.3 &
\cellcolor[HTML]{FEE382}57.7 &
\cellcolor[HTML]{E7E583}37.6 &
\cellcolor[HTML]{B3D580}71.5 &
\cellcolor[HTML]{E4E483}10.6 &
\cellcolor[HTML]{B8D780}8.0 &
\cellcolor[HTML]{C4DA81}46.4 &
\cellcolor[HTML]{F7E984}55.6 &
\cellcolor[HTML]{DBE182}9.5 &
\cellcolor[HTML]{E6E483}53.4 &
\cellcolor[HTML]{F4E884}11.1 &
\cellcolor[HTML]{F5E884}7.9 \\
Flickr30k &
145 K &
\cellcolor[HTML]{A6D27F}112.7 &
\cellcolor[HTML]{63BE7B}99.1 &
\cellcolor[HTML]{B0D580}80.0 &
\cellcolor[HTML]{BDD881}57.9 &
\cellcolor[HTML]{D9E082}65.2 &
\cellcolor[HTML]{E7E583}35.3 &
\cellcolor[HTML]{E9E583}57.7 &
\cellcolor[HTML]{CEDD82}42.5 &
\cellcolor[HTML]{ABD380}72.8 &
\cellcolor[HTML]{FDEB84}68.1 &
\cellcolor[HTML]{E6E483}40.3 &
\cellcolor[HTML]{FBEA84}46.5 &
\cellcolor[HTML]{EBE583}51.1 &
\cellcolor[HTML]{ACD380}66.2 &
\cellcolor[HTML]{ABD380}32.9 &
\cellcolor[HTML]{FEDF81}35.5 &
\cellcolor[HTML]{FCC37C}12.4 &
\cellcolor[HTML]{BAD780}25.7 &
\cellcolor[HTML]{DFE283}62.2 &
\cellcolor[HTML]{E8E583}37.5 &
\cellcolor[HTML]{8DCB7E}72.8 &
\cellcolor[HTML]{E5E483}10.4 &
\cellcolor[HTML]{ACD380}8.4 &
\cellcolor[HTML]{BCD881}47.0 &
\cellcolor[HTML]{ECE683}56.6 &
\cellcolor[HTML]{D5DF82}9.8 &
\cellcolor[HTML]{FEE683}52.8 &
\cellcolor[HTML]{B5D680}12.8 &
\cellcolor[HTML]{FEEB84}6.0 \\
Web CapFilt &
23,147 K &
\cellcolor[HTML]{73C37C}134.2 &
\cellcolor[HTML]{99CE7F}81.4 &
\cellcolor[HTML]{B4D680}78.9 &
\cellcolor[HTML]{ACD380}60.7 &
\cellcolor[HTML]{DEE283}65.0 &
\cellcolor[HTML]{CDDD82}38.5 &
\cellcolor[HTML]{ECE683}57.6 &
\cellcolor[HTML]{C6DB81}43.5 &
\cellcolor[HTML]{C3DA81}71.5 &
\cellcolor[HTML]{F9EA84}68.6 &
\cellcolor[HTML]{D4DF82}42.4 &
\cellcolor[HTML]{F3E884}47.1 &
\cellcolor[HTML]{F7E984}48.9 &
\cellcolor[HTML]{E6E483}63.9 &
\cellcolor[HTML]{C3DA81}31.6 &
\cellcolor[HTML]{FDCA7D}33.4 &
\cellcolor[HTML]{FCC37C}12.4 &
\cellcolor[HTML]{DBE182}21.9 &
\cellcolor[HTML]{E1E383}62.0 &
\cellcolor[HTML]{FEE983}33.7 &
\cellcolor[HTML]{AFD480}71.7 &
\cellcolor[HTML]{EBE583}9.5 &
\cellcolor[HTML]{C2DA81}7.7 &
\cellcolor[HTML]{D4DF82}45.3 &
\cellcolor[HTML]{F0E784}56.2 &
\cellcolor[HTML]{E0E383}9.3 &
\cellcolor[HTML]{FEDF81}52.5 &
\cellcolor[HTML]{6BC17C}14.7 &
\cellcolor[HTML]{EFE784}9.2 \\
TextCaps &
549 K &
\cellcolor[HTML]{FEDA80}65.5 &
\cellcolor[HTML]{FEE783}46.6 &
\cellcolor[HTML]{63BE7B}106.0 &
\cellcolor[HTML]{FDCD7E}36.7 &
\cellcolor[HTML]{FEE883}62.4 &
\cellcolor[HTML]{FEE783}31.3 &
\cellcolor[HTML]{FEE983}56.2 &
\cellcolor[HTML]{FBEA84}37.1 &
\cellcolor[HTML]{D4DF82}70.7 &
\cellcolor[HTML]{FCEB84}68.2 &
\cellcolor[HTML]{FEE182}34.7 &
\cellcolor[HTML]{FEE182}44.1 &
\cellcolor[HTML]{E4E483}52.2 &
\cellcolor[HTML]{FEDF81}61.5 &
\cellcolor[HTML]{CCDD82}31.1 &
\cellcolor[HTML]{FEE382}35.9 &
\cellcolor[HTML]{FCC37C}12.4 &
\cellcolor[HTML]{B8D780}26.0 &
\cellcolor[HTML]{9FD07F}66.3 &
\cellcolor[HTML]{EDE683}36.8 &
\cellcolor[HTML]{95CD7E}72.6 &
\cellcolor[HTML]{E5E483}10.4 &
\cellcolor[HTML]{A6D27F}8.6 &
\cellcolor[HTML]{D3DF82}45.4 &
\cellcolor[HTML]{F3E884}56.0 &
\cellcolor[HTML]{C8DB81}10.6 &
\cellcolor[HTML]{FDD17F}51.8 &
\cellcolor[HTML]{A3D17F}13.3 &
\cellcolor[HTML]{F6E984}7.7 \\
VQAv2 &
444 K &
\cellcolor[HTML]{FEE983}74.3 &
\cellcolor[HTML]{FEE883}46.9 &
\cellcolor[HTML]{FDD27F}45.3 &
\cellcolor[HTML]{63BE7B}72.4 &
\cellcolor[HTML]{63BE7B}71.2 &
\cellcolor[HTML]{7FC67D}48.2 &
\cellcolor[HTML]{9DCF7F}61.6 &
\cellcolor[HTML]{6FC27C}53.9 &
\cellcolor[HTML]{75C37C}75.5 &
\cellcolor[HTML]{FBEA84}68.4 &
\cellcolor[HTML]{A4D17F}47.9 &
\cellcolor[HTML]{C4DA81}50.3 &
\cellcolor[HTML]{D4DF82}55.1 &
\cellcolor[HTML]{63BE7B}69.1 &
\cellcolor[HTML]{63BE7B}36.8 &
\cellcolor[HTML]{A7D27F}39.6 &
\cellcolor[HTML]{FFEB84}12.5 &
\cellcolor[HTML]{8ACA7E}31.4 &
\cellcolor[HTML]{97CD7E}66.8 &
\cellcolor[HTML]{E5E483}37.9 &
\cellcolor[HTML]{C9DC81}70.8 &
\cellcolor[HTML]{F98871}1.3 &
\cellcolor[HTML]{77C47D}10.2 &
\cellcolor[HTML]{EBE683}43.7 &
\cellcolor[HTML]{DCE182}58.0 &
\cellcolor[HTML]{FBAF78}4.2 &
\cellcolor[HTML]{7AC57D}54.9 &
\cellcolor[HTML]{FED880}9.5 &
\cellcolor[HTML]{F97D6E}2.3 \\
OK-VQA &
9 K &
\cellcolor[HTML]{FEEB84}75.8 &
\cellcolor[HTML]{FDEB84}48.7 &
\cellcolor[HTML]{FEE582}51.6 &
\cellcolor[HTML]{BFD981}57.7 &
\cellcolor[HTML]{D9E082}65.2 &
\cellcolor[HTML]{63BE7B}51.5 &
\cellcolor[HTML]{63BE7B}64.4 &
\cellcolor[HTML]{C0D981}44.1 &
\cellcolor[HTML]{D6DF82}70.6 &
\cellcolor[HTML]{FDD17F}66.4 &
\cellcolor[HTML]{E4E483}40.5 &
\cellcolor[HTML]{F7E984}46.8 &
\cellcolor[HTML]{FEE482}47.0 &
\cellcolor[HTML]{FEE983}62.8 &
\cellcolor[HTML]{E4E483}29.8 &
\cellcolor[HTML]{FEE883}36.4 &
\cellcolor[HTML]{FBAF78}12.4 &
\cellcolor[HTML]{C1D981}25.0 &
\cellcolor[HTML]{EAE583}61.4 &
\cellcolor[HTML]{F0E784}36.3 &
\cellcolor[HTML]{F2E884}69.3 &
\cellcolor[HTML]{F97C6E}0.8 &
\cellcolor[HTML]{83C87D}9.8 &
\cellcolor[HTML]{ECE683}43.6 &
\cellcolor[HTML]{ECE683}56.6 &
\cellcolor[HTML]{FBAC77}4.0 &
\cellcolor[HTML]{96CD7E}54.5 &
\cellcolor[HTML]{FEE683}10.5 &
\cellcolor[HTML]{F87B6E}2.2 \\
A-OKVQA &
17 K &
\cellcolor[HTML]{FCB379}43.5 &
\cellcolor[HTML]{FCB679}28.7 &
\cellcolor[HTML]{FBA376}30.3 &
\cellcolor[HTML]{A2D17F}62.3 &
\cellcolor[HTML]{BDD881}66.7 &
\cellcolor[HTML]{7AC57D}48.8 &
\cellcolor[HTML]{8CCA7E}62.4 &
\cellcolor[HTML]{63BE7B}55.3 &
\cellcolor[HTML]{86C87D}74.7 &
\cellcolor[HTML]{FDEB84}68.1 &
\cellcolor[HTML]{D9E082}41.9 &
\cellcolor[HTML]{F8E984}46.7 &
\cellcolor[HTML]{F6E984}49.0 &
\cellcolor[HTML]{D9E082}64.4 &
\cellcolor[HTML]{96CD7E}34.1 &
\cellcolor[HTML]{C5DA81}38.6 &
\cellcolor[HTML]{FFEB84}12.5 &
\cellcolor[HTML]{BED881}25.3 &
\cellcolor[HTML]{E2E383}62.0 &
\cellcolor[HTML]{EFE784}36.5 &
\cellcolor[HTML]{FEE983}68.4 &
\cellcolor[HTML]{F87A6E}0.7 &
\cellcolor[HTML]{9ED07F}8.9 &
\cellcolor[HTML]{FEDF81}40.1 &
\cellcolor[HTML]{FEDD81}54.0 &
\cellcolor[HTML]{FA9172}2.5 &
\cellcolor[HTML]{F2E784}53.2 &
\cellcolor[HTML]{FDC97D}8.4 &
\cellcolor[HTML]{F8796E}2.2 \\
A-OKVQA (MC) &
17 K &
\cellcolor[HTML]{9FD07F}115.8 &
\cellcolor[HTML]{A4D17F}78.0 &
\cellcolor[HTML]{D0DE82}69.3 &
\cellcolor[HTML]{99CE7F}63.7 &
\cellcolor[HTML]{C8DB81}66.1 &
\cellcolor[HTML]{B2D580}41.8 &
\cellcolor[HTML]{DCE182}58.4 &
\cellcolor[HTML]{A7D27F}47.2 &
\cellcolor[HTML]{63BE7B}76.4 &
\cellcolor[HTML]{F2E884}69.7 &
\cellcolor[HTML]{CDDD82}43.2 &
\cellcolor[HTML]{EFE784}47.3 &
\cellcolor[HTML]{FBEA84}48.2 &
\cellcolor[HTML]{EEE683}63.6 &
\cellcolor[HTML]{73C37C}36.0 &
\cellcolor[HTML]{EBE683}37.3 &
\cellcolor[HTML]{FCB379}12.4 &
\cellcolor[HTML]{B0D480}27.0 &
\cellcolor[HTML]{D5DF82}62.8 &
\cellcolor[HTML]{EAE583}37.2 &
\cellcolor[HTML]{63BE7B}74.3 &
\cellcolor[HTML]{FBEA84}6.3 &
\cellcolor[HTML]{8AC97E}9.6 &
\cellcolor[HTML]{63BE7B}53.2 &
\cellcolor[HTML]{ECE683}56.6 &
\cellcolor[HTML]{F6E984}8.1 &
\cellcolor[HTML]{96CD7E}54.5 &
\cellcolor[HTML]{B3D680}12.8 &
\cellcolor[HTML]{FCC27C}4.4 \\
ScienceQA &
6 K &
\cellcolor[HTML]{93CC7E}120.7 &
\cellcolor[HTML]{AFD480}74.3 &
\cellcolor[HTML]{D0DE82}69.3 &
\cellcolor[HTML]{A3D17F}62.2 &
\cellcolor[HTML]{FEEA83}63.1 &
\cellcolor[HTML]{C2DA81}39.9 &
\cellcolor[HTML]{CFDE82}59.0 &
\cellcolor[HTML]{C3DA81}43.8 &
\cellcolor[HTML]{ECE683}69.4 &
\cellcolor[HTML]{63BE7B}91.5 &
\cellcolor[HTML]{CFDE82}42.9 &
\cellcolor[HTML]{E8E583}47.8 &
\cellcolor[HTML]{FEDE81}46.6 &
\cellcolor[HTML]{FEE582}62.2 &
\cellcolor[HTML]{B1D580}32.6 &
\cellcolor[HTML]{FCBC7A}31.9 &
\cellcolor[HTML]{F8696B}12.2 &
\cellcolor[HTML]{BDD881}25.4 &
\cellcolor[HTML]{F8E984}60.5 &
\cellcolor[HTML]{FEE783}33.0 &
\cellcolor[HTML]{B3D680}71.5 &
\cellcolor[HTML]{FCB579}3.2 &
\cellcolor[HTML]{FED880}4.9 &
\cellcolor[HTML]{E9E583}43.8 &
\cellcolor[HTML]{F3E884}56.0 &
\cellcolor[HTML]{FCBE7B}5.0 &
\cellcolor[HTML]{63BE7B}55.2 &
\cellcolor[HTML]{FA9C74}5.1 &
\cellcolor[HTML]{FBB279}3.9 \\
GQA &
943 K &
\cellcolor[HTML]{FA9273}25.1 &
\cellcolor[HTML]{FA9A74}18.6 &
\cellcolor[HTML]{F87A6E}17.2 &
\cellcolor[HTML]{AFD480}60.2 &
\cellcolor[HTML]{A0D07F}68.1 &
\cellcolor[HTML]{F4E884}33.8 &
\cellcolor[HTML]{B6D680}60.3 &
\cellcolor[HTML]{E2E383}40.1 &
\cellcolor[HTML]{F6E984}68.9 &
\cellcolor[HTML]{F7E984}68.9 &
\cellcolor[HTML]{63BE7B}55.3 &
\cellcolor[HTML]{63BE7B}57.0 &
\cellcolor[HTML]{FEE683}47.1 &
\cellcolor[HTML]{FEE382}62.0 &
\cellcolor[HTML]{C2DA81}31.7 &
\cellcolor[HTML]{85C87D}40.7 &
\cellcolor[HTML]{FFEB84}12.5 &
\cellcolor[HTML]{FEE082}16.1 &
\cellcolor[HTML]{FEE182}57.1 &
\cellcolor[HTML]{FDEB84}34.4 &
\cellcolor[HTML]{F2E884}69.3 &
\cellcolor[HTML]{F8716C}0.4 &
\cellcolor[HTML]{DDE283}6.8 &
\cellcolor[HTML]{90CB7E}50.1 &
\cellcolor[HTML]{FEE983}54.8 &
\cellcolor[HTML]{F97F6F}1.4 &
\cellcolor[HTML]{7FC77D}54.8 &
\cellcolor[HTML]{F86F6C}1.8 &
\cellcolor[HTML]{F8696B}1.7 \\
IconQA &
19 K &
\cellcolor[HTML]{8FCB7E}122.6 &
\cellcolor[HTML]{A6D27F}77.2 &
\cellcolor[HTML]{D4DF82}68.0 &
\cellcolor[HTML]{A8D27F}61.3 &
\cellcolor[HTML]{ECE683}64.2 &
\cellcolor[HTML]{F0E784}34.2 &
\cellcolor[HTML]{F9EA84}56.9 &
\cellcolor[HTML]{F0E784}38.5 &
\cellcolor[HTML]{FEE883}67.9 &
\cellcolor[HTML]{FEE983}67.6 &
\cellcolor[HTML]{CADC81}43.5 &
\cellcolor[HTML]{EAE583}47.7 &
\cellcolor[HTML]{63BE7B}75.3 &
\cellcolor[HTML]{E8E583}63.8 &
\cellcolor[HTML]{FEDC81}25.1 &
\cellcolor[HTML]{FDCF7E}33.9 &
\cellcolor[HTML]{63BE7B}12.8 &
\cellcolor[HTML]{CCDD82}23.6 &
\cellcolor[HTML]{FEE983}59.6 &
\cellcolor[HTML]{FAEA84}34.8 &
\cellcolor[HTML]{FEDE81}64.3 &
\cellcolor[HTML]{FBA376}2.5 &
\cellcolor[HTML]{E7E483}6.5 &
\cellcolor[HTML]{FDCB7D}36.2 &
\cellcolor[HTML]{FCC47C}52.4 &
\cellcolor[HTML]{FBA476}3.6 &
\cellcolor[HTML]{93CC7E}54.5 &
\cellcolor[HTML]{FBA476}5.6 &
\cellcolor[HTML]{FBA777}3.6 \\
VSR &
3 K &
\cellcolor[HTML]{B2D580}107.7 &
\cellcolor[HTML]{CADC81}65.4 &
\cellcolor[HTML]{DDE283}64.8 &
\cellcolor[HTML]{F98D71}13.2 &
\cellcolor[HTML]{FCB97A}46.8 &
\cellcolor[HTML]{F8776D}3.6 &
\cellcolor[HTML]{FBB078}44.1 &
\cellcolor[HTML]{F8746D}3.2 &
\cellcolor[HTML]{FCB579}59.7 &
\cellcolor[HTML]{F8696B}61.2 &
\cellcolor[HTML]{F98670}8.4 &
\cellcolor[HTML]{FED880}42.1 &
\cellcolor[HTML]{FCC47C}45.1 &
\cellcolor[HTML]{CBDC81}65.0 &
\cellcolor[HTML]{F8696B}0.2 &
\cellcolor[HTML]{FDCA7D}33.4 &
\cellcolor[HTML]{FFEB84}12.5 &
\cellcolor[HTML]{F98A71}4.5 &
\cellcolor[HTML]{FBAF78}42.2 &
\cellcolor[HTML]{FA9B74}13.3 &
\cellcolor[HTML]{FDD07E}59.4 &
\cellcolor[HTML]{FA9D75}2.2 &
\cellcolor[HTML]{F8726C}0.4 &
\cellcolor[HTML]{F8736D}19.4 &
\cellcolor[HTML]{FEE382}54.4 &
\cellcolor[HTML]{F86F6C}0.5 &
\cellcolor[HTML]{FEE883}52.9 &
\cellcolor[HTML]{FDD07E}8.9 &
\cellcolor[HTML]{F8716C}2.0 \\
TextVQA &
35 K &
\cellcolor[HTML]{F8696B}1.4 &
\cellcolor[HTML]{F8696B}0.5 &
\cellcolor[HTML]{F98470}20.2 &
\cellcolor[HTML]{F3E884}49.3 &
\cellcolor[HTML]{FEDE81}59.1 &
\cellcolor[HTML]{F2E884}33.9 &
\cellcolor[HTML]{EFE784}57.4 &
\cellcolor[HTML]{FEE983}36.0 &
\cellcolor[HTML]{FDD780}65.2 &
\cellcolor[HTML]{FDEB84}68.0 &
\cellcolor[HTML]{FEE783}36.3 &
\cellcolor[HTML]{FEE582}44.9 &
\cellcolor[HTML]{FDD37F}46.0 &
\cellcolor[HTML]{FBA977}55.2 &
\cellcolor[HTML]{C9DC81}31.3 &
\cellcolor[HTML]{FEE783}36.3 &
\cellcolor[HTML]{F97C6E}12.3 &
\cellcolor[HTML]{63BE7B}36.0 &
\cellcolor[HTML]{63BE7B}70.1 &
\cellcolor[HTML]{E7E483}37.7 &
\cellcolor[HTML]{9CCF7F}72.3 &
\cellcolor[HTML]{FA9573}1.9 &
\cellcolor[HTML]{63BE7B}10.8 &
\cellcolor[HTML]{B6D680}47.4 &
\cellcolor[HTML]{F5E984}55.8 &
\cellcolor[HTML]{FDCF7E}6.0 &
\cellcolor[HTML]{BFD981}53.9 &
\cellcolor[HTML]{91CC7E}13.7 &
\cellcolor[HTML]{F97C6E}2.3 \\
OCR-VQA &
802 K &
\cellcolor[HTML]{A4D17F}113.5 &
\cellcolor[HTML]{B9D780}70.9 &
\cellcolor[HTML]{FAEA84}55.0 &
\cellcolor[HTML]{FDCA7D}35.6 &
\cellcolor[HTML]{FCC57C}50.6 &
\cellcolor[HTML]{F98C71}8.9 &
\cellcolor[HTML]{FDC67C}48.8 &
\cellcolor[HTML]{FA9874}13.2 &
\cellcolor[HTML]{FA9673}54.5 &
\cellcolor[HTML]{FA9673}63.5 &
\cellcolor[HTML]{FDCB7D}28.3 &
\cellcolor[HTML]{FDD47F}41.2 &
\cellcolor[HTML]{FA9D75}42.9 &
\cellcolor[HTML]{FCC47C}58.4 &
\cellcolor[HTML]{FDCB7D}21.4 &
\cellcolor[HTML]{FDCD7E}33.6 &
\cellcolor[HTML]{FFEB84}12.5 &
\cellcolor[HTML]{FCBB7A}11.2 &
\cellcolor[HTML]{FBAB77}40.9 &
\cellcolor[HTML]{63BE7B}56.9 &
\cellcolor[HTML]{FEE683}67.3 &
\cellcolor[HTML]{F8696B}0.0 &
\cellcolor[HTML]{FBA276}2.5 &
\cellcolor[HTML]{FCBC7A}33.3 &
\cellcolor[HTML]{FEDA80}53.8 &
\cellcolor[HTML]{F8696B}0.1 &
\cellcolor[HTML]{FDD27F}51.9 &
\cellcolor[HTML]{F8696B}1.4 &
\cellcolor[HTML]{F86D6B}1.8 \\
OpenCQA &
6 K &
\cellcolor[HTML]{D3DF82}94.0 &
\cellcolor[HTML]{B6D680}72.0 &
\cellcolor[HTML]{E6E483}61.8 &
\cellcolor[HTML]{BED981}57.9 &
\cellcolor[HTML]{F9EA84}63.6 &
\cellcolor[HTML]{F7E984}33.3 &
\cellcolor[HTML]{FEE482}55.2 &
\cellcolor[HTML]{EEE683}38.7 &
\cellcolor[HTML]{EEE683}69.3 &
\cellcolor[HTML]{FCBF7B}65.5 &
\cellcolor[HTML]{D6E082}42.1 &
\cellcolor[HTML]{F4E884}47.0 &
\cellcolor[HTML]{FEDD81}46.6 &
\cellcolor[HTML]{FEDE81}61.4 &
\cellcolor[HTML]{FEDB81}25.0 &
\cellcolor[HTML]{63BE7B}41.8 &
\cellcolor[HTML]{FFEB84}12.5 &
\cellcolor[HTML]{FEE582}16.7 &
\cellcolor[HTML]{FEDF81}56.6 &
\cellcolor[HTML]{FBA576}16.0 &
\cellcolor[HTML]{FCBF7B}53.5 &
\cellcolor[HTML]{63BE7B}34.9 &
\cellcolor[HTML]{F8766D}0.6 &
\cellcolor[HTML]{FEE382}40.9 &
\cellcolor[HTML]{FBAC77}50.8 &
\cellcolor[HTML]{8CCA7E}13.8 &
\cellcolor[HTML]{FCB97A}50.7 &
\cellcolor[HTML]{63BE7B}14.9 &
\cellcolor[HTML]{EAE583}10.2 \\
HM &
9 K &
\cellcolor[HTML]{9FD07F}116.0 &
\cellcolor[HTML]{BAD881}70.5 &
\cellcolor[HTML]{E0E283}64.0 &
\cellcolor[HTML]{FCC17B}32.1 &
\cellcolor[HTML]{F6E984}63.7 &
\cellcolor[HTML]{F86B6B}0.6 &
\cellcolor[HTML]{F4E884}57.1 &
\cellcolor[HTML]{F86D6B}1.3 &
\cellcolor[HTML]{FEE082}66.7 &
\cellcolor[HTML]{FEDF81}67.1 &
\cellcolor[HTML]{FBB179}20.9 &
\cellcolor[HTML]{FEE282}44.2 &
\cellcolor[HTML]{FEEB84}47.6 &
\cellcolor[HTML]{DFE283}64.2 &
\cellcolor[HTML]{FDC97D}21.0 &
\cellcolor[HTML]{E6E483}37.5 &
\cellcolor[HTML]{F8696B}12.2 &
\cellcolor[HTML]{F98971}4.3 &
\cellcolor[HTML]{FEDB80}55.4 &
\cellcolor[HTML]{FEDC81}30.3 &
\cellcolor[HTML]{FDCD7E}58.4 &
\cellcolor[HTML]{F8696B}0.0 &
\cellcolor[HTML]{FA8E72}1.6 &
\cellcolor[HTML]{FDCD7E}36.7 &
\cellcolor[HTML]{63BE7B}68.4 &
\cellcolor[HTML]{F86D6B}0.4 &
\cellcolor[HTML]{C5DB81}53.8 &
\cellcolor[HTML]{F86E6C}1.7 &
\cellcolor[HTML]{F9816F}2.4 \\
\vlbenchmark{} &
7 K &
\cellcolor[HTML]{F97E6F}13.5 &
\cellcolor[HTML]{FA9573}16.5 &
\cellcolor[HTML]{F97B6E}17.6 &
\cellcolor[HTML]{FEE582}45.4 &
\cellcolor[HTML]{FEEA83}63.2 &
\cellcolor[HTML]{FBAA77}16.4 &
\cellcolor[HTML]{FEE983}56.3 &
\cellcolor[HTML]{FBB078}20.0 &
\cellcolor[HTML]{FEE582}67.5 &
\cellcolor[HTML]{FFEB84}67.7 &
\cellcolor[HTML]{F7E984}38.4 &
\cellcolor[HTML]{FBEA84}46.5 &
\cellcolor[HTML]{F2E884}49.8 &
\cellcolor[HTML]{BAD881}65.6 &
\cellcolor[HTML]{FEE482}26.8 &
\cellcolor[HTML]{ACD380}39.4 &
\cellcolor[HTML]{FFEB84}12.5 &
\cellcolor[HTML]{FDC97D}13.0 &
\cellcolor[HTML]{E4E483}61.8 &
\cellcolor[HTML]{FEE182}31.5 &
\cellcolor[HTML]{B5D680}71.4 &
\cellcolor[HTML]{E3E383}11.0 &
\cellcolor[HTML]{FDCD7E}4.4 &
\cellcolor[HTML]{A4D17F}48.7 &
\cellcolor[HTML]{FBAF78}51.0 &
\cellcolor[HTML]{63BE7B}15.9 &
\cellcolor[HTML]{F7E984}53.1 &
\cellcolor[HTML]{ADD480}13.0 &
\cellcolor[HTML]{66BF7C}38.1 \\
LLaVA Conversation &
57 K &
\cellcolor[HTML]{FDCF7E}59.5 &
\cellcolor[HTML]{FDD780}40.6 &
\cellcolor[HTML]{FDC87D}42.1 &
\cellcolor[HTML]{F97B6E}6.9 &
\cellcolor[HTML]{FED880}57.1 &
\cellcolor[HTML]{F86C6B}0.9 &
\cellcolor[HTML]{FCBE7B}47.2 &
\cellcolor[HTML]{F86D6B}1.2 &
\cellcolor[HTML]{FCC57C}62.3 &
\cellcolor[HTML]{F8756D}61.8 &
\cellcolor[HTML]{F97D6E}5.9 &
\cellcolor[HTML]{FEDA80}42.5 &
\cellcolor[HTML]{FEE182}46.8 &
\cellcolor[HTML]{FBA877}55.1 &
\cellcolor[HTML]{F87B6E}3.9 &
\cellcolor[HTML]{F8696B}23.4 &
\cellcolor[HTML]{FFEB84}12.5 &
\cellcolor[HTML]{F8796E}2.3 &
\cellcolor[HTML]{FDCA7D}50.3 &
\cellcolor[HTML]{F8766D}3.4 &
\cellcolor[HTML]{FBAC77}46.8 &
\cellcolor[HTML]{8ECB7E}26.8 &
\cellcolor[HTML]{F87A6E}0.8 &
\cellcolor[HTML]{FCC47C}34.8 &
\cellcolor[HTML]{FBA376}50.2 &
\cellcolor[HTML]{72C37C}15.1 &
\cellcolor[HTML]{F98871}48.5 &
\cellcolor[HTML]{DCE182}11.7 &
\cellcolor[HTML]{63BE7B}38.7 \\
LLaVA Reasoning &
77 K &
\cellcolor[HTML]{F8776D}9.4 &
\cellcolor[HTML]{F98570}11.0 &
\cellcolor[HTML]{F8696B}11.6 &
\cellcolor[HTML]{F8696B}0.0 &
\cellcolor[HTML]{FDD57F}56.1 &
\cellcolor[HTML]{F8696B}0.0 &
\cellcolor[HTML]{FCC27C}48.0 &
\cellcolor[HTML]{F8696B}0.0 &
\cellcolor[HTML]{FCC47C}62.0 &
\cellcolor[HTML]{FDD07E}66.3 &
\cellcolor[HTML]{F8696B}0.0 &
\cellcolor[HTML]{FEE382}44.4 &
\cellcolor[HTML]{FDC77D}45.3 &
\cellcolor[HTML]{FDCE7E}59.6 &
\cellcolor[HTML]{F86A6B}0.3 &
\cellcolor[HTML]{74C37C}41.3 &
\cellcolor[HTML]{FCC37C}12.4 &
\cellcolor[HTML]{F8696B}0.0 &
\cellcolor[HTML]{FBAF78}42.2 &
\cellcolor[HTML]{F8696B}0.0 &
\cellcolor[HTML]{FBB178}48.6 &
\cellcolor[HTML]{FDD780}4.7 &
\cellcolor[HTML]{F86C6B}0.2 &
\cellcolor[HTML]{FEE382}40.9 &
\cellcolor[HTML]{FBA376}50.2 &
\cellcolor[HTML]{69C07C}15.6 &
\cellcolor[HTML]{FED980}52.2 &
\cellcolor[HTML]{DCE182}11.7 &
\cellcolor[HTML]{F0E784}8.8 \\
LLaVA Description &
23 K &
\cellcolor[HTML]{F97D6E}13.1 &
\cellcolor[HTML]{FA9373}15.9 &
\cellcolor[HTML]{F8726C}14.7 &
\cellcolor[HTML]{F8696B}0.0 &
\cellcolor[HTML]{F8696B}19.7 &
\cellcolor[HTML]{F8696B}0.0 &
\cellcolor[HTML]{F8696B}28.9 &
\cellcolor[HTML]{F8696B}0.0 &
\cellcolor[HTML]{F8696B}47.0 &
\cellcolor[HTML]{F97D6F}62.2 &
\cellcolor[HTML]{F8696B}0.0 &
\cellcolor[HTML]{F8696B}16.9 &
\cellcolor[HTML]{F8696B}39.8 &
\cellcolor[HTML]{F8696B}47.7 &
\cellcolor[HTML]{F8696B}0.0 &
\cellcolor[HTML]{EDE683}37.3 &
\cellcolor[HTML]{FFEB84}12.5 &
\cellcolor[HTML]{F8696B}0.0 &
\cellcolor[HTML]{F8696B}21.0 &
\cellcolor[HTML]{F8696B}0.0 &
\cellcolor[HTML]{F8696B}23.3 &
\cellcolor[HTML]{FCBB7A}3.5 &
\cellcolor[HTML]{F8696B}0.0 &
\cellcolor[HTML]{F8696B}17.4 &
\cellcolor[HTML]{F8696B}46.4 &
\cellcolor[HTML]{7DC67D}14.5 &
\cellcolor[HTML]{F8696B}47.0 &
\cellcolor[HTML]{FCC37C}7.9 &
\cellcolor[HTML]{FAEA84}6.8 \\
VQAv2 QG &
444 K &
\cellcolor[HTML]{FA9D75}31.2 &
\cellcolor[HTML]{F98971}12.3 &
\cellcolor[HTML]{F98370}20.1 &
\cellcolor[HTML]{FDCF7E}37.2 &
\cellcolor[HTML]{FEDF81}59.5 &
\cellcolor[HTML]{FCB679}19.2 &
\cellcolor[HTML]{FEE482}55.2 &
\cellcolor[HTML]{FCC17C}25.0 &
\cellcolor[HTML]{FEE182}66.9 &
\cellcolor[HTML]{FCBA7A}65.2 &
\cellcolor[HTML]{FEDF81}34.0 &
\cellcolor[HTML]{FEE783}45.4 &
\cellcolor[HTML]{FDCC7E}45.6 &
\cellcolor[HTML]{FEE482}62.1 &
\cellcolor[HTML]{FEE282}26.5 &
\cellcolor[HTML]{ADD480}39.4 &
\cellcolor[HTML]{C9DC81}12.6 &
\cellcolor[HTML]{FEDF81}15.9 &
\cellcolor[HTML]{EAE583}61.5 &
\cellcolor[HTML]{FEDD81}30.6 &
\cellcolor[HTML]{FA9573}38.9 &
\cellcolor[HTML]{E7E483}10.2 &
\cellcolor[HTML]{FA9773}2.0 &
\cellcolor[HTML]{FDD07E}37.3 &
\cellcolor[HTML]{FCB579}51.4 &
\cellcolor[HTML]{FEE282}7.1 &
\cellcolor[HTML]{FDD07E}51.8 &
\cellcolor[HTML]{FEE683}10.5 &
\cellcolor[HTML]{F4E884}8.1 \\
OK-VQA QG &
9 K &
\cellcolor[HTML]{F98670}18.0 &
\cellcolor[HTML]{F98C71}13.6 &
\cellcolor[HTML]{F9816F}19.6 &
\cellcolor[HTML]{F86A6B}0.5 &
\cellcolor[HTML]{FBB279}44.2 &
\cellcolor[HTML]{F86D6B}1.2 &
\cellcolor[HTML]{FCB77A}45.7 &
\cellcolor[HTML]{F86F6C}1.7 &
\cellcolor[HTML]{FA9874}54.8 &
\cellcolor[HTML]{F86D6B}61.4 &
\cellcolor[HTML]{F86C6B}0.9 &
\cellcolor[HTML]{FCB479}34.0 &
\cellcolor[HTML]{FBAB77}43.7 &
\cellcolor[HTML]{FBAC77}55.6 &
\cellcolor[HTML]{F8696B}0.1 &
\cellcolor[HTML]{F8E984}36.9 &
\cellcolor[HTML]{FFEB84}12.5 &
\cellcolor[HTML]{F8746D}1.6 &
\cellcolor[HTML]{FCB87A}44.9 &
\cellcolor[HTML]{FCC27C}23.4 &
\cellcolor[HTML]{F98370}32.7 &
\cellcolor[HTML]{F8E984}6.8 &
\cellcolor[HTML]{FA9172}1.8 &
\cellcolor[HTML]{FCB87A}32.5 &
\cellcolor[HTML]{FBAC77}50.8 &
\cellcolor[HTML]{FDD680}6.4 &
\cellcolor[HTML]{FBA275}49.7 &
\cellcolor[HTML]{FEE081}10.0 &
\cellcolor[HTML]{F8E984}7.1 \\
A-OKVQA QG &
17 K &
\cellcolor[HTML]{FBA175}33.3 &
\cellcolor[HTML]{FA9E75}19.8 &
\cellcolor[HTML]{FBA977}32.2 &
\cellcolor[HTML]{F98670}10.7 &
\cellcolor[HTML]{FDD880}57.0 &
\cellcolor[HTML]{FA9C74}12.8 &
\cellcolor[HTML]{FDD67F}52.1 &
\cellcolor[HTML]{FBA276}16.2 &
\cellcolor[HTML]{FEE082}66.7 &
\cellcolor[HTML]{FCC17C}65.6 &
\cellcolor[HTML]{FA9B74}14.4 &
\cellcolor[HTML]{FEE182}44.1 &
\cellcolor[HTML]{F6E984}49.0 &
\cellcolor[HTML]{FCEB84}63.0 &
\cellcolor[HTML]{FA9673}9.9 &
\cellcolor[HTML]{FDD37F}34.3 &
\cellcolor[HTML]{FFEB84}12.5 &
\cellcolor[HTML]{FBB279}9.9 &
\cellcolor[HTML]{FDD27F}52.8 &
\cellcolor[HTML]{FED980}29.5 &
\cellcolor[HTML]{FA9874}39.8 &
\cellcolor[HTML]{F4E884}7.6 &
\cellcolor[HTML]{FBAD78}3.0 &
\cellcolor[HTML]{FDCA7D}36.0 &
\cellcolor[HTML]{F3E884}56.0 &
\cellcolor[HTML]{EEE784}8.5 &
\cellcolor[HTML]{F98A71}48.5 &
\cellcolor[HTML]{E6E483}11.5 &
\cellcolor[HTML]{FEEB84}6.0 \\
			\bottomrule
		\end{tabular}%
	}
	\caption
	{Unnormalized transfer learning performance of BLIP-2. Higher values indicate better performance. QG denotes question generation, MC denotes multiple-choice and G denotes open-ended generation. The color scale is normalized along each column. The colors represent values in descending order: green, yellow, orange and red.}
	\label{eval:affinity_matrix_raw_blip_landscape}
\end{table}
\end{landscape}

\begin{landscape}
\begin{table}[!tp]
	\resizebox{\columnwidth}{!}{%
		\begin{tabular}{c|c|ccccccccccccccccccccccccccccc}
			\toprule
			Source   Task &
			\setlength\extrarowheight{0pt}\begin{tabular}[c]{@{}c@{}}Dataset\\ Size \end{tabular} &
			\multicolumn{29}{c}{Target Task} \\
			&
			&
			\setlength\extrarowheight{0pt}\begin{tabular}[c]{@{}c@{}}COCO\\ Caption \end{tabular}  &
			\setlength\extrarowheight{0pt}\begin{tabular}[c]{@{}c@{}}Flickr\\ 30k\end{tabular}  &
			\setlength\extrarowheight{0pt}\begin{tabular}[c]{@{}c@{}}Text\\ Caps\end{tabular}  &
			\multicolumn{2}{c}{VQAv2} &
			\multicolumn{2}{c}{OK-VQA} &
			\multicolumn{2}{c}{A-OKVQA} &
			\setlength\extrarowheight{0pt}\begin{tabular}[c]{@{}c@{}}Science\\ QA\end{tabular} &
			\multicolumn{2}{c}{GQA} &
			\setlength\extrarowheight{0pt}\begin{tabular}[c]{@{}c@{}}Icon\\ QA \end{tabular} &
			VSR &
			\multicolumn{2}{c}{CLEVR} &
			\setlength\extrarowheight{0pt}\begin{tabular}[c]{@{}c@{}}RAVEN-\\ FAIR \end{tabular} &
			\multicolumn{2}{c}{\setlength\extrarowheight{0pt}\begin{tabular}[c]{@{}c@{}}Text\\ VQA\end{tabular}} &
			\multicolumn{2}{c}{\setlength\extrarowheight{0pt}\begin{tabular}[c]{@{}c@{}}OCR-\\ VQA\end{tabular}} &
			\setlength\extrarowheight{0pt}\begin{tabular}[c]{@{}c@{}}Open\\ CQA \end{tabular} &
			\multicolumn{2}{c}{\setlength\extrarowheight{0pt}\begin{tabular}[c]{@{}c@{}}Chart\\ QA\end{tabular}} &
			HM &
			\setlength\extrarowheight{0pt}\begin{tabular}[c]{@{}c@{}}NY\\ Explain \end{tabular} &
			\setlength\extrarowheight{0pt}\begin{tabular}[c]{@{}c@{}}NY\\ Rank \end{tabular} &
			MORE &
			\vlbenchmark \\
			&
			&
			&
			&
			&
			G &
			MC &
			G &
			MC &
			G &
			MC &
			&
			G &
			MC &
			&
			&
			G &
			MC &
			&
			G &
			MC &
			G &
			MC &
			&
			G &
			MC &
			&
			&
			&
			&
			\\
			\midrule
Zero-shot &
- &
\cellcolor[HTML]{F86F6C}9.5 &
\cellcolor[HTML]{F98370}14.2 &
\cellcolor[HTML]{F98570}13.3 &
\cellcolor[HTML]{FEDA80}35.7 &
\cellcolor[HTML]{FDCE7E}44.5 &
\cellcolor[HTML]{FAEA84}36.4 &
\cellcolor[HTML]{FCB579}40.9 &
\cellcolor[HTML]{FAEA84}36.5 &
\cellcolor[HTML]{FAA075}33.3 &
\cellcolor[HTML]{FDD47F}46.8 &
\cellcolor[HTML]{FEDC81}27.1 &
\cellcolor[HTML]{FEDE81}34.2 &
\cellcolor[HTML]{FBAE78}39.2 &
\cellcolor[HTML]{B4D680}58.7 &
\cellcolor[HTML]{EFE784}25.5 &
\cellcolor[HTML]{BBD881}35.8 &
\cellcolor[HTML]{FBEA84}12.5 &
\cellcolor[HTML]{FEE883}17.9 &
\cellcolor[HTML]{FCC07B}45.8 &
\cellcolor[HTML]{FCBC7B}14.8 &
\cellcolor[HTML]{FBAB77}27.3 &
\cellcolor[HTML]{FA9874}2.8 &
\cellcolor[HTML]{FEE683}3.7 &
\cellcolor[HTML]{96CD7E}49.0 &
\cellcolor[HTML]{FED880}53.6 &
\cellcolor[HTML]{FEDE81}8.8 &
\cellcolor[HTML]{DCE182}52.3 &
\cellcolor[HTML]{FDD07E}5.6 &
\cellcolor[HTML]{79C57D}29.6 \\
COCO Caption &
567 K &
\cellcolor[HTML]{63BE7B}133.9 &
\cellcolor[HTML]{8ACA7E}75.5 &
\cellcolor[HTML]{B4D680}65.2 &
\cellcolor[HTML]{F8766D}4.1 &
\cellcolor[HTML]{E7E483}50.6 &
\cellcolor[HTML]{F98971}9.0 &
\cellcolor[HTML]{F8E984}47.0 &
\cellcolor[HTML]{F9816F}6.8 &
\cellcolor[HTML]{FCBE7B}38.1 &
\cellcolor[HTML]{FDCD7E}45.9 &
\cellcolor[HTML]{F8726C}2.2 &
\cellcolor[HTML]{FED880}33.7 &
\cellcolor[HTML]{FEDD81}41.9 &
\cellcolor[HTML]{E0E283}55.6 &
\cellcolor[HTML]{F8726C}1.8 &
\cellcolor[HTML]{94CD7E}36.7 &
\cellcolor[HTML]{FEDC81}12.4 &
\cellcolor[HTML]{F9816F}3.5 &
\cellcolor[HTML]{F5E884}52.8 &
\cellcolor[HTML]{F8726C}1.7 &
\cellcolor[HTML]{FBEA84}36.2 &
\cellcolor[HTML]{B4D680}15.2 &
\cellcolor[HTML]{F98470}0.8 &
\cellcolor[HTML]{FCBA7A}29.6 &
\cellcolor[HTML]{E4E483}55.6 &
\cellcolor[HTML]{D4DF82}10.5 &
\cellcolor[HTML]{FA9773}49.7 &
\cellcolor[HTML]{9ACE7F}12.0 &
\cellcolor[HTML]{E1E383}16.4 \\
Flickr30k &
145 K &
\cellcolor[HTML]{9CCF7F}96.1 &
\cellcolor[HTML]{63BE7B}92.2 &
\cellcolor[HTML]{AAD380}71.4 &
\cellcolor[HTML]{FEE583}39.2 &
\cellcolor[HTML]{EAE583}50.2 &
\cellcolor[HTML]{FEE783}34.8 &
\cellcolor[HTML]{F1E784}47.6 &
\cellcolor[HTML]{FCEB84}36.2 &
\cellcolor[HTML]{FDEB84}45.7 &
\cellcolor[HTML]{F1E784}52.1 &
\cellcolor[HTML]{F4E884}32.2 &
\cellcolor[HTML]{E3E383}37.9 &
\cellcolor[HTML]{F6E984}44.1 &
\cellcolor[HTML]{ADD480}59.2 &
\cellcolor[HTML]{A1D07F}31.3 &
\cellcolor[HTML]{AFD480}36.1 &
\cellcolor[HTML]{FEE182}12.4 &
\cellcolor[HTML]{D5DF82}25.2 &
\cellcolor[HTML]{C4DA81}58.6 &
\cellcolor[HTML]{FAEA84}24.2 &
\cellcolor[HTML]{F1E784}38.7 &
\cellcolor[HTML]{C5DB81}13.3 &
\cellcolor[HTML]{CCDD82}6.5 &
\cellcolor[HTML]{DFE283}40.6 &
\cellcolor[HTML]{63BE7B}59.4 &
\cellcolor[HTML]{C3DA81}10.8 &
\cellcolor[HTML]{DBE182}52.3 &
\cellcolor[HTML]{A8D27F}11.3 &
\cellcolor[HTML]{C6DB81}19.8 \\
Web CapFilt &
23,147 K &
\cellcolor[HTML]{82C77D}113.7 &
\cellcolor[HTML]{87C97E}76.9 &
\cellcolor[HTML]{A2D17F}75.7 &
\cellcolor[HTML]{FA9C74}16.1 &
\cellcolor[HTML]{FEE983}47.3 &
\cellcolor[HTML]{FCB579}21.0 &
\cellcolor[HTML]{E9E583}48.4 &
\cellcolor[HTML]{FCC17B}24.2 &
\cellcolor[HTML]{D4DF82}53.7 &
\cellcolor[HTML]{F7E984}51.0 &
\cellcolor[HTML]{FA9B74}11.8 &
\cellcolor[HTML]{FEDA80}33.9 &
\cellcolor[HTML]{F6E984}44.1 &
\cellcolor[HTML]{FDD07E}52.4 &
\cellcolor[HTML]{FA9B74}9.4 &
\cellcolor[HTML]{FDD27F}31.8 &
\cellcolor[HTML]{EBE683}12.5 &
\cellcolor[HTML]{FCB679}10.9 &
\cellcolor[HTML]{E7E483}54.5 &
\cellcolor[HTML]{F8756D}2.3 &
\cellcolor[HTML]{FA9673}24.7 &
\cellcolor[HTML]{EBE583}9.2 &
\cellcolor[HTML]{F8726C}0.3 &
\cellcolor[HTML]{F8696B}17.4 &
\cellcolor[HTML]{FA9373}49.0 &
\cellcolor[HTML]{FDD67F}8.3 &
\cellcolor[HTML]{F8696B}48.7 &
\cellcolor[HTML]{8CCA7E}12.8 &
\cellcolor[HTML]{FDD17F}10.4 \\
TextCaps &
549 K &
\cellcolor[HTML]{D8E082}56.2 &
\cellcolor[HTML]{D8E082}42.0 &
\cellcolor[HTML]{63BE7B}111.8 &
\cellcolor[HTML]{FBB379}23.2 &
\cellcolor[HTML]{F8696B}33.8 &
\cellcolor[HTML]{FCC07B}24.1 &
\cellcolor[HTML]{F7E984}47.1 &
\cellcolor[HTML]{FDC97D}26.4 &
\cellcolor[HTML]{FEE182}43.6 &
\cellcolor[HTML]{FCC57C}45.0 &
\cellcolor[HTML]{FCB379}17.6 &
\cellcolor[HTML]{F8746D}25.3 &
\cellcolor[HTML]{FDD780}41.6 &
\cellcolor[HTML]{EBE583}54.8 &
\cellcolor[HTML]{FEDE81}22.0 &
\cellcolor[HTML]{FBA877}27.6 &
\cellcolor[HTML]{FDD37F}12.4 &
\cellcolor[HTML]{FED980}15.9 &
\cellcolor[HTML]{E0E283}55.3 &
\cellcolor[HTML]{FEEA83}22.9 &
\cellcolor[HTML]{FEE182}33.9 &
\cellcolor[HTML]{84C87D}20.4 &
\cellcolor[HTML]{FDCD7E}3.0 &
\cellcolor[HTML]{FA9373}23.8 &
\cellcolor[HTML]{FDC97D}52.6 &
\cellcolor[HTML]{8DCA7E}11.8 &
\cellcolor[HTML]{FA9D75}49.8 &
\cellcolor[HTML]{63BE7B}15.0 &
\cellcolor[HTML]{DAE182}17.3 \\
VQAv2 &
444 K &
\cellcolor[HTML]{D6DF82}57.9 &
\cellcolor[HTML]{E1E383}38.5 &
\cellcolor[HTML]{EDE683}32.5 &
\cellcolor[HTML]{63BE7B}72.5 &
\cellcolor[HTML]{63BE7B}67.6 &
\cellcolor[HTML]{6AC07C}55.5 &
\cellcolor[HTML]{76C47D}59.6 &
\cellcolor[HTML]{63BE7B}58.6 &
\cellcolor[HTML]{9BCE7F}65.2 &
\cellcolor[HTML]{D4DF82}57.6 &
\cellcolor[HTML]{8ACA7E}47.4 &
\cellcolor[HTML]{99CE7F}45.2 &
\cellcolor[HTML]{DCE182}48.0 &
\cellcolor[HTML]{7DC67D}62.5 &
\cellcolor[HTML]{63BE7B}35.9 &
\cellcolor[HTML]{82C77D}37.1 &
\cellcolor[HTML]{FEDE81}12.4 &
\cellcolor[HTML]{83C87D}38.3 &
\cellcolor[HTML]{8ECB7E}65.1 &
\cellcolor[HTML]{BFD981}38.5 &
\cellcolor[HTML]{FEDD81}33.4 &
\cellcolor[HTML]{FEE582}6.7 &
\cellcolor[HTML]{63BE7B}12.0 &
\cellcolor[HTML]{FCB579}29.0 &
\cellcolor[HTML]{6AC07C}59.2 &
\cellcolor[HTML]{F86D6B}1.2 &
\cellcolor[HTML]{FDD780}51.0 &
\cellcolor[HTML]{F8696B}2.1 &
\cellcolor[HTML]{F86C6B}2.0 \\
OK-VQA &
9 K &
\cellcolor[HTML]{DFE283}51.2 &
\cellcolor[HTML]{EBE683}34.0 &
\cellcolor[HTML]{F3E884}29.3 &
\cellcolor[HTML]{C7DB81}52.2 &
\cellcolor[HTML]{CBDC81}54.3 &
\cellcolor[HTML]{63BE7B}56.3 &
\cellcolor[HTML]{81C77D}58.5 &
\cellcolor[HTML]{B9D780}46.0 &
\cellcolor[HTML]{C3DA81}57.2 &
\cellcolor[HTML]{D4DF82}57.6 &
\cellcolor[HTML]{DFE283}35.2 &
\cellcolor[HTML]{C6DB81}40.8 &
\cellcolor[HTML]{E0E283}47.4 &
\cellcolor[HTML]{F8696B}48.5 &
\cellcolor[HTML]{FEDF81}22.1 &
\cellcolor[HTML]{C9DC81}35.5 &
\cellcolor[HTML]{EDE683}12.5 &
\cellcolor[HTML]{C7DB81}27.5 &
\cellcolor[HTML]{ACD380}61.6 &
\cellcolor[HTML]{FFEB84}23.0 &
\cellcolor[HTML]{EEE683}39.6 &
\cellcolor[HTML]{FCC27C}4.9 &
\cellcolor[HTML]{8ECB7E}9.8 &
\cellcolor[HTML]{FAEA84}37.6 &
\cellcolor[HTML]{C9DC81}56.4 &
\cellcolor[HTML]{FA9773}4.0 &
\cellcolor[HTML]{73C37C}54.8 &
\cellcolor[HTML]{D6E082}8.8 &
\cellcolor[HTML]{F86C6B}2.0 \\
A-OKVQA &
17 K &
\cellcolor[HTML]{EEE683}41.5 &
\cellcolor[HTML]{FAEA84}27.5 &
\cellcolor[HTML]{FCEB84}24.0 &
\cellcolor[HTML]{9ECF7F}60.6 &
\cellcolor[HTML]{88C97E}62.9 &
\cellcolor[HTML]{7BC57D}53.1 &
\cellcolor[HTML]{63BE7B}61.4 &
\cellcolor[HTML]{6DC17C}57.1 &
\cellcolor[HTML]{8FCB7E}67.5 &
\cellcolor[HTML]{D7E082}57.0 &
\cellcolor[HTML]{BAD780}40.5 &
\cellcolor[HTML]{AAD380}43.5 &
\cellcolor[HTML]{DBE182}48.1 &
\cellcolor[HTML]{F8696B}48.5 &
\cellcolor[HTML]{C7DB81}28.5 &
\cellcolor[HTML]{68C07C}37.6 &
\cellcolor[HTML]{63BE7B}13.0 &
\cellcolor[HTML]{B7D680}30.1 &
\cellcolor[HTML]{99CE7F}63.8 &
\cellcolor[HTML]{EDE683}27.3 &
\cellcolor[HTML]{F9826F}22.3 &
\cellcolor[HTML]{F8716C}0.9 &
\cellcolor[HTML]{96CD7E}9.4 &
\cellcolor[HTML]{FDD57F}33.7 &
\cellcolor[HTML]{C9DC81}56.4 &
\cellcolor[HTML]{F8696B}0.8 &
\cellcolor[HTML]{D9E082}52.3 &
\cellcolor[HTML]{FA9E75}3.9 &
\cellcolor[HTML]{F8696B}1.8 \\
A-OKVQA (MC) &
17 K &
\cellcolor[HTML]{FDD37F}25.9 &
\cellcolor[HTML]{FFEB84}25.5 &
\cellcolor[HTML]{FBEA84}24.7 &
\cellcolor[HTML]{DFE283}47.3 &
\cellcolor[HTML]{ABD380}58.4 &
\cellcolor[HTML]{C1DA81}43.9 &
\cellcolor[HTML]{76C47D}59.6 &
\cellcolor[HTML]{BED981}45.2 &
\cellcolor[HTML]{63BE7B}76.1 &
\cellcolor[HTML]{B4D680}63.8 &
\cellcolor[HTML]{FEE482}29.1 &
\cellcolor[HTML]{B8D780}42.2 &
\cellcolor[HTML]{E9E583}46.1 &
\cellcolor[HTML]{F98770}49.7 &
\cellcolor[HTML]{FEE482}23.1 &
\cellcolor[HTML]{A4D17F}36.3 &
\cellcolor[HTML]{D6E082}12.6 &
\cellcolor[HTML]{C4DA81}28.0 &
\cellcolor[HTML]{63BE7B}70.2 &
\cellcolor[HTML]{FBEA84}24.0 &
\cellcolor[HTML]{84C87D}67.2 &
\cellcolor[HTML]{DDE283}10.7 &
\cellcolor[HTML]{B5D680}7.7 &
\cellcolor[HTML]{63BE7B}54.8 &
\cellcolor[HTML]{E4E483}55.6 &
\cellcolor[HTML]{FEE683}9.3 &
\cellcolor[HTML]{C2DA81}52.9 &
\cellcolor[HTML]{FEE883}6.4 &
\cellcolor[HTML]{FAEA84}13.2 \\
ScienceQA &
6 K &
\cellcolor[HTML]{FBA175}17.8 &
\cellcolor[HTML]{FBAB77}18.5 &
\cellcolor[HTML]{FDD37F}20.1 &
\cellcolor[HTML]{D8E082}48.7 &
\cellcolor[HTML]{D4DF82}53.1 &
\cellcolor[HTML]{C3DA81}43.6 &
\cellcolor[HTML]{AED480}54.1 &
\cellcolor[HTML]{C1D981}44.9 &
\cellcolor[HTML]{A8D27F}62.4 &
\cellcolor[HTML]{63BE7B}79.1 &
\cellcolor[HTML]{E1E383}34.8 &
\cellcolor[HTML]{E1E383}38.2 &
\cellcolor[HTML]{E8E583}46.2 &
\cellcolor[HTML]{FEE282}53.0 &
\cellcolor[HTML]{D6E082}27.4 &
\cellcolor[HTML]{FDD27F}31.9 &
\cellcolor[HTML]{99CE7F}12.8 &
\cellcolor[HTML]{D4DF82}25.4 &
\cellcolor[HTML]{A3D17F}62.6 &
\cellcolor[HTML]{FBEA84}24.1 &
\cellcolor[HTML]{9ECF7F}60.6 &
\cellcolor[HTML]{F3E884}8.3 &
\cellcolor[HTML]{9ACE7F}9.2 &
\cellcolor[HTML]{FCEA84}37.3 &
\cellcolor[HTML]{FDD57F}53.4 &
\cellcolor[HTML]{FFEB84}9.7 &
\cellcolor[HTML]{63BE7B}55.2 &
\cellcolor[HTML]{FEE282}6.2 &
\cellcolor[HTML]{63BE7B}32.3 \\
GQA &
943 K &
\cellcolor[HTML]{E7E583}46.0 &
\cellcolor[HTML]{FAEA84}27.7 &
\cellcolor[HTML]{FDD47F}20.2 &
\cellcolor[HTML]{AAD380}58.1 &
\cellcolor[HTML]{8BCA7E}62.5 &
\cellcolor[HTML]{EFE784}37.8 &
\cellcolor[HTML]{9DCF7F}55.8 &
\cellcolor[HTML]{D5DF82}41.9 &
\cellcolor[HTML]{A4D17F}63.3 &
\cellcolor[HTML]{DDE283}55.9 &
\cellcolor[HTML]{63BE7B}52.8 &
\cellcolor[HTML]{63BE7B}50.3 &
\cellcolor[HTML]{EEE784}45.3 &
\cellcolor[HTML]{91CC7E}61.1 &
\cellcolor[HTML]{A4D17F}31.1 &
\cellcolor[HTML]{63BE7B}37.7 &
\cellcolor[HTML]{FEDE81}12.4 &
\cellcolor[HTML]{FEE482}17.4 &
\cellcolor[HTML]{FEE983}51.4 &
\cellcolor[HTML]{E6E483}29.1 &
\cellcolor[HTML]{E5E483}41.9 &
\cellcolor[HTML]{F8696B}0.5 &
\cellcolor[HTML]{F4E884}4.4 &
\cellcolor[HTML]{FEDF81}35.2 &
\cellcolor[HTML]{A1D07F}57.6 &
\cellcolor[HTML]{F5E984}9.9 &
\cellcolor[HTML]{96CD7E}54.0 &
\cellcolor[HTML]{FFEB84}6.5 &
\cellcolor[HTML]{F86D6B}2.1 \\
IconQA &
19 K &
\cellcolor[HTML]{FA9473}15.6 &
\cellcolor[HTML]{FA9D75}17.0 &
\cellcolor[HTML]{FDD47F}20.2 &
\cellcolor[HTML]{E5E483}46.1 &
\cellcolor[HTML]{C5DB81}55.1 &
\cellcolor[HTML]{CCDD82}42.5 &
\cellcolor[HTML]{E3E383}49.1 &
\cellcolor[HTML]{CDDD82}43.1 &
\cellcolor[HTML]{D3DF82}54.1 &
\cellcolor[HTML]{DCE182}56.1 &
\cellcolor[HTML]{E9E583}33.8 &
\cellcolor[HTML]{D9E082}39.0 &
\cellcolor[HTML]{63BE7B}65.9 &
\cellcolor[HTML]{FEEB84}53.4 &
\cellcolor[HTML]{D6E082}27.4 &
\cellcolor[HTML]{FEDE81}33.0 &
\cellcolor[HTML]{FA9E75}12.2 &
\cellcolor[HTML]{D9E082}24.5 &
\cellcolor[HTML]{D5DF82}56.6 &
\cellcolor[HTML]{F7E984}25.1 &
\cellcolor[HTML]{9CCF7F}61.1 &
\cellcolor[HTML]{EEE683}8.9 &
\cellcolor[HTML]{A8D27F}8.4 &
\cellcolor[HTML]{FED880}34.2 &
\cellcolor[HTML]{FFEB84}54.8 &
\cellcolor[HTML]{ECE683}10.0 &
\cellcolor[HTML]{97CD7E}53.9 &
\cellcolor[HTML]{FEEA83}6.5 &
\cellcolor[HTML]{8ACA7E}27.4 \\
VSR &
3 K &
\cellcolor[HTML]{F8716C}9.9 &
\cellcolor[HTML]{F9816F}14.0 &
\cellcolor[HTML]{FA9072}14.3 &
\cellcolor[HTML]{CBDC81}51.4 &
\cellcolor[HTML]{FCEB84}47.9 &
\cellcolor[HTML]{E7E583}38.9 &
\cellcolor[HTML]{FCC37C}42.3 &
\cellcolor[HTML]{DBE182}41.1 &
\cellcolor[HTML]{FBAA77}34.8 &
\cellcolor[HTML]{FEDA80}47.4 &
\cellcolor[HTML]{C7DB81}38.6 &
\cellcolor[HTML]{F8E984}35.9 &
\cellcolor[HTML]{FDCC7E}40.9 &
\cellcolor[HTML]{63BE7B}64.3 &
\cellcolor[HTML]{BAD881}29.5 &
\cellcolor[HTML]{BBD881}35.8 &
\cellcolor[HTML]{F9EA84}12.5 &
\cellcolor[HTML]{ECE683}21.4 &
\cellcolor[HTML]{FDC77D}46.7 &
\cellcolor[HTML]{DBE182}31.8 &
\cellcolor[HTML]{DEE283}43.9 &
\cellcolor[HTML]{FBAB77}3.8 &
\cellcolor[HTML]{F6E984}4.3 &
\cellcolor[HTML]{7EC67D}51.7 &
\cellcolor[HTML]{FBA276}50.0 &
\cellcolor[HTML]{FEDC81}8.7 &
\cellcolor[HTML]{E6E483}52.0 &
\cellcolor[HTML]{FDD57F}5.8 &
\cellcolor[HTML]{76C47D}30.0 \\
TextVQA &
35 K &
\cellcolor[HTML]{B3D680}80.6 &
\cellcolor[HTML]{B4D680}57.7 &
\cellcolor[HTML]{BCD881}61.0 &
\cellcolor[HTML]{A9D380}58.3 &
\cellcolor[HTML]{B2D580}57.4 &
\cellcolor[HTML]{BDD881}44.5 &
\cellcolor[HTML]{A8D27F}54.7 &
\cellcolor[HTML]{BBD881}45.8 &
\cellcolor[HTML]{A7D27F}62.6 &
\cellcolor[HTML]{CCDD82}59.1 &
\cellcolor[HTML]{C2DA81}39.3 &
\cellcolor[HTML]{CADC81}40.4 &
\cellcolor[HTML]{ECE683}45.6 &
\cellcolor[HTML]{C6DB81}57.4 &
\cellcolor[HTML]{81C77D}33.7 &
\cellcolor[HTML]{FEEA83}34.3 &
\cellcolor[HTML]{F9EA84}12.5 &
\cellcolor[HTML]{63BE7B}43.5 &
\cellcolor[HTML]{7FC67D}66.9 &
\cellcolor[HTML]{BCD881}39.2 &
\cellcolor[HTML]{D6E082}45.9 &
\cellcolor[HTML]{FCBE7B}4.7 &
\cellcolor[HTML]{66BF7C}11.9 &
\cellcolor[HTML]{FDD17F}33.0 &
\cellcolor[HTML]{9ACE7F}57.8 &
\cellcolor[HTML]{FA9B74}4.3 &
\cellcolor[HTML]{F86C6B}48.8 &
\cellcolor[HTML]{FCC47C}5.2 &
\cellcolor[HTML]{F8796E}3.1 \\
OCR-VQA &
802 K &
\cellcolor[HTML]{D4DF82}58.6 &
\cellcolor[HTML]{D3DF82}44.5 &
\cellcolor[HTML]{C8DC81}53.8 &
\cellcolor[HTML]{BBD881}54.6 &
\cellcolor[HTML]{FDD07E}44.7 &
\cellcolor[HTML]{D7E082}41.0 &
\cellcolor[HTML]{FDCF7E}43.5 &
\cellcolor[HTML]{C8DC81}43.8 &
\cellcolor[HTML]{E2E383}51.1 &
\cellcolor[HTML]{DBE182}56.4 &
\cellcolor[HTML]{B8D780}40.7 &
\cellcolor[HTML]{CCDD82}40.2 &
\cellcolor[HTML]{FEE182}42.1 &
\cellcolor[HTML]{E9E583}54.9 &
\cellcolor[HTML]{8ACA7E}33.1 &
\cellcolor[HTML]{F8696B}21.2 &
\cellcolor[HTML]{E8E583}12.5 &
\cellcolor[HTML]{B0D580}31.1 &
\cellcolor[HTML]{FCBA7A}44.9 &
\cellcolor[HTML]{63BE7B}60.6 &
\cellcolor[HTML]{63BE7B}75.8 &
\cellcolor[HTML]{F6E984}8.0 &
\cellcolor[HTML]{8DCB7E}9.8 &
\cellcolor[HTML]{DEE283}40.8 &
\cellcolor[HTML]{8CCA7E}58.2 &
\cellcolor[HTML]{D0DE82}10.6 &
\cellcolor[HTML]{FBB078}50.2 &
\cellcolor[HTML]{E7E483}7.9 &
\cellcolor[HTML]{F9816F}3.8 \\
OpenCQA &
6 K &
\cellcolor[HTML]{FEE182}28.3 &
\cellcolor[HTML]{FDD47F}22.8 &
\cellcolor[HTML]{F6E984}27.7 &
\cellcolor[HTML]{FEE282}38.1 &
\cellcolor[HTML]{FCBA7A}42.4 &
\cellcolor[HTML]{FEE582}34.0 &
\cellcolor[HTML]{FEE382}45.5 &
\cellcolor[HTML]{FEDE81}32.2 &
\cellcolor[HTML]{FEE783}44.6 &
\cellcolor[HTML]{DEE283}55.7 &
\cellcolor[HTML]{FEDC81}27.2 &
\cellcolor[HTML]{FDD27F}33.1 &
\cellcolor[HTML]{F7E984}44.0 &
\cellcolor[HTML]{FEE883}53.3 &
\cellcolor[HTML]{FDD17F}19.5 &
\cellcolor[HTML]{FEE683}33.8 &
\cellcolor[HTML]{FEE783}12.4 &
\cellcolor[HTML]{F1E784}20.6 &
\cellcolor[HTML]{FEDE81}49.9 &
\cellcolor[HTML]{FDD07E}18.3 &
\cellcolor[HTML]{FCEA84}36.0 &
\cellcolor[HTML]{63BE7B}24.0 &
\cellcolor[HTML]{FEEB84}3.9 &
\cellcolor[HTML]{DAE182}41.2 &
\cellcolor[HTML]{FBAB77}50.6 &
\cellcolor[HTML]{97CD7E}11.6 &
\cellcolor[HTML]{FDD780}51.0 &
\cellcolor[HTML]{9DCF7F}11.9 &
\cellcolor[HTML]{C2DA81}20.4 \\
HM &
9 K &
\cellcolor[HTML]{F8716C}9.9 &
\cellcolor[HTML]{F98370}14.2 &
\cellcolor[HTML]{F98C71}13.9 &
\cellcolor[HTML]{F1E784}43.8 &
\cellcolor[HTML]{FDD07E}44.7 &
\cellcolor[HTML]{FEE983}35.3 &
\cellcolor[HTML]{FBAC77}40.0 &
\cellcolor[HTML]{FEE582}34.2 &
\cellcolor[HTML]{FA9D75}32.8 &
\cellcolor[HTML]{FCB97A}43.6 &
\cellcolor[HTML]{F3E884}32.4 &
\cellcolor[HTML]{FDD57F}33.4 &
\cellcolor[HTML]{FEDA80}41.7 &
\cellcolor[HTML]{FCC57C}52.0 &
\cellcolor[HTML]{F0E784}25.4 &
\cellcolor[HTML]{FEEA83}34.3 &
\cellcolor[HTML]{FDEB84}12.4 &
\cellcolor[HTML]{FEE182}17.1 &
\cellcolor[HTML]{FA9272}39.4 &
\cellcolor[HTML]{FEE482}21.8 &
\cellcolor[HTML]{F0E784}39.0 &
\cellcolor[HTML]{FBA175}3.3 &
\cellcolor[HTML]{FEE783}3.7 &
\cellcolor[HTML]{B0D580}46.0 &
\cellcolor[HTML]{DEE283}55.8 &
\cellcolor[HTML]{FEDD81}8.7 &
\cellcolor[HTML]{FCBF7B}50.5 &
\cellcolor[HTML]{FDCC7E}5.5 &
\cellcolor[HTML]{67BF7C}31.9 \\
\vlbenchmark{} &
7 K &
\cellcolor[HTML]{F8696B}8.6 &
\cellcolor[HTML]{F86E6C}12.0 &
\cellcolor[HTML]{F98D72}14.0 &
\cellcolor[HTML]{FCC47C}28.5 &
\cellcolor[HTML]{FBB078}41.3 &
\cellcolor[HTML]{FBA376}16.2 &
\cellcolor[HTML]{FA9D75}38.6 &
\cellcolor[HTML]{FBA476}16.2 &
\cellcolor[HTML]{FBA977}34.7 &
\cellcolor[HTML]{FDCD7E}45.9 &
\cellcolor[HTML]{FCC07B}20.6 &
\cellcolor[HTML]{FDCD7E}32.8 &
\cellcolor[HTML]{FBA376}38.5 &
\cellcolor[HTML]{F98A71}49.8 &
\cellcolor[HTML]{FBA776}11.6 &
\cellcolor[HTML]{FA9C74}26.3 &
\cellcolor[HTML]{FBA977}12.2 &
\cellcolor[HTML]{FCBF7B}12.3 &
\cellcolor[HTML]{FCB479}44.2 &
\cellcolor[HTML]{FBA476}10.5 &
\cellcolor[HTML]{FCB579}28.5 &
\cellcolor[HTML]{FDEB84}7.2 &
\cellcolor[HTML]{F97E6F}0.6 &
\cellcolor[HTML]{FDEB84}37.2 &
\cellcolor[HTML]{FBA576}50.2 &
\cellcolor[HTML]{E6E483}10.1 &
\cellcolor[HTML]{EFE784}51.8 &
\cellcolor[HTML]{FFEB84}6.5 &
\cellcolor[HTML]{71C27C}30.6 \\
LLaVA Conversation &
57 K &
\cellcolor[HTML]{FBB078}20.3 &
\cellcolor[HTML]{FA9373}15.9 &
\cellcolor[HTML]{FEDB80}21.1 &
\cellcolor[HTML]{F8726C}3.0 &
\cellcolor[HTML]{FCB379}41.7 &
\cellcolor[HTML]{FA9773}12.6 &
\cellcolor[HTML]{FDCE7E}43.4 &
\cellcolor[HTML]{F98770}8.2 &
\cellcolor[HTML]{FBAD78}35.3 &
\cellcolor[HTML]{FEE883}50.3 &
\cellcolor[HTML]{F8706C}1.7 &
\cellcolor[HTML]{FBA276}29.1 &
\cellcolor[HTML]{FBA977}38.8 &
\cellcolor[HTML]{E0E283}55.6 &
\cellcolor[HTML]{F86D6B}0.8 &
\cellcolor[HTML]{FBAF78}28.3 &
\cellcolor[HTML]{FEE282}12.4 &
\cellcolor[HTML]{FA9673}6.4 &
\cellcolor[HTML]{FCB87A}44.6 &
\cellcolor[HTML]{F86B6B}0.4 &
\cellcolor[HTML]{FBAE78}27.7 &
\cellcolor[HTML]{ECE683}9.7 &
\cellcolor[HTML]{F9816F}0.7 &
\cellcolor[HTML]{FBEA84}37.4 &
\cellcolor[HTML]{81C77D}58.6 &
\cellcolor[HTML]{FEE081}9.0 &
\cellcolor[HTML]{F8696B}48.7 &
\cellcolor[HTML]{FDD07E}5.6 &
\cellcolor[HTML]{67C07C}31.8 \\
LLaVA Reasoning &
77 K &
\cellcolor[HTML]{F8756D}10.5 &
\cellcolor[HTML]{F97E6F}13.7 &
\cellcolor[HTML]{F86B6B}11.0 &
\cellcolor[HTML]{FDEB84}41.2 &
\cellcolor[HTML]{FCC07B}43.0 &
\cellcolor[HTML]{FDD47F}29.4 &
\cellcolor[HTML]{FBAA77}39.8 &
\cellcolor[HTML]{FDD47F}29.6 &
\cellcolor[HTML]{FA9072}30.7 &
\cellcolor[HTML]{FCC37C}44.7 &
\cellcolor[HTML]{F9EA84}31.5 &
\cellcolor[HTML]{F98C71}27.3 &
\cellcolor[HTML]{F8726C}35.7 &
\cellcolor[HTML]{FDCE7E}52.3 &
\cellcolor[HTML]{FED980}21.1 &
\cellcolor[HTML]{F6E984}34.5 &
\cellcolor[HTML]{FEDA80}12.4 &
\cellcolor[HTML]{FDEB84}18.8 &
\cellcolor[HTML]{FA9C74}40.8 &
\cellcolor[HTML]{F8696B}0.0 &
\cellcolor[HTML]{F8696B}19.2 &
\cellcolor[HTML]{F98A71}2.1 &
\cellcolor[HTML]{F86A6B}0.0 &
\cellcolor[HTML]{C6DB81}43.5 &
\cellcolor[HTML]{FFEB84}54.8 &
\cellcolor[HTML]{FEE382}9.1 &
\cellcolor[HTML]{FCB579}50.3 &
\cellcolor[HTML]{FEDC81}6.0 &
\cellcolor[HTML]{F97C6E}3.4 \\
LLaVA Description &
23 K &
\cellcolor[HTML]{F8696B}8.5 &
\cellcolor[HTML]{F8696B}11.4 &
\cellcolor[HTML]{F8696B}10.8 &
\cellcolor[HTML]{F8696B}0.0 &
\cellcolor[HTML]{F86E6C}34.5 &
\cellcolor[HTML]{F8696B}0.0 &
\cellcolor[HTML]{F8696B}33.3 &
\cellcolor[HTML]{F8696B}0.0 &
\cellcolor[HTML]{F8696B}24.4 &
\cellcolor[HTML]{F8696B}34.3 &
\cellcolor[HTML]{F8696B}0.0 &
\cellcolor[HTML]{F8696B}24.4 &
\cellcolor[HTML]{F8696B}35.1 &
\cellcolor[HTML]{FCBE7B}51.7 &
\cellcolor[HTML]{F8696B}0.0 &
\cellcolor[HTML]{FDC97D}31.0 &
\cellcolor[HTML]{F8696B}12.0 &
\cellcolor[HTML]{F8696B}0.0 &
\cellcolor[HTML]{F8696B}33.8 &
\cellcolor[HTML]{F8696B}0.0 &
\cellcolor[HTML]{FEE382}34.1 &
\cellcolor[HTML]{F98370}1.8 &
\cellcolor[HTML]{F8696B}0.0 &
\cellcolor[HTML]{FCB87A}29.3 &
\cellcolor[HTML]{FBAB77}50.6 &
\cellcolor[HTML]{FEEA83}9.7 &
\cellcolor[HTML]{FCBF7B}50.5 &
\cellcolor[HTML]{FDD07E}5.6 &
\cellcolor[HTML]{F86D6B}2.1 \\
VQAv2 QG &
444 K &
\cellcolor[HTML]{D5DF82}58.5 &
\cellcolor[HTML]{CCDD82}47.5 &
\cellcolor[HTML]{FAEA84}25.4 &
\cellcolor[HTML]{FA9C74}16.0 &
\cellcolor[HTML]{FEEB84}47.7 &
\cellcolor[HTML]{FCBD7B}23.3 &
\cellcolor[HTML]{FDC67D}42.6 &
\cellcolor[HTML]{FCC27C}24.6 &
\cellcolor[HTML]{FBEA84}46.0 &
\cellcolor[HTML]{FEDC81}47.6 &
\cellcolor[HTML]{FA9673}10.7 &
\cellcolor[HTML]{D4DF82}39.4 &
\cellcolor[HTML]{FEE081}42.1 &
\cellcolor[HTML]{FDC77D}52.0 &
\cellcolor[HTML]{FA9473}8.1 &
\cellcolor[HTML]{FDEB84}34.4 &
\cellcolor[HTML]{FEDE81}12.4 &
\cellcolor[HTML]{FCBF7B}12.2 &
\cellcolor[HTML]{FEEB84}51.8 &
\cellcolor[HTML]{FA9072}6.9 &
\cellcolor[HTML]{FCC37C}30.2 &
\cellcolor[HTML]{EFE784}8.7 &
\cellcolor[HTML]{FA9F75}1.6 &
\cellcolor[HTML]{FEE182}35.5 &
\cellcolor[HTML]{EBE683}55.4 &
\cellcolor[HTML]{B0D480}11.2 &
\cellcolor[HTML]{FCB579}50.3 &
\cellcolor[HTML]{B2D580}10.7 &
\cellcolor[HTML]{FCB87A}8.3 \\
OK-VQA QG &
9 K &
\cellcolor[HTML]{FDEB84}31.4 &
\cellcolor[HTML]{FEE883}25.0 &
\cellcolor[HTML]{FCB87A}17.7 &
\cellcolor[HTML]{FA9272}12.9 &
\cellcolor[HTML]{FBA175}39.7 &
\cellcolor[HTML]{FBAC78}18.6 &
\cellcolor[HTML]{FBA276}39.0 &
\cellcolor[HTML]{FBB279}20.3 &
\cellcolor[HTML]{FCB77A}36.9 &
\cellcolor[HTML]{FBB279}42.7 &
\cellcolor[HTML]{F98A71}7.8 &
\cellcolor[HTML]{FA9673}28.2 &
\cellcolor[HTML]{FCEA84}43.2 &
\cellcolor[HTML]{F87A6E}49.2 &
\cellcolor[HTML]{FBA676}11.6 &
\cellcolor[HTML]{FEDA80}32.7 &
\cellcolor[HTML]{FAA075}12.2 &
\cellcolor[HTML]{FBA175}8.0 &
\cellcolor[HTML]{FA9E75}41.1 &
\cellcolor[HTML]{F97D6F}3.7 &
\cellcolor[HTML]{F98B71}23.4 &
\cellcolor[HTML]{F2E884}8.4 &
\cellcolor[HTML]{F98871}0.9 &
\cellcolor[HTML]{FEE883}36.6 &
\cellcolor[HTML]{FBA877}50.4 &
\cellcolor[HTML]{63BE7B}12.6 &
\cellcolor[HTML]{E2E383}52.1 &
\cellcolor[HTML]{AAD380}11.2 &
\cellcolor[HTML]{FEE282}11.8 \\
A-OKVQA QG &
17 K &
\cellcolor[HTML]{FBAB77}19.3 &
\cellcolor[HTML]{FA9573}16.2 &
\cellcolor[HTML]{FA9D75}15.4 &
\cellcolor[HTML]{FA8F72}12.0 &
\cellcolor[HTML]{FCB579}41.9 &
\cellcolor[HTML]{FA9673}12.5 &
\cellcolor[HTML]{FBAC78}40.1 &
\cellcolor[HTML]{FAA075}15.1 &
\cellcolor[HTML]{FA9072}30.7 &
\cellcolor[HTML]{FBA777}41.5 &
\cellcolor[HTML]{F98971}7.7 &
\cellcolor[HTML]{FCB479}30.6 &
\cellcolor[HTML]{FCBF7B}40.1 &
\cellcolor[HTML]{F9E984}53.8 &
\cellcolor[HTML]{FBA175}10.5 &
\cellcolor[HTML]{FEE983}34.2 &
\cellcolor[HTML]{BCD881}12.7 &
\cellcolor[HTML]{FA9673}6.5 &
\cellcolor[HTML]{F98D72}38.8 &
\cellcolor[HTML]{F87A6E}3.1 &
\cellcolor[HTML]{F98370}22.4 &
\cellcolor[HTML]{FDC97D}5.3 &
\cellcolor[HTML]{FA9172}1.2 &
\cellcolor[HTML]{FCC37C}31.0 &
\cellcolor[HTML]{F8696B}46.2 &
\cellcolor[HTML]{DBE182}10.4 &
\cellcolor[HTML]{FA9573}49.7 &
\cellcolor[HTML]{B8D780}10.4 &
\cellcolor[HTML]{FBA877}7.0
			\\
			\bottomrule
		\end{tabular}%
	}
	\caption
	{Unnormalized transfer learning  performance of LLaVA. Higher values indicate better performance. QG denotes question generation, MC denotes multiple-choice and G denotes open-ended generation. The color scale is normalized along each column. The colors represent values in descending order: green, yellow, orange and red. 
	}
	\label{eval:affinity_matrix_raw_llava_landscape}
\end{table}
\end{landscape}

\begin{landscape}
\begin{table}[!tp]
	\resizebox{\columnwidth}{!}{%
		\begin{tabular}{c|c|ccccccccccccccccccccccccccccc}
			\toprule
			Source   Task &
			\setlength\extrarowheight{0pt}\begin{tabular}[c]{@{}c@{}}Dataset\\ Size \end{tabular} &
			\multicolumn{29}{c}{Target Task} \\
			&
			&
			\setlength\extrarowheight{0pt}\begin{tabular}[c]{@{}c@{}}COCO\\ Caption \end{tabular}  &
			\setlength\extrarowheight{0pt}\begin{tabular}[c]{@{}c@{}}Flickr\\ 30k\end{tabular}  &
			\setlength\extrarowheight{0pt}\begin{tabular}[c]{@{}c@{}}Text\\ Caps\end{tabular}  &
			\multicolumn{2}{c}{VQAv2} &
			\multicolumn{2}{c}{OK-VQA} &
			\multicolumn{2}{c}{A-OKVQA} &
			\setlength\extrarowheight{0pt}\begin{tabular}[c]{@{}c@{}}Science\\ QA\end{tabular} &
			\multicolumn{2}{c}{GQA} &
			\setlength\extrarowheight{0pt}\begin{tabular}[c]{@{}c@{}}Icon\\ QA \end{tabular} &
			VSR &
			\multicolumn{2}{c}{CLEVR} &
			\setlength\extrarowheight{0pt}\begin{tabular}[c]{@{}c@{}}RAVEN-\\ FAIR \end{tabular} &
			\multicolumn{2}{c}{\setlength\extrarowheight{0pt}\begin{tabular}[c]{@{}c@{}}Text\\ VQA\end{tabular}} &
			\multicolumn{2}{c}{\setlength\extrarowheight{0pt}\begin{tabular}[c]{@{}c@{}}OCR-\\ VQA\end{tabular}} &
			\setlength\extrarowheight{0pt}\begin{tabular}[c]{@{}c@{}}Open\\ CQA \end{tabular} &
			\multicolumn{2}{c}{\setlength\extrarowheight{0pt}\begin{tabular}[c]{@{}c@{}}Chart\\ QA\end{tabular}} &
			HM &
			\setlength\extrarowheight{0pt}\begin{tabular}[c]{@{}c@{}}NY\\ Explain \end{tabular} &
			\setlength\extrarowheight{0pt}\begin{tabular}[c]{@{}c@{}}NY\\ Rank \end{tabular} &
			MORE &
			\vlbenchmark \\
			&
			&
			&
			&
			&
			G &
			MC &
			G &
			MC &
			G &
			MC &
			&
			G &
			MC &
			&
			&
			G &
			MC &
			&
			G &
			MC &
			G &
			MC &
			&
			G &
			MC &
			&
			&
			&
			&
			\\
			\midrule
			Zero-shot &
			- &
			\cellcolor[HTML]{FA9373}13.3 &
			\cellcolor[HTML]{FCB579}14.5 &
			\cellcolor[HTML]{F8696B}12.4 &
			\cellcolor[HTML]{F86E6C}0.2 &
			\cellcolor[HTML]{FEE582}48.6 &
			\cellcolor[HTML]{F86E6C}0.1 &
			\cellcolor[HTML]{FDC97D}39.5 &
			\cellcolor[HTML]{F8696B}0.0 &
			\cellcolor[HTML]{FCB77A}33.4 &
			\cellcolor[HTML]{FDC97D}43.6 &
			\cellcolor[HTML]{F8696B}0.0 &
			\cellcolor[HTML]{FEEA83}35.8 &
			\cellcolor[HTML]{FCC07B}41.2 &
			\cellcolor[HTML]{FCBE7B}52.7 &
			\cellcolor[HTML]{F8696B}0.0 &
			\cellcolor[HTML]{FEDF81}34.1 &
			\cellcolor[HTML]{FFEB84}12.4 &
			\cellcolor[HTML]{F98370}0.6 &
			\cellcolor[HTML]{FEDF81}47.7 &
			\cellcolor[HTML]{F8706C}0.0 &
			\cellcolor[HTML]{FEDA80}48.0 &
			\cellcolor[HTML]{D1DE82}15.8 &
			\cellcolor[HTML]{FA9874}1.9 &
			\cellcolor[HTML]{DFE283}41.8 &
			\cellcolor[HTML]{FBAC77}49.8 &
			\cellcolor[HTML]{E6E483}11.3 &
			\cellcolor[HTML]{E6E483}49.0 &
			\cellcolor[HTML]{FA9573}5.8 &
			\cellcolor[HTML]{F8796E}1.6 \\
			COCO Caption &
			567 K &
			\cellcolor[HTML]{63BE7B}136.1 &
			\cellcolor[HTML]{7FC67D}80.4 &
			\cellcolor[HTML]{AAD380}68.1 &
			\cellcolor[HTML]{F8696B}0.0 &
			\cellcolor[HTML]{FDD27F}44.9 &
			\cellcolor[HTML]{F86A6B}0.0 &
			\cellcolor[HTML]{FDD17F}39.9 &
			\cellcolor[HTML]{F8696B}0.0 &
			\cellcolor[HTML]{FBB179}32.4 &
			\cellcolor[HTML]{FDD47F}44.3 &
			\cellcolor[HTML]{F8696B}0.0 &
			\cellcolor[HTML]{FEDD81}33.9 &
			\cellcolor[HTML]{F9EA84}43.6 &
			\cellcolor[HTML]{FBA676}51.4 &
			\cellcolor[HTML]{F8696B}0.0 &
			\cellcolor[HTML]{EAE583}35.0 &
			\cellcolor[HTML]{F0E784}12.5 &
			\cellcolor[HTML]{F8796E}0.4 &
			\cellcolor[HTML]{FDCE7E}45.5 &
			\cellcolor[HTML]{FBA376}0.3 &
			\cellcolor[HTML]{FDD680}47.3 &
			\cellcolor[HTML]{FCEA84}10.4 &
			\cellcolor[HTML]{FA9473}1.7 &
			\cellcolor[HTML]{91CC7E}46.4 &
			\cellcolor[HTML]{FCC57C}50.4 &
			\cellcolor[HTML]{FDD57F}9.7 &
			\cellcolor[HTML]{FBA977}47.7 &
			\cellcolor[HTML]{F7E984}10.0 &
			\cellcolor[HTML]{EEE784}9.9 \\
			Flickr30k &
			145 K &
			\cellcolor[HTML]{8DCA7E}109.4 &
			\cellcolor[HTML]{63BE7B}92.2 &
			\cellcolor[HTML]{ADD480}67.0 &
			\cellcolor[HTML]{F97E6F}1.0 &
			\cellcolor[HTML]{F1E784}51.3 &
			\cellcolor[HTML]{FDC67C}1.4 &
			\cellcolor[HTML]{D7E082}46.6 &
			\cellcolor[HTML]{FBA576}0.8 &
			\cellcolor[HTML]{DEE283}49.0 &
			\cellcolor[HTML]{FEE983}45.7 &
			\cellcolor[HTML]{F97E6F}0.7 &
			\cellcolor[HTML]{EAE583}37.5 &
			\cellcolor[HTML]{FEDF81}42.2 &
			\cellcolor[HTML]{B5D680}59.1 &
			\cellcolor[HTML]{FCC47C}0.1 &
			\cellcolor[HTML]{FDCC7E}33.7 &
			\cellcolor[HTML]{B2D580}12.7 &
			\cellcolor[HTML]{FFEB84}3.3 &
			\cellcolor[HTML]{F3E884}50.3 &
			\cellcolor[HTML]{FDD57F}0.6 &
			\cellcolor[HTML]{F5E984}52.2 &
			\cellcolor[HTML]{E9E583}12.8 &
			\cellcolor[HTML]{FBEA84}5.2 &
			\cellcolor[HTML]{CCDD82}42.9 &
			\cellcolor[HTML]{FBAC77}49.8 &
			\cellcolor[HTML]{FEEA83}10.8 &
			\cellcolor[HTML]{FCBC7A}48.0 &
			\cellcolor[HTML]{EFE784}10.3 &
			\cellcolor[HTML]{E8E583}11.0 \\
			Web CapFilt &
			23,147 K &
			\cellcolor[HTML]{71C27C}127.4 &
			\cellcolor[HTML]{86C87D}77.3 &
			\cellcolor[HTML]{A4D17F}71.2 &
			\cellcolor[HTML]{FA9072}1.8 &
			\cellcolor[HTML]{DCE182}53.5 &
			\cellcolor[HTML]{FDD07E}1.6 &
			\cellcolor[HTML]{C5DB81}48.9 &
			\cellcolor[HTML]{FBAD78}1.0 &
			\cellcolor[HTML]{CCDD82}52.4 &
			\cellcolor[HTML]{F7E984}47.5 &
			\cellcolor[HTML]{FBA676}1.9 &
			\cellcolor[HTML]{D0DE82}39.6 &
			\cellcolor[HTML]{FCBE7B}41.2 &
			\cellcolor[HTML]{FA8F72}50.2 &
			\cellcolor[HTML]{F8696B}0.0 &
			\cellcolor[HTML]{FEDD81}34.0 &
			\cellcolor[HTML]{F0E784}12.5 &
			\cellcolor[HTML]{FCBE7B}2.1 &
			\cellcolor[HTML]{F0E784}50.5 &
			\cellcolor[HTML]{FFEB84}0.9 &
			\cellcolor[HTML]{F97C6E}30.9 &
			\cellcolor[HTML]{FDD37F}8.2 &
			\cellcolor[HTML]{F8696B}0.0 &
			\cellcolor[HTML]{FBAF78}30.6 &
			\cellcolor[HTML]{F8716C}48.4 &
			\cellcolor[HTML]{FCB579}8.0 &
			\cellcolor[HTML]{FCC17C}48.1 &
			\cellcolor[HTML]{FEE583}9.6 &
			\cellcolor[HTML]{F1E784}9.4 \\
			TextCaps &
			549 K &
			\cellcolor[HTML]{A9D380}91.3 &
			\cellcolor[HTML]{A9D380}62.0 &
			\cellcolor[HTML]{63BE7B}99.7 &
			\cellcolor[HTML]{F8776D}0.7 &
			\cellcolor[HTML]{FEE382}48.3 &
			\cellcolor[HTML]{FBAA77}1.0 &
			\cellcolor[HTML]{EAE583}44.2 &
			\cellcolor[HTML]{FBA376}0.8 &
			\cellcolor[HTML]{FFEB84}42.4 &
			\cellcolor[HTML]{FDCA7D}43.7 &
			\cellcolor[HTML]{F8766D}0.4 &
			\cellcolor[HTML]{F3E884}36.8 &
			\cellcolor[HTML]{F97D6F}39.3 &
			\cellcolor[HTML]{FBA676}51.4 &
			\cellcolor[HTML]{F8696B}0.0 &
			\cellcolor[HTML]{FEE382}34.1 &
			\cellcolor[HTML]{79C57D}13.0 &
			\cellcolor[HTML]{FA9473}1.1 &
			\cellcolor[HTML]{DFE283}51.8 &
			\cellcolor[HTML]{F8766D}0.1 &
			\cellcolor[HTML]{F8696B}27.3 &
			\cellcolor[HTML]{EAE583}12.7 &
			\cellcolor[HTML]{F8696B}0.0 &
			\cellcolor[HTML]{F8696B}19.4 &
			\cellcolor[HTML]{FBA376}49.6 &
			\cellcolor[HTML]{FEDB81}10.0 &
			\cellcolor[HTML]{F8696B}46.8 &
			\cellcolor[HTML]{63BE7B}14.0 &
			\cellcolor[HTML]{EFE784}9.8 \\
			VQAv2 &
			444 K &
			\cellcolor[HTML]{FEDB81}31.7 &
			\cellcolor[HTML]{FEE282}22.9 &
			\cellcolor[HTML]{FDD57F}27.2 &
			\cellcolor[HTML]{63BE7B}68.7 &
			\cellcolor[HTML]{63BE7B}66.3 &
			\cellcolor[HTML]{6FC27C}50.2 &
			\cellcolor[HTML]{7EC67D}58.0 &
			\cellcolor[HTML]{6BC17C}52.5 &
			\cellcolor[HTML]{8ECB7E}64.7 &
			\cellcolor[HTML]{DBE182}53.3 &
			\cellcolor[HTML]{76C47D}44.8 &
			\cellcolor[HTML]{97CD7E}44.0 &
			\cellcolor[HTML]{FBEA84}43.3 &
			\cellcolor[HTML]{67BF7C}63.3 &
			\cellcolor[HTML]{6AC07C}34.8 &
			\cellcolor[HTML]{E8E583}35.0 &
			\cellcolor[HTML]{F5E884}12.4 &
			\cellcolor[HTML]{83C87D}27.0 &
			\cellcolor[HTML]{94CC7E}57.5 &
			\cellcolor[HTML]{A4D17F}31.5 &
			\cellcolor[HTML]{CEDD82}57.0 &
			\cellcolor[HTML]{FA9874}3.8 &
			\cellcolor[HTML]{7EC67D}9.3 &
			\cellcolor[HTML]{FEE582}39.0 &
			\cellcolor[HTML]{FBEA84}51.8 &
			\cellcolor[HTML]{FA9F75}6.9 &
			\cellcolor[HTML]{FCC47C}48.1 &
			\cellcolor[HTML]{FA9E75}6.3 &
			\cellcolor[HTML]{F98B71}2.5 \\
			OK-VQA &
			9 K &
			\cellcolor[HTML]{FBA776}18.2 &
			\cellcolor[HTML]{FBA476}11.3 &
			\cellcolor[HTML]{FBA476}20.5 &
			\cellcolor[HTML]{83C77D}56.2 &
			\cellcolor[HTML]{CBDC81}55.3 &
			\cellcolor[HTML]{63BE7B}54.1 &
			\cellcolor[HTML]{73C37C}59.5 &
			\cellcolor[HTML]{82C77D}44.6 &
			\cellcolor[HTML]{A7D27F}59.7 &
			\cellcolor[HTML]{E0E283}52.2 &
			\cellcolor[HTML]{92CC7E}36.6 &
			\cellcolor[HTML]{FEEB84}36.0 &
			\cellcolor[HTML]{FCB479}40.9 &
			\cellcolor[HTML]{94CC7E}60.9 &
			\cellcolor[HTML]{6FC27C}33.6 &
			\cellcolor[HTML]{FBA977}32.9 &
			\cellcolor[HTML]{F8696B}12.0 &
			\cellcolor[HTML]{8CCA7E}25.4 &
			\cellcolor[HTML]{9ECF7F}56.7 &
			\cellcolor[HTML]{D5DF82}15.1 &
			\cellcolor[HTML]{FEEA83}50.9 &
			\cellcolor[HTML]{F97F6F}2.0 &
			\cellcolor[HTML]{91CC7E}8.6 &
			\cellcolor[HTML]{F4E884}40.6 &
			\cellcolor[HTML]{F2E884}53.0 &
			\cellcolor[HTML]{FCB679}8.1 &
			\cellcolor[HTML]{9ACE7F}50.1 &
			\cellcolor[HTML]{FDD37F}8.7 &
			\cellcolor[HTML]{F97F6F}1.9 \\
			A-OKVQA &
			17 K &
			\cellcolor[HTML]{FDCD7E}28.0 &
			\cellcolor[HTML]{FDCD7E}18.8 &
			\cellcolor[HTML]{FBB279}22.5 &
			\cellcolor[HTML]{7AC57D}59.7 &
			\cellcolor[HTML]{ABD380}58.8 &
			\cellcolor[HTML]{6DC17C}50.7 &
			\cellcolor[HTML]{63BE7B}61.4 &
			\cellcolor[HTML]{63BE7B}55.1 &
			\cellcolor[HTML]{91CC7E}64.1 &
			\cellcolor[HTML]{DCE182}53.0 &
			\cellcolor[HTML]{8AC97E}39.0 &
			\cellcolor[HTML]{E2E383}38.2 &
			\cellcolor[HTML]{F5E984}44.2 &
			\cellcolor[HTML]{77C47D}62.4 &
			\cellcolor[HTML]{77C47D}31.7 &
			\cellcolor[HTML]{FCBA7A}33.3 &
			\cellcolor[HTML]{FEE382}12.4 &
			\cellcolor[HTML]{89C97E}26.0 &
			\cellcolor[HTML]{B0D580}55.4 &
			\cellcolor[HTML]{C5DB81}20.2 &
			\cellcolor[HTML]{D5DF82}56.1 &
			\cellcolor[HTML]{F98370}2.3 &
			\cellcolor[HTML]{86C97E}9.0 &
			\cellcolor[HTML]{FDEB84}40.0 &
			\cellcolor[HTML]{F5E884}52.6 &
			\cellcolor[HTML]{FDCE7E}9.3 &
			\cellcolor[HTML]{FDCE7E}48.2 &
			\cellcolor[HTML]{FDC97D}8.2 &
			\cellcolor[HTML]{F98871}2.3 \\
			A-OKVQA (MC) &
			17 K &
			\cellcolor[HTML]{F9EA84}39.6 &
			\cellcolor[HTML]{F9EA84}27.2 &
			\cellcolor[HTML]{F5E984}34.7 &
			\cellcolor[HTML]{7EC67D}58.2 &
			\cellcolor[HTML]{B4D680}57.7 &
			\cellcolor[HTML]{89C97E}41.7 &
			\cellcolor[HTML]{78C47D}58.8 &
			\cellcolor[HTML]{87C97E}42.9 &
			\cellcolor[HTML]{63BE7B}73.2 &
			\cellcolor[HTML]{C8DC81}57.1 &
			\cellcolor[HTML]{97CD7E}35.1 &
			\cellcolor[HTML]{FEE582}35.0 &
			\cellcolor[HTML]{EFE784}45.3 &
			\cellcolor[HTML]{FED880}54.1 &
			\cellcolor[HTML]{E8E583}5.7 &
			\cellcolor[HTML]{FCC27C}33.5 &
			\cellcolor[HTML]{FFEB84}12.4 &
			\cellcolor[HTML]{ACD380}19.3 &
			\cellcolor[HTML]{63BE7B}61.1 &
			\cellcolor[HTML]{FCEA84}2.0 &
			\cellcolor[HTML]{DCE182}55.4 &
			\cellcolor[HTML]{EAE583}12.7 &
			\cellcolor[HTML]{D1DE82}6.6 &
			\cellcolor[HTML]{CCDD82}42.9 &
			\cellcolor[HTML]{FEE683}51.2 &
			\cellcolor[HTML]{94CD7E}12.7 &
			\cellcolor[HTML]{C5DB81}49.5 &
			\cellcolor[HTML]{93CC7E}12.7 &
			\cellcolor[HTML]{D5DF82}14.3 \\
			ScienceQA &
			6 K &
			\cellcolor[HTML]{DAE182}60.0 &
			\cellcolor[HTML]{CFDD82}45.5 &
			\cellcolor[HTML]{E9E583}40.2 &
			\cellcolor[HTML]{95CD7E}48.7 &
			\cellcolor[HTML]{FDD57F}45.6 &
			\cellcolor[HTML]{A6D27F}32.0 &
			\cellcolor[HTML]{C0D981}49.6 &
			\cellcolor[HTML]{B0D580}28.8 &
			\cellcolor[HTML]{C3DA81}54.2 &
			\cellcolor[HTML]{63BE7B}77.8 &
			\cellcolor[HTML]{9ECF7F}32.9 &
			\cellcolor[HTML]{FDD47F}32.5 &
			\cellcolor[HTML]{F7E984}44.0 &
			\cellcolor[HTML]{FCBB7A}52.5 &
			\cellcolor[HTML]{93CC7E}25.1 &
			\cellcolor[HTML]{F8696B}31.6 &
			\cellcolor[HTML]{63BE7B}13.1 &
			\cellcolor[HTML]{B3D580}17.8 &
			\cellcolor[HTML]{9DCF7F}56.8 &
			\cellcolor[HTML]{F7E984}3.6 &
			\cellcolor[HTML]{FFEB84}51.0 &
			\cellcolor[HTML]{FBA276}4.6 &
			\cellcolor[HTML]{F0E784}5.6 &
			\cellcolor[HTML]{FCC37C}33.7 &
			\cellcolor[HTML]{FEDE81}51.0 &
			\cellcolor[HTML]{63BE7B}13.6 &
			\cellcolor[HTML]{63BE7B}50.8 &
			\cellcolor[HTML]{70C27C}13.7 &
			\cellcolor[HTML]{F86E6B}1.1 \\
			GQA &
			943 K &
			\cellcolor[HTML]{FA9F75}16.2 &
			\cellcolor[HTML]{FA9573}8.7 &
			\cellcolor[HTML]{FBA877}21.1 &
			\cellcolor[HTML]{78C47D}60.5 &
			\cellcolor[HTML]{86C97E}62.6 &
			\cellcolor[HTML]{94CC7E}38.0 &
			\cellcolor[HTML]{96CD7E}55.0 &
			\cellcolor[HTML]{86C87D}43.3 &
			\cellcolor[HTML]{A8D27F}59.6 &
			\cellcolor[HTML]{F2E784}48.6 &
			\cellcolor[HTML]{63BE7B}50.3 &
			\cellcolor[HTML]{63BE7B}48.1 &
			\cellcolor[HTML]{FBEA84}43.3 &
			\cellcolor[HTML]{FDC77D}53.2 &
			\cellcolor[HTML]{63BE7B}36.2 &
			\cellcolor[HTML]{FEE683}34.2 &
			\cellcolor[HTML]{FFEB84}12.4 &
			\cellcolor[HTML]{ACD480}19.1 &
			\cellcolor[HTML]{FEE883}49.0 &
			\cellcolor[HTML]{B7D780}25.2 &
			\cellcolor[HTML]{F4E884}52.4 &
			\cellcolor[HTML]{F98A71}2.9 &
			\cellcolor[HTML]{9FD07F}8.2 &
			\cellcolor[HTML]{FEE182}38.4 &
			\cellcolor[HTML]{FFEB84}51.4 &
			\cellcolor[HTML]{FED980}9.9 &
			\cellcolor[HTML]{FBA977}47.7 &
			\cellcolor[HTML]{FCC17C}7.9 &
			\cellcolor[HTML]{F98770}2.3 \\
			IconQA &
			19 K &
			\cellcolor[HTML]{E1E383}55.2 &
			\cellcolor[HTML]{E3E383}36.8 &
			\cellcolor[HTML]{EFE784}37.4 &
			\cellcolor[HTML]{E3E383}17.2 &
			\cellcolor[HTML]{FEDD81}47.2 &
			\cellcolor[HTML]{ECE683}8.4 &
			\cellcolor[HTML]{FDD27F}40.0 &
			\cellcolor[HTML]{ECE683}8.4 &
			\cellcolor[HTML]{FEE983}42.1 &
			\cellcolor[HTML]{FEDD81}44.9 &
			\cellcolor[HTML]{E9E583}10.7 &
			\cellcolor[HTML]{FEE783}35.3 &
			\cellcolor[HTML]{63BE7B}68.5 &
			\cellcolor[HTML]{FDD780}54.0 &
			\cellcolor[HTML]{AFD480}18.6 &
			\cellcolor[HTML]{C2DA81}36.2 &
			\cellcolor[HTML]{85C87D}13.0 &
			\cellcolor[HTML]{ECE683}6.9 &
			\cellcolor[HTML]{FCC17C}43.8 &
			\cellcolor[HTML]{EEE784}6.5 &
			\cellcolor[HTML]{F8726C}29.0 &
			\cellcolor[HTML]{FBA376}4.7 &
			\cellcolor[HTML]{E1E383}6.1 &
			\cellcolor[HTML]{FCBE7B}33.0 &
			\cellcolor[HTML]{F8696B}48.2 &
			\cellcolor[HTML]{9ACE7F}12.6 &
			\cellcolor[HTML]{7FC67D}50.5 &
			\cellcolor[HTML]{C1D981}11.5 &
			\cellcolor[HTML]{C0D981}18.0 \\
			VSR &
			3 K &
			\cellcolor[HTML]{E9E583}50.0 &
			\cellcolor[HTML]{E2E383}37.0 &
			\cellcolor[HTML]{FCB579}22.9 &
			\cellcolor[HTML]{FCBE7B}3.9 &
			\cellcolor[HTML]{DDE283}53.4 &
			\cellcolor[HTML]{FFEB84}2.1 &
			\cellcolor[HTML]{FDD780}40.3 &
			\cellcolor[HTML]{FEEB84}2.2 &
			\cellcolor[HTML]{F6E984}44.2 &
			\cellcolor[HTML]{FEE583}45.4 &
			\cellcolor[HTML]{FDD37F}3.3 &
			\cellcolor[HTML]{FFEB84}35.9 &
			\cellcolor[HTML]{F3E884}44.6 &
			\cellcolor[HTML]{63BE7B}63.5 &
			\cellcolor[HTML]{F8696B}0.0 &
			\cellcolor[HTML]{6FC27C}38.7 &
			\cellcolor[HTML]{F9EA84}12.4 &
			\cellcolor[HTML]{FAA075}1.3 &
			\cellcolor[HTML]{F5E884}50.2 &
			\cellcolor[HTML]{F97C6E}0.1 &
			\cellcolor[HTML]{FDEB84}51.3 &
			\cellcolor[HTML]{B2D580}19.8 &
			\cellcolor[HTML]{FBB178}2.8 &
			\cellcolor[HTML]{EBE583}41.1 &
			\cellcolor[HTML]{FDEB84}51.6 &
			\cellcolor[HTML]{ABD380}12.3 &
			\cellcolor[HTML]{FCBC7A}48.0 &
			\cellcolor[HTML]{FEE683}9.6 &
			\cellcolor[HTML]{FEE783}6.8 \\
			TextVQA &
			35 K &
			\cellcolor[HTML]{F8696B}2.2 &
			\cellcolor[HTML]{F8696B}0.5 &
			\cellcolor[HTML]{FA9773}18.8 &
			\cellcolor[HTML]{BBD881}33.7 &
			\cellcolor[HTML]{CEDD82}55.0 &
			\cellcolor[HTML]{A4D17F}32.4 &
			\cellcolor[HTML]{BAD780}50.4 &
			\cellcolor[HTML]{9ECF7F}35.1 &
			\cellcolor[HTML]{D1DE82}51.4 &
			\cellcolor[HTML]{D4DF82}54.6 &
			\cellcolor[HTML]{B8D780}25.1 &
			\cellcolor[HTML]{CFDE82}39.6 &
			\cellcolor[HTML]{F8E984}43.7 &
			\cellcolor[HTML]{A5D17F}60.0 &
			\cellcolor[HTML]{7AC57D}31.0 &
			\cellcolor[HTML]{F8766D}31.8 &
			\cellcolor[HTML]{F0E784}12.5 &
			\cellcolor[HTML]{63BE7B}33.1 &
			\cellcolor[HTML]{80C77D}59.0 &
			\cellcolor[HTML]{A1D07F}32.4 &
			\cellcolor[HTML]{FA9874}36.0 &
			\cellcolor[HTML]{F8696B}0.4 &
			\cellcolor[HTML]{63BE7B}10.1 &
			\cellcolor[HTML]{FCBE7B}32.9 &
			\cellcolor[HTML]{F8E984}52.2 &
			\cellcolor[HTML]{F8696B}4.0 &
			\cellcolor[HTML]{FEDE81}48.5 &
			\cellcolor[HTML]{F8696B}3.7 &
			\cellcolor[HTML]{F98770}2.3 \\
			OCR-VQA &
			802 K &
			\cellcolor[HTML]{A7D27F}92.4 &
			\cellcolor[HTML]{B6D680}56.3 &
			\cellcolor[HTML]{D7E082}48.5 &
			\cellcolor[HTML]{E5E483}16.8 &
			\cellcolor[HTML]{BBD881}57.1 &
			\cellcolor[HTML]{DEE283}13.1 &
			\cellcolor[HTML]{DEE283}45.8 &
			\cellcolor[HTML]{E1E383}12.2 &
			\cellcolor[HTML]{F4E884}44.5 &
			\cellcolor[HTML]{DEE283}52.6 &
			\cellcolor[HTML]{EFE784}9.0 &
			\cellcolor[HTML]{BFD981}40.9 &
			\cellcolor[HTML]{FDCD7E}41.6 &
			\cellcolor[HTML]{79C57D}62.4 &
			\cellcolor[HTML]{E7E483}5.9 &
			\cellcolor[HTML]{D0DE82}35.8 &
			\cellcolor[HTML]{FFEB84}12.4 &
			\cellcolor[HTML]{BAD881}16.5 &
			\cellcolor[HTML]{FBEA84}49.6 &
			\cellcolor[HTML]{63BE7B}53.1 &
			\cellcolor[HTML]{63BE7B}70.1 &
			\cellcolor[HTML]{FDD27F}8.1 &
			\cellcolor[HTML]{9DCF7F}8.3 &
			\cellcolor[HTML]{E4E483}41.5 &
			\cellcolor[HTML]{FCBC7B}50.2 &
			\cellcolor[HTML]{A9D380}12.3 &
			\cellcolor[HTML]{FED980}48.4 &
			\cellcolor[HTML]{D9E082}10.9 &
			\cellcolor[HTML]{FFEB84}7.1 \\
			OpenCQA &
			6 K &
			\cellcolor[HTML]{FDCF7E}28.5 &
			\cellcolor[HTML]{FDD57F}20.5 &
			\cellcolor[HTML]{FBAD78}21.8 &
			\cellcolor[HTML]{F86C6B}0.2 &
			\cellcolor[HTML]{F5E884}50.9 &
			\cellcolor[HTML]{F8766D}0.2 &
			\cellcolor[HTML]{FCEA84}42.0 &
			\cellcolor[HTML]{F86F6C}0.1 &
			\cellcolor[HTML]{FDD27F}38.1 &
			\cellcolor[HTML]{E6E483}51.0 &
			\cellcolor[HTML]{F86A6B}0.0 &
			\cellcolor[HTML]{E1E383}38.2 &
			\cellcolor[HTML]{FCBA7A}41.1 &
			\cellcolor[HTML]{FCBB7A}52.5 &
			\cellcolor[HTML]{F8696B}0.0 &
			\cellcolor[HTML]{E5E483}35.1 &
			\cellcolor[HTML]{FFEB84}12.4 &
			\cellcolor[HTML]{F98870}0.8 &
			\cellcolor[HTML]{CEDD82}53.1 &
			\cellcolor[HTML]{FA9072}0.2 &
			\cellcolor[HTML]{F7E984}52.0 &
			\cellcolor[HTML]{63BE7B}29.8 &
			\cellcolor[HTML]{F8696B}0.0 &
			\cellcolor[HTML]{63BE7B}49.0 &
			\cellcolor[HTML]{F2E884}53.0 &
			\cellcolor[HTML]{B4D680}12.2 &
			\cellcolor[HTML]{A5D17F}49.9 &
			\cellcolor[HTML]{F0E784}10.3 &
			\cellcolor[HTML]{BFD981}18.3 \\
			HM &
			9 K &
			\cellcolor[HTML]{97CD7E}103.1 &
			\cellcolor[HTML]{A1D07F}65.7 &
			\cellcolor[HTML]{D3DF82}50.1 &
			\cellcolor[HTML]{97CD7E}48.1 &
			\cellcolor[HTML]{D5DF82}54.2 &
			\cellcolor[HTML]{C9DC81}20.3 &
			\cellcolor[HTML]{FEE282}41.0 &
			\cellcolor[HTML]{CDDD82}19.0 &
			\cellcolor[HTML]{FEDB81}39.7 &
			\cellcolor[HTML]{FFEB84}45.8 &
			\cellcolor[HTML]{A1D07F}31.9 &
			\cellcolor[HTML]{E0E283}38.4 &
			\cellcolor[HTML]{FDD07E}41.7 &
			\cellcolor[HTML]{EDE683}56.1 &
			\cellcolor[HTML]{9ECF7F}22.8 &
			\cellcolor[HTML]{FEE382}34.1 &
			\cellcolor[HTML]{E8E583}12.5 &
			\cellcolor[HTML]{CDDD82}12.8 &
			\cellcolor[HTML]{FDD37F}46.2 &
			\cellcolor[HTML]{C0D981}22.0 &
			\cellcolor[HTML]{FDD780}47.4 &
			\cellcolor[HTML]{FDCB7E}7.6 &
			\cellcolor[HTML]{99CE7F}8.4 &
			\cellcolor[HTML]{FEDB80}37.4 &
			\cellcolor[HTML]{63BE7B}70.6 &
			\cellcolor[HTML]{FCEA84}10.9 &
			\cellcolor[HTML]{FA8E72}47.3 &
			\cellcolor[HTML]{A3D17F}12.3 &
			\cellcolor[HTML]{D6DF82}14.2 \\
			\vlbenchmark{} &
			7 K &
			\cellcolor[HTML]{FA9974}14.7 &
			\cellcolor[HTML]{FCC17B}16.7 &
			\cellcolor[HTML]{FA9E75}19.8 &
			\cellcolor[HTML]{F8696B}0.0 &
			\cellcolor[HTML]{FDCF7E}44.3 &
			\cellcolor[HTML]{F86C6B}0.1 &
			\cellcolor[HTML]{FCBD7B}38.8 &
			\cellcolor[HTML]{F86B6B}0.0 &
			\cellcolor[HTML]{FCBB7A}34.1 &
			\cellcolor[HTML]{FDD27F}44.2 &
			\cellcolor[HTML]{F86A6B}0.0 &
			\cellcolor[HTML]{FEDC81}33.7 &
			\cellcolor[HTML]{FEE382}42.3 &
			\cellcolor[HTML]{C6DB81}58.2 &
			\cellcolor[HTML]{F8696B}0.0 &
			\cellcolor[HTML]{99CE7F}37.5 &
			\cellcolor[HTML]{FFEB84}12.4 &
			\cellcolor[HTML]{F97E6F}0.5 &
			\cellcolor[HTML]{FDD07E}45.9 &
			\cellcolor[HTML]{F8746D}0.1 &
			\cellcolor[HTML]{FEE082}49.1 &
			\cellcolor[HTML]{D5DF82}15.4 &
			\cellcolor[HTML]{F8726C}0.4 &
			\cellcolor[HTML]{FEDE81}37.9 &
			\cellcolor[HTML]{F7E984}52.4 &
			\cellcolor[HTML]{FEE983}10.7 &
			\cellcolor[HTML]{8FCB7E}50.2 &
			\cellcolor[HTML]{FCBF7B}7.8 &
			\cellcolor[HTML]{63BE7B}34.1 \\
			LLaVA Conversation &
			57 K &
			\cellcolor[HTML]{FDD47F}29.9 &
			\cellcolor[HTML]{FEE482}23.1 &
			\cellcolor[HTML]{FEDD81}28.4 &
			\cellcolor[HTML]{F8696B}0.0 &
			\cellcolor[HTML]{F8696B}24.3 &
			\cellcolor[HTML]{F8696B}0.0 &
			\cellcolor[HTML]{F98770}35.5 &
			\cellcolor[HTML]{F8696B}0.0 &
			\cellcolor[HTML]{FA9874}28.0 &
			\cellcolor[HTML]{FA9D75}40.8 &
			\cellcolor[HTML]{F8696B}0.0 &
			\cellcolor[HTML]{F8696B}16.6 &
			\cellcolor[HTML]{FDCA7D}41.5 &
			\cellcolor[HTML]{E6E483}56.5 &
			\cellcolor[HTML]{F8696B}0.0 &
			\cellcolor[HTML]{63BE7B}39.1 &
			\cellcolor[HTML]{FFEB84}12.4 &
			\cellcolor[HTML]{F86E6B}0.1 &
			\cellcolor[HTML]{FBA877}40.5 &
			\cellcolor[HTML]{F8696B}0.0 &
			\cellcolor[HTML]{FEE482}49.8 &
			\cellcolor[HTML]{C9DC81}16.9 &
			\cellcolor[HTML]{F8746D}0.4 &
			\cellcolor[HTML]{B4D680}44.3 &
			\cellcolor[HTML]{E5E483}54.6 &
			\cellcolor[HTML]{FDEB84}10.9 &
			\cellcolor[HTML]{F3E884}48.8 &
			\cellcolor[HTML]{FDD37F}8.7 &
			\cellcolor[HTML]{F8716C}1.2 \\
			LLaVA Reasoning &
			77 K &
			\cellcolor[HTML]{FA8F72}12.0 &
			\cellcolor[HTML]{FCBD7B}16.1 &
			\cellcolor[HTML]{F97B6E}15.0 &
			\cellcolor[HTML]{F8696B}0.0 &
			\cellcolor[HTML]{FCBF7B}41.2 &
			\cellcolor[HTML]{F8696B}0.0 &
			\cellcolor[HTML]{FCB87A}38.5 &
			\cellcolor[HTML]{F8696B}0.0 &
			\cellcolor[HTML]{FBA576}30.3 &
			\cellcolor[HTML]{FA9272}40.1 &
			\cellcolor[HTML]{F8696B}0.0 &
			\cellcolor[HTML]{FCC47C}30.1 &
			\cellcolor[HTML]{FBAB77}40.6 &
			\cellcolor[HTML]{E4E483}56.5 &
			\cellcolor[HTML]{F8696B}0.0 &
			\cellcolor[HTML]{7EC67D}38.3 &
			\cellcolor[HTML]{FFEB84}12.4 &
			\cellcolor[HTML]{F8706C}0.2 &
			\cellcolor[HTML]{FBAA77}40.7 &
			\cellcolor[HTML]{F8696B}0.0 &
			\cellcolor[HTML]{FDCB7E}45.3 &
			\cellcolor[HTML]{FEE482}9.4 &
			\cellcolor[HTML]{F8716C}0.3 &
			\cellcolor[HTML]{FEE082}38.3 &
			\cellcolor[HTML]{FBA376}49.6 &
			\cellcolor[HTML]{FDD47F}9.7 &
			\cellcolor[HTML]{A7D27F}49.9 &
			\cellcolor[HTML]{FBA276}6.4 &
			\cellcolor[HTML]{F8696B}0.9 \\
			LLaVA Description &
			23 K &
			\cellcolor[HTML]{F98770}10.0 &
			\cellcolor[HTML]{FBB178}13.8 &
			\cellcolor[HTML]{F8706C}13.4 &
			\cellcolor[HTML]{F8696B}0.0 &
			\cellcolor[HTML]{FBA877}36.7 &
			\cellcolor[HTML]{F8696B}0.0 &
			\cellcolor[HTML]{F8696B}33.7 &
			\cellcolor[HTML]{F8696B}0.0 &
			\cellcolor[HTML]{F8696B}19.8 &
			\cellcolor[HTML]{F8696B}37.4 &
			\cellcolor[HTML]{F8696B}0.0 &
			\cellcolor[HTML]{FCB479}27.8 &
			\cellcolor[HTML]{F8696B}38.6 &
			\cellcolor[HTML]{F8696B}48.1 &
			\cellcolor[HTML]{F8696B}0.0 &
			\cellcolor[HTML]{FDEB84}34.4 &
			\cellcolor[HTML]{C3DA81}12.7 &
			\cellcolor[HTML]{F8696B}0.0 &
			\cellcolor[HTML]{F8696B}32.0 &
			\cellcolor[HTML]{F86E6C}0.0 &
			\cellcolor[HTML]{FBAB77}39.4 &
			\cellcolor[HTML]{FEDE81}9.0 &
			\cellcolor[HTML]{F8696B}0.0 &
			\cellcolor[HTML]{FDD37F}36.2 &
			\cellcolor[HTML]{F8796E}48.6 &
			\cellcolor[HTML]{FEE482}10.5 &
			\cellcolor[HTML]{FDC97D}48.2 &
			\cellcolor[HTML]{FBB379}7.2 &
			\cellcolor[HTML]{FA9072}2.7 \\
			VQAv2 QG &
			444 K &
			\cellcolor[HTML]{B2D580}85.3 &
			\cellcolor[HTML]{BFD981}52.4 &
			\cellcolor[HTML]{D1DE82}50.8 &
			\cellcolor[HTML]{EEE683}13.1 &
			\cellcolor[HTML]{FDCC7E}43.8 &
			\cellcolor[HTML]{DCE182}13.8 &
			\cellcolor[HTML]{FCB479}38.2 &
			\cellcolor[HTML]{DDE182}13.7 &
			\cellcolor[HTML]{FBA676}30.5 &
			\cellcolor[HTML]{FEEB84}46.1 &
			\cellcolor[HTML]{F7E984}6.5 &
			\cellcolor[HTML]{FDD37F}32.3 &
			\cellcolor[HTML]{FBEA84}43.3 &
			\cellcolor[HTML]{FBB078}52.0 &
			\cellcolor[HTML]{F5E984}2.5 &
			\cellcolor[HTML]{FEE082}34.1 &
			\cellcolor[HTML]{FFEB84}12.4 &
			\cellcolor[HTML]{EDE683}6.8 &
			\cellcolor[HTML]{FDCA7D}44.9 &
			\cellcolor[HTML]{FFEB84}0.9 &
			\cellcolor[HTML]{E9E583}53.8 &
			\cellcolor[HTML]{E0E283}14.0 &
			\cellcolor[HTML]{FEE783}5.0 &
			\cellcolor[HTML]{ECE683}41.0 &
			\cellcolor[HTML]{FCBC7B}50.2 &
			\cellcolor[HTML]{B2D580}12.2 &
			\cellcolor[HTML]{E3E383}49.0 &
			\cellcolor[HTML]{82C77D}13.2 &
			\cellcolor[HTML]{FBAE78}4.1 \\
			OK-VQA QG &
			9 K &
			\cellcolor[HTML]{FDCB7E}27.7 &
			\cellcolor[HTML]{FEDD81}21.8 &
			\cellcolor[HTML]{FBEA84}32.0 &
			\cellcolor[HTML]{F9806F}1.1 &
			\cellcolor[HTML]{FDCE7E}44.2 &
			\cellcolor[HTML]{FEE482}1.9 &
			\cellcolor[HTML]{FA9D75}36.8 &
			\cellcolor[HTML]{FDD17F}1.5 &
			\cellcolor[HTML]{FCB679}33.2 &
			\cellcolor[HTML]{FEDC81}44.8 &
			\cellcolor[HTML]{F8766D}0.4 &
			\cellcolor[HTML]{FDD780}33.0 &
			\cellcolor[HTML]{F6E984}44.1 &
			\cellcolor[HTML]{FBAC77}51.7 &
			\cellcolor[HTML]{F86F6C}0.0 &
			\cellcolor[HTML]{C2DA81}36.2 &
			\cellcolor[HTML]{FFEB84}12.4 &
			\cellcolor[HTML]{FDD37F}2.6 &
			\cellcolor[HTML]{FEE683}48.7 &
			\cellcolor[HTML]{FA8E72}0.2 &
			\cellcolor[HTML]{EBE683}53.4 &
			\cellcolor[HTML]{D2DE82}15.8 &
			\cellcolor[HTML]{FEE282}4.8 &
			\cellcolor[HTML]{FDEB84}40.0 &
			\cellcolor[HTML]{EBE683}53.8 &
			\cellcolor[HTML]{C0D981}11.9 &
			\cellcolor[HTML]{D3DF82}49.3 &
			\cellcolor[HTML]{BDD881}11.6 &
			\cellcolor[HTML]{D1DE82}15.1 \\
			A-OKVQA QG &
			17 K &
			\cellcolor[HTML]{F9E984}40.1 &
			\cellcolor[HTML]{FDEB84}25.6 &
			\cellcolor[HTML]{EFE784}37.5 &
			\cellcolor[HTML]{FAEA84}8.0 &
			\cellcolor[HTML]{FDD57F}45.6 &
			\cellcolor[HTML]{FA9673}0.7 &
			\cellcolor[HTML]{FDCE7E}39.8 &
			\cellcolor[HTML]{FBAD78}1.0 &
			\cellcolor[HTML]{FBB078}32.1 &
			\cellcolor[HTML]{FEE582}45.4 &
			\cellcolor[HTML]{FDEB84}4.7 &
			\cellcolor[HTML]{FED880}33.1 &
			\cellcolor[HTML]{FEEB84}42.7 &
			\cellcolor[HTML]{E6E483}56.5 &
			\cellcolor[HTML]{FFEB84}0.2 &
			\cellcolor[HTML]{B3D680}36.6 &
			\cellcolor[HTML]{FFEB84}12.4 &
			\cellcolor[HTML]{FEE683}3.1 &
			\cellcolor[HTML]{FCB479}42.1 &
			\cellcolor[HTML]{FCB87A}0.4 &
			\cellcolor[HTML]{E6E483}54.1 &
			\cellcolor[HTML]{F1E784}11.8 &
			\cellcolor[HTML]{EAE583}5.8 &
			\cellcolor[HTML]{FEEA83}39.8 &
			\cellcolor[HTML]{FFEB84}51.4 &
			\cellcolor[HTML]{BAD780}12.1 &
			\cellcolor[HTML]{D0DE82}49.3 &
			\cellcolor[HTML]{B0D580}11.9 &
			\cellcolor[HTML]{CFDD82}15.4
 \\
			\bottomrule
		\end{tabular}%
	}
	\caption
	{Unnormalized transfer learning performance of MiniGPT-4. Higher values indicate better performance. QG denotes question generation, MC denotes multiple-choice and G denotes open-ended generation. The color scale is normalized along each column. The colors represent values in descending order: green, yellow, orange and red.
	}
	\label{eval:affinity_matrix_raw_minigpt_landscape}
\end{table}
\end{landscape}

\begin{landscape}
\begin{table}[!tp]
	\resizebox{\columnwidth}{!}{%
		\begin{tabular}{c|c|ccccccccccccccccccccccccccccc}
			\toprule
			Source   Task &
			\setlength\extrarowheight{0pt}\begin{tabular}[c]{@{}c@{}}Dataset\\ Size \end{tabular} &
			\multicolumn{29}{c}{Target Task} \\
			&
			&
			\setlength\extrarowheight{0pt}\begin{tabular}[c]{@{}c@{}}COCO\\ Caption \end{tabular}  &
			\setlength\extrarowheight{0pt}\begin{tabular}[c]{@{}c@{}}Flickr\\ 30k\end{tabular}  &
			\setlength\extrarowheight{0pt}\begin{tabular}[c]{@{}c@{}}Text\\ Caps\end{tabular}  &
			\multicolumn{2}{c}{VQAv2} &
			\multicolumn{2}{c}{OK-VQA} &
			\multicolumn{2}{c}{A-OKVQA} &
			\setlength\extrarowheight{0pt}\begin{tabular}[c]{@{}c@{}}Science\\ QA\end{tabular} &
			\multicolumn{2}{c}{GQA} &
			\setlength\extrarowheight{0pt}\begin{tabular}[c]{@{}c@{}}Icon\\ QA \end{tabular} &
			VSR &
			\multicolumn{2}{c}{CLEVR} &
			\setlength\extrarowheight{0pt}\begin{tabular}[c]{@{}c@{}}RAVEN-\\ FAIR \end{tabular} &
			\multicolumn{2}{c}{\setlength\extrarowheight{0pt}\begin{tabular}[c]{@{}c@{}}Text\\ VQA\end{tabular}} &
			\multicolumn{2}{c}{\setlength\extrarowheight{0pt}\begin{tabular}[c]{@{}c@{}}OCR-\\ VQA\end{tabular}} &
			\setlength\extrarowheight{0pt}\begin{tabular}[c]{@{}c@{}}Open\\ CQA \end{tabular} &
			\multicolumn{2}{c}{\setlength\extrarowheight{0pt}\begin{tabular}[c]{@{}c@{}}Chart\\ QA\end{tabular}} &
			HM &
			\setlength\extrarowheight{0pt}\begin{tabular}[c]{@{}c@{}}NY\\ Explain \end{tabular} &
			\setlength\extrarowheight{0pt}\begin{tabular}[c]{@{}c@{}}NY\\ Rank \end{tabular} &
			MORE &
			\vlbenchmark \\
			&
			&
			&
			&
			&
			G &
			MC &
			G &
			MC &
			G &
			MC &
			&
			G &
			MC &
			&
			&
			G &
			MC &
			&
			G &
			MC &
			G &
			MC &
			&
			G &
			MC &
			&
			&
			&
			&
			\\
			\midrule
			Zero-shot &
			- &
			\cellcolor[HTML]{F87A6E}16.6 &
			\cellcolor[HTML]{F8756D}15.2 &
			\cellcolor[HTML]{FA9A74}26.2 &
			\cellcolor[HTML]{F8696B}0.0 &
			\cellcolor[HTML]{FDCD7E}48.0 &
			\cellcolor[HTML]{F8696B}0.0 &
			\cellcolor[HTML]{FBA676}38.7 &
			\cellcolor[HTML]{F8696B}0.0 &
			\cellcolor[HTML]{FA9773}35.5 &
			\cellcolor[HTML]{FEE582}43.4 &
			\cellcolor[HTML]{F8696B}0.0 &
			\cellcolor[HTML]{FEE282}35.5 &
			\cellcolor[HTML]{FCBF7B}40.9 &
			\cellcolor[HTML]{D6DF82}55.9 &
			\cellcolor[HTML]{F8696B}0.0 &
			\cellcolor[HTML]{FDC77D}31.4 &
			\cellcolor[HTML]{FFEB84}12.4 &
			\cellcolor[HTML]{F8696B}0.0 &
			\cellcolor[HTML]{FCC07B}47.9 &
			\cellcolor[HTML]{F8696B}0.0 &
			\cellcolor[HTML]{DBE182}51.8 &
			\cellcolor[HTML]{F8706C}0.6 &
			\cellcolor[HTML]{F8696B}0.0 &
			\cellcolor[HTML]{D1DE82}42.6 &
			\cellcolor[HTML]{FFEB84}50.2 &
			\cellcolor[HTML]{FBEA84}10.8 &
			\cellcolor[HTML]{E3E383}50.6 &
			\cellcolor[HTML]{FEDB80}8.0 &
			\cellcolor[HTML]{FA8E72}4.4 \\
			COCO Caption &
			567 K &
			\cellcolor[HTML]{63BE7B}123.9 &
			\cellcolor[HTML]{83C87D}75.4 &
			\cellcolor[HTML]{C5DB81}69.8 &
			\cellcolor[HTML]{F8696B}0.0 &
			\cellcolor[HTML]{FDD37F}49.0 &
			\cellcolor[HTML]{F86B6B}0.1 &
			\cellcolor[HTML]{FCB479}40.1 &
			\cellcolor[HTML]{F86B6B}0.1 &
			\cellcolor[HTML]{FED980}43.8 &
			\cellcolor[HTML]{FCC07B}41.6 &
			\cellcolor[HTML]{F8696B}0.0 &
			\cellcolor[HTML]{FEE883}36.2 &
			\cellcolor[HTML]{FDC97D}41.5 &
			\cellcolor[HTML]{FEDD81}52.0 &
			\cellcolor[HTML]{F8696B}0.0 &
			\cellcolor[HTML]{E4E483}35.5 &
			\cellcolor[HTML]{FEE482}12.4 &
			\cellcolor[HTML]{F8706C}0.6 &
			\cellcolor[HTML]{C7DB81}58.8 &
			\cellcolor[HTML]{FA9673}2.4 &
			\cellcolor[HTML]{FEE783}44.1 &
			\cellcolor[HTML]{D7E082}14.8 &
			\cellcolor[HTML]{F8746D}0.2 &
			\cellcolor[HTML]{63BE7B}52.5 &
			\cellcolor[HTML]{F8696B}50.0 &
			\cellcolor[HTML]{E8E583}11.0 &
			\cellcolor[HTML]{FCBB7A}49.1 &
			\cellcolor[HTML]{ADD480}12.8 &
			\cellcolor[HTML]{ACD380}26.9 \\
			Flickr30k &
			145 K &
			\cellcolor[HTML]{A9D380}96.7 &
			\cellcolor[HTML]{63BE7B}83.2 &
			\cellcolor[HTML]{BCD881}73.3 &
			\cellcolor[HTML]{FA9B74}3.2 &
			\cellcolor[HTML]{FFEB84}52.9 &
			\cellcolor[HTML]{FDD27F}6.8 &
			\cellcolor[HTML]{FEDF81}44.2 &
			\cellcolor[HTML]{FEE582}7.5 &
			\cellcolor[HTML]{F3E884}48.4 &
			\cellcolor[HTML]{FFEB84}43.8 &
			\cellcolor[HTML]{FBA276}2.8 &
			\cellcolor[HTML]{F7E984}37.0 &
			\cellcolor[HTML]{F7E984}44.2 &
			\cellcolor[HTML]{FBAE78}50.5 &
			\cellcolor[HTML]{F86E6C}0.1 &
			\cellcolor[HTML]{FEE482}34.0 &
			\cellcolor[HTML]{CDDD82}12.6 &
			\cellcolor[HTML]{FBAB77}5.4 &
			\cellcolor[HTML]{CDDD82}58.3 &
			\cellcolor[HTML]{FDD680}5.8 &
			\cellcolor[HTML]{ECE683}48.4 &
			\cellcolor[HTML]{CCDD82}16.3 &
			\cellcolor[HTML]{FEE382}1.7 &
			\cellcolor[HTML]{BAD881}44.6 &
			\cellcolor[HTML]{F4E884}51.8 &
			\cellcolor[HTML]{C2DA81}11.3 &
			\cellcolor[HTML]{FCBD7B}49.2 &
			\cellcolor[HTML]{B5D680}12.4 &
			\cellcolor[HTML]{A8D27F}27.6 \\
			Web CapFilt &
			23,147 K &
			\cellcolor[HTML]{78C47D}116.0 &
			\cellcolor[HTML]{7FC77D}76.4 &
			\cellcolor[HTML]{A4D17F}83.3 &
			\cellcolor[HTML]{F7E984}11.6 &
			\cellcolor[HTML]{F6E984}53.7 &
			\cellcolor[HTML]{E7E483}16.4 &
			\cellcolor[HTML]{EFE784}47.3 &
			\cellcolor[HTML]{DDE182}19.1 &
			\cellcolor[HTML]{F2E884}48.5 &
			\cellcolor[HTML]{FEEA83}43.7 &
			\cellcolor[HTML]{F4E884}9.6 &
			\cellcolor[HTML]{CADC81}40.5 &
			\cellcolor[HTML]{F8696B}36.4 &
			\cellcolor[HTML]{F8696B}48.2 &
			\cellcolor[HTML]{FEEB84}2.0 &
			\cellcolor[HTML]{FDD17F}32.2 &
			\cellcolor[HTML]{F9816F}12.1 &
			\cellcolor[HTML]{FEDB81}9.3 &
			\cellcolor[HTML]{F0E784}54.9 &
			\cellcolor[HTML]{FCC27C}4.7 &
			\cellcolor[HTML]{F98670}27.0 &
			\cellcolor[HTML]{EFE784}11.7 &
			\cellcolor[HTML]{FEEB84}1.9 &
			\cellcolor[HTML]{F8696B}16.5 &
			\cellcolor[HTML]{FBEA84}50.8 &
			\cellcolor[HTML]{FDD680}9.4 &
			\cellcolor[HTML]{F8696B}47.6 &
			\cellcolor[HTML]{84C87D}14.7 &
			\cellcolor[HTML]{F2E884}13.9 \\
			TextCaps &
			549 K &
			\cellcolor[HTML]{E9E583}71.5 &
			\cellcolor[HTML]{E8E583}50.7 &
			\cellcolor[HTML]{63BE7B}109.4 &
			\cellcolor[HTML]{FCB679}4.9 &
			\cellcolor[HTML]{FCB679}44.3 &
			\cellcolor[HTML]{FFEB84}8.5 &
			\cellcolor[HTML]{FDC87D}41.9 &
			\cellcolor[HTML]{FEEB84}8.2 &
			\cellcolor[HTML]{FBB279}39.0 &
			\cellcolor[HTML]{FEDD81}43.1 &
			\cellcolor[HTML]{FBA977}3.1 &
			\cellcolor[HTML]{FEDE81}35.1 &
			\cellcolor[HTML]{FA9673}38.8 &
			\cellcolor[HTML]{FAEA84}52.9 &
			\cellcolor[HTML]{F8706C}0.1 &
			\cellcolor[HTML]{DDE182}35.7 &
			\cellcolor[HTML]{8DCA7E}12.8 &
			\cellcolor[HTML]{FEEB84}10.9 &
			\cellcolor[HTML]{FDEB84}53.7 &
			\cellcolor[HTML]{FED980}6.0 &
			\cellcolor[HTML]{FDC87D}38.7 &
			\cellcolor[HTML]{CBDC81}16.5 &
			\cellcolor[HTML]{FCB379}1.0 &
			\cellcolor[HTML]{BAD780}44.7 &
			\cellcolor[HTML]{F8696B}50.0 &
			\cellcolor[HTML]{BCD881}11.4 &
			\cellcolor[HTML]{FCBF7B}49.2 &
			\cellcolor[HTML]{63BE7B}16.2 &
			\cellcolor[HTML]{B8D780}24.7 \\
			VQAv2 &
			444 K &
			\cellcolor[HTML]{FEE683}61.0 &
			\cellcolor[HTML]{FDD780}40.0 &
			\cellcolor[HTML]{FDCF7E}39.0 &
			\cellcolor[HTML]{63BE7B}67.5 &
			\cellcolor[HTML]{63BE7B}65.3 &
			\cellcolor[HTML]{73C37C}54.2 &
			\cellcolor[HTML]{79C57D}61.8 &
			\cellcolor[HTML]{69C07C}56.7 &
			\cellcolor[HTML]{8BCA7E}66.8 &
			\cellcolor[HTML]{C0D981}58.1 &
			\cellcolor[HTML]{78C47D}44.0 &
			\cellcolor[HTML]{A4D17F}43.3 &
			\cellcolor[HTML]{E9E583}45.8 &
			\cellcolor[HTML]{80C77D}62.9 &
			\cellcolor[HTML]{63BE7B}39.7 &
			\cellcolor[HTML]{84C87D}38.6 &
			\cellcolor[HTML]{A0D07F}12.7 &
			\cellcolor[HTML]{7DC67D}37.6 &
			\cellcolor[HTML]{65BF7C}68.3 &
			\cellcolor[HTML]{99CE7F}42.0 &
			\cellcolor[HTML]{EDE683}48.2 &
			\cellcolor[HTML]{F86C6B}0.3 &
			\cellcolor[HTML]{65BF7C}12.4 &
			\cellcolor[HTML]{FA9874}24.5 &
			\cellcolor[HTML]{DDE283}54.8 &
			\cellcolor[HTML]{F98C71}4.4 &
			\cellcolor[HTML]{C2DA81}51.3 &
			\cellcolor[HTML]{FBA576}5.0 &
			\cellcolor[HTML]{F8706C}2.1 \\
			OK-VQA &
			9 K &
			\cellcolor[HTML]{FDD57F}54.0 &
			\cellcolor[HTML]{FCBC7B}33.2 &
			\cellcolor[HTML]{FCB97A}33.8 &
			\cellcolor[HTML]{7DC67D}57.7 &
			\cellcolor[HTML]{BDD881}58.2 &
			\cellcolor[HTML]{63BE7B}59.4 &
			\cellcolor[HTML]{70C27C}63.0 &
			\cellcolor[HTML]{87C97E}46.9 &
			\cellcolor[HTML]{ACD380}60.9 &
			\cellcolor[HTML]{C6DB81}56.8 &
			\cellcolor[HTML]{8DCA7E}38.2 &
			\cellcolor[HTML]{CCDD82}40.3 &
			\cellcolor[HTML]{F0E784}44.9 &
			\cellcolor[HTML]{FA9874}49.8 &
			\cellcolor[HTML]{86C87D}31.3 &
			\cellcolor[HTML]{A4D17F}37.6 &
			\cellcolor[HTML]{FDCC7E}12.3 &
			\cellcolor[HTML]{97CD7E}32.3 &
			\cellcolor[HTML]{A1D07F}62.5 &
			\cellcolor[HTML]{A3D17F}38.3 &
			\cellcolor[HTML]{FBA676}32.7 &
			\cellcolor[HTML]{F86A6B}0.2 &
			\cellcolor[HTML]{6BC17C}12.0 &
			\cellcolor[HTML]{FCB379}29.1 &
			\cellcolor[HTML]{F8696B}50.0 &
			\cellcolor[HTML]{F8726C}2.6 &
			\cellcolor[HTML]{A6D27F}51.9 &
			\cellcolor[HTML]{FA9B74}4.4 &
			\cellcolor[HTML]{F8696B}1.6 \\
			A-OKVQA &
			17 K &
			\cellcolor[HTML]{FDCD7E}50.6 &
			\cellcolor[HTML]{FCBF7B}34.1 &
			\cellcolor[HTML]{FAA075}27.6 &
			\cellcolor[HTML]{71C27C}62.2 &
			\cellcolor[HTML]{7BC57D}63.4 &
			\cellcolor[HTML]{71C27C}54.9 &
			\cellcolor[HTML]{63BE7B}64.5 &
			\cellcolor[HTML]{63BE7B}58.3 &
			\cellcolor[HTML]{8FCB7E}66.0 &
			\cellcolor[HTML]{C2DA81}57.7 &
			\cellcolor[HTML]{80C77D}41.9 &
			\cellcolor[HTML]{B7D780}41.9 &
			\cellcolor[HTML]{DEE283}47.0 &
			\cellcolor[HTML]{FA8E72}49.4 &
			\cellcolor[HTML]{6BC17C}38.0 &
			\cellcolor[HTML]{63BE7B}39.7 &
			\cellcolor[HTML]{FEE081}12.4 &
			\cellcolor[HTML]{8FCB7E}33.9 &
			\cellcolor[HTML]{77C47D}66.5 &
			\cellcolor[HTML]{9FD07F}39.8 &
			\cellcolor[HTML]{E3E383}50.3 &
			\cellcolor[HTML]{F86A6B}0.2 &
			\cellcolor[HTML]{84C87D}10.3 &
			\cellcolor[HTML]{F4E884}39.5 &
			\cellcolor[HTML]{F8696B}50.0 &
			\cellcolor[HTML]{F8696B}2.0 &
			\cellcolor[HTML]{63BE7B}53.2 &
			\cellcolor[HTML]{FA9C74}4.5 &
			\cellcolor[HTML]{F86C6B}1.8 \\
			A-OKVQA (MC) &
			17 K &
			\cellcolor[HTML]{BDD881}88.9 &
			\cellcolor[HTML]{C7DB81}58.7 &
			\cellcolor[HTML]{F5E884}50.1 &
			\cellcolor[HTML]{81C77D}56.4 &
			\cellcolor[HTML]{FAEA84}53.3 &
			\cellcolor[HTML]{93CC7E}43.8 &
			\cellcolor[HTML]{77C47D}62.0 &
			\cellcolor[HTML]{8ECB7E}44.7 &
			\cellcolor[HTML]{63BE7B}73.7 &
			\cellcolor[HTML]{ACD380}62.8 &
			\cellcolor[HTML]{A1D07F}32.5 &
			\cellcolor[HTML]{FDC87D}32.8 &
			\cellcolor[HTML]{E0E283}46.8 &
			\cellcolor[HTML]{B1D580}58.9 &
			\cellcolor[HTML]{A9D380}22.7 &
			\cellcolor[HTML]{97CD7E}38.0 &
			\cellcolor[HTML]{FEE482}12.4 &
			\cellcolor[HTML]{9FD07F}30.6 &
			\cellcolor[HTML]{63BE7B}68.4 &
			\cellcolor[HTML]{BCD881}29.9 &
			\cellcolor[HTML]{8FCB7E}66.7 &
			\cellcolor[HTML]{FA9B74}3.8 &
			\cellcolor[HTML]{7AC57D}11.0 &
			\cellcolor[HTML]{ACD380}46.0 &
			\cellcolor[HTML]{FFEB84}50.2 &
			\cellcolor[HTML]{99CE7F}11.7 &
			\cellcolor[HTML]{ABD380}51.8 &
			\cellcolor[HTML]{EFE784}9.7 &
			\cellcolor[HTML]{F6E984}13.2 \\
			ScienceQA &
			6 K &
			\cellcolor[HTML]{EFE784}69.4 &
			\cellcolor[HTML]{ECE683}49.8 &
			\cellcolor[HTML]{ECE683}53.9 &
			\cellcolor[HTML]{C3DA81}31.1 &
			\cellcolor[HTML]{C4DA81}57.6 &
			\cellcolor[HTML]{BCD881}30.5 &
			\cellcolor[HTML]{B0D480}55.1 &
			\cellcolor[HTML]{C0D981}28.5 &
			\cellcolor[HTML]{7FC67D}68.9 &
			\cellcolor[HTML]{63BE7B}79.3 &
			\cellcolor[HTML]{B6D680}26.9 &
			\cellcolor[HTML]{F1E784}37.5 &
			\cellcolor[HTML]{F2E884}44.8 &
			\cellcolor[HTML]{B2D580}58.8 &
			\cellcolor[HTML]{B1D580}20.9 &
			\cellcolor[HTML]{BCD881}36.8 &
			\cellcolor[HTML]{87C97E}12.8 &
			\cellcolor[HTML]{AED480}27.4 &
			\cellcolor[HTML]{92CC7E}64.0 &
			\cellcolor[HTML]{B5D680}32.4 &
			\cellcolor[HTML]{97CD7E}65.3 &
			\cellcolor[HTML]{FEDB81}8.5 &
			\cellcolor[HTML]{BED981}6.3 &
			\cellcolor[HTML]{FBAF78}28.4 &
			\cellcolor[HTML]{F8696B}50.0 &
			\cellcolor[HTML]{9ED07F}11.7 &
			\cellcolor[HTML]{E8E583}50.5 &
			\cellcolor[HTML]{B2D580}12.5 &
			\cellcolor[HTML]{FA9473}4.8 \\
			GQA &
			943 K &
			\cellcolor[HTML]{FDD680}54.5 &
			\cellcolor[HTML]{FDC67C}35.7 &
			\cellcolor[HTML]{FA9172}24.0 &
			\cellcolor[HTML]{7AC57D}58.8 &
			\cellcolor[HTML]{7FC67D}63.1 &
			\cellcolor[HTML]{9ACE7F}41.5 &
			\cellcolor[HTML]{73C37C}62.6 &
			\cellcolor[HTML]{8ECB7E}44.7 &
			\cellcolor[HTML]{98CE7F}64.5 &
			\cellcolor[HTML]{CFDE82}54.7 &
			\cellcolor[HTML]{63BE7B}49.7 &
			\cellcolor[HTML]{63BE7B}48.2 &
			\cellcolor[HTML]{F7E984}44.2 &
			\cellcolor[HTML]{F3E884}53.5 &
			\cellcolor[HTML]{7DC67D}33.4 &
			\cellcolor[HTML]{BCD881}36.8 &
			\cellcolor[HTML]{63BE7B}13.0 &
			\cellcolor[HTML]{CFDD82}20.7 &
			\cellcolor[HTML]{DCE182}56.8 &
			\cellcolor[HTML]{B5D680}32.2 &
			\cellcolor[HTML]{FEE382}43.3 &
			\cellcolor[HTML]{F8696B}0.0 &
			\cellcolor[HTML]{AAD380}7.7 &
			\cellcolor[HTML]{FDC67D}32.3 &
			\cellcolor[HTML]{F8696B}50.0 &
			\cellcolor[HTML]{F87B6E}3.2 &
			\cellcolor[HTML]{7DC67D}52.7 &
			\cellcolor[HTML]{F8766D}2.3 &
			\cellcolor[HTML]{F8696B}1.5 \\
			IconQA &
			19 K &
			\cellcolor[HTML]{FEE482}60.2 &
			\cellcolor[HTML]{FEE482}43.4 &
			\cellcolor[HTML]{FEEB84}46.4 &
			\cellcolor[HTML]{98CE7F}47.4 &
			\cellcolor[HTML]{CFDD82}56.8 &
			\cellcolor[HTML]{B8D780}31.7 &
			\cellcolor[HTML]{D2DE82}50.8 &
			\cellcolor[HTML]{B7D780}31.2 &
			\cellcolor[HTML]{C3DA81}56.9 &
			\cellcolor[HTML]{C4DA81}57.4 &
			\cellcolor[HTML]{9BCF7F}34.3 &
			\cellcolor[HTML]{FEE482}35.7 &
			\cellcolor[HTML]{63BE7B}61.0 &
			\cellcolor[HTML]{BAD780}58.2 &
			\cellcolor[HTML]{77C47D}35.0 &
			\cellcolor[HTML]{FDD57F}32.6 &
			\cellcolor[HTML]{6FC27C}12.9 &
			\cellcolor[HTML]{9BCE7F}31.4 &
			\cellcolor[HTML]{FDD37F}50.4 &
			\cellcolor[HTML]{A0D07F}39.5 &
			\cellcolor[HTML]{FEE983}44.4 &
			\cellcolor[HTML]{FEE582}9.3 &
			\cellcolor[HTML]{68C07C}12.2 &
			\cellcolor[HTML]{F8796E}19.3 &
			\cellcolor[HTML]{E3E383}54.0 &
			\cellcolor[HTML]{BFD981}11.4 &
			\cellcolor[HTML]{E5E483}50.6 &
			\cellcolor[HTML]{FAEA84}9.1 &
			\cellcolor[HTML]{FEE883}11.2 \\
			VSR &
			3 K &
			\cellcolor[HTML]{FEDB81}56.5 &
			\cellcolor[HTML]{FEE382}43.0 &
			\cellcolor[HTML]{FEE883}45.2 &
			\cellcolor[HTML]{ACD380}39.9 &
			\cellcolor[HTML]{B3D680}58.9 &
			\cellcolor[HTML]{E5E483}17.1 &
			\cellcolor[HTML]{D2DE82}50.9 &
			\cellcolor[HTML]{EAE583}14.8 &
			\cellcolor[HTML]{D6DF82}53.5 &
			\cellcolor[HTML]{E3E383}50.2 &
			\cellcolor[HTML]{AED480}28.9 &
			\cellcolor[HTML]{C7DB81}40.6 &
			\cellcolor[HTML]{F3E884}44.7 &
			\cellcolor[HTML]{63BE7B}65.2 &
			\cellcolor[HTML]{CFDD82}13.5 &
			\cellcolor[HTML]{FEDF81}33.5 &
			\cellcolor[HTML]{FDD37F}12.3 &
			\cellcolor[HTML]{C8DB81}22.1 &
			\cellcolor[HTML]{D8E082}57.2 &
			\cellcolor[HTML]{D3DF82}22.0 &
			\cellcolor[HTML]{74C37C}72.0 &
			\cellcolor[HTML]{D9E082}14.6 &
			\cellcolor[HTML]{EAE583}3.3 &
			\cellcolor[HTML]{F7E984}39.2 &
			\cellcolor[HTML]{FEEB84}50.4 &
			\cellcolor[HTML]{FBEA84}10.8 &
			\cellcolor[HTML]{C2DA81}51.3 &
			\cellcolor[HTML]{FEE683}8.6 &
			\cellcolor[HTML]{FEEB84}11.6 \\
			TextVQA &
			35 K &
			\cellcolor[HTML]{B5D680}92.1 &
			\cellcolor[HTML]{B3D580}63.8 &
			\cellcolor[HTML]{EAE583}54.5 &
			\cellcolor[HTML]{94CD7E}48.9 &
			\cellcolor[HTML]{E4E483}55.1 &
			\cellcolor[HTML]{A6D27F}37.7 &
			\cellcolor[HTML]{A7D27F}56.2 &
			\cellcolor[HTML]{A4D17F}37.5 &
			\cellcolor[HTML]{BED981}57.7 &
			\cellcolor[HTML]{DCE182}51.9 &
			\cellcolor[HTML]{A3D17F}32.1 &
			\cellcolor[HTML]{DDE283}39.0 &
			\cellcolor[HTML]{F9EA84}43.9 &
			\cellcolor[HTML]{C1D981}57.6 &
			\cellcolor[HTML]{86C87D}31.4 &
			\cellcolor[HTML]{E5E483}35.4 &
			\cellcolor[HTML]{FBA877}12.2 &
			\cellcolor[HTML]{63BE7B}42.8 &
			\cellcolor[HTML]{C1DA81}59.4 &
			\cellcolor[HTML]{99CE7F}41.9 &
			\cellcolor[HTML]{FA9473}29.5 &
			\cellcolor[HTML]{F8766D}1.1 &
			\cellcolor[HTML]{63BE7B}12.5 &
			\cellcolor[HTML]{FCC17C}31.5 &
			\cellcolor[HTML]{D9E082}55.4 &
			\cellcolor[HTML]{F8776D}3.0 &
			\cellcolor[HTML]{F98D71}48.2 &
			\cellcolor[HTML]{F87B6E}2.6 &
			\cellcolor[HTML]{F8716C}2.2 \\
			OCR-VQA &
			802 K &
			\cellcolor[HTML]{A2D07F}99.6 &
			\cellcolor[HTML]{9DCF7F}69.2 &
			\cellcolor[HTML]{DDE283}59.8 &
			\cellcolor[HTML]{8CCA7E}52.0 &
			\cellcolor[HTML]{A8D27F}59.8 &
			\cellcolor[HTML]{B2D580}33.6 &
			\cellcolor[HTML]{CFDE82}51.2 &
			\cellcolor[HTML]{9ECF7F}39.4 &
			\cellcolor[HTML]{C0D981}57.3 &
			\cellcolor[HTML]{D1DE82}54.3 &
			\cellcolor[HTML]{92CC7E}36.9 &
			\cellcolor[HTML]{C0D981}41.2 &
			\cellcolor[HTML]{FEDD81}42.5 &
			\cellcolor[HTML]{ECE683}54.1 &
			\cellcolor[HTML]{68C07C}38.8 &
			\cellcolor[HTML]{EEE683}35.1 &
			\cellcolor[HTML]{83C87D}12.8 &
			\cellcolor[HTML]{A3D17F}29.7 &
			\cellcolor[HTML]{FDC57C}48.6 &
			\cellcolor[HTML]{63BE7B}60.1 &
			\cellcolor[HTML]{63BE7B}75.4 &
			\cellcolor[HTML]{F98570}2.1 &
			\cellcolor[HTML]{6CC17C}11.9 &
			\cellcolor[HTML]{C3DA81}43.8 &
			\cellcolor[HTML]{ABD380}61.6 &
			\cellcolor[HTML]{CADC81}11.3 &
			\cellcolor[HTML]{9BCE7F}52.1 &
			\cellcolor[HTML]{F8696B}1.6 &
			\cellcolor[HTML]{F8726C}2.2 \\
			OpenCQA &
			6 K &
			\cellcolor[HTML]{FBEA84}64.7 &
			\cellcolor[HTML]{F9EA84}46.5 &
			\cellcolor[HTML]{DCE182}60.2 &
			\cellcolor[HTML]{FBA776}3.9 &
			\cellcolor[HTML]{FEDA80}50.3 &
			\cellcolor[HTML]{FEE983}8.3 &
			\cellcolor[HTML]{F7E984}46.3 &
			\cellcolor[HTML]{FCBD7B}5.1 &
			\cellcolor[HTML]{FDD57F}43.4 &
			\cellcolor[HTML]{D8E082}52.8 &
			\cellcolor[HTML]{FA8E72}1.8 &
			\cellcolor[HTML]{FFEB84}36.4 &
			\cellcolor[HTML]{F2E884}44.8 &
			\cellcolor[HTML]{FBA676}50.2 &
			\cellcolor[HTML]{F8736D}0.1 &
			\cellcolor[HTML]{FEDA80}33.0 &
			\cellcolor[HTML]{F8696B}12.0 &
			\cellcolor[HTML]{EFE784}14.0 &
			\cellcolor[HTML]{FEE983}53.2 &
			\cellcolor[HTML]{FDD880}5.9 &
			\cellcolor[HTML]{BCD881}58.0 &
			\cellcolor[HTML]{63BE7B}29.7 &
			\cellcolor[HTML]{FEDB80}1.6 &
			\cellcolor[HTML]{DFE283}41.4 &
			\cellcolor[HTML]{D9E082}55.4 &
			\cellcolor[HTML]{63BE7B}12.2 &
			\cellcolor[HTML]{F9EA84}50.2 &
			\cellcolor[HTML]{83C87D}14.7 &
			\cellcolor[HTML]{A9D27F}27.5 \\
			HM &
			9 K &
			\cellcolor[HTML]{FDC87D}48.6 &
			\cellcolor[HTML]{FCB97A}32.5 &
			\cellcolor[HTML]{FCEA84}47.2 &
			\cellcolor[HTML]{C3DA81}31.3 &
			\cellcolor[HTML]{FEEA83}52.9 &
			\cellcolor[HTML]{FA9573}2.9 &
			\cellcolor[HTML]{FDCC7E}42.3 &
			\cellcolor[HTML]{FA9673}2.7 &
			\cellcolor[HTML]{FDC97D}41.8 &
			\cellcolor[HTML]{FEDA80}42.9 &
			\cellcolor[HTML]{C9DC81}21.4 &
			\cellcolor[HTML]{F8E984}37.0 &
			\cellcolor[HTML]{EEE683}45.2 &
			\cellcolor[HTML]{F87A6E}48.8 &
			\cellcolor[HTML]{B1D580}20.8 &
			\cellcolor[HTML]{FDD680}32.7 &
			\cellcolor[HTML]{EBE683}12.5 &
			\cellcolor[HTML]{FEE683}10.1 &
			\cellcolor[HTML]{FBA175}43.9 &
			\cellcolor[HTML]{AED480}34.7 &
			\cellcolor[HTML]{FEE482}43.5 &
			\cellcolor[HTML]{CFDE82}15.9 &
			\cellcolor[HTML]{F2E884}2.8 &
			\cellcolor[HTML]{FDCC7E}33.3 &
			\cellcolor[HTML]{63BE7B}71.2 &
			\cellcolor[HTML]{E0E283}11.1 &
			\cellcolor[HTML]{FEE382}49.9 &
			\cellcolor[HTML]{E3E383}10.2 &
			\cellcolor[HTML]{C0D981}23.2 \\
			\vlbenchmark{} &
			7 K &
			\cellcolor[HTML]{F8766D}14.9 &
			\cellcolor[HTML]{F8736D}14.8 &
			\cellcolor[HTML]{F98D72}23.1 &
			\cellcolor[HTML]{F8696B}0.0 &
			\cellcolor[HTML]{FDC97D}47.4 &
			\cellcolor[HTML]{F8696B}0.0 &
			\cellcolor[HTML]{FDC97D}42.1 &
			\cellcolor[HTML]{F8696B}0.0 &
			\cellcolor[HTML]{FBAE78}38.4 &
			\cellcolor[HTML]{FBB179}40.9 &
			\cellcolor[HTML]{F8696B}0.0 &
			\cellcolor[HTML]{FEEA83}36.3 &
			\cellcolor[HTML]{FEDB81}42.4 &
			\cellcolor[HTML]{D5DF82}56.0 &
			\cellcolor[HTML]{F8696B}0.0 &
			\cellcolor[HTML]{F86E6C}23.2 &
			\cellcolor[HTML]{FCB97A}12.2 &
			\cellcolor[HTML]{F86A6B}0.1 &
			\cellcolor[HTML]{FA9F75}43.6 &
			\cellcolor[HTML]{F86A6B}0.1 &
			\cellcolor[HTML]{DEE283}51.2 &
			\cellcolor[HTML]{CFDE82}15.8 &
			\cellcolor[HTML]{F8696B}0.0 &
			\cellcolor[HTML]{FEE482}37.4 &
			\cellcolor[HTML]{F8696B}50.0 &
			\cellcolor[HTML]{FEEA83}10.7 &
			\cellcolor[HTML]{FBA977}48.8 &
			\cellcolor[HTML]{FEE683}8.6 &
			\cellcolor[HTML]{63BE7B}40.3 \\
			LLaVA Conversation &
			57 K &
			\cellcolor[HTML]{FDD680}54.4 &
			\cellcolor[HTML]{FDCC7E}37.1 &
			\cellcolor[HTML]{FDD780}41.1 &
			\cellcolor[HTML]{F8696B}0.0 &
			\cellcolor[HTML]{FDCB7E}47.8 &
			\cellcolor[HTML]{F8696B}0.0 &
			\cellcolor[HTML]{FA9C74}37.8 &
			\cellcolor[HTML]{F8696B}0.0 &
			\cellcolor[HTML]{F98B71}34.0 &
			\cellcolor[HTML]{FCB87A}41.2 &
			\cellcolor[HTML]{F8696B}0.0 &
			\cellcolor[HTML]{FDCF7E}33.6 &
			\cellcolor[HTML]{FBAB77}39.9 &
			\cellcolor[HTML]{F98670}49.2 &
			\cellcolor[HTML]{F8696B}0.0 &
			\cellcolor[HTML]{F8696B}22.7 &
			\cellcolor[HTML]{FFEB84}12.4 &
			\cellcolor[HTML]{F8696B}0.0 &
			\cellcolor[HTML]{FBAD78}45.5 &
			\cellcolor[HTML]{F8696B}0.0 &
			\cellcolor[HTML]{E5E483}49.9 &
			\cellcolor[HTML]{D4DF82}15.2 &
			\cellcolor[HTML]{F8696B}0.0 &
			\cellcolor[HTML]{F2E884}39.6 &
			\cellcolor[HTML]{F8696B}50.0 &
			\cellcolor[HTML]{FEE482}10.3 &
			\cellcolor[HTML]{FDD17F}49.5 &
			\cellcolor[HTML]{FDD47F}7.6 &
			\cellcolor[HTML]{FA9874}5.2 \\
			LLaVA Reasoning &
			77 K &
			\cellcolor[HTML]{F86E6C}11.5 &
			\cellcolor[HTML]{F86F6C}13.7 &
			\cellcolor[HTML]{F8746D}16.9 &
			\cellcolor[HTML]{F8696B}0.0 &
			\cellcolor[HTML]{FBB178}43.4 &
			\cellcolor[HTML]{F8696B}0.0 &
			\cellcolor[HTML]{F8696B}32.9 &
			\cellcolor[HTML]{F8696B}0.0 &
			\cellcolor[HTML]{F8716C}30.7 &
			\cellcolor[HTML]{F8696B}37.3 &
			\cellcolor[HTML]{F8696B}0.0 &
			\cellcolor[HTML]{FDC97D}32.9 &
			\cellcolor[HTML]{FA9E75}39.2 &
			\cellcolor[HTML]{DDE283}55.3 &
			\cellcolor[HTML]{F8696B}0.0 &
			\cellcolor[HTML]{FDCE7E}31.9 &
			\cellcolor[HTML]{FBEA84}12.4 &
			\cellcolor[HTML]{F8696B}0.0 &
			\cellcolor[HTML]{F8696B}36.6 &
			\cellcolor[HTML]{F8696B}0.0 &
			\cellcolor[HTML]{FEEB84}44.9 &
			\cellcolor[HTML]{FBAD78}5.1 &
			\cellcolor[HTML]{F8696B}0.0 &
			\cellcolor[HTML]{FEE683}37.6 &
			\cellcolor[HTML]{F8696B}50.0 &
			\cellcolor[HTML]{FEDA80}9.7 &
			\cellcolor[HTML]{FDC97D}49.4 &
			\cellcolor[HTML]{FCC27C}6.6 &
			\cellcolor[HTML]{F2E784}14.0 \\
			LLaVA Description &
			23 K &
			\cellcolor[HTML]{F8696B}9.2 &
			\cellcolor[HTML]{F8696B}12.0 &
			\cellcolor[HTML]{F8696B}14.2 &
			\cellcolor[HTML]{F8696B}0.0 &
			\cellcolor[HTML]{FBA977}42.1 &
			\cellcolor[HTML]{F8696B}0.0 &
			\cellcolor[HTML]{FA9D75}37.9 &
			\cellcolor[HTML]{F8696B}0.0 &
			\cellcolor[HTML]{F8696B}29.6 &
			\cellcolor[HTML]{F8746D}37.8 &
			\cellcolor[HTML]{F8696B}0.0 &
			\cellcolor[HTML]{FDC77D}32.7 &
			\cellcolor[HTML]{F98370}37.8 &
			\cellcolor[HTML]{FDCC7E}51.5 &
			\cellcolor[HTML]{F8696B}0.0 &
			\cellcolor[HTML]{FCBB7A}30.2 &
			\cellcolor[HTML]{FBAF78}12.2 &
			\cellcolor[HTML]{F8696B}0.0 &
			\cellcolor[HTML]{F8736D}38.0 &
			\cellcolor[HTML]{F86A6B}0.1 &
			\cellcolor[HTML]{FCB679}35.5 &
			\cellcolor[HTML]{FBA676}4.6 &
			\cellcolor[HTML]{F8696B}0.0 &
			\cellcolor[HTML]{AFD480}45.7 &
			\cellcolor[HTML]{F8696B}50.0 &
			\cellcolor[HTML]{FEE883}10.6 &
			\cellcolor[HTML]{FBAB77}48.8 &
			\cellcolor[HTML]{FED980}7.9 &
			\cellcolor[HTML]{F98971}4.0 \\
			VQAv2 QG &
			444 K &
			\cellcolor[HTML]{A9D380}96.6 &
			\cellcolor[HTML]{AED480}64.9 &
			\cellcolor[HTML]{FEDF81}43.1 &
			\cellcolor[HTML]{FA9974}3.1 &
			\cellcolor[HTML]{FA9874}39.3 &
			\cellcolor[HTML]{F86E6C}0.4 &
			\cellcolor[HTML]{F8796E}34.4 &
			\cellcolor[HTML]{F86A6B}0.1 &
			\cellcolor[HTML]{FBA476}37.1 &
			\cellcolor[HTML]{F97F6F}38.4 &
			\cellcolor[HTML]{F98B71}1.7 &
			\cellcolor[HTML]{FCB679}31.0 &
			\cellcolor[HTML]{FBA276}39.5 &
			\cellcolor[HTML]{FDCC7E}51.5 &
			\cellcolor[HTML]{FDC67D}1.1 &
			\cellcolor[HTML]{A7D27F}37.5 &
			\cellcolor[HTML]{FFEB84}12.4 &
			\cellcolor[HTML]{F8776D}1.2 &
			\cellcolor[HTML]{FBB279}46.1 &
			\cellcolor[HTML]{FDEB84}7.8 &
			\cellcolor[HTML]{F98C71}28.2 &
			\cellcolor[HTML]{FCEB84}10.1 &
			\cellcolor[HTML]{FBA275}0.8 &
			\cellcolor[HTML]{EEE683}40.0 &
			\cellcolor[HTML]{F8696B}50.0 &
			\cellcolor[HTML]{D7E082}11.1 &
			\cellcolor[HTML]{D4DF82}50.9 &
			\cellcolor[HTML]{CBDC81}11.4 &
			\cellcolor[HTML]{DAE182}18.3 \\
			OK-VQA QG &
			9 K &
			\cellcolor[HTML]{AFD480}94.2 &
			\cellcolor[HTML]{B9D780}62.3 &
			\cellcolor[HTML]{FBEA84}47.8 &
			\cellcolor[HTML]{F8696B}0.1 &
			\cellcolor[HTML]{F8696B}31.5 &
			\cellcolor[HTML]{F8696B}0.1 &
			\cellcolor[HTML]{FDCC7E}42.3 &
			\cellcolor[HTML]{F8696B}0.0 &
			\cellcolor[HTML]{FA9E75}36.4 &
			\cellcolor[HTML]{F98971}38.9 &
			\cellcolor[HTML]{F8696B}0.0 &
			\cellcolor[HTML]{F8696B}22.9 &
			\cellcolor[HTML]{FCB479}40.4 &
			\cellcolor[HTML]{F8726C}48.5 &
			\cellcolor[HTML]{F8696B}0.0 &
			\cellcolor[HTML]{FED980}33.0 &
			\cellcolor[HTML]{FEE983}12.4 &
			\cellcolor[HTML]{F86F6C}0.5 &
			\cellcolor[HTML]{FA9C74}43.3 &
			\cellcolor[HTML]{F97C6E}1.0 &
			\cellcolor[HTML]{F8696B}21.9 &
			\cellcolor[HTML]{EEE683}11.9 &
			\cellcolor[HTML]{F97F6F}0.3 &
			\cellcolor[HTML]{FDC67C}32.2 &
			\cellcolor[HTML]{F7E984}51.4 &
			\cellcolor[HTML]{FA9172}4.7 &
			\cellcolor[HTML]{FCB379}49.0 &
			\cellcolor[HTML]{E0E383}10.4 &
			\cellcolor[HTML]{FA9473}4.8 \\
			A-OKVQA QG &
			17 K &
			\cellcolor[HTML]{EBE583}71.1 &
			\cellcolor[HTML]{ECE683}49.6 &
			\cellcolor[HTML]{FEE182}43.5 &
			\cellcolor[HTML]{FA9874}3.0 &
			\cellcolor[HTML]{F98B71}37.3 &
			\cellcolor[HTML]{F8696B}0.0 &
			\cellcolor[HTML]{FEDD81}43.9 &
			\cellcolor[HTML]{F86A6B}0.1 &
			\cellcolor[HTML]{FBA175}36.8 &
			\cellcolor[HTML]{FCB679}41.2 &
			\cellcolor[HTML]{F8736D}0.5 &
			\cellcolor[HTML]{FCB87A}31.1 &
			\cellcolor[HTML]{FBA777}39.7 &
			\cellcolor[HTML]{FDD07E}51.6 &
			\cellcolor[HTML]{F8776D}0.2 &
			\cellcolor[HTML]{81C77D}38.7 &
			\cellcolor[HTML]{FCB479}12.2 &
			\cellcolor[HTML]{F8766D}1.1 &
			\cellcolor[HTML]{FCB87A}46.9 &
			\cellcolor[HTML]{FDCF7E}5.4 &
			\cellcolor[HTML]{FDCC7E}39.3 &
			\cellcolor[HTML]{F4E884}11.1 &
			\cellcolor[HTML]{F8746D}0.2 &
			\cellcolor[HTML]{FDD57F}34.8 &
			\cellcolor[HTML]{E6E483}53.6 &
			\cellcolor[HTML]{FDD27F}9.1 &
			\cellcolor[HTML]{FCC57C}49.3 &
			\cellcolor[HTML]{CEDD82}11.2 &
			\cellcolor[HTML]{E9E583}15.6
			\\
			\bottomrule
		\end{tabular}%
	}
	\caption
	{Unnormalized transfer learning performance of mPLUG-Owl. Higher values indicate better performance. QG denotes question generation, MC denotes multiple-choice and G denotes open-ended generation. The color scale is normalized along each column. The colors represent values in descending order: green, yellow, orange and red.}
	\label{eval:affinity_matrix_raw_mplugowl_landscape}
\end{table}
\end{landscape}

\begin{landscape}
\begin{table}[!tp]
	\resizebox{\columnwidth}{!}{%
		\begin{tabular}{c|c|c|ccccccccccccccccccccccccccccc}
			\toprule
			Source   Task &
			\setlength\extrarowheight{0pt}\begin{tabular}[c]{@{}c@{}}Dataset\\ Size \end{tabular} &
			\setlength\extrarowheight{0pt}\begin{tabular}[c]{@{}c@{}}AHP\\ Ranking \\ Score \end{tabular}   &
			\multicolumn{29}{c}{Target Task} \\
			&
			&
			&
			\setlength\extrarowheight{0pt}\begin{tabular}[c]{@{}c@{}}COCO\\ Caption \end{tabular}  &
			\setlength\extrarowheight{0pt}\begin{tabular}[c]{@{}c@{}}Flickr\\ 30k\end{tabular}  &
			\setlength\extrarowheight{0pt}\begin{tabular}[c]{@{}c@{}}Text\\ Caps\end{tabular}  &
			\multicolumn{2}{c}{VQAv2} &
			\multicolumn{2}{c}{OK-VQA} &
			\multicolumn{2}{c}{A-OKVQA} &
			\setlength\extrarowheight{0pt}\begin{tabular}[c]{@{}c@{}}Science\\ QA\end{tabular} &
			\multicolumn{2}{c}{GQA} &
			\setlength\extrarowheight{0pt}\begin{tabular}[c]{@{}c@{}}Icon\\ QA \end{tabular} &
			VSR &
			\multicolumn{2}{c}{CLEVR} &
			\setlength\extrarowheight{0pt}\begin{tabular}[c]{@{}c@{}}RAVEN-\\ FAIR \end{tabular} &
			\multicolumn{2}{c}{\setlength\extrarowheight{0pt}\begin{tabular}[c]{@{}c@{}}Text\\ VQA\end{tabular}} &
			\multicolumn{2}{c}{\setlength\extrarowheight{0pt}\begin{tabular}[c]{@{}c@{}}OCR-\\ VQA\end{tabular}} &
			\setlength\extrarowheight{0pt}\begin{tabular}[c]{@{}c@{}}Open\\ CQA \end{tabular} &
			\multicolumn{2}{c}{\setlength\extrarowheight{0pt}\begin{tabular}[c]{@{}c@{}}Chart\\ QA\end{tabular}} &
			HM &
			\setlength\extrarowheight{0pt}\begin{tabular}[c]{@{}c@{}}NY\\ Explain \end{tabular} &
			\setlength\extrarowheight{0pt}\begin{tabular}[c]{@{}c@{}}NY\\ Rank \end{tabular} &
			MORE &
			\vlbenchmark \\
			 &
			&
			&
			&
			&
			&
			G &
			MC &
			G &
			MC &
			G &
			MC &
			&
			G &
			MC &
			&
			&
			G &
			MC &
			&
			G &
			MC &
			G &
			MC &
			&
			G &
			MC &
			&
			&
			&
			&
			\\
			\midrule
A-OKVQA (MC) &
  17 K &
  10.3 &
  \cellcolor[HTML]{9ECF7F}-10.6 &
  \cellcolor[HTML]{A3D17F}-0.6 &
  \cellcolor[HTML]{CDDD82}-0.6 &
  \cellcolor[HTML]{96CD7E}0.7 &
  \cellcolor[HTML]{C7DB81}2.2 &
  \cellcolor[HTML]{AED480}0.8 &
  \cellcolor[HTML]{D7E082}-1.7 &
  \cellcolor[HTML]{A5D17F}3.2 &
  \cellcolor[HTML]{63BE7B}10.0 &
  \cellcolor[HTML]{F2E884}0.0 &
  \cellcolor[HTML]{C8DB81}-0.7 &
  \cellcolor[HTML]{EBE683}0.5 &
  \cellcolor[HTML]{FAEA84}0.1 &
  \cellcolor[HTML]{EBE683}-6.3 &
  \cellcolor[HTML]{71C27C}7.1 &
  \cellcolor[HTML]{E5E483}0.9 &
  \cellcolor[HTML]{FCB379}-0.8 &
  \cellcolor[HTML]{ADD480}0.4 &
  \cellcolor[HTML]{D0DE82}0.1 &
  \cellcolor[HTML]{E8E583}0.2 &
  \cellcolor[HTML]{63BE7B}10.0 &
  \cellcolor[HTML]{F7E984}-0.2 &
  \cellcolor[HTML]{84C87D}2.8 &
  \cellcolor[HTML]{63BE7B}10.0 &
  \cellcolor[HTML]{EBE683}1.2 &
  \cellcolor[HTML]{EDE683}-0.4 &
  \cellcolor[HTML]{93CC7E}5.9 &
  \cellcolor[HTML]{AED480}-28.7 &
  \cellcolor[HTML]{FCBA7A}-0.3 \\
VQAv2 &
  444 K &
  9.2 &
  \cellcolor[HTML]{FFEB84}-44.8 &
  \cellcolor[HTML]{FFEB84}-16.3 &
  \cellcolor[HTML]{FDD680}-7.5 &
  \cellcolor[HTML]{63BE7B}10.0 &
  \cellcolor[HTML]{63BE7B}10.0 &
  \cellcolor[HTML]{7DC67D}6.8 &
  \cellcolor[HTML]{9ACE7F}4.5 &
  \cellcolor[HTML]{6FC27C}8.8 &
  \cellcolor[HTML]{74C37C}8.6 &
  \cellcolor[HTML]{FAEA84}-0.6 &
  \cellcolor[HTML]{A1D07F}3.5 &
  \cellcolor[HTML]{C2DA81}3.4 &
  \cellcolor[HTML]{D3DF82}2.6 &
  \cellcolor[HTML]{63BE7B}10.0 &
  \cellcolor[HTML]{63BE7B}10.0 &
  \cellcolor[HTML]{A4D17F}5.5 &
  \cellcolor[HTML]{FFEB84}1.9 &
  \cellcolor[HTML]{88C97E}5.2 &
  \cellcolor[HTML]{94CD7E}5.6 &
  \cellcolor[HTML]{E3E383}0.5 &
  \cellcolor[HTML]{C1D981}-11.2 &
  \cellcolor[HTML]{F98D72}-2.0 &
  \cellcolor[HTML]{74C37C}6.3 &
  \cellcolor[HTML]{DCE182}-34.1 &
  \cellcolor[HTML]{DBE182}2.2 &
  \cellcolor[HTML]{FCB479}-5.6 &
  \cellcolor[HTML]{79C57D}8.2 &
  \cellcolor[HTML]{FEDC81}-90.0 &
  \cellcolor[HTML]{F97B6E}-0.9 \\
Web CapFilt &
  23,147 K &
  8.0 &
  \cellcolor[HTML]{73C37C}4.4 &
  \cellcolor[HTML]{98CE7F}1.1 &
  \cellcolor[HTML]{B1D580}2.2 &
  \cellcolor[HTML]{A7D27F}-2.4 &
  \cellcolor[HTML]{DCE182}0.5 &
  \cellcolor[HTML]{C8DB81}-2.2 &
  \cellcolor[HTML]{E6E483}-3.2 &
  \cellcolor[HTML]{C3DA81}0.0 &
  \cellcolor[HTML]{BED881}2.1 &
  \cellcolor[HTML]{F9EA84}-0.5 &
  \cellcolor[HTML]{CEDD82}-1.4 &
  \cellcolor[HTML]{EFE784}0.3 &
  \cellcolor[HTML]{F6E984}0.3 &
  \cellcolor[HTML]{E3E383}-5.4 &
  \cellcolor[HTML]{B5D680}-7.5 &
  \cellcolor[HTML]{FDCC7E}-7.2 &
  \cellcolor[HTML]{FCC37C}0.0 &
  \cellcolor[HTML]{D6E082}-5.0 &
  \cellcolor[HTML]{DCE182}-1.0 &
  \cellcolor[HTML]{FFEB84}-1.6 &
  \cellcolor[HTML]{A9D380}-5.8 &
  \cellcolor[HTML]{E7E483}0.9 &
  \cellcolor[HTML]{B5D680}-7.9 &
  \cellcolor[HTML]{C8DB81}-26.7 &
  \cellcolor[HTML]{EFE784}0.9 &
  \cellcolor[HTML]{D9E082}1.1 &
  \cellcolor[HTML]{FEE182}-5.9 &
  \cellcolor[HTML]{6BC17C}6.1 &
  \cellcolor[HTML]{F0E784}1.2 \\
Flickr30k &
  145 K &
  6.2 &
  \cellcolor[HTML]{A6D27F}-13.2 &
  \cellcolor[HTML]{63BE7B}10.0 &
  \cellcolor[HTML]{AED480}2.5 &
  \cellcolor[HTML]{B7D780}-5.4 &
  \cellcolor[HTML]{D8E082}0.8 &
  \cellcolor[HTML]{E1E383}-5.3 &
  \cellcolor[HTML]{E4E483}-2.9 &
  \cellcolor[HTML]{CBDC81}-0.9 &
  \cellcolor[HTML]{A7D27F}4.1 &
  \cellcolor[HTML]{FCEB84}-0.7 &
  \cellcolor[HTML]{DFE283}-3.2 &
  \cellcolor[HTML]{F7E984}-0.3 &
  \cellcolor[HTML]{E9E583}1.1 &
  \cellcolor[HTML]{AAD380}1.5 &
  \cellcolor[HTML]{A1D07F}-3.2 &
  \cellcolor[HTML]{FEE182}-2.9 &
  \cellcolor[HTML]{FCC37C}0.0 &
  \cellcolor[HTML]{B7D680}-0.9 &
  \cellcolor[HTML]{D9E082}-0.7 &
  \cellcolor[HTML]{E6E483}0.3 &
  \cellcolor[HTML]{8ACA7E}1.3 &
  \cellcolor[HTML]{E2E383}1.3 &
  \cellcolor[HTML]{A2D17F}-3.7 &
  \cellcolor[HTML]{B3D580}-18.9 &
  \cellcolor[HTML]{EBE683}1.2 &
  \cellcolor[HTML]{CFDD82}1.9 &
  \cellcolor[HTML]{FEE883}-4.1 &
  \cellcolor[HTML]{B0D480}-29.8 &
  \cellcolor[HTML]{FFEB84}0.2 \\
ScienceQA &
  6 K &
  5.8 &
  \cellcolor[HTML]{93CC7E}-6.6 &
  \cellcolor[HTML]{ADD480}-2.5 &
  \cellcolor[HTML]{CDDD82}-0.6 &
  \cellcolor[HTML]{9ECF7F}-0.8 &
  \cellcolor[HTML]{FEEA83}-2.4 &
  \cellcolor[HTML]{BED881}-1.0 &
  \cellcolor[HTML]{CBDC81}-0.4 &
  \cellcolor[HTML]{C0D981}0.2 &
  \cellcolor[HTML]{E4E483}-1.3 &
  \cellcolor[HTML]{63BE7B}10.0 &
  \cellcolor[HTML]{CADC81}-0.9 &
  \cellcolor[HTML]{E4E483}1.0 &
  \cellcolor[HTML]{FEE282}-0.5 &
  \cellcolor[HTML]{FEE683}-10.5 &
  \cellcolor[HTML]{A6D27F}-4.3 &
  \cellcolor[HTML]{FCBD7B}-10.3 &
  \cellcolor[HTML]{F8696B}-4.3 &
  \cellcolor[HTML]{BAD780}-1.3 &
  \cellcolor[HTML]{F2E884}-2.9 &
  \cellcolor[HTML]{FEE883}-1.9 &
  \cellcolor[HTML]{ADD480}-6.7 &
  \cellcolor[HTML]{FCC27C}-1.3 &
  \cellcolor[HTML]{FFEB84}-24.2 &
  \cellcolor[HTML]{DBE182}-33.5 &
  \cellcolor[HTML]{F2E884}0.7 &
  \cellcolor[HTML]{FCC47C}-4.5 &
  \cellcolor[HTML]{63BE7B}10.0 &
  \cellcolor[HTML]{FA9E75}-169.7 &
  \cellcolor[HTML]{FBAC77}-0.4 \\
A-OKVQA &
  17 K &
  5.7 &
  \cellcolor[HTML]{FCB479}-70.1 &
  \cellcolor[HTML]{FCB77A}-25.5 &
  \cellcolor[HTML]{FBA576}-11.9 &
  \cellcolor[HTML]{9ECF7F}-0.7 &
  \cellcolor[HTML]{BCD881}3.0 &
  \cellcolor[HTML]{79C57D}7.4 &
  \cellcolor[HTML]{8BCA7E}6.1 &
  \cellcolor[HTML]{63BE7B}10.0 &
  \cellcolor[HTML]{84C87D}7.2 &
  \cellcolor[HTML]{FCEB84}-0.7 &
  \cellcolor[HTML]{D2DE82}-1.8 &
  \cellcolor[HTML]{F4E884}-0.1 &
  \cellcolor[HTML]{F5E884}0.4 &
  \cellcolor[HTML]{D7E082}-3.9 &
  \cellcolor[HTML]{8ECB7E}0.8 &
  \cellcolor[HTML]{C0D981}3.5 &
  \cellcolor[HTML]{FFEB84}1.9 &
  \cellcolor[HTML]{BAD780}-1.3 &
  \cellcolor[HTML]{DCE182}-1.0 &
  \cellcolor[HTML]{EDE683}-0.2 &
  \cellcolor[HTML]{FFEB84}-25.5 &
  \cellcolor[HTML]{F97D6E}-2.2 &
  \cellcolor[HTML]{96CD7E}-1.2 &
  \cellcolor[HTML]{FEE683}-50.7 &
  \cellcolor[HTML]{FEDE81}-0.7 &
  \cellcolor[HTML]{FA9473}-7.9 &
  \cellcolor[HTML]{EAE583}-1.6 &
  \cellcolor[HTML]{FDCC7E}-110.0 &
  \cellcolor[HTML]{F8786D}-0.9 \\
OpenCQA &
  6K &
  5.7 &
  \cellcolor[HTML]{D1DE82}-28.6 &
  \cellcolor[HTML]{B4D680}-3.6 &
  \cellcolor[HTML]{E2E383}-2.8 &
  \cellcolor[HTML]{B7D780}-5.5 &
  \cellcolor[HTML]{F8E984}-1.7 &
  \cellcolor[HTML]{F0E784}-7.2 &
  \cellcolor[HTML]{FEE583}-7.8 &
  \cellcolor[HTML]{EAE583}-4.1 &
  \cellcolor[HTML]{E6E483}-1.4 &
  \cellcolor[HTML]{FCC07B}-1.9 &
  \cellcolor[HTML]{D0DE82}-1.6 &
  \cellcolor[HTML]{F0E784}0.2 &
  \cellcolor[HTML]{FEE182}-0.5 &
  \cellcolor[HTML]{FEDE81}-12.9 &
  \cellcolor[HTML]{FEE282}-29.6 &
  \cellcolor[HTML]{FDD57F}-5.4 &
  \cellcolor[HTML]{FFEB84}1.9 &
  \cellcolor[HTML]{FFEB84}-10.4 &
  \cellcolor[HTML]{FEE082}-8.2 &
  \cellcolor[HTML]{FBA676}-10.4 &
  \cellcolor[HTML]{FCC07B}-115.6 &
  \cellcolor[HTML]{63BE7B}10.0 &
  \cellcolor[HTML]{F8786E}-49.3 &
  \cellcolor[HTML]{FFEB84}-47.0 &
  \cellcolor[HTML]{FBAD78}-3.1 &
  \cellcolor[HTML]{89C97E}7.1 &
  \cellcolor[HTML]{FCBA7A}-16.4 &
  \cellcolor[HTML]{63BE7B}10.0 &
  \cellcolor[HTML]{EBE683}1.5 \\
IconQA &
  19 K &
  5.3 &
  \cellcolor[HTML]{8FCB7E}-5.1 &
  \cellcolor[HTML]{A5D17F}-1.0 &
  \cellcolor[HTML]{D0DE82}-1.0 &
  \cellcolor[HTML]{A3D17F}-1.8 &
  \cellcolor[HTML]{EBE683}-0.7 &
  \cellcolor[HTML]{E9E583}-6.3 &
  \cellcolor[HTML]{F3E884}-4.5 &
  \cellcolor[HTML]{ECE683}-4.3 &
  \cellcolor[HTML]{FFEB84}-3.7 &
  \cellcolor[HTML]{FFEB84}-0.9 &
  \cellcolor[HTML]{C5DB81}-0.4 &
  \cellcolor[HTML]{E6E483}0.9 &
  \cellcolor[HTML]{63BE7B}10.0 &
  \cellcolor[HTML]{E5E483}-5.6 &
  \cellcolor[HTML]{FEE282}-29.3 &
  \cellcolor[HTML]{FDD17F}-6.2 &
  \cellcolor[HTML]{63BE7B}10.0 &
  \cellcolor[HTML]{C8DB81}-3.1 &
  \cellcolor[HTML]{FFEB84}-4.2 &
  \cellcolor[HTML]{F8E984}-1.0 &
  \cellcolor[HTML]{FEDF81}-50.1 &
  \cellcolor[HTML]{FBAD78}-1.6 &
  \cellcolor[HTML]{D5DF82}-14.9 &
  \cellcolor[HTML]{FDD17F}-68.7 &
  \cellcolor[HTML]{FDC57C}-1.9 &
  \cellcolor[HTML]{FBA977}-6.5 &
  \cellcolor[HTML]{90CB7E}6.1 &
  \cellcolor[HTML]{FBA676}-159.8 &
  \cellcolor[HTML]{FBA276}-0.5 \\
GQA &
  943 K &
  5.0 &
  \cellcolor[HTML]{FA9373}-85.3 &
  \cellcolor[HTML]{FA9B74}-30.6 &
  \cellcolor[HTML]{F87B6E}-15.6 &
  \cellcolor[HTML]{AAD380}-3.0 &
  \cellcolor[HTML]{9FD07F}5.3 &
  \cellcolor[HTML]{EDE683}-6.7 &
  \cellcolor[HTML]{B3D580}2.0 &
  \cellcolor[HTML]{DEE283}-2.9 &
  \cellcolor[HTML]{EEE683}-2.1 &
  \cellcolor[HTML]{F7E984}-0.3 &
  \cellcolor[HTML]{63BE7B}10.0 &
  \cellcolor[HTML]{63BE7B}10.0 &
  \cellcolor[HTML]{FFEB84}-0.3 &
  \cellcolor[HTML]{FEE482}-11.0 &
  \cellcolor[HTML]{B4D680}-7.3 &
  \cellcolor[HTML]{84C87D}7.7 &
  \cellcolor[HTML]{FFEB84}1.9 &
  \cellcolor[HTML]{FEE683}-11.1 &
  \cellcolor[HTML]{FEE282}-7.6 &
  \cellcolor[HTML]{FBEA84}-1.2 &
  \cellcolor[HTML]{E7E483}-19.8 &
  \cellcolor[HTML]{F8736D}-2.3 &
  \cellcolor[HTML]{CDDD82}-13.0 &
  \cellcolor[HTML]{8BCA7E}-4.4 &
  \cellcolor[HTML]{FFEB84}-0.1 &
  \cellcolor[HTML]{F9816F}-9.3 &
  \cellcolor[HTML]{7EC67D}7.7 &
  \cellcolor[HTML]{F86F6C}-230.0 &
  \cellcolor[HTML]{F8696B}-1.1 \\
OK-VQA &
  9 K &
  4.9 &
  \cellcolor[HTML]{FCEB84}-43.6 &
  \cellcolor[HTML]{FAEA84}-15.4 &
  \cellcolor[HTML]{FFEB84}-5.7 &
  \cellcolor[HTML]{B8D780}-5.7 &
  \cellcolor[HTML]{D8E082}0.8 &
  \cellcolor[HTML]{63BE7B}10.0 &
  \cellcolor[HTML]{63BE7B}10.0 &
  \cellcolor[HTML]{BED981}0.5 &
  \cellcolor[HTML]{CFDE82}0.6 &
  \cellcolor[HTML]{FDD27F}-1.5 &
  \cellcolor[HTML]{DEE283}-3.0 &
  \cellcolor[HTML]{F3E884}0.0 &
  \cellcolor[HTML]{FEE983}-0.4 &
  \cellcolor[HTML]{FFEB84}-8.8 &
  \cellcolor[HTML]{D1DE82}-13.6 &
  \cellcolor[HTML]{FFEB84}-1.0 &
  \cellcolor[HTML]{FBAF78}-0.9 &
  \cellcolor[HTML]{BDD881}-1.7 &
  \cellcolor[HTML]{E4E483}-1.7 &
  \cellcolor[HTML]{EEE784}-0.3 &
  \cellcolor[HTML]{E7E483}-19.9 &
  \cellcolor[HTML]{F97F6F}-2.2 &
  \cellcolor[HTML]{7FC67D}4.0 &
  \cellcolor[HTML]{DDE283}-34.4 &
  \cellcolor[HTML]{EBE683}1.2 &
  \cellcolor[HTML]{FBB178}-5.9 &
  \cellcolor[HTML]{93CC7E}5.9 &
  \cellcolor[HTML]{FFEB84}-71.5 &
  \cellcolor[HTML]{F8796E}-0.9 \\
COCO Caption &
  567 K &
  4.8 &
  \cellcolor[HTML]{63BE7B}10.0 &
  \cellcolor[HTML]{94CC7E}1.9 &
  \cellcolor[HTML]{BCD881}1.1 &
  \cellcolor[HTML]{FDC67C}-32.6 &
  \cellcolor[HTML]{FDEB84}-2.1 &
  \cellcolor[HTML]{FDCB7D}-16.1 &
  \cellcolor[HTML]{FEEA83}-6.0 &
  \cellcolor[HTML]{FDCB7D}-13.9 &
  \cellcolor[HTML]{CEDD82}0.7 &
  \cellcolor[HTML]{F7E984}-0.4 &
  \cellcolor[HTML]{FEDF81}-9.5 &
  \cellcolor[HTML]{FFEB84}-0.9 &
  \cellcolor[HTML]{E5E483}1.4 &
  \cellcolor[HTML]{AAD380}1.5 &
  \cellcolor[HTML]{BED981}-9.5 &
  \cellcolor[HTML]{FEDB81}-4.1 &
  \cellcolor[HTML]{FCC37C}0.0 &
  \cellcolor[HTML]{F3E884}-8.8 &
  \cellcolor[HTML]{FEE482}-6.7 &
  \cellcolor[HTML]{E6E483}0.4 &
  \cellcolor[HTML]{ADD480}-6.7 &
  \cellcolor[HTML]{E1E383}1.3 &
  \cellcolor[HTML]{ADD480}-6.0 &
  \cellcolor[HTML]{BAD780}-21.5 &
  \cellcolor[HTML]{F6E984}0.4 &
  \cellcolor[HTML]{D4DF82}1.5 &
  \cellcolor[HTML]{E0E283}-0.7 &
  \cellcolor[HTML]{E9E583}-59.9 &
  \cellcolor[HTML]{F6E984}0.8 \\
TextCaps &
  549 K &
  4.7 &
  \cellcolor[HTML]{FEDB80}-52.0 &
  \cellcolor[HTML]{FEE983}-16.5 &
  \cellcolor[HTML]{63BE7B}10.0 &
  \cellcolor[HTML]{FDD27F}-28.0 &
  \cellcolor[HTML]{FEE883}-3.5 &
  \cellcolor[HTML]{FFEB84}-9.0 &
  \cellcolor[HTML]{FEEA83}-5.7 &
  \cellcolor[HTML]{F7E984}-5.5 &
  \cellcolor[HTML]{CEDD82}0.7 &
  \cellcolor[HTML]{FCEA84}-0.7 &
  \cellcolor[HTML]{FEE582}-8.1 &
  \cellcolor[HTML]{FEE282}-2.6 &
  \cellcolor[HTML]{E3E383}1.5 &
  \cellcolor[HTML]{FEE081}-12.4 &
  \cellcolor[HTML]{BCD881}-9.1 &
  \cellcolor[HTML]{FEE583}-2.1 &
  \cellcolor[HTML]{FCC37C}0.0 &
  \cellcolor[HTML]{B5D680}-0.6 &
  \cellcolor[HTML]{9CCF7F}4.8 &
  \cellcolor[HTML]{EBE683}0.0 &
  \cellcolor[HTML]{91CC7E}-0.3 &
  \cellcolor[HTML]{E2E383}1.3 &
  \cellcolor[HTML]{9DCF7F}-2.6 &
  \cellcolor[HTML]{C6DB81}-26.1 &
  \cellcolor[HTML]{F2E884}0.7 &
  \cellcolor[HTML]{C2DA81}2.9 &
  \cellcolor[HTML]{FDD37F}-9.8 &
  \cellcolor[HTML]{9ECF7F}-20.7 &
  \cellcolor[HTML]{F7E984}0.7 \\
TextVQA &
  35 K &
  4.7 &
  \cellcolor[HTML]{F8696B}-104.8 &
  \cellcolor[HTML]{F8696B}-39.7 &
  \cellcolor[HTML]{F98570}-14.8 &
  \cellcolor[HTML]{E9E583}-14.7 &
  \cellcolor[HTML]{FEDE81}-8.6 &
  \cellcolor[HTML]{EBE683}-6.6 &
  \cellcolor[HTML]{E9E583}-3.5 &
  \cellcolor[HTML]{FFEB84}-6.4 &
  \cellcolor[HTML]{FEDA80}-8.0 &
  \cellcolor[HTML]{FDEB84}-0.7 &
  \cellcolor[HTML]{FFEB84}-6.7 &
  \cellcolor[HTML]{FEE683}-1.8 &
  \cellcolor[HTML]{FDD780}-0.7 &
  \cellcolor[HTML]{FBA977}-31.2 &
  \cellcolor[HTML]{BAD780}-8.6 &
  \cellcolor[HTML]{FEE983}-1.2 &
  \cellcolor[HTML]{F97C6E}-3.4 &
  \cellcolor[HTML]{63BE7B}10.0 &
  \cellcolor[HTML]{63BE7B}10.0 &
  \cellcolor[HTML]{E5E483}0.4 &
  \cellcolor[HTML]{98CE7F}-1.9 &
  \cellcolor[HTML]{FA9D75}-1.8 &
  \cellcolor[HTML]{63BE7B}10.0 &
  \cellcolor[HTML]{ADD480}-16.9 &
  \cellcolor[HTML]{F4E884}0.6 &
  \cellcolor[HTML]{FDD680}-3.2 &
  \cellcolor[HTML]{BBD881}2.5 &
  \cellcolor[HTML]{8ECB7E}-12.0 &
  \cellcolor[HTML]{F87A6E}-0.9 \\
\vlbenchmark &
  7K &
  3.9 &
  \cellcolor[HTML]{F97E6F}-94.8 &
  \cellcolor[HTML]{FA9573}-31.6 &
  \cellcolor[HTML]{F97C6E}-15.5 &
  \cellcolor[HTML]{FFEB84}-18.8 &
  \cellcolor[HTML]{FFEB84}-2.3 &
  \cellcolor[HTML]{FBAD78}-23.1 &
  \cellcolor[HTML]{FFEB84}-5.7 &
  \cellcolor[HTML]{FBB178}-20.1 &
  \cellcolor[HTML]{FEE883}-4.4 &
  \cellcolor[HTML]{FFEB84}-0.9 &
  \cellcolor[HTML]{EFE784}-4.9 &
  \cellcolor[HTML]{F6E984}-0.3 &
  \cellcolor[HTML]{F1E784}0.7 &
  \cellcolor[HTML]{B9D780}-0.2 &
  \cellcolor[HTML]{FFEB84}-23.6 &
  \cellcolor[HTML]{F1E784}0.0 &
  \cellcolor[HTML]{FFEB84}1.9 &
  \cellcolor[HTML]{FDCD7E}-14.4 &
  \cellcolor[HTML]{DEE283}-1.2 &
  \cellcolor[HTML]{FEE282}-2.7 &
  \cellcolor[HTML]{AFD480}-7.2 &
  \cellcolor[HTML]{DFE283}1.4 &
  \cellcolor[HTML]{FEDE81}-27.0 &
  \cellcolor[HTML]{9DCF7F}-10.9 &
  \cellcolor[HTML]{FBB078}-3.0 &
  \cellcolor[HTML]{63BE7B}10.0 &
  \cellcolor[HTML]{F0E784}-2.0 &
  \cellcolor[HTML]{A8D27F}-25.7 &
  \cellcolor[HTML]{66BF7C}9.8 \\
HM &
  9 K &
  3.3 &
  \cellcolor[HTML]{9ECF7F}-10.5 &
  \cellcolor[HTML]{B9D780}-4.4 &
  \cellcolor[HTML]{DCE182}-2.1 &
  \cellcolor[HTML]{FCC47C}-33.0 &
  \cellcolor[HTML]{F5E884}-1.5 &
  \cellcolor[HTML]{F86B6B}-37.9 &
  \cellcolor[HTML]{EFE784}-4.0 &
  \cellcolor[HTML]{F86D6B}-36.1 &
  \cellcolor[HTML]{FEE382}-5.6 &
  \cellcolor[HTML]{FEE082}-1.2 &
  \cellcolor[HTML]{FCB379}-20.2 &
  \cellcolor[HTML]{FEE382}-2.5 &
  \cellcolor[HTML]{FDEB84}-0.1 &
  \cellcolor[HTML]{DDE283}-4.6 &
  \cellcolor[HTML]{FDCE7E}-43.0 &
  \cellcolor[HTML]{63BE7B}10.0 &
  \cellcolor[HTML]{F8696B}-4.3 &
  \cellcolor[HTML]{F98A71}-23.6 &
  \cellcolor[HTML]{FEDC81}-9.9 &
  \cellcolor[HTML]{FEDD81}-3.3 &
  \cellcolor[HTML]{FDCE7E}-85.7 &
  \cellcolor[HTML]{F8696B}-2.5 &
  \cellcolor[HTML]{FA9473}-43.3 &
  \cellcolor[HTML]{FDD37F}-66.5 &
  \cellcolor[HTML]{63BE7B}10.0 &
  \cellcolor[HTML]{F86D6B}-10.7 &
  \cellcolor[HTML]{C0D981}2.0 &
  \cellcolor[HTML]{F86E6C}-231.3 &
  \cellcolor[HTML]{F97F6F}-0.8 \\
VQAv2 QG &
  444 K &
  2.9 &
  \cellcolor[HTML]{FA9E75}-80.2 &
  \cellcolor[HTML]{F98A71}-33.7 &
  \cellcolor[HTML]{F98470}-14.8 &
  \cellcolor[HTML]{FDD37F}-27.5 &
  \cellcolor[HTML]{FEE081}-7.8 &
  \cellcolor[HTML]{FCB87A}-20.4 &
  \cellcolor[HTML]{FEE582}-7.8 &
  \cellcolor[HTML]{FCC37C}-15.8 &
  \cellcolor[HTML]{FEE482}-5.4 &
  \cellcolor[HTML]{FCBB7A}-2.0 &
  \cellcolor[HTML]{FEE282}-8.8 &
  \cellcolor[HTML]{FEE883}-1.4 &
  \cellcolor[HTML]{FDD07E}-0.9 &
  \cellcolor[HTML]{FEE582}-10.7 &
  \cellcolor[HTML]{FEE983}-24.6 &
  \cellcolor[HTML]{73C37C}8.9 &
  \cellcolor[HTML]{C9DC81}4.7 &
  \cellcolor[HTML]{FEE482}-11.3 &
  \cellcolor[HTML]{E4E383}-1.7 &
  \cellcolor[HTML]{FEDE81}-3.1 &
  \cellcolor[HTML]{FA9673}-203.9 &
  \cellcolor[HTML]{E3E383}1.2 &
  \cellcolor[HTML]{FA9E75}-40.9 &
  \cellcolor[HTML]{FDD680}-63.9 &
  \cellcolor[HTML]{FCB679}-2.7 &
  \cellcolor[HTML]{FFEB84}-1.8 &
  \cellcolor[HTML]{FDD27F}-10.0 &
  \cellcolor[HTML]{FEEA83}-71.7 &
  \cellcolor[HTML]{F5E984}0.8 \\
LLaVA Conversation &
  57 K &
  2.4 &
  \cellcolor[HTML]{FDD07E}-56.9 &
  \cellcolor[HTML]{FED980}-19.5 &
  \cellcolor[HTML]{FDCC7E}-8.4 &
  \cellcolor[HTML]{F97C6E}-59.9 &
  \cellcolor[HTML]{FED980}-11.5 &
  \cellcolor[HTML]{F86C6B}-37.7 &
  \cellcolor[HTML]{FCBF7B}-23.2 &
  \cellcolor[HTML]{F86D6B}-36.1 &
  \cellcolor[HTML]{FDC77D}-12.8 &
  \cellcolor[HTML]{F8766D}-3.6 &
  \cellcolor[HTML]{F97E6F}-33.4 &
  \cellcolor[HTML]{FEDC81}-4.2 &
  \cellcolor[HTML]{FEE683}-0.4 &
  \cellcolor[HTML]{FBA877}-31.7 &
  \cellcolor[HTML]{F97C6E}-100.1 &
  \cellcolor[HTML]{E0E283}1.2 &
  \cellcolor[HTML]{FFEB84}1.9 &
  \cellcolor[HTML]{F87A6E}-25.8 &
  \cellcolor[HTML]{FDCB7D}-16.8 &
  \cellcolor[HTML]{F8766D}-16.7 &
  \cellcolor[HTML]{FBAC78}-155.9 &
  \cellcolor[HTML]{8DCB7E}7.1 &
  \cellcolor[HTML]{F97D6E}-48.4 &
  \cellcolor[HTML]{FDC97D}-75.2 &
  \cellcolor[HTML]{FBA376}-3.6 &
  \cellcolor[HTML]{72C37C}8.9 &
  \cellcolor[HTML]{F98971}-29.8 &
  \cellcolor[HTML]{D4DF82}-48.7 &
  \cellcolor[HTML]{63BE7B}10.0 \\
A-OKVQA QG &
  17 K &
  2.3 &
  \cellcolor[HTML]{FBA175}-78.5 &
  \cellcolor[HTML]{FA9F75}-30.0 &
  \cellcolor[HTML]{FBAC77}-11.3 &
  \cellcolor[HTML]{F98770}-55.8 &
  \cellcolor[HTML]{FED880}-11.8 &
  \cellcolor[HTML]{FA9E75}-26.4 &
  \cellcolor[HTML]{FDD780}-13.6 &
  \cellcolor[HTML]{FBA376}-23.3 &
  \cellcolor[HTML]{FEE382}-5.6 &
  \cellcolor[HTML]{FCC27C}-1.8 &
  \cellcolor[HTML]{FA9C74}-25.9 &
  \cellcolor[HTML]{FEE382}-2.6 &
  \cellcolor[HTML]{F5E884}0.4 &
  \cellcolor[HTML]{F9EA84}-8.0 &
  \cellcolor[HTML]{FA9974}-80.1 &
  \cellcolor[HTML]{A9D380}5.1 &
  \cellcolor[HTML]{FFEB84}1.9 &
  \cellcolor[HTML]{FCB579}-17.7 &
  \cellcolor[HTML]{FDD37F}-13.5 &
  \cellcolor[HTML]{FEDA80}-3.7 &
  \cellcolor[HTML]{FA9874}-198.4 &
  \cellcolor[HTML]{F1E784}0.2 &
  \cellcolor[HTML]{FCB87A}-35.3 &
  \cellcolor[HTML]{FDD07E}-69.6 &
  \cellcolor[HTML]{F2E884}0.7 &
  \cellcolor[HTML]{E6E483}0.1 &
  \cellcolor[HTML]{F98A71}-29.3 &
  \cellcolor[HTML]{DDE182}-53.3 &
  \cellcolor[HTML]{FFEB84}0.2 \\
OCR-VQA &
  802 K &
  2.3 &
  \cellcolor[HTML]{A4D17F}-12.6 &
  \cellcolor[HTML]{B8D780}-4.2 &
  \cellcolor[HTML]{F6E984}-4.7 &
  \cellcolor[HTML]{FDCF7E}-29.2 &
  \cellcolor[HTML]{FCC57C}-21.5 &
  \cellcolor[HTML]{F98E72}-30.1 &
  \cellcolor[HTML]{FDC77D}-20.1 &
  \cellcolor[HTML]{FA9874}-25.8 &
  \cellcolor[HTML]{FA9773}-25.4 &
  \cellcolor[HTML]{FA9773}-2.8 &
  \cellcolor[HTML]{FDCE7E}-13.8 &
  \cellcolor[HTML]{FDD67F}-5.5 &
  \cellcolor[HTML]{FA9F75}-1.9 &
  \cellcolor[HTML]{FCC57C}-21.7 &
  \cellcolor[HTML]{FDD07E}-41.6 &
  \cellcolor[HTML]{FDCE7E}-6.8 &
  \cellcolor[HTML]{FFEB84}1.9 &
  \cellcolor[HTML]{FCBF7B}-16.3 &
  \cellcolor[HTML]{FBAC77}-29.5 &
  \cellcolor[HTML]{63BE7B}10.0 &
  \cellcolor[HTML]{FEE783}-32.4 &
  \cellcolor[HTML]{F8696B}-2.5 &
  \cellcolor[HTML]{FBAB77}-38.1 &
  \cellcolor[HTML]{FCC07B}-82.2 &
  \cellcolor[HTML]{FEDB81}-0.9 &
  \cellcolor[HTML]{F8696B}-11.0 &
  \cellcolor[HTML]{FDD57F}-9.3 &
  \cellcolor[HTML]{F8696B}-238.2 &
  \cellcolor[HTML]{F86D6B}-1.0 \\
VSR &
  3 K &
  2.0 &
  \cellcolor[HTML]{B1D580}-17.3 &
  \cellcolor[HTML]{C8DC81}-7.0 &
  \cellcolor[HTML]{D9E182}-1.9 &
  \cellcolor[HTML]{FA8E72}-53.2 &
  \cellcolor[HTML]{FCBA7A}-27.3 &
  \cellcolor[HTML]{F8776D}-35.1 &
  \cellcolor[HTML]{FBB178}-29.0 &
  \cellcolor[HTML]{F8746D}-34.4 &
  \cellcolor[HTML]{FCB77A}-17.0 &
  \cellcolor[HTML]{F8696B}-3.8 &
  \cellcolor[HTML]{F98670}-31.3 &
  \cellcolor[HTML]{FED980}-4.6 &
  \cellcolor[HTML]{FDC77D}-1.1 &
  \cellcolor[HTML]{C9DC81}-2.2 &
  \cellcolor[HTML]{F8696B}-112.7 &
  \cellcolor[HTML]{FDCC7E}-7.2 &
  \cellcolor[HTML]{FFEB84}1.9 &
  \cellcolor[HTML]{F98C71}-23.3 &
  \cellcolor[HTML]{FBB078}-27.8 &
  \cellcolor[HTML]{FA9C74}-11.8 &
  \cellcolor[HTML]{FDD17F}-79.8 &
  \cellcolor[HTML]{FBA676}-1.7 &
  \cellcolor[HTML]{F8736D}-50.5 &
  \cellcolor[HTML]{F8746D}-146.7 &
  \cellcolor[HTML]{FEE482}-0.4 &
  \cellcolor[HTML]{F8706C}-10.5 &
  \cellcolor[HTML]{FFEB84}-3.4 &
  \cellcolor[HTML]{FDD47F}-100.5 &
  \cellcolor[HTML]{F8716C}-1.0 \\
LLaVA Reasoning &
  77 K &
  2.0 &
  \cellcolor[HTML]{F8776D}-98.1 &
  \cellcolor[HTML]{F98670}-34.4 &
  \cellcolor[HTML]{F8696B}-17.3 &
  \cellcolor[HTML]{F8696B}-67.2 &
  \cellcolor[HTML]{FDD67F}-13.0 &
  \cellcolor[HTML]{F8696B}-38.5 &
  \cellcolor[HTML]{FCC37C}-21.6 &
  \cellcolor[HTML]{F8696B}-37.1 &
  \cellcolor[HTML]{FDC67C}-13.2 &
  \cellcolor[HTML]{FDD17F}-1.5 &
  \cellcolor[HTML]{F8696B}-38.6 &
  \cellcolor[HTML]{FEE482}-2.3 &
  \cellcolor[HTML]{FDCB7D}-1.0 &
  \cellcolor[HTML]{FDCF7E}-18.3 &
  \cellcolor[HTML]{F86A6B}-112.3 &
  \cellcolor[HTML]{A9D27F}5.1 &
  \cellcolor[HTML]{FCC37C}0.0 &
  \cellcolor[HTML]{F8696B}-28.2 &
  \cellcolor[HTML]{FBB078}-27.7 &
  \cellcolor[HTML]{F8696B}-18.4 &
  \cellcolor[HTML]{FBB179}-145.4 &
  \cellcolor[HTML]{FFEB84}-0.8 &
  \cellcolor[HTML]{F86D6B}-51.9 &
  \cellcolor[HTML]{FEEA83}-47.2 &
  \cellcolor[HTML]{FBA376}-3.6 &
  \cellcolor[HTML]{69C07C}9.6 &
  \cellcolor[HTML]{FEDB81}-7.5 &
  \cellcolor[HTML]{D3DF82}-48.5 &
  \cellcolor[HTML]{F2E884}1.1 \\
OK-VQA QG &
  9 K &
  1.5 &
  \cellcolor[HTML]{F98670}-91.1 &
  \cellcolor[HTML]{F98D72}-33.1 &
  \cellcolor[HTML]{F98370}-15.0 &
  \cellcolor[HTML]{F86A6B}-66.7 &
  \cellcolor[HTML]{FBB279}-31.3 &
  \cellcolor[HTML]{F86E6B}-37.3 &
  \cellcolor[HTML]{FCB87A}-26.1 &
  \cellcolor[HTML]{F86F6C}-35.7 &
  \cellcolor[HTML]{FA9974}-24.9 &
  \cellcolor[HTML]{F86E6B}-3.7 &
  \cellcolor[HTML]{F86C6B}-37.8 &
  \cellcolor[HTML]{FCB579}-12.6 &
  \cellcolor[HTML]{FBAD78}-1.6 &
  \cellcolor[HTML]{FBAC78}-30.2 &
  \cellcolor[HTML]{F8696B}-113.0 &
  \cellcolor[HTML]{E6E483}0.7 &
  \cellcolor[HTML]{FFEB84}1.9 &
  \cellcolor[HTML]{F8756D}-26.5 &
  \cellcolor[HTML]{FCB97A}-24.1 &
  \cellcolor[HTML]{FCC37C}-6.7 &
  \cellcolor[HTML]{F98470}-241.5 &
  \cellcolor[HTML]{F4E884}0.0 &
  \cellcolor[HTML]{FA9874}-42.3 &
  \cellcolor[HTML]{FCBC7B}-85.9 &
  \cellcolor[HTML]{FBAD78}-3.1 &
  \cellcolor[HTML]{FEDF81}-2.6 &
  \cellcolor[HTML]{FBA376}-22.7 &
  \cellcolor[HTML]{FEE482}-79.6 &
  \cellcolor[HTML]{FAEA84}0.6 \\
LLaVA Description &
  23 K &
  1.3 &
  \cellcolor[HTML]{F97D6F}-95.2 &
  \cellcolor[HTML]{FA9473}-31.9 &
  \cellcolor[HTML]{F8736C}-16.4 &
  \cellcolor[HTML]{F8696B}-67.2 &
  \cellcolor[HTML]{F8696B}-68.7 &
  \cellcolor[HTML]{F8696B}-38.5 &
  \cellcolor[HTML]{F8696B}-58.4 &
  \cellcolor[HTML]{F8696B}-37.1 &
  \cellcolor[HTML]{F8696B}-37.5 &
  \cellcolor[HTML]{F97E6F}-3.4 &
  \cellcolor[HTML]{F8696B}-38.6 &
  \cellcolor[HTML]{F8696B}-29.3 &
  \cellcolor[HTML]{F8696B}-3.0 &
  \cellcolor[HTML]{F8696B}-53.7 &
  \cellcolor[HTML]{F8696B}-113.2 &
  \cellcolor[HTML]{F8696B}-27.6 &
  \cellcolor[HTML]{FFEB84}1.9 &
  \cellcolor[HTML]{F8696B}-28.1 &
  \cellcolor[HTML]{F8696B}-56.4 &
  \cellcolor[HTML]{F8696B}-18.4 &
  \cellcolor[HTML]{F8696B}-298.5 &
  \cellcolor[HTML]{FDCA7D}-1.2 &
  \cellcolor[HTML]{F8696B}-52.8 &
  \cellcolor[HTML]{F8696B}-155.9 &
  \cellcolor[HTML]{F8696B}-6.4 &
  \cellcolor[HTML]{7CC67D}8.2 &
  \cellcolor[HTML]{F8696B}-38.4 &
  \cellcolor[HTML]{FDC67D}-118.0 &
  \cellcolor[HTML]{FBEA84}0.5
           \\
			\bottomrule
		\end{tabular}%
	}
	\caption
	{Normalized transfer learning performance of BLIP-2. Higher values indicate better transferability. The rows are sorted in descending order of average performance. We multiply the values by a factor of 10 to aid visualization. The highest performance in each column is 10. 
  QG denotes question generation, MC denotes multiple-choice and G denotes open-ended generation. The color scale is normalized along each column. The colors represent values in descending order: green, yellow, orange and red. 
	}
	\label{eval:affinity_matrix_norm_blip_landscape}
\end{table}
\end{landscape}
\begin{landscape}
\begin{table}[!tp]
	\resizebox{\columnwidth}{!}{%
		\begin{tabular}{c|c|c|ccccccccccccccccccccccccccccc}
			\toprule
			Source   Task &
			\setlength\extrarowheight{0pt}\begin{tabular}[c]{@{}c@{}}Dataset\\ Size \end{tabular} &
			\setlength\extrarowheight{0pt}\begin{tabular}[c]{@{}c@{}}AHP\\ Ranking \\ Score \end{tabular}   &
			\multicolumn{29}{c}{Target Task} \\
			&
			&
			&
			\setlength\extrarowheight{0pt}\begin{tabular}[c]{@{}c@{}}COCO\\ Caption \end{tabular}  &
			\setlength\extrarowheight{0pt}\begin{tabular}[c]{@{}c@{}}Flickr\\ 30k\end{tabular}  &
			\setlength\extrarowheight{0pt}\begin{tabular}[c]{@{}c@{}}Text\\ Caps\end{tabular}  &
			\multicolumn{2}{c}{VQAv2} &
			\multicolumn{2}{c}{OK-VQA} &
			\multicolumn{2}{c}{A-OKVQA} &
			\setlength\extrarowheight{0pt}\begin{tabular}[c]{@{}c@{}}Science\\ QA\end{tabular} &
			\multicolumn{2}{c}{GQA} &
			\setlength\extrarowheight{0pt}\begin{tabular}[c]{@{}c@{}}Icon\\ QA \end{tabular} &
			VSR &
			\multicolumn{2}{c}{CLEVR} &
			\setlength\extrarowheight{0pt}\begin{tabular}[c]{@{}c@{}}RAVEN-\\ FAIR \end{tabular} &
			\multicolumn{2}{c}{\setlength\extrarowheight{0pt}\begin{tabular}[c]{@{}c@{}}Text\\ VQA\end{tabular}} &
			\multicolumn{2}{c}{\setlength\extrarowheight{0pt}\begin{tabular}[c]{@{}c@{}}OCR-\\ VQA\end{tabular}} &
			\setlength\extrarowheight{0pt}\begin{tabular}[c]{@{}c@{}}Open\\ CQA \end{tabular} &
			\multicolumn{2}{c}{\setlength\extrarowheight{0pt}\begin{tabular}[c]{@{}c@{}}Chart\\ QA\end{tabular}} &
			HM &
			\setlength\extrarowheight{0pt}\begin{tabular}[c]{@{}c@{}}NY\\ Explain \end{tabular} &
			\setlength\extrarowheight{0pt}\begin{tabular}[c]{@{}c@{}}NY\\ Rank \end{tabular} &
			MORE &
			\vlbenchmark \\
			 &
			&
			&
			&
			&
			&
			G &
			MC &
			G &
			MC &
			G &
			MC &
			&
			G &
			MC &
			&
			&
			G &
			MC &
			&
			G &
			MC &
			G &
			MC &
			&
			G &
			MC &
			&
			&
			&
			&
			\\
			\midrule
VQAv2 &
  444 K &
  6.9 &
  \cellcolor[HTML]{D7E082}3.9 &
  \cellcolor[HTML]{E1E383}3.1 &
  \cellcolor[HTML]{F1E784}1.9 &
  \cellcolor[HTML]{63BE7B}10.0 &
  \cellcolor[HTML]{63BE7B}10.0 &
  \cellcolor[HTML]{69C07C}9.6 &
  \cellcolor[HTML]{77C47D}9.1 &
  \cellcolor[HTML]{63BE7B}10.0 &
  \cellcolor[HTML]{9CCF7F}7.4 &
  \cellcolor[HTML]{DBE182}3.4 &
  \cellcolor[HTML]{8BCA7E}7.9 &
  \cellcolor[HTML]{9BCF7F}6.8 &
  \cellcolor[HTML]{DFE283}3.3 &
  \cellcolor[HTML]{7DC67D}6.8 &
  \cellcolor[HTML]{63BE7B}10.0 &
  \cellcolor[HTML]{81C77D}6.5 &
  \cellcolor[HTML]{FEE883}-1.0 &
  \cellcolor[HTML]{84C87D}8.0 &
  \cellcolor[HTML]{8FCB7E}7.9 &
  \cellcolor[HTML]{BFD981}5.2 &
  \cellcolor[HTML]{FDD780}1.3 &
  \cellcolor[HTML]{FDD37F}1.8 &
  \cellcolor[HTML]{63BE7B}10.0 &
  \cellcolor[HTML]{FCB77A}-34.9 &
  \cellcolor[HTML]{6BC17C}9.7 &
  \cellcolor[HTML]{F86D6B}-20.3 &
  \cellcolor[HTML]{FFEB84}-4.3 &
  \cellcolor[HTML]{F8696B}-3.8 &
  \cellcolor[HTML]{F86C6B}-101.0 \\
A-OKVQA (MC) &
  17 K &
  6.5 &
  \cellcolor[HTML]{FDCC7E}1.3 &
  \cellcolor[HTML]{FFEB84}1.4 &
  \cellcolor[HTML]{FEEB84}1.2 &
  \cellcolor[HTML]{E1E383}3.2 &
  \cellcolor[HTML]{ACD380}6.0 &
  \cellcolor[HTML]{BED981}3.7 &
  \cellcolor[HTML]{77C47D}9.1 &
  \cellcolor[HTML]{B9D780}4.0 &
  \cellcolor[HTML]{63BE7B}10.0 &
  \cellcolor[HTML]{B9D780}5.3 &
  \cellcolor[HTML]{FEE082}0.8 &
  \cellcolor[HTML]{BCD881}5.0 &
  \cellcolor[HTML]{ECE683}2.6 &
  \cellcolor[HTML]{F98871}-15.9 &
  \cellcolor[HTML]{F4E884}-2.3 &
  \cellcolor[HTML]{A3D17F}2.6 &
  \cellcolor[HTML]{D0DE82}2.4 &
  \cellcolor[HTML]{C6DB81}3.9 &
  \cellcolor[HTML]{63BE7B}10.0 &
  \cellcolor[HTML]{FBEA84}2.0 &
  \cellcolor[HTML]{85C87D}8.2 &
  \cellcolor[HTML]{E6E483}3.7 &
  \cellcolor[HTML]{B6D680}4.9 &
  \cellcolor[HTML]{63BE7B}10.0 &
  \cellcolor[HTML]{F8E984}3.4 &
  \cellcolor[HTML]{FEE683}1.3 &
  \cellcolor[HTML]{B9D780}2.2 &
  \cellcolor[HTML]{FEE883}0.9 &
  \cellcolor[HTML]{F5E884}-60.0 \\
ScienceQA &
  6 K &
  6.3 &
  \cellcolor[HTML]{FA9E75}0.7 &
  \cellcolor[HTML]{FBAA77}0.5 &
  \cellcolor[HTML]{FCC37C}0.7 &
  \cellcolor[HTML]{DAE182}3.5 &
  \cellcolor[HTML]{D5DF82}3.7 &
  \cellcolor[HTML]{C0D981}3.6 &
  \cellcolor[HTML]{B2D580}6.5 &
  \cellcolor[HTML]{BBD881}3.8 &
  \cellcolor[HTML]{A9D380}6.8 &
  \cellcolor[HTML]{63BE7B}10.0 &
  \cellcolor[HTML]{E7E483}3.0 &
  \cellcolor[HTML]{E7E483}2.5 &
  \cellcolor[HTML]{EBE683}2.6 &
  \cellcolor[HTML]{FEE482}-10.0 &
  \cellcolor[HTML]{C4DA81}1.8 &
  \cellcolor[HTML]{FDD37F}-20.6 &
  \cellcolor[HTML]{96CD7E}6.5 &
  \cellcolor[HTML]{D6DF82}2.9 &
  \cellcolor[HTML]{A4D17F}6.9 &
  \cellcolor[HTML]{FBEA84}2.0 &
  \cellcolor[HTML]{9FD07F}6.9 &
  \cellcolor[HTML]{FCEB84}2.6 &
  \cellcolor[HTML]{9ACE7F}6.6 &
  \cellcolor[HTML]{F9EA84}-20.3 &
  \cellcolor[HTML]{FDCE7E}-0.3 &
  \cellcolor[HTML]{FFEB84}2.2 &
  \cellcolor[HTML]{63BE7B}10.0 &
  \cellcolor[HTML]{FEE282}0.6 &
  \cellcolor[HTML]{63BE7B}10.0 \\
GQA &
  943 K &
  5.8 &
  \cellcolor[HTML]{E9E583}2.9 &
  \cellcolor[HTML]{FAEA84}1.7 &
  \cellcolor[HTML]{FCC57C}0.7 &
  \cellcolor[HTML]{ABD380}6.1 &
  \cellcolor[HTML]{8BCA7E}7.8 &
  \cellcolor[HTML]{EAE583}0.7 &
  \cellcolor[HTML]{A0D07F}7.3 &
  \cellcolor[HTML]{CFDD82}2.4 &
  \cellcolor[HTML]{A5D17F}7.0 &
  \cellcolor[HTML]{E4E483}2.8 &
  \cellcolor[HTML]{63BE7B}10.0 &
  \cellcolor[HTML]{63BE7B}10.0 &
  \cellcolor[HTML]{F2E784}2.3 &
  \cellcolor[HTML]{91CB7E}4.3 &
  \cellcolor[HTML]{9ACE7F}5.4 &
  \cellcolor[HTML]{63BE7B}10.0 &
  \cellcolor[HTML]{FEE883}-1.0 &
  \cellcolor[HTML]{FEE182}-0.2 &
  \cellcolor[HTML]{FEE783}2.3 &
  \cellcolor[HTML]{E6E483}3.1 &
  \cellcolor[HTML]{E8E583}3.0 &
  \cellcolor[HTML]{F8696B}-1.1 &
  \cellcolor[HTML]{F5E984}0.9 &
  \cellcolor[HTML]{FEE182}-24.0 &
  \cellcolor[HTML]{AAD380}6.9 &
  \cellcolor[HTML]{F5E984}2.7 &
  \cellcolor[HTML]{91CC7E}5.8 &
  \cellcolor[HTML]{FFEB84}1.0 &
  \cellcolor[HTML]{F86D6B}-100.7 \\
A-OKVQA &
  17 K &
  5.8 &
  \cellcolor[HTML]{F0E784}2.6 &
  \cellcolor[HTML]{FBEA84}1.7 &
  \cellcolor[HTML]{FFEB84}1.1 &
  \cellcolor[HTML]{9FD07F}6.8 &
  \cellcolor[HTML]{88C97E}8.0 &
  \cellcolor[HTML]{7AC57D}8.4 &
  \cellcolor[HTML]{63BE7B}10.0 &
  \cellcolor[HTML]{6DC17C}9.3 &
  \cellcolor[HTML]{8FCB7E}8.0 &
  \cellcolor[HTML]{DEE283}3.2 &
  \cellcolor[HTML]{BED881}5.2 &
  \cellcolor[HTML]{ADD480}5.8 &
  \cellcolor[HTML]{DEE283}3.4 &
  \cellcolor[HTML]{F8696B}-18.0 &
  \cellcolor[HTML]{B7D780}2.9 &
  \cellcolor[HTML]{68C07C}9.5 &
  \cellcolor[HTML]{63BE7B}10.0 &
  \cellcolor[HTML]{B8D780}4.8 &
  \cellcolor[HTML]{99CE7F}7.4 &
  \cellcolor[HTML]{EDE683}2.7 &
  \cellcolor[HTML]{F9806F}-1.0 &
  \cellcolor[HTML]{F8706C}-0.9 &
  \cellcolor[HTML]{96CD7E}6.8 &
  \cellcolor[HTML]{FDD780}-26.6 &
  \cellcolor[HTML]{D9E082}4.8 &
  \cellcolor[HTML]{F8696B}-21.2 &
  \cellcolor[HTML]{CEDD82}0.3 &
  \cellcolor[HTML]{FA9E75}-1.8 &
  \cellcolor[HTML]{F8696B}-102.1 \\
TextVQA &
  35 K &
  5.4 &
  \cellcolor[HTML]{B5D680}5.7 &
  \cellcolor[HTML]{B4D680}5.6 &
  \cellcolor[HTML]{BED981}4.8 &
  \cellcolor[HTML]{AAD380}6.1 &
  \cellcolor[HTML]{B3D580}5.6 &
  \cellcolor[HTML]{B9D780}4.1 &
  \cellcolor[HTML]{ACD380}6.7 &
  \cellcolor[HTML]{B5D680}4.2 &
  \cellcolor[HTML]{A9D27F}6.9 &
  \cellcolor[HTML]{D2DE82}3.8 &
  \cellcolor[HTML]{C6DB81}4.7 &
  \cellcolor[HTML]{CFDE82}3.8 &
  \cellcolor[HTML]{EFE784}2.4 &
  \cellcolor[HTML]{C6DB81}-2.3 &
  \cellcolor[HTML]{7CC67D}7.9 &
  \cellcolor[HTML]{FFEB84}-8.0 &
  \cellcolor[HTML]{F1E784}0.1 &
  \cellcolor[HTML]{63BE7B}10.0 &
  \cellcolor[HTML]{7FC77D}8.7 &
  \cellcolor[HTML]{BCD881}5.3 &
  \cellcolor[HTML]{D9E082}3.8 &
  \cellcolor[HTML]{FBB279}0.9 &
  \cellcolor[HTML]{66BF7C}9.9 &
  \cellcolor[HTML]{FDD27F}-27.8 &
  \cellcolor[HTML]{A2D17F}7.2 &
  \cellcolor[HTML]{FA9B74}-12.1 &
  \cellcolor[HTML]{F8716C}-11.7 &
  \cellcolor[HTML]{FCC47C}-0.4 &
  \cellcolor[HTML]{F87A6E}-97.1 \\
VSR &
  3 K &
  5.4 &
  \cellcolor[HTML]{F8706C}0.0 &
  \cellcolor[HTML]{F9806F}0.0 &
  \cellcolor[HTML]{F98A71}0.1 &
  \cellcolor[HTML]{CCDD82}4.3 &
  \cellcolor[HTML]{FEEB84}1.5 &
  \cellcolor[HTML]{E2E383}1.2 &
  \cellcolor[HTML]{FCBE7B}0.7 &
  \cellcolor[HTML]{D4DF82}2.1 &
  \cellcolor[HTML]{FBA877}0.4 &
  \cellcolor[HTML]{FDCF7E}0.2 &
  \cellcolor[HTML]{CBDC81}4.5 &
  \cellcolor[HTML]{FFEB84}1.1 &
  \cellcolor[HTML]{FDC57C}0.7 &
  \cellcolor[HTML]{63BE7B}10.0 &
  \cellcolor[HTML]{ACD480}3.8 &
  \cellcolor[HTML]{B9D780}0.1 &
  \cellcolor[HTML]{F1E784}0.1 &
  \cellcolor[HTML]{EFE784}1.4 &
  \cellcolor[HTML]{FDC67C}0.4 &
  \cellcolor[HTML]{DBE182}3.7 &
  \cellcolor[HTML]{E1E383}3.4 &
  \cellcolor[HTML]{FBA276}0.5 &
  \cellcolor[HTML]{F8E984}0.8 &
  \cellcolor[HTML]{7EC67D}4.7 &
  \cellcolor[HTML]{FA9E75}-6.2 &
  \cellcolor[HTML]{FEDC81}-0.4 &
  \cellcolor[HTML]{D9E082}-0.8 &
  \cellcolor[HTML]{FDD57F}0.2 &
  \cellcolor[HTML]{76C47D}1.3 \\
Flickr30k &
  145 K &
  5.4 &
  \cellcolor[HTML]{9DCF7F}7.0 &
  \cellcolor[HTML]{63BE7B}10.0 &
  \cellcolor[HTML]{ABD380}5.9 &
  \cellcolor[HTML]{FEE482}1.0 &
  \cellcolor[HTML]{EBE683}2.5 &
  \cellcolor[HTML]{FFEB84}-0.8 &
  \cellcolor[HTML]{F9EA84}3.3 &
  \cellcolor[HTML]{F3E884}-0.2 &
  \cellcolor[HTML]{FFEB84}2.9 &
  \cellcolor[HTML]{F9EA84}1.7 &
  \cellcolor[HTML]{FAEA84}2.0 &
  \cellcolor[HTML]{EAE583}2.3 &
  \cellcolor[HTML]{FAEA84}1.9 &
  \cellcolor[HTML]{ACD480}0.9 &
  \cellcolor[HTML]{97CD7E}5.6 &
  \cellcolor[HTML]{AED480}1.4 &
  \cellcolor[HTML]{FFEB84}-0.9 &
  \cellcolor[HTML]{D7E082}2.8 &
  \cellcolor[HTML]{C6DB81}5.3 &
  \cellcolor[HTML]{FAEA84}2.1 &
  \cellcolor[HTML]{F5E884}2.3 &
  \cellcolor[HTML]{CBDC81}5.0 &
  \cellcolor[HTML]{CDDD82}3.4 &
  \cellcolor[HTML]{DDE283}-14.6 &
  \cellcolor[HTML]{63BE7B}10.0 &
  \cellcolor[HTML]{C4DA81}5.2 &
  \cellcolor[HTML]{CFDE82}0.1 &
  \cellcolor[HTML]{A8D27F}6.0 &
  \cellcolor[HTML]{C3DA81}-35.9 \\
OCR-VQA &
  802 K &
  5.0 &
  \cellcolor[HTML]{D6E082}3.9 &
  \cellcolor[HTML]{D3DF82}3.9 &
  \cellcolor[HTML]{CADC81}4.1 &
  \cellcolor[HTML]{BDD881}5.1 &
  \cellcolor[HTML]{FDCE7E}0.1 &
  \cellcolor[HTML]{D3DF82}2.3 &
  \cellcolor[HTML]{FDC97D}1.3 &
  \cellcolor[HTML]{C2DA81}3.3 &
  \cellcolor[HTML]{E4E383}4.2 &
  \cellcolor[HTML]{E2E383}3.0 &
  \cellcolor[HTML]{BCD881}5.3 &
  \cellcolor[HTML]{D1DE82}3.8 &
  \cellcolor[HTML]{FED980}1.1 &
  \cellcolor[HTML]{E8E583}-6.7 &
  \cellcolor[HTML]{84C87D}7.3 &
  \cellcolor[HTML]{F8696B}-76.7 &
  \cellcolor[HTML]{E1E383}1.3 &
  \cellcolor[HTML]{B1D580}5.2 &
  \cellcolor[HTML]{FCB97A}-0.4 &
  \cellcolor[HTML]{63BE7B}10.0 &
  \cellcolor[HTML]{63BE7B}10.0 &
  \cellcolor[HTML]{FFEB84}2.5 &
  \cellcolor[HTML]{8ECB7E}7.4 &
  \cellcolor[HTML]{DCE182}-14.3 &
  \cellcolor[HTML]{92CC7E}7.9 &
  \cellcolor[HTML]{D0DE82}4.6 &
  \cellcolor[HTML]{FCBE7B}-7.0 &
  \cellcolor[HTML]{E7E483}2.4 &
  \cellcolor[HTML]{F98370}-94.7 \\
OK-VQA &
  9 K &
  5.0 &
  \cellcolor[HTML]{E1E383}3.3 &
  \cellcolor[HTML]{ECE683}2.5 &
  \cellcolor[HTML]{F6E984}1.6 &
  \cellcolor[HTML]{C8DC81}4.5 &
  \cellcolor[HTML]{CCDD82}4.2 &
  \cellcolor[HTML]{63BE7B}10.0 &
  \cellcolor[HTML]{82C77D}8.6 &
  \cellcolor[HTML]{B4D680}4.3 &
  \cellcolor[HTML]{C4DA81}5.6 &
  \cellcolor[HTML]{DBE182}3.4 &
  \cellcolor[HTML]{E5E483}3.1 &
  \cellcolor[HTML]{CBDC81}4.1 &
  \cellcolor[HTML]{E3E383}3.1 &
  \cellcolor[HTML]{F8696B}-18.0 &
  \cellcolor[HTML]{FFEB84}-3.3 &
  \cellcolor[HTML]{C8DB81}-1.6 &
  \cellcolor[HTML]{E6E483}0.9 &
  \cellcolor[HTML]{C8DC81}3.7 &
  \cellcolor[HTML]{ACD480}6.5 &
  \cellcolor[HTML]{FFEB84}1.8 &
  \cellcolor[HTML]{F1E784}2.5 &
  \cellcolor[HTML]{FCB579}1.0 &
  \cellcolor[HTML]{8FCB7E}7.3 &
  \cellcolor[HTML]{F7E984}-19.9 &
  \cellcolor[HTML]{D9E082}4.8 &
  \cellcolor[HTML]{FA9773}-12.8 &
  \cellcolor[HTML]{72C37C}8.7 &
  \cellcolor[HTML]{D6E082}3.3 &
  \cellcolor[HTML]{F86C6B}-101.2 \\
IconQA &
  19 K &
  4.9 &
  \cellcolor[HTML]{FA9172}0.5 &
  \cellcolor[HTML]{FA9C74}0.4 &
  \cellcolor[HTML]{FCC47C}0.7 &
  \cellcolor[HTML]{E7E483}2.8 &
  \cellcolor[HTML]{C6DB81}4.6 &
  \cellcolor[HTML]{C8DB81}3.0 &
  \cellcolor[HTML]{E9E583}4.0 &
  \cellcolor[HTML]{C6DB81}3.0 &
  \cellcolor[HTML]{D4DF82}4.9 &
  \cellcolor[HTML]{E3E383}2.9 &
  \cellcolor[HTML]{EFE784}2.6 &
  \cellcolor[HTML]{DFE283}3.0 &
  \cellcolor[HTML]{63BE7B}10.0 &
  \cellcolor[HTML]{FDEB84}-9.3 &
  \cellcolor[HTML]{C4DA81}1.8 &
  \cellcolor[HTML]{FEDE81}-14.5 &
  \cellcolor[HTML]{FBA276}-4.7 &
  \cellcolor[HTML]{DBE182}2.6 &
  \cellcolor[HTML]{D6E082}4.4 &
  \cellcolor[HTML]{F7E984}2.2 &
  \cellcolor[HTML]{9DCF7F}7.0 &
  \cellcolor[HTML]{F7E984}2.9 &
  \cellcolor[HTML]{A9D380}5.7 &
  \cellcolor[HTML]{FEDA80}-25.8 &
  \cellcolor[HTML]{FEE282}2.1 &
  \cellcolor[HTML]{ECE683}3.2 &
  \cellcolor[HTML]{92CC7E}5.7 &
  \cellcolor[HTML]{FEEA83}0.9 &
  \cellcolor[HTML]{89C97E}-7.9 \\
HM &
  9 K &
  4.7 &
  \cellcolor[HTML]{F8706C}0.0 &
  \cellcolor[HTML]{F9826F}0.0 &
  \cellcolor[HTML]{F98670}0.1 &
  \cellcolor[HTML]{F2E884}2.2 &
  \cellcolor[HTML]{FDCE7E}0.1 &
  \cellcolor[HTML]{FCEB84}-0.6 &
  \cellcolor[HTML]{FBA877}-0.4 &
  \cellcolor[HTML]{FFEB84}-1.0 &
  \cellcolor[HTML]{FA9C74}-0.1 &
  \cellcolor[HTML]{FBB178}-1.0 &
  \cellcolor[HTML]{F9EA84}2.0 &
  \cellcolor[HTML]{FDCE7E}-0.5 &
  \cellcolor[HTML]{FDD27F}1.0 &
  \cellcolor[HTML]{FDC77D}-11.9 &
  \cellcolor[HTML]{DAE182}-0.1 &
  \cellcolor[HTML]{FFEB84}-8.1 &
  \cellcolor[HTML]{F5E884}-0.1 &
  \cellcolor[HTML]{FEDF81}-0.3 &
  \cellcolor[HTML]{FA9172}-2.6 &
  \cellcolor[HTML]{FEE482}1.5 &
  \cellcolor[HTML]{F4E884}2.4 &
  \cellcolor[HTML]{FA9974}0.2 &
  \cellcolor[HTML]{FEE482}0.0 &
  \cellcolor[HTML]{AFD480}-5.3 &
  \cellcolor[HTML]{F0E784}3.8 &
  \cellcolor[HTML]{FEDD81}-0.3 &
  \cellcolor[HTML]{FDCF7E}-6.0 &
  \cellcolor[HTML]{FDCC7E}-0.2 &
  \cellcolor[HTML]{66BF7C}8.6 \\
OpenCQA &
  6K &
  3.9 &
  \cellcolor[HTML]{FED980}1.5 &
  \cellcolor[HTML]{FDD27F}1.1 &
  \cellcolor[HTML]{F9EA84}1.5 &
  \cellcolor[HTML]{FEE182}0.7 &
  \cellcolor[HTML]{FCB97A}-0.9 &
  \cellcolor[HTML]{FEE783}-1.2 &
  \cellcolor[HTML]{FEDC81}2.3 &
  \cellcolor[HTML]{FEE382}-2.0 &
  \cellcolor[HTML]{FEE482}2.7 &
  \cellcolor[HTML]{E6E483}2.8 &
  \cellcolor[HTML]{FED980}0.0 &
  \cellcolor[HTML]{FDCB7D}-0.7 &
  \cellcolor[HTML]{FAEA84}1.8 &
  \cellcolor[HTML]{FFEB84}-9.6 &
  \cellcolor[HTML]{FEDB81}-5.8 &
  \cellcolor[HTML]{FEE683}-10.4 &
  \cellcolor[HTML]{FAEA84}-0.5 &
  \cellcolor[HTML]{F4E884}1.1 &
  \cellcolor[HTML]{FEDD81}1.7 &
  \cellcolor[HTML]{FDD07E}0.8 &
  \cellcolor[HTML]{FFEB84}1.8 &
  \cellcolor[HTML]{63BE7B}10.0 &
  \cellcolor[HTML]{FFEB84}0.3 &
  \cellcolor[HTML]{D8E082}-13.5 &
  \cellcolor[HTML]{FBA776}-5.2 &
  \cellcolor[HTML]{97CD7E}7.4 &
  \cellcolor[HTML]{FFEB84}-4.3 &
  \cellcolor[HTML]{9DCF7F}6.6 &
  \cellcolor[HTML]{BED981}-33.8 \\
LLaVA Conversation &
  57 K &
  3.8 &
  \cellcolor[HTML]{FBAC77}0.9 &
  \cellcolor[HTML]{FA9273}0.2 &
  \cellcolor[HTML]{FDCE7E}0.8 &
  \cellcolor[HTML]{F8726C}-8.9 &
  \cellcolor[HTML]{FBB279}-1.2 &
  \cellcolor[HTML]{FA9774}-12.0 &
  \cellcolor[HTML]{FDC87D}1.2 &
  \cellcolor[HTML]{F98870}-12.8 &
  \cellcolor[HTML]{FBAB77}0.5 &
  \cellcolor[HTML]{FEE582}1.1 &
  \cellcolor[HTML]{F8706C}-9.9 &
  \cellcolor[HTML]{FA9E75}-3.2 &
  \cellcolor[HTML]{FBA476}-0.1 &
  \cellcolor[HTML]{DFE283}-5.5 &
  \cellcolor[HTML]{F86D6B}-23.7 &
  \cellcolor[HTML]{FBAF78}-39.3 &
  \cellcolor[HTML]{FEE683}-1.1 &
  \cellcolor[HTML]{FA9573}-4.5 &
  \cellcolor[HTML]{FCB77A}-0.5 &
  \cellcolor[HTML]{F86B6B}-3.1 &
  \cellcolor[HTML]{FBAA77}0.1 &
  \cellcolor[HTML]{EFE784}3.2 &
  \cellcolor[HTML]{F9806F}-3.6 &
  \cellcolor[HTML]{F8E984}-20.1 &
  \cellcolor[HTML]{83C87D}8.6 &
  \cellcolor[HTML]{FEE081}0.3 &
  \cellcolor[HTML]{F8696B}-12.2 &
  \cellcolor[HTML]{FDD07E}0.0 &
  \cellcolor[HTML]{67C07C}8.2 \\
COCO Caption &
  567 K &
  3.8 &
  \cellcolor[HTML]{63BE7B}10.0 &
  \cellcolor[HTML]{8BCA7E}7.9 &
  \cellcolor[HTML]{B6D680}5.3 &
  \cellcolor[HTML]{F8756D}-8.6 &
  \cellcolor[HTML]{E9E583}2.6 &
  \cellcolor[HTML]{F98A71}-13.8 &
  \cellcolor[HTML]{FFEB84}3.0 &
  \cellcolor[HTML]{F9826F}-13.4 &
  \cellcolor[HTML]{FCBC7B}1.1 &
  \cellcolor[HTML]{FCC37C}-0.3 &
  \cellcolor[HTML]{F8716C}-9.7 &
  \cellcolor[HTML]{FDD17F}-0.3 &
  \cellcolor[HTML]{FDD57F}1.0 &
  \cellcolor[HTML]{DFE283}-5.5 &
  \cellcolor[HTML]{F8736D}-22.7 &
  \cellcolor[HTML]{93CC7E}4.5 &
  \cellcolor[HTML]{FEE683}-1.1 &
  \cellcolor[HTML]{F9816F}-5.7 &
  \cellcolor[HTML]{F7E984}2.9 &
  \cellcolor[HTML]{F8726C}-2.9 &
  \cellcolor[HTML]{FFEB84}1.8 &
  \cellcolor[HTML]{BAD780}5.8 &
  \cellcolor[HTML]{F98370}-3.5 &
  \cellcolor[HTML]{FCBB7A}-33.7 &
  \cellcolor[HTML]{F8E984}3.4 &
  \cellcolor[HTML]{D4DF82}4.4 &
  \cellcolor[HTML]{FBA276}-8.7 &
  \cellcolor[HTML]{9ACE7F}6.8 &
  \cellcolor[HTML]{DCE182}-48.4 \\
TextCaps &
  549 K &
  3.8 &
  \cellcolor[HTML]{DAE182}3.8 &
  \cellcolor[HTML]{D9E082}3.6 &
  \cellcolor[HTML]{63BE7B}10.0 &
  \cellcolor[HTML]{FBB279}-3.4 &
  \cellcolor[HTML]{F8696B}-4.6 &
  \cellcolor[HTML]{FCC27C}-6.2 &
  \cellcolor[HTML]{FEEB84}3.0 &
  \cellcolor[HTML]{FDCD7E}-4.6 &
  \cellcolor[HTML]{FEDE81}2.4 &
  \cellcolor[HTML]{FCBC7A}-0.6 &
  \cellcolor[HTML]{FBB178}-3.7 &
  \cellcolor[HTML]{F8736C}-5.5 &
  \cellcolor[HTML]{FDD07E}0.9 &
  \cellcolor[HTML]{EAE583}-6.8 &
  \cellcolor[HTML]{FEEA83}-3.4 &
  \cellcolor[HTML]{FBA877}-43.1 &
  \cellcolor[HTML]{FEDC81}-1.6 &
  \cellcolor[HTML]{FDD680}-0.8 &
  \cellcolor[HTML]{E2E383}3.9 &
  \cellcolor[HTML]{FFEB84}1.8 &
  \cellcolor[HTML]{FEDA80}1.3 &
  \cellcolor[HTML]{87C97E}8.3 &
  \cellcolor[HTML]{FDCB7D}-0.9 &
  \cellcolor[HTML]{FA9373}-43.9 &
  \cellcolor[HTML]{FCC37C}-1.7 &
  \cellcolor[HTML]{8DCA7E}7.9 &
  \cellcolor[HTML]{FBA877}-8.3 &
  \cellcolor[HTML]{63BE7B}10.0 &
  \cellcolor[HTML]{D6DF82}-45.1 \\
VQAv2 QG &
  444 K &
  3.6 &
  \cellcolor[HTML]{D6E082}3.9 &
  \cellcolor[HTML]{CCDD82}4.3 &
  \cellcolor[HTML]{FDEB84}1.2 &
  \cellcolor[HTML]{FA9B74}-5.3 &
  \cellcolor[HTML]{FFEB84}1.4 &
  \cellcolor[HTML]{FCBF7B}-6.6 &
  \cellcolor[HTML]{FCC17C}0.8 &
  \cellcolor[HTML]{FDC67C}-5.4 &
  \cellcolor[HTML]{FEEB84}3.0 &
  \cellcolor[HTML]{FDD07E}0.3 &
  \cellcolor[HTML]{FA9473}-6.4 &
  \cellcolor[HTML]{DAE182}3.2 &
  \cellcolor[HTML]{FDD880}1.1 &
  \cellcolor[HTML]{FDC97D}-11.7 &
  \cellcolor[HTML]{FA9874}-16.7 &
  \cellcolor[HTML]{FBEA84}-7.6 &
  \cellcolor[HTML]{FEE883}-1.0 &
  \cellcolor[HTML]{FCBD7B}-2.3 &
  \cellcolor[HTML]{FFEB84}2.5 &
  \cellcolor[HTML]{FA9072}-1.7 &
  \cellcolor[HTML]{FCBE7B}0.6 &
  \cellcolor[HTML]{F8E984}2.8 &
  \cellcolor[HTML]{FA9E75}-2.5 &
  \cellcolor[HTML]{FEE382}-23.5 &
  \cellcolor[HTML]{FFEB84}3.1 &
  \cellcolor[HTML]{B0D480}6.2 &
  \cellcolor[HTML]{FCC47C}-6.6 &
  \cellcolor[HTML]{B2D580}5.4 &
  \cellcolor[HTML]{FCBD7B}-78.1 \\
Web CapFilt &
  23,147 K &
  3.6 &
  \cellcolor[HTML]{82C77D}8.4 &
  \cellcolor[HTML]{87C97E}8.0 &
  \cellcolor[HTML]{A4D17F}6.3 &
  \cellcolor[HTML]{FA9B74}-5.3 &
  \cellcolor[HTML]{FEE783}1.2 &
  \cellcolor[HTML]{FCB77A}-7.8 &
  \cellcolor[HTML]{F0E784}3.7 &
  \cellcolor[HTML]{FCC47C}-5.6 &
  \cellcolor[HTML]{D6E082}4.8 &
  \cellcolor[HTML]{FFEB84}1.3 &
  \cellcolor[HTML]{FA9974}-5.9 &
  \cellcolor[HTML]{FDD37F}-0.2 &
  \cellcolor[HTML]{F9EA84}1.9 &
  \cellcolor[HTML]{FDD27F}-11.2 &
  \cellcolor[HTML]{FAA075}-15.5 &
  \cellcolor[HTML]{FDD27F}-21.0 &
  \cellcolor[HTML]{E4E483}1.0 &
  \cellcolor[HTML]{FCB479}-2.8 &
  \cellcolor[HTML]{E8E583}3.6 &
  \cellcolor[HTML]{F8756D}-2.7 &
  \cellcolor[HTML]{FA9373}-0.5 &
  \cellcolor[HTML]{F4E884}3.0 &
  \cellcolor[HTML]{F8726C}-4.1 &
  \cellcolor[HTML]{F8696B}-54.9 &
  \cellcolor[HTML]{FA9072}-7.9 &
  \cellcolor[HTML]{FDD67F}-1.5 &
  \cellcolor[HTML]{F86D6B}-11.9 &
  \cellcolor[HTML]{8CCA7E}7.7 &
  \cellcolor[HTML]{FED880}-70.5 \\
LLaVA Reasoning &
  77 K &
  3.4 &
  \cellcolor[HTML]{F8746D}0.1 &
  \cellcolor[HTML]{F97D6F}-0.1 &
  \cellcolor[HTML]{F86A6B}-0.2 &
  \cellcolor[HTML]{FFEB84}1.5 &
  \cellcolor[HTML]{FCBF7B}-0.6 &
  \cellcolor[HTML]{FDD680}-3.5 &
  \cellcolor[HTML]{FBA676}-0.5 &
  \cellcolor[HTML]{FED980}-3.1 &
  \cellcolor[HTML]{FA8F72}-0.6 &
  \cellcolor[HTML]{FCBA7A}-0.6 &
  \cellcolor[HTML]{FFEB84}1.7 &
  \cellcolor[HTML]{F98A71}-4.3 &
  \cellcolor[HTML]{F8726C}-1.3 &
  \cellcolor[HTML]{FDD07E}-11.3 &
  \cellcolor[HTML]{FEE582}-4.3 &
  \cellcolor[HTML]{F4E884}-6.7 &
  \cellcolor[HTML]{FEE382}-1.3 &
  \cellcolor[HTML]{FFEB84}0.3 &
  \cellcolor[HTML]{FA9B74}-2.0 &
  \cellcolor[HTML]{F8696B}-3.2 &
  \cellcolor[HTML]{F8696B}-1.7 &
  \cellcolor[HTML]{F98570}-0.3 &
  \cellcolor[HTML]{F86A6B}-4.4 &
  \cellcolor[HTML]{C4DA81}-9.6 &
  \cellcolor[HTML]{FEE282}2.1 &
  \cellcolor[HTML]{FEE382}0.8 &
  \cellcolor[HTML]{FCC47C}-6.6 &
  \cellcolor[HTML]{FEDC81}0.4 &
  \cellcolor[HTML]{F97E6F}-96.2 \\
OK-VQA QG &
  9 K &
  3.1 &
  \cellcolor[HTML]{FFEB84}1.8 &
  \cellcolor[HTML]{FEE683}1.4 &
  \cellcolor[HTML]{FBAC78}0.4 &
  \cellcolor[HTML]{FA9172}-6.2 &
  \cellcolor[HTML]{FAA075}-2.1 &
  \cellcolor[HTML]{FBAE78}-9.0 &
  \cellcolor[HTML]{FA9F75}-0.9 &
  \cellcolor[HTML]{FCB679}-7.4 &
  \cellcolor[HTML]{FCB579}0.9 &
  \cellcolor[HTML]{FBAA77}-1.2 &
  \cellcolor[HTML]{F98971}-7.5 &
  \cellcolor[HTML]{FA9373}-3.7 &
  \cellcolor[HTML]{FFEB84}1.5 &
  \cellcolor[HTML]{F87A6E}-16.8 &
  \cellcolor[HTML]{FBAD78}-13.4 &
  \cellcolor[HTML]{FEDB80}-16.4 &
  \cellcolor[HTML]{FBA576}-4.6 &
  \cellcolor[HTML]{FAA075}-3.9 &
  \cellcolor[HTML]{FA9D75}-1.9 &
  \cellcolor[HTML]{F97D6F}-2.4 &
  \cellcolor[HTML]{F98971}-0.8 &
  \cellcolor[HTML]{FBEA84}2.7 &
  \cellcolor[HTML]{F98770}-3.3 &
  \cellcolor[HTML]{FFEB84}-21.6 &
  \cellcolor[HTML]{FBA476}-5.5 &
  \cellcolor[HTML]{63BE7B}10.0 &
  \cellcolor[HTML]{D6E082}-0.5 &
  \cellcolor[HTML]{AAD380}5.9 &
  \cellcolor[HTML]{FFEB84}-65.3 \\
\vlbenchmark &
  7K &
  3.0 &
  \cellcolor[HTML]{F8696B}-0.1 &
  \cellcolor[HTML]{F86E6C}-0.3 &
  \cellcolor[HTML]{F98870}0.1 &
  \cellcolor[HTML]{FCC37C}-1.9 &
  \cellcolor[HTML]{FBAF78}-1.4 &
  \cellcolor[HTML]{FBA576}-10.2 &
  \cellcolor[HTML]{FA9A74}-1.1 &
  \cellcolor[HTML]{FBA676}-9.2 &
  \cellcolor[HTML]{FBA777}0.3 &
  \cellcolor[HTML]{FCC37C}-0.3 &
  \cellcolor[HTML]{FCBE7B}-2.5 &
  \cellcolor[HTML]{FDC77D}-0.9 &
  \cellcolor[HTML]{FAA075}-0.2 &
  \cellcolor[HTML]{F98A71}-15.8 &
  \cellcolor[HTML]{FBAD78}-13.4 &
  \cellcolor[HTML]{FA9C74}-49.6 &
  \cellcolor[HTML]{FBAE78}-4.1 &
  \cellcolor[HTML]{FCBE7B}-2.2 &
  \cellcolor[HTML]{FCB379}-0.7 &
  \cellcolor[HTML]{FBA476}-0.9 &
  \cellcolor[HTML]{FBB178}0.3 &
  \cellcolor[HTML]{FEDD81}2.1 &
  \cellcolor[HTML]{F97E6F}-3.7 &
  \cellcolor[HTML]{FAEA84}-20.5 &
  \cellcolor[HTML]{FBA175}-5.9 &
  \cellcolor[HTML]{E6E483}3.5 &
  \cellcolor[HTML]{E2E383}-1.6 &
  \cellcolor[HTML]{FFEB84}1.0 &
  \cellcolor[HTML]{71C27C}3.5 \\
A-OKVQA QG &
  17 K &
  2.9 &
  \cellcolor[HTML]{FBA676}0.8 &
  \cellcolor[HTML]{FA9473}0.2 &
  \cellcolor[HTML]{FA9673}0.2 &
  \cellcolor[HTML]{FA8E72}-6.4 &
  \cellcolor[HTML]{FCB479}-1.1 &
  \cellcolor[HTML]{FA9773}-12.1 &
  \cellcolor[HTML]{FBA977}-0.4 &
  \cellcolor[HTML]{FBA276}-9.7 &
  \cellcolor[HTML]{FA8F72}-0.6 &
  \cellcolor[HTML]{FBA175}-1.6 &
  \cellcolor[HTML]{F98871}-7.6 &
  \cellcolor[HTML]{FBAF78}-2.2 &
  \cellcolor[HTML]{FCB97A}0.4 &
  \cellcolor[HTML]{F7E984}-8.6 &
  \cellcolor[HTML]{FBA676}-14.4 &
  \cellcolor[HTML]{FEEA83}-8.5 &
  \cellcolor[HTML]{B7D780}4.2 &
  \cellcolor[HTML]{FA9573}-4.5 &
  \cellcolor[HTML]{F98D71}-2.9 &
  \cellcolor[HTML]{F87A6E}-2.5 &
  \cellcolor[HTML]{F9816F}-1.0 &
  \cellcolor[HTML]{FCBC7B}1.2 &
  \cellcolor[HTML]{FA9072}-3.0 &
  \cellcolor[HTML]{FCC47C}-31.4 &
  \cellcolor[HTML]{F8696B}-12.8 &
  \cellcolor[HTML]{DBE182}4.0 &
  \cellcolor[HTML]{FAA075}-8.8 &
  \cellcolor[HTML]{B8D780}5.1 &
  \cellcolor[HTML]{FBAD78}-82.7 \\
LLaVA Description &
  23 K &
  2.3 &
  \cellcolor[HTML]{F8696B}-0.1 &
  \cellcolor[HTML]{F8696B}-0.4 &
  \cellcolor[HTML]{F8696B}-0.3 &
  \cellcolor[HTML]{F8696B}-9.7 &
  \cellcolor[HTML]{F86E6C}-4.3 &
  \cellcolor[HTML]{F8696B}-18.4 &
  \cellcolor[HTML]{F8696B}-3.7 &
  \cellcolor[HTML]{F8696B}-16.5 &
  \cellcolor[HTML]{F8696B}-2.1 &
  \cellcolor[HTML]{F8696B}-3.9 &
  \cellcolor[HTML]{F8696B}-10.5 &
  \cellcolor[HTML]{F8696B}-6.1 &
  \cellcolor[HTML]{F8696B}-1.5 &
  \cellcolor[HTML]{FCC07B}-12.3 &
  \cellcolor[HTML]{F8696B}-24.5 &
  \cellcolor[HTML]{FDCA7D}-25.5 &
  \cellcolor[HTML]{F8696B}-7.7 &
  \cellcolor[HTML]{F8696B}-7.0 &
  \cellcolor[HTML]{F8696B}-4.9 &
  \cellcolor[HTML]{F8696B}-3.2 &
  \cellcolor[HTML]{FEDC81}1.4 &
  \cellcolor[HTML]{F9806F}-0.5 &
  \cellcolor[HTML]{F8696B}-4.4 &
  \cellcolor[HTML]{FCB97A}-34.2 &
  \cellcolor[HTML]{FBA776}-5.2 &
  \cellcolor[HTML]{FEEA83}2.2 &
  \cellcolor[HTML]{FDCF7E}-6.0 &
  \cellcolor[HTML]{FDD07E}0.0 &
  \cellcolor[HTML]{F86D6B}-100.7
           \\
			\bottomrule
		\end{tabular}%
	}
	\caption
	{Normalized transfer learning performance of LLaVA. Higher values indicate better transferability. The rows are sorted in descending order of average performance. We multiply the values by a factor of 10 to aid visualization. The highest performance in each column is 10. 
  QG denotes question generation, MC denotes multiple-choice and G denotes open-ended generation. The color scale is normalized along each column. The colors represent values in descending order: green, yellow, orange and red. }
	\label{eval:affinity_matrix_norm_llava_landscape}
\end{table}
\end{landscape}

\begin{landscape}
\begin{table}[!tp]
	\resizebox{\columnwidth}{!}{%
		\begin{tabular}{c|c|c|ccccccccccccccccccccccccccccc}
			\toprule
			Source   Task &
			\setlength\extrarowheight{0pt}\begin{tabular}[c]{@{}c@{}}Dataset\\ Size \end{tabular} &
			\setlength\extrarowheight{0pt}\begin{tabular}[c]{@{}c@{}}AHP\\ Ranking \\ Score \end{tabular}  &
			\multicolumn{29}{c}{Target Task} \\
			&
			&
			&
			\setlength\extrarowheight{0pt}\begin{tabular}[c]{@{}c@{}}COCO\\ Caption \end{tabular}  &
			\setlength\extrarowheight{0pt}\begin{tabular}[c]{@{}c@{}}Flickr\\ 30k\end{tabular}  &
			\setlength\extrarowheight{0pt}\begin{tabular}[c]{@{}c@{}}Text\\ Caps\end{tabular}  &
			\multicolumn{2}{c}{VQAv2} &
			\multicolumn{2}{c}{OK-VQA} &
			\multicolumn{2}{c}{A-OKVQA} &
			\setlength\extrarowheight{0pt}\begin{tabular}[c]{@{}c@{}}Science\\ QA\end{tabular} &
			\multicolumn{2}{c}{GQA} &
			\setlength\extrarowheight{0pt}\begin{tabular}[c]{@{}c@{}}Icon\\ QA \end{tabular} &
			VSR &
			\multicolumn{2}{c}{CLEVR} &
			\setlength\extrarowheight{0pt}\begin{tabular}[c]{@{}c@{}}RAVEN-\\ FAIR \end{tabular} &
			\multicolumn{2}{c}{\setlength\extrarowheight{0pt}\begin{tabular}[c]{@{}c@{}}Text\\ VQA\end{tabular}} &
			\multicolumn{2}{c}{\setlength\extrarowheight{0pt}\begin{tabular}[c]{@{}c@{}}OCR-\\ VQA\end{tabular}} &
			\setlength\extrarowheight{0pt}\begin{tabular}[c]{@{}c@{}}Open\\ CQA \end{tabular} &
			\multicolumn{2}{c}{\setlength\extrarowheight{0pt}\begin{tabular}[c]{@{}c@{}}Chart\\ QA\end{tabular}} &
			HM &
			\setlength\extrarowheight{0pt}\begin{tabular}[c]{@{}c@{}}NY\\ Explain \end{tabular} &
			\setlength\extrarowheight{0pt}\begin{tabular}[c]{@{}c@{}}NY\\ Rank \end{tabular} &
			MORE &
			\vlbenchmark \\
			 &
			&
			&
			&
			&
			&
			G &
			MC &
			G &
			MC &
			G &
			MC &
			&
			G &
			MC &
			&
			&
			G &
			MC &
			&
			G &
			MC &
			G &
			MC &
			&
			G &
			MC &
			&
			&
			&
			&
			\\
			\midrule
VQAv2 &
  444 K &
  5.7 &
  \cellcolor[HTML]{FDCF7E}1.5 &
  \cellcolor[HTML]{FEDC81}1.1 &
  \cellcolor[HTML]{FDC97D}1.7 &
  \cellcolor[HTML]{63BE7B}10.0 &
  \cellcolor[HTML]{63BE7B}10.0 &
  \cellcolor[HTML]{6FC27C}9.3 &
  \cellcolor[HTML]{7EC67D}8.5 &
  \cellcolor[HTML]{6BC17C}9.5 &
  \cellcolor[HTML]{8ECB7E}7.9 &
  \cellcolor[HTML]{DBE182}2.8 &
  \cellcolor[HTML]{77C47D}8.9 &
  \cellcolor[HTML]{98CE7F}6.7 &
  \cellcolor[HTML]{FCEB84}0.8 &
  \cellcolor[HTML]{67BF7C}9.8 &
  \cellcolor[HTML]{6AC07C}9.6 &
  \cellcolor[HTML]{EAE583}1.9 &
  \cellcolor[HTML]{F5E884}0.7 &
  \cellcolor[HTML]{84C87D}8.1 &
  \cellcolor[HTML]{95CD7E}7.3 &
  \cellcolor[HTML]{A4D17F}5.9 &
  \cellcolor[HTML]{CFDD82}4.1 &
  \cellcolor[HTML]{FA9A74}-8.5 &
  \cellcolor[HTML]{7EC67D}9.0 &
  \cellcolor[HTML]{FEE583}-3.9 &
  \cellcolor[HTML]{FCEB84}1.0 &
  \cellcolor[HTML]{FA9F75}-19.5 &
  \cellcolor[HTML]{FDCD7E}-5.0 &
  \cellcolor[HTML]{FA9C74}0.5 &
  \cellcolor[HTML]{F98B71}0.3 \\
A-OKVQA (MC) &
  17 K &
  5.6 &
  \cellcolor[HTML]{FFEB84}2.1 &
  \cellcolor[HTML]{FCEA84}1.6 &
  \cellcolor[HTML]{F9EA84}2.6 &
  \cellcolor[HTML]{7EC67D}8.5 &
  \cellcolor[HTML]{BBD881}5.2 &
  \cellcolor[HTML]{89C97E}7.7 &
  \cellcolor[HTML]{78C47D}8.8 &
  \cellcolor[HTML]{87C97E}7.8 &
  \cellcolor[HTML]{63BE7B}10.0 &
  \cellcolor[HTML]{C8DC81}4.0 &
  \cellcolor[HTML]{97CD7E}7.0 &
  \cellcolor[HTML]{FEE582}-0.6 &
  \cellcolor[HTML]{F0E784}1.5 &
  \cellcolor[HTML]{FDCA7D}1.3 &
  \cellcolor[HTML]{E8E583}1.6 &
  \cellcolor[HTML]{FCBF7B}-1.2 &
  \cellcolor[HTML]{FFEB84}0.0 &
  \cellcolor[HTML]{ACD380}5.7 &
  \cellcolor[HTML]{63BE7B}10.0 &
  \cellcolor[HTML]{FCEA84}0.4 &
  \cellcolor[HTML]{DCE182}3.3 &
  \cellcolor[HTML]{E7E483}-2.2 &
  \cellcolor[HTML]{D4DF82}5.7 &
  \cellcolor[HTML]{CADC81}1.6 &
  \cellcolor[HTML]{FEE282}0.7 &
  \cellcolor[HTML]{94CC7E}6.3 &
  \cellcolor[HTML]{BED981}2.5 &
  \cellcolor[HTML]{96CD7E}8.4 &
  \cellcolor[HTML]{D6DF82}3.9 \\
ScienceQA &
  6 K &
  5.5 &
  \cellcolor[HTML]{DFE283}3.8 &
  \cellcolor[HTML]{D1DE82}4.0 &
  \cellcolor[HTML]{EDE683}3.2 &
  \cellcolor[HTML]{97CD7E}7.1 &
  \cellcolor[HTML]{FDD07E}-1.7 &
  \cellcolor[HTML]{A6D27F}5.9 &
  \cellcolor[HTML]{C2DA81}4.6 &
  \cellcolor[HTML]{B1D580}5.2 &
  \cellcolor[HTML]{C4DA81}5.2 &
  \cellcolor[HTML]{63BE7B}10.0 &
  \cellcolor[HTML]{9FD07F}6.5 &
  \cellcolor[HTML]{FDD47F}-2.7 &
  \cellcolor[HTML]{F8E984}1.0 &
  \cellcolor[HTML]{FBB178}-0.2 &
  \cellcolor[HTML]{93CC7E}6.9 &
  \cellcolor[HTML]{F8696B}-4.9 &
  \cellcolor[HTML]{63BE7B}10.0 &
  \cellcolor[HTML]{B3D680}5.3 &
  \cellcolor[HTML]{9FD07F}6.7 &
  \cellcolor[HTML]{F7E984}0.7 &
  \cellcolor[HTML]{FFEB84}1.4 &
  \cellcolor[HTML]{FBA576}-8.0 &
  \cellcolor[HTML]{F4E884}4.5 &
  \cellcolor[HTML]{FCC47C}-11.2 &
  \cellcolor[HTML]{FEDA80}0.6 &
  \cellcolor[HTML]{63BE7B}10.0 &
  \cellcolor[HTML]{63BE7B}10.0 &
  \cellcolor[HTML]{70C27C}9.6 &
  \cellcolor[HTML]{F86D6B}-0.2 \\
A-OKVQA &
  17 K &
  5.3 &
  \cellcolor[HTML]{FCC27C}1.2 &
  \cellcolor[HTML]{FDC87D}0.6 &
  \cellcolor[HTML]{FBA877}1.2 &
  \cellcolor[HTML]{7BC57D}8.7 &
  \cellcolor[HTML]{B0D580}5.7 &
  \cellcolor[HTML]{6EC17C}9.4 &
  \cellcolor[HTML]{63BE7B}10.0 &
  \cellcolor[HTML]{63BE7B}10.0 &
  \cellcolor[HTML]{92CC7E}7.7 &
  \cellcolor[HTML]{DCE182}2.8 &
  \cellcolor[HTML]{8ACA7E}7.8 &
  \cellcolor[HTML]{E2E383}2.0 &
  \cellcolor[HTML]{F7E984}1.1 &
  \cellcolor[HTML]{7AC57D}9.0 &
  \cellcolor[HTML]{77C47D}8.8 &
  \cellcolor[HTML]{FCB77A}-1.6 &
  \cellcolor[HTML]{FEE382}-0.3 &
  \cellcolor[HTML]{89C97E}7.8 &
  \cellcolor[HTML]{B2D580}5.7 &
  \cellcolor[HTML]{C6DB81}3.8 &
  \cellcolor[HTML]{D6DF82}3.7 &
  \cellcolor[HTML]{F98470}-9.6 &
  \cellcolor[HTML]{87C97E}8.6 &
  \cellcolor[HTML]{FBEA84}-2.4 &
  \cellcolor[HTML]{F6E984}1.3 &
  \cellcolor[HTML]{FDCF7E}-8.5 &
  \cellcolor[HTML]{FED980}-4.2 &
  \cellcolor[HTML]{FCC57C}3.0 &
  \cellcolor[HTML]{F98770}0.2 \\
OK-VQA &
  9 K &
  5.1 &
  \cellcolor[HTML]{FAA075}0.4 &
  \cellcolor[HTML]{FBA175}-0.4 &
  \cellcolor[HTML]{FA9A74}0.9 &
  \cellcolor[HTML]{84C87D}8.2 &
  \cellcolor[HTML]{D3DF82}3.8 &
  \cellcolor[HTML]{63BE7B}10.0 &
  \cellcolor[HTML]{73C37C}9.1 &
  \cellcolor[HTML]{83C77D}8.1 &
  \cellcolor[HTML]{A8D27F}6.6 &
  \cellcolor[HTML]{E1E383}2.5 &
  \cellcolor[HTML]{92CC7E}7.3 &
  \cellcolor[HTML]{FFEB84}0.1 &
  \cellcolor[HTML]{FBB078}-0.1 &
  \cellcolor[HTML]{9ACE7F}7.6 &
  \cellcolor[HTML]{6FC27C}9.3 &
  \cellcolor[HTML]{FBA776}-2.3 &
  \cellcolor[HTML]{F8696B}-5.1 &
  \cellcolor[HTML]{8CCA7E}7.6 &
  \cellcolor[HTML]{9FD07F}6.7 &
  \cellcolor[HTML]{D5DF82}2.8 &
  \cellcolor[HTML]{FEEA83}1.3 &
  \cellcolor[HTML]{F9806F}-9.8 &
  \cellcolor[HTML]{93CC7E}8.2 &
  \cellcolor[HTML]{F2E884}-1.7 &
  \cellcolor[HTML]{F2E884}1.5 &
  \cellcolor[HTML]{FCB77A}-14.0 &
  \cellcolor[HTML]{96CD7E}5.8 &
  \cellcolor[HTML]{FDCF7E}3.5 &
  \cellcolor[HTML]{F97E6F}0.1 \\
OCR-VQA &
  802 K &
  5.0 &
  \cellcolor[HTML]{AAD380}6.4 &
  \cellcolor[HTML]{B8D780}5.4 &
  \cellcolor[HTML]{DAE182}4.1 &
  \cellcolor[HTML]{E9E583}2.4 &
  \cellcolor[HTML]{C1DA81}4.8 &
  \cellcolor[HTML]{DFE283}2.4 &
  \cellcolor[HTML]{E1E383}2.9 &
  \cellcolor[HTML]{E2E383}2.2 &
  \cellcolor[HTML]{F5E984}2.8 &
  \cellcolor[HTML]{DEE283}2.6 &
  \cellcolor[HTML]{F1E784}1.8 &
  \cellcolor[HTML]{BFD981}4.2 &
  \cellcolor[HTML]{FDC87D}0.2 &
  \cellcolor[HTML]{7CC57D}8.9 &
  \cellcolor[HTML]{E7E483}1.6 &
  \cellcolor[HTML]{D2DE82}3.4 &
  \cellcolor[HTML]{FFEB84}0.0 &
  \cellcolor[HTML]{BBD881}4.9 &
  \cellcolor[HTML]{FFEB84}1.4 &
  \cellcolor[HTML]{63BE7B}10.0 &
  \cellcolor[HTML]{63BE7B}10.0 &
  \cellcolor[HTML]{FDD780}-5.5 &
  \cellcolor[HTML]{9ECF7F}7.8 &
  \cellcolor[HTML]{E2E383}-0.4 &
  \cellcolor[HTML]{FCBA7A}0.2 &
  \cellcolor[HTML]{A8D27F}4.7 &
  \cellcolor[HTML]{FEE582}-3.3 &
  \cellcolor[HTML]{E0E283}6.2 &
  \cellcolor[HTML]{FFEB84}1.7 \\
GQA &
  943 K &
  4.9 &
  \cellcolor[HTML]{FA9974}0.2 &
  \cellcolor[HTML]{FA9373}-0.7 &
  \cellcolor[HTML]{FA9E75}1.0 &
  \cellcolor[HTML]{79C57D}8.8 &
  \cellcolor[HTML]{89C97E}7.9 &
  \cellcolor[HTML]{94CC7E}7.0 &
  \cellcolor[HTML]{97CD7E}7.1 &
  \cellcolor[HTML]{86C97E}7.9 &
  \cellcolor[HTML]{A9D27F}6.6 &
  \cellcolor[HTML]{F2E884}1.5 &
  \cellcolor[HTML]{63BE7B}10.0 &
  \cellcolor[HTML]{63BE7B}10.0 &
  \cellcolor[HTML]{FCEB84}0.8 &
  \cellcolor[HTML]{FCBC7A}0.5 &
  \cellcolor[HTML]{63BE7B}10.0 &
  \cellcolor[HTML]{FEE282}0.3 &
  \cellcolor[HTML]{FFEB84}0.0 &
  \cellcolor[HTML]{ADD480}5.7 &
  \cellcolor[HTML]{FEE683}1.0 &
  \cellcolor[HTML]{B7D780}4.7 &
  \cellcolor[HTML]{F4E884}2.0 &
  \cellcolor[HTML]{F98C71}-9.2 &
  \cellcolor[HTML]{A1D07F}7.7 &
  \cellcolor[HTML]{FEE282}-4.7 &
  \cellcolor[HTML]{FFEB84}0.8 &
  \cellcolor[HTML]{FEDA80}-6.0 &
  \cellcolor[HTML]{FBAF78}-7.1 &
  \cellcolor[HTML]{FCBE7B}2.5 &
  \cellcolor[HTML]{F98670}0.2 \\
HM &
  9 K &
  4.8 &
  \cellcolor[HTML]{99CE7F}7.3 &
  \cellcolor[HTML]{A2D07F}6.6 &
  \cellcolor[HTML]{D6DF82}4.3 &
  \cellcolor[HTML]{98CE7F}7.0 &
  \cellcolor[HTML]{DEE283}3.2 &
  \cellcolor[HTML]{C9DC81}3.7 &
  \cellcolor[HTML]{FEDB81}0.7 &
  \cellcolor[HTML]{CEDD82}3.4 &
  \cellcolor[HTML]{FEDA80}1.6 &
  \cellcolor[HTML]{FFEB84}0.7 &
  \cellcolor[HTML]{A2D17F}6.3 &
  \cellcolor[HTML]{E0E283}2.1 &
  \cellcolor[HTML]{FDCA7D}0.2 &
  \cellcolor[HTML]{FFEB84}3.1 &
  \cellcolor[HTML]{9ECF7F}6.3 &
  \cellcolor[HTML]{FEDF81}0.2 &
  \cellcolor[HTML]{E8E583}1.5 &
  \cellcolor[HTML]{CEDD82}3.7 &
  \cellcolor[HTML]{FDD17F}-1.1 &
  \cellcolor[HTML]{C0D981}4.1 &
  \cellcolor[HTML]{FDD780}-0.3 &
  \cellcolor[HTML]{FDD17F}-5.8 &
  \cellcolor[HTML]{9ACE7F}7.9 &
  \cellcolor[HTML]{FEDC81}-6.1 &
  \cellcolor[HTML]{63BE7B}10.0 &
  \cellcolor[HTML]{FAEA84}-1.6 &
  \cellcolor[HTML]{FA9272}-9.2 &
  \cellcolor[HTML]{A7D27F}7.9 &
  \cellcolor[HTML]{D6E082}3.9 \\
OpenCQA &
  6K &
  4.6 &
  \cellcolor[HTML]{FCC47C}1.2 &
  \cellcolor[HTML]{FDD07E}0.8 &
  \cellcolor[HTML]{FBA376}1.1 &
  \cellcolor[HTML]{F86B6B}0.0 &
  \cellcolor[HTML]{FFEB84}1.3 &
  \cellcolor[HTML]{F8766D}0.0 &
  \cellcolor[HTML]{FFEB84}1.1 &
  \cellcolor[HTML]{F86E6B}0.0 &
  \cellcolor[HTML]{FDD17F}1.2 &
  \cellcolor[HTML]{E6E483}2.2 &
  \cellcolor[HTML]{F8696B}0.0 &
  \cellcolor[HTML]{E2E383}2.0 &
  \cellcolor[HTML]{FCB579}-0.1 &
  \cellcolor[HTML]{FBB178}-0.2 &
  \cellcolor[HTML]{F8696B}0.0 &
  \cellcolor[HTML]{E7E483}2.1 &
  \cellcolor[HTML]{FFEB84}0.0 &
  \cellcolor[HTML]{F98770}0.0 &
  \cellcolor[HTML]{D1DE82}4.0 &
  \cellcolor[HTML]{F98A71}0.0 &
  \cellcolor[HTML]{F7E984}1.8 &
  \cellcolor[HTML]{63BE7B}10.0 &
  \cellcolor[HTML]{F8696B}-2.3 &
  \cellcolor[HTML]{63BE7B}10.0 &
  \cellcolor[HTML]{F2E884}1.5 &
  \cellcolor[HTML]{B2D580}3.9 &
  \cellcolor[HTML]{A0D07F}5.0 &
  \cellcolor[HTML]{F7E984}5.4 &
  \cellcolor[HTML]{BFD981}5.1 \\
Flickr30k &
  145 K &
  4.5 &
  \cellcolor[HTML]{8FCB7E}7.8 &
  \cellcolor[HTML]{63BE7B}10.0 &
  \cellcolor[HTML]{AFD480}6.2 &
  \cellcolor[HTML]{F8786E}0.1 &
  \cellcolor[HTML]{FCEA84}1.5 &
  \cellcolor[HTML]{FCC17B}0.2 &
  \cellcolor[HTML]{DAE182}3.2 &
  \cellcolor[HTML]{FA9B74}0.2 &
  \cellcolor[HTML]{DEE283}3.9 &
  \cellcolor[HTML]{FEE883}0.6 &
  \cellcolor[HTML]{F87B6E}0.1 &
  \cellcolor[HTML]{EBE683}1.4 &
  \cellcolor[HTML]{FED880}0.3 &
  \cellcolor[HTML]{C0D981}5.9 &
  \cellcolor[HTML]{FBB078}0.0 &
  \cellcolor[HTML]{FDC97D}-0.8 &
  \cellcolor[HTML]{B2D580}5.0 &
  \cellcolor[HTML]{FFEB84}0.8 &
  \cellcolor[HTML]{F7E984}1.9 &
  \cellcolor[HTML]{FDC67C}0.1 &
  \cellcolor[HTML]{F6E984}1.9 &
  \cellcolor[HTML]{E6E483}-2.2 &
  \cellcolor[HTML]{FFEB84}4.1 &
  \cellcolor[HTML]{CADC81}1.6 &
  \cellcolor[HTML]{FBAA77}0.0 &
  \cellcolor[HTML]{FFEB84}-2.1 &
  \cellcolor[HTML]{FCC47C}-5.6 &
  \cellcolor[HTML]{F7E984}5.4 &
  \cellcolor[HTML]{E9E583}2.9 \\
IconQA &
  19 K &
  4.5 &
  \cellcolor[HTML]{E6E483}3.4 &
  \cellcolor[HTML]{E5E483}2.9 &
  \cellcolor[HTML]{F3E884}2.9 &
  \cellcolor[HTML]{E8E583}2.5 &
  \cellcolor[HTML]{FED880}-0.8 &
  \cellcolor[HTML]{ECE683}1.5 &
  \cellcolor[HTML]{FDCB7D}0.2 &
  \cellcolor[HTML]{EDE683}1.5 &
  \cellcolor[HTML]{FEE883}2.2 &
  \cellcolor[HTML]{FEDC81}0.4 &
  \cellcolor[HTML]{EBE683}2.1 &
  \cellcolor[HTML]{FEE783}-0.4 &
  \cellcolor[HTML]{63BE7B}10.0 &
  \cellcolor[HTML]{FDC97D}1.2 &
  \cellcolor[HTML]{B0D480}5.2 &
  \cellcolor[HTML]{C4DA81}4.2 &
  \cellcolor[HTML]{85C87D}7.8 &
  \cellcolor[HTML]{EDE683}1.9 &
  \cellcolor[HTML]{FCC07B}-2.9 &
  \cellcolor[HTML]{EFE784}1.2 &
  \cellcolor[HTML]{F8726C}-8.6 &
  \cellcolor[HTML]{FBA676}-7.9 &
  \cellcolor[HTML]{E5E483}5.1 &
  \cellcolor[HTML]{FCBF7B}-12.3 &
  \cellcolor[HTML]{F8696B}-0.8 &
  \cellcolor[HTML]{9ACE7F}5.8 &
  \cellcolor[HTML]{7DC67D}7.9 &
  \cellcolor[HTML]{C6DB81}7.0 &
  \cellcolor[HTML]{C1D981}5.0 \\
VSR &
  3 K &
  4.5 &
  \cellcolor[HTML]{EFE784}3.0 &
  \cellcolor[HTML]{E5E483}2.9 &
  \cellcolor[HTML]{FBAB77}1.2 &
  \cellcolor[HTML]{FBA877}0.5 &
  \cellcolor[HTML]{E7E483}2.7 &
  \cellcolor[HTML]{FFEB84}0.4 &
  \cellcolor[HTML]{FDD17F}0.4 &
  \cellcolor[HTML]{FFEB84}0.4 &
  \cellcolor[HTML]{F7E984}2.7 &
  \cellcolor[HTML]{FEE482}0.5 &
  \cellcolor[HTML]{FCC37C}0.7 &
  \cellcolor[HTML]{FFEB84}0.1 &
  \cellcolor[HTML]{F4E884}1.3 &
  \cellcolor[HTML]{63BE7B}10.0 &
  \cellcolor[HTML]{F8696B}0.0 &
  \cellcolor[HTML]{6FC27C}9.3 &
  \cellcolor[HTML]{F9EA84}0.4 &
  \cellcolor[HTML]{FA9E75}0.2 &
  \cellcolor[HTML]{F9EA84}1.8 &
  \cellcolor[HTML]{F8796E}0.0 &
  \cellcolor[HTML]{FDEB84}1.5 &
  \cellcolor[HTML]{B0D580}2.9 &
  \cellcolor[HTML]{FBAF78}1.2 &
  \cellcolor[HTML]{E9E583}-0.9 &
  \cellcolor[HTML]{FEEB84}0.9 &
  \cellcolor[HTML]{AAD380}4.5 &
  \cellcolor[HTML]{FCC47C}-5.6 &
  \cellcolor[HTML]{FEE182}4.6 &
  \cellcolor[HTML]{FEE482}1.6 \\
TextVQA &
  35 K &
  4.5 &
  \cellcolor[HTML]{F8696B}-0.9 &
  \cellcolor[HTML]{F8696B}-1.8 &
  \cellcolor[HTML]{FA8E72}0.7 &
  \cellcolor[HTML]{BDD881}4.9 &
  \cellcolor[HTML]{D6DF82}3.6 &
  \cellcolor[HTML]{A4D17F}6.0 &
  \cellcolor[HTML]{BCD881}5.0 &
  \cellcolor[HTML]{9FD07F}6.4 &
  \cellcolor[HTML]{D2DE82}4.5 &
  \cellcolor[HTML]{D5DF82}3.2 &
  \cellcolor[HTML]{BAD780}5.0 &
  \cellcolor[HTML]{D0DE82}3.1 &
  \cellcolor[HTML]{F9EA84}0.9 &
  \cellcolor[HTML]{ADD480}6.7 &
  \cellcolor[HTML]{7AC57D}8.6 &
  \cellcolor[HTML]{F8756D}-4.4 &
  \cellcolor[HTML]{F0E784}1.0 &
  \cellcolor[HTML]{63BE7B}10.0 &
  \cellcolor[HTML]{81C77D}8.4 &
  \cellcolor[HTML]{A1D07F}6.1 &
  \cellcolor[HTML]{FA9874}-5.4 &
  \cellcolor[HTML]{F8696B}-11.0 &
  \cellcolor[HTML]{63BE7B}10.0 &
  \cellcolor[HTML]{FCBF7B}-12.3 &
  \cellcolor[HTML]{F9EA84}1.2 &
  \cellcolor[HTML]{F8696B}-32.0 &
  \cellcolor[HTML]{FFEB84}-2.9 &
  \cellcolor[HTML]{F8696B}-2.5 &
  \cellcolor[HTML]{F98670}0.2 \\
VQAv2 QG &
  444 K &
  4.3 &
  \cellcolor[HTML]{B6D680}5.9 &
  \cellcolor[HTML]{C1D981}4.9 &
  \cellcolor[HTML]{D4DF82}4.4 &
  \cellcolor[HTML]{F2E884}1.9 &
  \cellcolor[HTML]{FDC87D}-2.7 &
  \cellcolor[HTML]{DCE182}2.5 &
  \cellcolor[HTML]{FBAF78}-0.6 &
  \cellcolor[HTML]{DEE283}2.5 &
  \cellcolor[HTML]{FBA676}-0.7 &
  \cellcolor[HTML]{FEEB84}0.7 &
  \cellcolor[HTML]{FAEA84}1.3 &
  \cellcolor[HTML]{FDD27F}-2.8 &
  \cellcolor[HTML]{FCEA84}0.8 &
  \cellcolor[HTML]{FBA777}-0.7 &
  \cellcolor[HTML]{F6E984}0.7 &
  \cellcolor[HTML]{FEDD81}0.1 &
  \cellcolor[HTML]{FFEB84}0.0 &
  \cellcolor[HTML]{EDE683}1.9 &
  \cellcolor[HTML]{FDC87D}-2.1 &
  \cellcolor[HTML]{FFEB84}0.2 &
  \cellcolor[HTML]{E9E583}2.6 &
  \cellcolor[HTML]{DDE182}-1.3 &
  \cellcolor[HTML]{FEE482}3.7 &
  \cellcolor[HTML]{EAE583}-1.1 &
  \cellcolor[HTML]{FCBA7A}0.2 &
  \cellcolor[HTML]{B1D580}4.0 &
  \cellcolor[HTML]{DAE182}0.2 &
  \cellcolor[HTML]{83C87D}9.0 &
  \cellcolor[HTML]{FBAD78}0.8 \\
A-OKVQA QG &
  17 K &
  4.2 &
  \cellcolor[HTML]{FFEB84}2.2 &
  \cellcolor[HTML]{FFEB84}1.4 &
  \cellcolor[HTML]{F3E884}2.9 &
  \cellcolor[HTML]{FFEB84}1.1 &
  \cellcolor[HTML]{FDD17F}-1.7 &
  \cellcolor[HTML]{FA9373}0.1 &
  \cellcolor[HTML]{FDC87D}0.2 &
  \cellcolor[HTML]{FBA275}0.2 &
  \cellcolor[HTML]{FBAF78}-0.3 &
  \cellcolor[HTML]{FEE482}0.5 &
  \cellcolor[HTML]{FFEB84}0.9 &
  \cellcolor[HTML]{FDD880}-2.2 &
  \cellcolor[HTML]{FFEB84}0.6 &
  \cellcolor[HTML]{F7E984}3.5 &
  \cellcolor[HTML]{FFEB84}0.1 &
  \cellcolor[HTML]{B5D680}5.1 &
  \cellcolor[HTML]{FFEB84}0.0 &
  \cellcolor[HTML]{FEE382}0.7 &
  \cellcolor[HTML]{FCB379}-4.2 &
  \cellcolor[HTML]{FBAC78}0.1 &
  \cellcolor[HTML]{E6E483}2.8 &
  \cellcolor[HTML]{EEE683}-2.9 &
  \cellcolor[HTML]{EEE683}4.8 &
  \cellcolor[HTML]{FFEB84}-2.8 &
  \cellcolor[HTML]{FFEB84}0.8 &
  \cellcolor[HTML]{B8D780}3.4 &
  \cellcolor[HTML]{C8DC81}1.7 &
  \cellcolor[HTML]{B5D680}7.5 &
  \cellcolor[HTML]{CFDE82}4.2 \\
OK-VQA QG &
  9 K &
  4.2 &
  \cellcolor[HTML]{FCC17C}1.2 &
  \cellcolor[HTML]{FDD780}0.9 &
  \cellcolor[HTML]{FFEB84}2.2 &
  \cellcolor[HTML]{F87A6E}0.1 &
  \cellcolor[HTML]{FDCA7D}-2.5 &
  \cellcolor[HTML]{FEDD81}0.3 &
  \cellcolor[HTML]{FA9A74}-1.2 &
  \cellcolor[HTML]{FCBF7B}0.3 &
  \cellcolor[HTML]{FCB579}0.0 &
  \cellcolor[HTML]{FEDB81}0.4 &
  \cellcolor[HTML]{F8746D}0.1 &
  \cellcolor[HTML]{FDD780}-2.3 &
  \cellcolor[HTML]{F7E984}1.0 &
  \cellcolor[HTML]{FBA376}-0.9 &
  \cellcolor[HTML]{F86E6C}0.0 &
  \cellcolor[HTML]{C4DA81}4.2 &
  \cellcolor[HTML]{FFEB84}0.0 &
  \cellcolor[HTML]{FDD07E}0.6 &
  \cellcolor[HTML]{FEE482}0.7 &
  \cellcolor[HTML]{F98971}0.0 &
  \cellcolor[HTML]{ECE683}2.5 &
  \cellcolor[HTML]{CFDE82}0.0 &
  \cellcolor[HTML]{FEDF81}3.5 &
  \cellcolor[HTML]{FBEA84}-2.4 &
  \cellcolor[HTML]{ECE683}1.9 &
  \cellcolor[HTML]{BFD981}2.9 &
  \cellcolor[HTML]{CBDC81}1.5 &
  \cellcolor[HTML]{C2DA81}7.1 &
  \cellcolor[HTML]{D1DE82}4.1 \\
COCO Caption &
  567 K &
  4.1 &
  \cellcolor[HTML]{63BE7B}10.0 &
  \cellcolor[HTML]{7FC67D}8.5 &
  \cellcolor[HTML]{ACD380}6.4 &
  \cellcolor[HTML]{F8696B}0.0 &
  \cellcolor[HTML]{FDCD7E}-2.0 &
  \cellcolor[HTML]{F86A6B}0.0 &
  \cellcolor[HTML]{FDCA7D}0.2 &
  \cellcolor[HTML]{F8696B}0.0 &
  \cellcolor[HTML]{FBB178}-0.2 &
  \cellcolor[HTML]{FDD37F}0.2 &
  \cellcolor[HTML]{F8696B}0.0 &
  \cellcolor[HTML]{FEDD81}-1.6 &
  \cellcolor[HTML]{FAEA84}0.9 &
  \cellcolor[HTML]{FA9E75}-1.2 &
  \cellcolor[HTML]{F8696B}0.0 &
  \cellcolor[HTML]{ECE683}1.8 &
  \cellcolor[HTML]{F0E784}1.0 &
  \cellcolor[HTML]{F8786E}-0.1 &
  \cellcolor[HTML]{FDCC7E}-1.7 &
  \cellcolor[HTML]{FA9B74}0.1 &
  \cellcolor[HTML]{FDD680}-0.3 &
  \cellcolor[HTML]{F8E984}-3.8 &
  \cellcolor[HTML]{FA9373}-0.2 &
  \cellcolor[HTML]{90CB7E}6.3 &
  \cellcolor[HTML]{FCC27C}0.3 &
  \cellcolor[HTML]{FDD680}-6.8 &
  \cellcolor[HTML]{FBAF78}-7.1 &
  \cellcolor[HTML]{FFEB84}5.2 &
  \cellcolor[HTML]{EFE784}2.6 \\
Web CapFilt &
  23,147 K &
  4.1 &
  \cellcolor[HTML]{72C37C}9.3 &
  \cellcolor[HTML]{86C97E}8.1 &
  \cellcolor[HTML]{A5D17F}6.7 &
  \cellcolor[HTML]{F98670}0.2 &
  \cellcolor[HTML]{E5E483}2.8 &
  \cellcolor[HTML]{FDCA7D}0.3 &
  \cellcolor[HTML]{C7DB81}4.3 &
  \cellcolor[HTML]{FBA275}0.2 &
  \cellcolor[HTML]{CDDD82}4.8 &
  \cellcolor[HTML]{F7E984}1.1 &
  \cellcolor[HTML]{FA9D75}0.4 &
  \cellcolor[HTML]{D0DE82}3.1 &
  \cellcolor[HTML]{FCB97A}0.0 &
  \cellcolor[HTML]{F98A71}-2.3 &
  \cellcolor[HTML]{F8696B}0.0 &
  \cellcolor[HTML]{FED980}-0.1 &
  \cellcolor[HTML]{F0E784}1.0 &
  \cellcolor[HTML]{FCBB7A}0.4 &
  \cellcolor[HTML]{F4E884}2.1 &
  \cellcolor[HTML]{FFEB84}0.2 &
  \cellcolor[HTML]{F97C6E}-7.8 &
  \cellcolor[HTML]{FED880}-5.4 &
  \cellcolor[HTML]{F8696B}-2.3 &
  \cellcolor[HTML]{FBB078}-15.6 &
  \cellcolor[HTML]{F8716C}-0.7 &
  \cellcolor[HTML]{FCB679}-14.3 &
  \cellcolor[HTML]{FDCA7D}-5.2 &
  \cellcolor[HTML]{FEE182}4.6 &
  \cellcolor[HTML]{F2E884}2.4 \\
\vlbenchmark &
  7K &
  4.0 &
  \cellcolor[HTML]{FA9473}0.1 &
  \cellcolor[HTML]{FCBC7B}0.3 &
  \cellcolor[HTML]{FA9573}0.8 &
  \cellcolor[HTML]{F8696B}0.0 &
  \cellcolor[HTML]{FDCA7D}-2.4 &
  \cellcolor[HTML]{F86C6B}0.0 &
  \cellcolor[HTML]{FCB87A}-0.3 &
  \cellcolor[HTML]{F86A6B}0.0 &
  \cellcolor[HTML]{FCBB7A}0.2 &
  \cellcolor[HTML]{FDD17F}0.2 &
  \cellcolor[HTML]{F86A6B}0.0 &
  \cellcolor[HTML]{FEDB81}-1.8 &
  \cellcolor[HTML]{FEDC81}0.4 &
  \cellcolor[HTML]{D3DF82}5.1 &
  \cellcolor[HTML]{F8696B}0.0 &
  \cellcolor[HTML]{9ACE7F}6.7 &
  \cellcolor[HTML]{FFEB84}0.0 &
  \cellcolor[HTML]{F97D6E}0.0 &
  \cellcolor[HTML]{FDCF7E}-1.4 &
  \cellcolor[HTML]{F8726C}0.0 &
  \cellcolor[HTML]{FEE082}0.5 &
  \cellcolor[HTML]{D2DE82}-0.3 &
  \cellcolor[HTML]{F8716C}-1.8 &
  \cellcolor[HTML]{FEDF81}-5.4 &
  \cellcolor[HTML]{F7E984}1.3 &
  \cellcolor[HTML]{FEE983}-2.4 &
  \cellcolor[HTML]{8CCA7E}6.7 &
  \cellcolor[HTML]{FCBC7B}2.4 &
  \cellcolor[HTML]{63BE7B}10.0 \\
TextCaps &
  549 K &
  4.0 &
  \cellcolor[HTML]{ACD380}6.4 &
  \cellcolor[HTML]{AAD380}6.1 &
  \cellcolor[HTML]{63BE7B}10.0 &
  \cellcolor[HTML]{F8746D}0.1 &
  \cellcolor[HTML]{FEDE81}-0.1 &
  \cellcolor[HTML]{FBA676}0.2 &
  \cellcolor[HTML]{EEE683}2.2 &
  \cellcolor[HTML]{FA9974}0.1 &
  \cellcolor[HTML]{FFEB84}2.3 &
  \cellcolor[HTML]{FDC97D}0.0 &
  \cellcolor[HTML]{F8746D}0.1 &
  \cellcolor[HTML]{F4E884}0.8 &
  \cellcolor[HTML]{F97C6E}-0.7 &
  \cellcolor[HTML]{FA9E75}-1.2 &
  \cellcolor[HTML]{F8696B}0.0 &
  \cellcolor[HTML]{FEDF81}0.2 &
  \cellcolor[HTML]{79C57D}8.6 &
  \cellcolor[HTML]{FA9273}0.1 &
  \cellcolor[HTML]{E3E383}3.0 &
  \cellcolor[HTML]{F8746D}0.0 &
  \cellcolor[HTML]{F8696B}-9.4 &
  \cellcolor[HTML]{E7E483}-2.2 &
  \cellcolor[HTML]{F8696B}-2.3 &
  \cellcolor[HTML]{F8696B}-31.1 &
  \cellcolor[HTML]{FBA175}-0.1 &
  \cellcolor[HTML]{FEDC81}-5.4 &
  \cellcolor[HTML]{F8696B}-12.1 &
  \cellcolor[HTML]{63BE7B}10.0 &
  \cellcolor[HTML]{F0E784}2.5 \\
LLaVA Conversation &
  57 K &
  3.9 &
  \cellcolor[HTML]{FDC97D}1.4 &
  \cellcolor[HTML]{FEDE81}1.1 &
  \cellcolor[HTML]{FDD17F}1.8 &
  \cellcolor[HTML]{F8696B}0.0 &
  \cellcolor[HTML]{F8696B}-13.6 &
  \cellcolor[HTML]{F8696B}0.0 &
  \cellcolor[HTML]{F98570}-1.8 &
  \cellcolor[HTML]{F8696B}0.0 &
  \cellcolor[HTML]{FA9874}-1.3 &
  \cellcolor[HTML]{FA9D75}-0.8 &
  \cellcolor[HTML]{F8696B}0.0 &
  \cellcolor[HTML]{F8696B}-15.7 &
  \cellcolor[HTML]{FCC47C}0.1 &
  \cellcolor[HTML]{F7E984}3.5 &
  \cellcolor[HTML]{F8696B}0.0 &
  \cellcolor[HTML]{63BE7B}10.0 &
  \cellcolor[HTML]{FFEB84}0.0 &
  \cellcolor[HTML]{F86D6B}-0.2 &
  \cellcolor[HTML]{FBA777}-5.4 &
  \cellcolor[HTML]{F8696B}0.0 &
  \cellcolor[HTML]{FEE482}0.8 &
  \cellcolor[HTML]{C7DB81}0.8 &
  \cellcolor[HTML]{F8736D}-1.7 &
  \cellcolor[HTML]{B3D580}3.4 &
  \cellcolor[HTML]{E5E483}2.3 &
  \cellcolor[HTML]{FBEA84}-1.7 &
  \cellcolor[HTML]{E9E583}-1.0 &
  \cellcolor[HTML]{FDCF7E}3.6 &
  \cellcolor[HTML]{F8716C}-0.1 \\
LLaVA Reasoning &
  77 K &
  3.6 &
  \cellcolor[HTML]{F98B71}-0.1 &
  \cellcolor[HTML]{FCB97A}0.2 &
  \cellcolor[HTML]{F8736D}0.3 &
  \cellcolor[HTML]{F8696B}0.0 &
  \cellcolor[HTML]{FCBB7A}-4.1 &
  \cellcolor[HTML]{F8696B}0.0 &
  \cellcolor[HTML]{FCB379}-0.5 &
  \cellcolor[HTML]{F8696B}0.0 &
  \cellcolor[HTML]{FBA576}-0.8 &
  \cellcolor[HTML]{FA9172}-1.0 &
  \cellcolor[HTML]{F8696B}0.0 &
  \cellcolor[HTML]{FCC37C}-4.7 &
  \cellcolor[HTML]{FBA777}-0.2 &
  \cellcolor[HTML]{F5E984}3.6 &
  \cellcolor[HTML]{F8696B}0.0 &
  \cellcolor[HTML]{7EC67D}8.4 &
  \cellcolor[HTML]{FFEB84}0.0 &
  \cellcolor[HTML]{F8706C}-0.1 &
  \cellcolor[HTML]{FBA977}-5.3 &
  \cellcolor[HTML]{F8696B}0.0 &
  \cellcolor[HTML]{FDCB7D}-1.2 &
  \cellcolor[HTML]{FFEB84}-4.5 &
  \cellcolor[HTML]{F8706C}-1.9 &
  \cellcolor[HTML]{FEE182}-4.9 &
  \cellcolor[HTML]{FBA175}-0.1 &
  \cellcolor[HTML]{FDD47F}-7.2 &
  \cellcolor[HTML]{A2D17F}4.8 &
  \cellcolor[HTML]{FAA075}0.8 &
  \cellcolor[HTML]{F8696B}-0.2 \\
LLaVA Description &
  23 K &
  3.1 &
  \cellcolor[HTML]{F98470}-0.3 &
  \cellcolor[HTML]{FBAD78}-0.1 &
  \cellcolor[HTML]{F8696B}0.1 &
  \cellcolor[HTML]{F8696B}0.0 &
  \cellcolor[HTML]{FBA576}-6.6 &
  \cellcolor[HTML]{F8696B}0.0 &
  \cellcolor[HTML]{F8696B}-2.6 &
  \cellcolor[HTML]{F8696B}0.0 &
  \cellcolor[HTML]{F8696B}-3.4 &
  \cellcolor[HTML]{F8696B}-1.8 &
  \cellcolor[HTML]{F8696B}0.0 &
  \cellcolor[HTML]{FCB479}-6.6 &
  \cellcolor[HTML]{F8696B}-0.9 &
  \cellcolor[HTML]{F8696B}-4.2 &
  \cellcolor[HTML]{F8696B}0.0 &
  \cellcolor[HTML]{FFEB84}0.7 &
  \cellcolor[HTML]{C3DA81}3.9 &
  \cellcolor[HTML]{F8696B}-0.2 &
  \cellcolor[HTML]{F8696B}-11.8 &
  \cellcolor[HTML]{F86D6B}0.0 &
  \cellcolor[HTML]{FBAB77}-3.9 &
  \cellcolor[HTML]{FEE582}-4.8 &
  \cellcolor[HTML]{F8696B}-2.3 &
  \cellcolor[HTML]{FDD47F}-7.8 &
  \cellcolor[HTML]{F8796E}-0.6 &
  \cellcolor[HTML]{FEE582}-3.3 &
  \cellcolor[HTML]{FDD37F}-4.6 &
  \cellcolor[HTML]{FBB078}1.7 &
  \cellcolor[HTML]{FA8F72}0.3
           \\
			\bottomrule
		\end{tabular}%
	}
	\caption
	{Normalized transfer learning performance of MiniGPT-4. Higher values indicate better transferability. The rows are sorted in descending order of average performance. We multiply the values by a factor of 10 to aid visualization. The highest performance in each column is 10. 
  QG denotes question generation, MC denotes multiple-choice and G denotes open-ended generation. The color scale is normalized along each column. The colors represent values in descending order: green, yellow, orange and red. 
	}
	\label{eval:affinity_matrix_norm_minigpt_landscape}
\end{table}
\end{landscape}

\begin{landscape}
\begin{table}[!tp]
	\resizebox{\columnwidth}{!}{%
		\begin{tabular}{c|c|c|ccccccccccccccccccccccccccccc}
			\toprule
			Source   Task &
			\setlength\extrarowheight{0pt}\begin{tabular}[c]{@{}c@{}}Dataset\\ Size \end{tabular} &
			\setlength\extrarowheight{0pt}\begin{tabular}[c]{@{}c@{}}AHP\\ Ranking \\ Score \end{tabular}   &
			\multicolumn{29}{c}{Target Task} \\
			&
			&
			&
			\setlength\extrarowheight{0pt}\begin{tabular}[c]{@{}c@{}}COCO\\ Caption \end{tabular}  &
			\setlength\extrarowheight{0pt}\begin{tabular}[c]{@{}c@{}}Flickr\\ 30k\end{tabular}  &
			\setlength\extrarowheight{0pt}\begin{tabular}[c]{@{}c@{}}Text\\ Caps\end{tabular}  &
			\multicolumn{2}{c}{VQAv2} &
			\multicolumn{2}{c}{OK-VQA} &
			\multicolumn{2}{c}{A-OKVQA} &
			\setlength\extrarowheight{0pt}\begin{tabular}[c]{@{}c@{}}Science\\ QA\end{tabular} &
			\multicolumn{2}{c}{GQA} &
			\setlength\extrarowheight{0pt}\begin{tabular}[c]{@{}c@{}}Icon\\ QA \end{tabular} &
			VSR &
			\multicolumn{2}{c}{CLEVR} &
			\setlength\extrarowheight{0pt}\begin{tabular}[c]{@{}c@{}}RAVEN-\\ FAIR \end{tabular} &
			\multicolumn{2}{c}{\setlength\extrarowheight{0pt}\begin{tabular}[c]{@{}c@{}}Text\\ VQA\end{tabular}} &
			\multicolumn{2}{c}{\setlength\extrarowheight{0pt}\begin{tabular}[c]{@{}c@{}}OCR-\\ VQA\end{tabular}} &
			\setlength\extrarowheight{0pt}\begin{tabular}[c]{@{}c@{}}Open\\ CQA \end{tabular} &
			\multicolumn{2}{c}{\setlength\extrarowheight{0pt}\begin{tabular}[c]{@{}c@{}}Chart\\ QA\end{tabular}} &
			HM &
			\setlength\extrarowheight{0pt}\begin{tabular}[c]{@{}c@{}}NY\\ Explain \end{tabular} &
			\setlength\extrarowheight{0pt}\begin{tabular}[c]{@{}c@{}}NY\\ Rank \end{tabular} &
			MORE &
			\vlbenchmark \\
			 &
			&
			&
			&
			&
			&
			G &
			MC &
			G &
			MC &
			G &
			MC &
			&
			G &
			MC &
			&
			&
			G &
			MC &
			&
			G &
			MC &
			G &
			MC &
			&
			G &
			MC &
			&
			&
			&
			&
			\\
			\midrule
A-OKVQA (MC) &
  17 K &
  6.0 &
  \cellcolor[HTML]{C0D981}6.7 &
  \cellcolor[HTML]{CCDD82}6.4 &
  \cellcolor[HTML]{F6E984}2.9 &
  \cellcolor[HTML]{82C77D}8.4 &
  \cellcolor[HTML]{FBEA84}3.1 &
  \cellcolor[HTML]{93CC7E}7.4 &
  \cellcolor[HTML]{79C57D}9.0 &
  \cellcolor[HTML]{8ECB7E}7.7 &
  \cellcolor[HTML]{63BE7B}10.0 &
  \cellcolor[HTML]{ACD380}5.4 &
  \cellcolor[HTML]{A6D27F}6.5 &
  \cellcolor[HTML]{FDC77D}-2.2 &
  \cellcolor[HTML]{E5E483}2.9 &
  \cellcolor[HTML]{AED480}3.2 &
  \cellcolor[HTML]{AAD380}5.7 &
  \cellcolor[HTML]{9DCF7F}8.0 &
  \cellcolor[HTML]{FEE683}-0.4 &
  \cellcolor[HTML]{9FD07F}7.1 &
  \cellcolor[HTML]{63BE7B}10.0 &
  \cellcolor[HTML]{BED881}5.0 &
  \cellcolor[HTML]{8FCB7E}6.3 &
  \cellcolor[HTML]{FA9974}1.1 &
  \cellcolor[HTML]{7AC57D}8.8 &
  \cellcolor[HTML]{A8D27F}3.4 &
  \cellcolor[HTML]{FFEB84}0.0 &
  \cellcolor[HTML]{97CD7E}6.5 &
  \cellcolor[HTML]{A9D27F}4.3 &
  \cellcolor[HTML]{F3E884}2.0 &
  \cellcolor[HTML]{F7E984}2.5 \\
VQAv2 &
  444 K &
  5.9 &
  \cellcolor[HTML]{FEE282}4.1 &
  \cellcolor[HTML]{FDD27F}3.6 &
  \cellcolor[HTML]{FDCD7E}1.5 &
  \cellcolor[HTML]{63BE7B}10.0 &
  \cellcolor[HTML]{63BE7B}10.0 &
  \cellcolor[HTML]{73C37C}9.1 &
  \cellcolor[HTML]{7AC57D}9.0 &
  \cellcolor[HTML]{69C07C}9.7 &
  \cellcolor[HTML]{8ECB7E}8.2 &
  \cellcolor[HTML]{C1D981}4.1 &
  \cellcolor[HTML]{7AC57D}8.8 &
  \cellcolor[HTML]{A4D17F}6.1 &
  \cellcolor[HTML]{EEE683}2.4 &
  \cellcolor[HTML]{7FC67D}7.5 &
  \cellcolor[HTML]{63BE7B}10.0 &
  \cellcolor[HTML]{89C97E}8.7 &
  \cellcolor[HTML]{A0D07F}6.1 &
  \cellcolor[HTML]{7DC67D}8.8 &
  \cellcolor[HTML]{65BF7C}9.9 &
  \cellcolor[HTML]{9ACE7F}7.0 &
  \cellcolor[HTML]{ECE683}-1.5 &
  \cellcolor[HTML]{F86C6B}-0.1 &
  \cellcolor[HTML]{65BF7C}9.9 &
  \cellcolor[HTML]{FA9A74}-18.4 &
  \cellcolor[HTML]{DDE283}2.2 &
  \cellcolor[HTML]{F98C71}-45.6 &
  \cellcolor[HTML]{BED981}2.6 &
  \cellcolor[HTML]{FBA376}-3.6 &
  \cellcolor[HTML]{F8706C}-0.6 \\
OCR-VQA &
  802 K &
  5.7 &
  \cellcolor[HTML]{A3D17F}7.7 &
  \cellcolor[HTML]{9FD07F}7.9 &
  \cellcolor[HTML]{DEE283}4.0 &
  \cellcolor[HTML]{8FCB7E}7.7 &
  \cellcolor[HTML]{A9D27F}6.8 &
  \cellcolor[HTML]{B2D580}5.7 &
  \cellcolor[HTML]{D6DF82}4.8 &
  \cellcolor[HTML]{9ECF7F}6.8 &
  \cellcolor[HTML]{C9DC81}5.7 &
  \cellcolor[HTML]{D1DE82}3.0 &
  \cellcolor[HTML]{95CD7E}7.4 &
  \cellcolor[HTML]{C0D981}4.5 &
  \cellcolor[HTML]{FDD37F}0.8 &
  \cellcolor[HTML]{E7E583}-1.9 &
  \cellcolor[HTML]{68C07C}9.8 &
  \cellcolor[HTML]{FFEB84}4.6 &
  \cellcolor[HTML]{83C77D}8.0 &
  \cellcolor[HTML]{A4D17F}6.9 &
  \cellcolor[HTML]{FCC47C}0.4 &
  \cellcolor[HTML]{63BE7B}10.0 &
  \cellcolor[HTML]{63BE7B}10.0 &
  \cellcolor[HTML]{F98370}0.5 &
  \cellcolor[HTML]{6CC17C}9.5 &
  \cellcolor[HTML]{BED981}1.2 &
  \cellcolor[HTML]{ABD380}5.4 &
  \cellcolor[HTML]{C7DB81}3.2 &
  \cellcolor[HTML]{99CE7F}5.7 &
  \cellcolor[HTML]{F8696B}-7.8 &
  \cellcolor[HTML]{F8716C}-0.6 \\
A-OKVQA &
  17 K &
  5.6 &
  \cellcolor[HTML]{FDC97D}3.2 &
  \cellcolor[HTML]{FCBC7A}2.8 &
  \cellcolor[HTML]{FA9F75}0.2 &
  \cellcolor[HTML]{72C37C}9.2 &
  \cellcolor[HTML]{7BC57D}8.9 &
  \cellcolor[HTML]{71C27C}9.2 &
  \cellcolor[HTML]{63BE7B}10.0 &
  \cellcolor[HTML]{63BE7B}10.0 &
  \cellcolor[HTML]{93CC7E}8.0 &
  \cellcolor[HTML]{C2DA81}4.0 &
  \cellcolor[HTML]{82C77D}8.4 &
  \cellcolor[HTML]{B7D780}5.0 &
  \cellcolor[HTML]{E3E383}3.0 &
  \cellcolor[HTML]{FA9272}-6.9 &
  \cellcolor[HTML]{6BC17C}9.5 &
  \cellcolor[HTML]{63BE7B}10.0 &
  \cellcolor[HTML]{FEE182}-0.6 &
  \cellcolor[HTML]{8FCB7E}7.9 &
  \cellcolor[HTML]{78C47D}9.1 &
  \cellcolor[HTML]{A0D07F}6.6 &
  \cellcolor[HTML]{E2E383}-0.7 &
  \cellcolor[HTML]{F86A6B}-0.2 &
  \cellcolor[HTML]{84C87D}8.2 &
  \cellcolor[HTML]{ECE683}-3.2 &
  \cellcolor[HTML]{F8696B}-0.1 &
  \cellcolor[HTML]{F8696B}-62.5 &
  \cellcolor[HTML]{63BE7B}10.0 &
  \cellcolor[HTML]{FA9A74}-4.3 &
  \cellcolor[HTML]{F86C6B}-0.7 \\
ScienceQA &
  6 K &
  5.5 &
  \cellcolor[HTML]{F3E884}4.9 &
  \cellcolor[HTML]{F1E784}5.1 &
  \cellcolor[HTML]{EDE683}3.3 &
  \cellcolor[HTML]{C9DC81}4.6 &
  \cellcolor[HTML]{C4DA81}5.6 &
  \cellcolor[HTML]{BCD881}5.1 &
  \cellcolor[HTML]{B4D680}6.4 &
  \cellcolor[HTML]{C0D981}4.9 &
  \cellcolor[HTML]{81C77D}8.7 &
  \cellcolor[HTML]{63BE7B}10.0 &
  \cellcolor[HTML]{BCD881}5.4 &
  \cellcolor[HTML]{F2E884}1.5 &
  \cellcolor[HTML]{F8E984}1.9 &
  \cellcolor[HTML]{AFD480}3.2 &
  \cellcolor[HTML]{B1D580}5.3 &
  \cellcolor[HTML]{C8DB81}6.5 &
  \cellcolor[HTML]{87C97E}7.7 &
  \cellcolor[HTML]{AFD480}6.4 &
  \cellcolor[HTML]{93CC7E}7.8 &
  \cellcolor[HTML]{B6D680}5.4 &
  \cellcolor[HTML]{96CD7E}5.7 &
  \cellcolor[HTML]{FDD680}2.7 &
  \cellcolor[HTML]{BFD981}5.0 &
  \cellcolor[HTML]{FBB279}-14.5 &
  \cellcolor[HTML]{F8696B}-0.1 &
  \cellcolor[HTML]{9DCF7F}6.1 &
  \cellcolor[HTML]{E3E383}-0.4 &
  \cellcolor[HTML]{B5D680}5.5 &
  \cellcolor[HTML]{FA9373}0.1 \\
IconQA &
  19 K &
  5.2 &
  \cellcolor[HTML]{FEE081}4.1 &
  \cellcolor[HTML]{FEDF81}4.2 &
  \cellcolor[HTML]{FFEB84}2.4 &
  \cellcolor[HTML]{9CCF7F}7.0 &
  \cellcolor[HTML]{CFDE82}5.1 &
  \cellcolor[HTML]{B8D780}5.3 &
  \cellcolor[HTML]{D9E082}4.7 &
  \cellcolor[HTML]{B8D780}5.3 &
  \cellcolor[HTML]{CBDC81}5.6 &
  \cellcolor[HTML]{C4DA81}3.9 &
  \cellcolor[HTML]{A0D07F}6.9 &
  \cellcolor[HTML]{FEE382}0.1 &
  \cellcolor[HTML]{63BE7B}10.0 &
  \cellcolor[HTML]{B7D780}2.5 &
  \cellcolor[HTML]{77C47D}8.8 &
  \cellcolor[HTML]{FDD07E}1.5 &
  \cellcolor[HTML]{6FC27C}9.2 &
  \cellcolor[HTML]{9BCF7F}7.3 &
  \cellcolor[HTML]{FDD17F}1.2 &
  \cellcolor[HTML]{A1D07F}6.6 &
  \cellcolor[HTML]{FFEB84}-3.1 &
  \cellcolor[HTML]{FEE081}3.0 &
  \cellcolor[HTML]{68C07C}9.8 &
  \cellcolor[HTML]{F87A6E}-23.7 &
  \cellcolor[HTML]{E3E383}1.8 &
  \cellcolor[HTML]{BCD881}4.0 &
  \cellcolor[HTML]{E0E283}-0.1 &
  \cellcolor[HTML]{FFEB84}1.4 &
  \cellcolor[HTML]{FEE683}1.9 \\
GQA &
  943 K &
  5.2 &
  \cellcolor[HTML]{FDD27F}3.5 &
  \cellcolor[HTML]{FCC27C}3.0 &
  \cellcolor[HTML]{FA9072}-0.3 &
  \cellcolor[HTML]{7CC57D}8.7 &
  \cellcolor[HTML]{7FC67D}8.7 &
  \cellcolor[HTML]{9ACE7F}7.0 &
  \cellcolor[HTML]{74C37C}9.3 &
  \cellcolor[HTML]{8ECB7E}7.7 &
  \cellcolor[HTML]{9DCF7F}7.6 &
  \cellcolor[HTML]{D0DE82}3.1 &
  \cellcolor[HTML]{63BE7B}10.0 &
  \cellcolor[HTML]{63BE7B}10.0 &
  \cellcolor[HTML]{FDEB84}1.6 &
  \cellcolor[HTML]{EEE683}-2.5 &
  \cellcolor[HTML]{7EC67D}8.4 &
  \cellcolor[HTML]{C8DB81}6.5 &
  \cellcolor[HTML]{63BE7B}10.0 &
  \cellcolor[HTML]{D0DE82}4.8 &
  \cellcolor[HTML]{DEE283}4.3 &
  \cellcolor[HTML]{B7D780}5.4 &
  \cellcolor[HTML]{FEE482}-3.6 &
  \cellcolor[HTML]{F8696B}-0.2 &
  \cellcolor[HTML]{ABD380}6.1 &
  \cellcolor[HTML]{FDCA7D}-10.5 &
  \cellcolor[HTML]{F8696B}-0.1 &
  \cellcolor[HTML]{F87B6E}-53.7 &
  \cellcolor[HTML]{7CC67D}8.0 &
  \cellcolor[HTML]{F8756D}-6.9 &
  \cellcolor[HTML]{F8696B}-0.8 \\
VSR &
  3 K &
  5.0 &
  \cellcolor[HTML]{FDD780}3.7 &
  \cellcolor[HTML]{FEDD81}4.1 &
  \cellcolor[HTML]{FEE683}2.3 &
  \cellcolor[HTML]{B0D580}5.9 &
  \cellcolor[HTML]{B4D680}6.3 &
  \cellcolor[HTML]{E5E483}2.9 &
  \cellcolor[HTML]{D8E082}4.7 &
  \cellcolor[HTML]{EBE683}2.5 &
  \cellcolor[HTML]{E0E283}4.7 &
  \cellcolor[HTML]{E3E383}1.9 &
  \cellcolor[HTML]{B4D680}5.8 &
  \cellcolor[HTML]{C8DB81}4.0 &
  \cellcolor[HTML]{F8E984}1.9 &
  \cellcolor[HTML]{63BE7B}10.0 &
  \cellcolor[HTML]{D0DE82}3.4 &
  \cellcolor[HTML]{FEDA80}2.6 &
  \cellcolor[HTML]{FDD47F}-1.3 &
  \cellcolor[HTML]{C9DC81}5.1 &
  \cellcolor[HTML]{DAE182}4.5 &
  \cellcolor[HTML]{D5DF82}3.7 &
  \cellcolor[HTML]{74C37C}8.6 &
  \cellcolor[HTML]{DCE182}4.8 &
  \cellcolor[HTML]{EBE683}2.7 &
  \cellcolor[HTML]{EFE784}-3.5 &
  \cellcolor[HTML]{FEEB84}0.1 &
  \cellcolor[HTML]{F6E984}0.0 &
  \cellcolor[HTML]{BED981}2.6 &
  \cellcolor[HTML]{FEE282}0.8 &
  \cellcolor[HTML]{FFEB84}2.0 \\
OK-VQA &
  9 K &
  4.8 &
  \cellcolor[HTML]{FDD17F}3.5 &
  \cellcolor[HTML]{FCB87A}2.7 &
  \cellcolor[HTML]{FCB87A}0.9 &
  \cellcolor[HTML]{7FC67D}8.6 &
  \cellcolor[HTML]{BDD881}5.9 &
  \cellcolor[HTML]{63BE7B}10.0 &
  \cellcolor[HTML]{70C27C}9.4 &
  \cellcolor[HTML]{87C97E}8.0 &
  \cellcolor[HTML]{B3D580}6.6 &
  \cellcolor[HTML]{C6DB81}3.7 &
  \cellcolor[HTML]{90CB7E}7.7 &
  \cellcolor[HTML]{CDDD82}3.7 &
  \cellcolor[HTML]{F6E984}2.0 &
  \cellcolor[HTML]{FA9D75}-6.6 &
  \cellcolor[HTML]{86C97E}7.9 &
  \cellcolor[HTML]{ACD380}7.5 &
  \cellcolor[HTML]{FDCD7E}-1.6 &
  \cellcolor[HTML]{97CD7E}7.5 &
  \cellcolor[HTML]{A2D17F}7.1 &
  \cellcolor[HTML]{A5D17F}6.4 &
  \cellcolor[HTML]{FBA777}-8.1 &
  \cellcolor[HTML]{F86A6B}-0.1 &
  \cellcolor[HTML]{6BC17C}9.6 &
  \cellcolor[HTML]{FCB679}-13.8 &
  \cellcolor[HTML]{F8696B}-0.1 &
  \cellcolor[HTML]{F8726C}-58.1 &
  \cellcolor[HTML]{A3D17F}4.8 &
  \cellcolor[HTML]{FA9A74}-4.3 &
  \cellcolor[HTML]{F8696B}-0.8 \\
TextVQA &
  35 K &
  4.7 &
  \cellcolor[HTML]{B7D780}7.0 &
  \cellcolor[HTML]{B6D680}7.1 &
  \cellcolor[HTML]{EBE683}3.4 &
  \cellcolor[HTML]{97CD7E}7.3 &
  \cellcolor[HTML]{E4E483}4.1 &
  \cellcolor[HTML]{A6D27F}6.3 &
  \cellcolor[HTML]{AAD380}6.8 &
  \cellcolor[HTML]{A4D17F}6.4 &
  \cellcolor[HTML]{C6DB81}5.8 &
  \cellcolor[HTML]{DCE182}2.4 &
  \cellcolor[HTML]{A8D27F}6.5 &
  \cellcolor[HTML]{DEE283}2.7 &
  \cellcolor[HTML]{FFEB84}1.5 &
  \cellcolor[HTML]{BED881}1.8 &
  \cellcolor[HTML]{86C87D}7.9 &
  \cellcolor[HTML]{F6E984}4.9 &
  \cellcolor[HTML]{FBA877}-3.5 &
  \cellcolor[HTML]{63BE7B}10.0 &
  \cellcolor[HTML]{C3DA81}5.6 &
  \cellcolor[HTML]{9ACE7F}7.0 &
  \cellcolor[HTML]{FA9473}-9.5 &
  \cellcolor[HTML]{F8766D}0.2 &
  \cellcolor[HTML]{63BE7B}10.0 &
  \cellcolor[HTML]{FCC57C}-11.3 &
  \cellcolor[HTML]{D9E082}2.5 &
  \cellcolor[HTML]{F8776D}-55.5 &
  \cellcolor[HTML]{FA8F72}-9.0 &
  \cellcolor[HTML]{F87A6E}-6.5 &
  \cellcolor[HTML]{F8716C}-0.6 \\
OpenCQA &
  6 K &
  4.6 &
  \cellcolor[HTML]{FFEB84}4.5 &
  \cellcolor[HTML]{FFEB84}4.6 &
  \cellcolor[HTML]{DDE283}4.1 &
  \cellcolor[HTML]{FA9573}0.6 &
  \cellcolor[HTML]{FEDA80}1.3 &
  \cellcolor[HTML]{FEE783}1.4 &
  \cellcolor[HTML]{FFEB84}2.9 &
  \cellcolor[HTML]{FCBA7A}0.9 &
  \cellcolor[HTML]{FDC87D}2.1 &
  \cellcolor[HTML]{D8E082}2.6 &
  \cellcolor[HTML]{F9816F}0.4 &
  \cellcolor[HTML]{FFEB84}0.7 &
  \cellcolor[HTML]{F8E984}1.9 &
  \cellcolor[HTML]{FBAE78}-6.1 &
  \cellcolor[HTML]{F8716C}0.0 &
  \cellcolor[HTML]{FDD47F}2.0 &
  \cellcolor[HTML]{F8696B}-6.8 &
  \cellcolor[HTML]{F0E784}3.3 &
  \cellcolor[HTML]{FEE783}2.6 &
  \cellcolor[HTML]{FDCB7D}1.0 &
  \cellcolor[HTML]{BBD881}2.6 &
  \cellcolor[HTML]{63BE7B}10.0 &
  \cellcolor[HTML]{FDD57F}1.3 &
  \cellcolor[HTML]{D8E082}-1.3 &
  \cellcolor[HTML]{D9E082}2.5 &
  \cellcolor[HTML]{63BE7B}10.0 &
  \cellcolor[HTML]{F3E884}-1.7 &
  \cellcolor[HTML]{84C87D}8.2 &
  \cellcolor[HTML]{A9D380}6.4 \\
COCO Caption &
  567 K &
  4.6 &
  \cellcolor[HTML]{63BE7B}10.0 &
  \cellcolor[HTML]{85C87D}8.9 &
  \cellcolor[HTML]{C6DB81}5.2 &
  \cellcolor[HTML]{F8696B}0.0 &
  \cellcolor[HTML]{FDD37F}0.6 &
  \cellcolor[HTML]{F86A6B}0.0 &
  \cellcolor[HTML]{FBAE78}0.5 &
  \cellcolor[HTML]{F86B6B}0.0 &
  \cellcolor[HTML]{FDCB7D}2.2 &
  \cellcolor[HTML]{FCC07B}-0.5 &
  \cellcolor[HTML]{F8696B}0.0 &
  \cellcolor[HTML]{FEE883}0.5 &
  \cellcolor[HTML]{FCC07B}0.3 &
  \cellcolor[HTML]{FFEB84}-4.1 &
  \cellcolor[HTML]{F8696B}0.0 &
  \cellcolor[HTML]{F4E884}4.9 &
  \cellcolor[HTML]{FEE683}-0.4 &
  \cellcolor[HTML]{F86F6C}0.1 &
  \cellcolor[HTML]{C9DC81}5.3 &
  \cellcolor[HTML]{FA9172}0.4 &
  \cellcolor[HTML]{FEE983}-3.3 &
  \cellcolor[HTML]{DAE182}4.9 &
  \cellcolor[HTML]{F8736D}0.1 &
  \cellcolor[HTML]{63BE7B}10.0 &
  \cellcolor[HTML]{F8696B}-0.1 &
  \cellcolor[HTML]{E4E483}1.2 &
  \cellcolor[HTML]{FCC07B}-5.7 &
  \cellcolor[HTML]{B0D480}5.8 &
  \cellcolor[HTML]{ACD380}6.3 \\
Flickr30k &
  145 K &
  4.5 &
  \cellcolor[HTML]{ABD380}7.5 &
  \cellcolor[HTML]{63BE7B}10.0 &
  \cellcolor[HTML]{BDD881}5.7 &
  \cellcolor[HTML]{F98C71}0.5 &
  \cellcolor[HTML]{FFEB84}2.9 &
  \cellcolor[HTML]{FDD17F}1.1 &
  \cellcolor[HTML]{FDD680}2.1 &
  \cellcolor[HTML]{FEE081}1.3 &
  \cellcolor[HTML]{FFEB84}3.4 &
  \cellcolor[HTML]{FFEB84}0.1 &
  \cellcolor[HTML]{FA8F72}0.6 &
  \cellcolor[HTML]{F8E984}1.1 &
  \cellcolor[HTML]{FDEB84}1.6 &
  \cellcolor[HTML]{FCB679}-5.8 &
  \cellcolor[HTML]{F86D6B}0.0 &
  \cellcolor[HTML]{FEDE81}3.2 &
  \cellcolor[HTML]{CDDD82}3.2 &
  \cellcolor[HTML]{FBA977}1.3 &
  \cellcolor[HTML]{CFDE82}5.0 &
  \cellcolor[HTML]{FDC97D}1.0 &
  \cellcolor[HTML]{EBE683}-1.5 &
  \cellcolor[HTML]{CEDD82}5.4 &
  \cellcolor[HTML]{FEDD81}1.4 &
  \cellcolor[HTML]{B6D680}2.0 &
  \cellcolor[HTML]{F4E884}0.8 &
  \cellcolor[HTML]{BFD981}3.8 &
  \cellcolor[HTML]{FCC27C}-5.5 &
  \cellcolor[HTML]{B8D780}5.3 &
  \cellcolor[HTML]{A9D27F}6.5 \\
TextCaps &
  549 K &
  4.5 &
  \cellcolor[HTML]{EEE683}5.1 &
  \cellcolor[HTML]{EEE683}5.2 &
  \cellcolor[HTML]{63BE7B}10.0 &
  \cellcolor[HTML]{FA9F75}0.7 &
  \cellcolor[HTML]{FCB679}-2.1 &
  \cellcolor[HTML]{FFEB84}1.4 &
  \cellcolor[HTML]{FCC07B}1.3 &
  \cellcolor[HTML]{FFEB84}1.4 &
  \cellcolor[HTML]{FBA977}0.9 &
  \cellcolor[HTML]{FEDD81}-0.1 &
  \cellcolor[HTML]{FA9373}0.6 &
  \cellcolor[HTML]{FEDE81}-0.3 &
  \cellcolor[HTML]{FA9272}-1.1 &
  \cellcolor[HTML]{F5E884}-3.2 &
  \cellcolor[HTML]{F86E6C}0.0 &
  \cellcolor[HTML]{ECE683}5.2 &
  \cellcolor[HTML]{8CCA7E}7.3 &
  \cellcolor[HTML]{FFEB84}2.5 &
  \cellcolor[HTML]{FFEB84}2.8 &
  \cellcolor[HTML]{FDCC7E}1.0 &
  \cellcolor[HTML]{FDC97D}-5.6 &
  \cellcolor[HTML]{CDDD82}5.5 &
  \cellcolor[HTML]{FBAF78}0.8 &
  \cellcolor[HTML]{B5D680}2.1 &
  \cellcolor[HTML]{F8696B}-0.1 &
  \cellcolor[HTML]{B9D780}4.2 &
  \cellcolor[HTML]{FCC47C}-5.4 &
  \cellcolor[HTML]{63BE7B}10.0 &
  \cellcolor[HTML]{B8D780}5.7 \\
HM &
  9 K &
  4.1 &
  \cellcolor[HTML]{FCC57C}3.0 &
  \cellcolor[HTML]{FCB679}2.6 &
  \cellcolor[HTML]{FDEB84}2.5 &
  \cellcolor[HTML]{C9DC81}4.6 &
  \cellcolor[HTML]{FEEA83}2.8 &
  \cellcolor[HTML]{FA9573}0.5 &
  \cellcolor[HTML]{FCC47C}1.4 &
  \cellcolor[HTML]{FA9473}0.5 &
  \cellcolor[HTML]{FCBD7B}1.6 &
  \cellcolor[HTML]{FED980}-0.2 &
  \cellcolor[HTML]{D1DE82}4.3 &
  \cellcolor[HTML]{F9E984}1.1 &
  \cellcolor[HTML]{F3E884}2.1 &
  \cellcolor[HTML]{F97C6E}-7.6 &
  \cellcolor[HTML]{B2D580}5.2 &
  \cellcolor[HTML]{FDD17F}1.6 &
  \cellcolor[HTML]{EAE583}1.3 &
  \cellcolor[HTML]{FEE182}2.4 &
  \cellcolor[HTML]{FAA075}-1.9 &
  \cellcolor[HTML]{AFD480}5.8 &
  \cellcolor[HTML]{FEE582}-3.5 &
  \cellcolor[HTML]{D1DE82}5.3 &
  \cellcolor[HTML]{F3E884}2.2 &
  \cellcolor[HTML]{FDD07E}-9.5 &
  \cellcolor[HTML]{63BE7B}10.0 &
  \cellcolor[HTML]{DCE182}1.8 &
  \cellcolor[HTML]{FFEB84}-2.8 &
  \cellcolor[HTML]{E8E583}2.7 &
  \cellcolor[HTML]{C0D981}5.3 \\
Web CapFilt &
  23,147 K &
  4.0 &
  \cellcolor[HTML]{78C47D}9.3 &
  \cellcolor[HTML]{81C77D}9.0 &
  \cellcolor[HTML]{A4D17F}6.9 &
  \cellcolor[HTML]{FFEB84}1.7 &
  \cellcolor[HTML]{F6E984}3.3 &
  \cellcolor[HTML]{E7E483}2.8 &
  \cellcolor[HTML]{F7E984}3.3 &
  \cellcolor[HTML]{DEE283}3.3 &
  \cellcolor[HTML]{FFEB84}3.4 &
  \cellcolor[HTML]{FEE983}0.1 &
  \cellcolor[HTML]{FFEB84}1.9 &
  \cellcolor[HTML]{CADC81}3.9 &
  \cellcolor[HTML]{F8696B}-2.2 &
  \cellcolor[HTML]{F8696B}-8.2 &
  \cellcolor[HTML]{FFEB84}0.5 &
  \cellcolor[HTML]{FDCC7E}1.0 &
  \cellcolor[HTML]{F9816F}-5.6 &
  \cellcolor[HTML]{FDD780}2.2 &
  \cellcolor[HTML]{F3E884}3.4 &
  \cellcolor[HTML]{FCB87A}0.8 &
  \cellcolor[HTML]{F98670}-10.6 &
  \cellcolor[HTML]{F2E884}3.8 &
  \cellcolor[HTML]{FFEB84}1.5 &
  \cellcolor[HTML]{F8696B}-26.6 &
  \cellcolor[HTML]{FBEA84}0.3 &
  \cellcolor[HTML]{FDD780}-10.0 &
  \cellcolor[HTML]{F8696B}-11.6 &
  \cellcolor[HTML]{85C87D}8.1 &
  \cellcolor[HTML]{F3E884}2.7 \\
VQAv2 QG &
  444 K &
  3.9 &
  \cellcolor[HTML]{ABD380}7.5 &
  \cellcolor[HTML]{B1D580}7.3 &
  \cellcolor[HTML]{FEDD81}2.0 &
  \cellcolor[HTML]{F98B71}0.5 &
  \cellcolor[HTML]{FA9874}-5.1 &
  \cellcolor[HTML]{F86E6C}0.1 &
  \cellcolor[HTML]{F8776D}-1.7 &
  \cellcolor[HTML]{F86A6B}0.0 &
  \cellcolor[HTML]{FA9C74}0.4 &
  \cellcolor[HTML]{F97E6F}-1.4 &
  \cellcolor[HTML]{F97F6F}0.3 &
  \cellcolor[HTML]{FCB679}-3.6 &
  \cellcolor[HTML]{FA9D75}-0.7 &
  \cellcolor[HTML]{FDD780}-4.7 &
  \cellcolor[HTML]{FBB279}0.3 &
  \cellcolor[HTML]{B0D480}7.3 &
  \cellcolor[HTML]{FEEB84}0.0 &
  \cellcolor[HTML]{F8766D}0.3 &
  \cellcolor[HTML]{FBB178}-0.9 &
  \cellcolor[HTML]{FFEB84}1.3 &
  \cellcolor[HTML]{F98D71}-10.1 &
  \cellcolor[HTML]{FFEB84}3.3 &
  \cellcolor[HTML]{FA9F75}0.6 &
  \cellcolor[HTML]{E7E483}-2.7 &
  \cellcolor[HTML]{F8696B}-0.1 &
  \cellcolor[HTML]{D4DF82}2.3 &
  \cellcolor[HTML]{D0DE82}1.2 &
  \cellcolor[HTML]{CFDD82}4.1 &
  \cellcolor[HTML]{DBE182}3.9 \\
A-OKVQA QG &
  17 K &
  3.6 &
  \cellcolor[HTML]{EFE784}5.1 &
  \cellcolor[HTML]{F2E884}5.1 &
  \cellcolor[HTML]{FEDF81}2.1 &
  \cellcolor[HTML]{F98B71}0.4 &
  \cellcolor[HTML]{F98B71}-6.2 &
  \cellcolor[HTML]{F8696B}0.0 &
  \cellcolor[HTML]{FDD47F}2.0 &
  \cellcolor[HTML]{F86A6B}0.0 &
  \cellcolor[HTML]{FA9A74}0.3 &
  \cellcolor[HTML]{FCB679}-0.6 &
  \cellcolor[HTML]{F8706C}0.1 &
  \cellcolor[HTML]{FCB87A}-3.5 &
  \cellcolor[HTML]{FBA275}-0.6 &
  \cellcolor[HTML]{FEDD81}-4.6 &
  \cellcolor[HTML]{F8746D}0.0 &
  \cellcolor[HTML]{85C87D}8.8 &
  \cellcolor[HTML]{FCB579}-2.9 &
  \cellcolor[HTML]{F8756D}0.2 &
  \cellcolor[HTML]{FCB77A}-0.5 &
  \cellcolor[HTML]{FCC37C}0.9 &
  \cellcolor[HTML]{FDCD7E}-5.3 &
  \cellcolor[HTML]{F7E984}3.6 &
  \cellcolor[HTML]{F8736D}0.1 &
  \cellcolor[HTML]{FED980}-8.0 &
  \cellcolor[HTML]{E6E483}1.6 &
  \cellcolor[HTML]{FDD27F}-12.2 &
  \cellcolor[HTML]{FDCB7D}-4.9 &
  \cellcolor[HTML]{D1DE82}3.9 &
  \cellcolor[HTML]{EAE583}3.1 \\
\vlbenchmark &
  7 K &
  3.5 &
  \cellcolor[HTML]{F8766D}-0.2 &
  \cellcolor[HTML]{F8736C}-0.1 &
  \cellcolor[HTML]{F98C71}-0.4 &
  \cellcolor[HTML]{F8696B}0.0 &
  \cellcolor[HTML]{FDC97D}-0.4 &
  \cellcolor[HTML]{F8696B}0.0 &
  \cellcolor[HTML]{FCC17C}1.3 &
  \cellcolor[HTML]{F8696B}0.0 &
  \cellcolor[HTML]{FBA676}0.8 &
  \cellcolor[HTML]{FBB178}-0.7 &
  \cellcolor[HTML]{F8696B}0.0 &
  \cellcolor[HTML]{FEEA83}0.6 &
  \cellcolor[HTML]{FDD17F}0.7 &
  \cellcolor[HTML]{D1DE82}0.1 &
  \cellcolor[HTML]{F8696B}0.0 &
  \cellcolor[HTML]{F86E6C}-9.8 &
  \cellcolor[HTML]{FCB97A}-2.7 &
  \cellcolor[HTML]{F86A6B}0.0 &
  \cellcolor[HTML]{FA9E75}-2.1 &
  \cellcolor[HTML]{F8696B}0.0 &
  \cellcolor[HTML]{DDE283}-0.3 &
  \cellcolor[HTML]{D2DE82}5.2 &
  \cellcolor[HTML]{F8696B}0.0 &
  \cellcolor[HTML]{FEE983}-5.3 &
  \cellcolor[HTML]{F8696B}-0.1 &
  \cellcolor[HTML]{FFEB84}-0.6 &
  \cellcolor[HTML]{FBAD78}-7.0 &
  \cellcolor[HTML]{FEE282}0.8 &
  \cellcolor[HTML]{63BE7B}10.0 \\
LLaVA Conversation &
  57 K &
  3.4 &
  \cellcolor[HTML]{FDD27F}3.5 &
  \cellcolor[HTML]{FDC77D}3.2 &
  \cellcolor[HTML]{FDD57F}1.8 &
  \cellcolor[HTML]{F8696B}0.0 &
  \cellcolor[HTML]{FDCB7D}-0.1 &
  \cellcolor[HTML]{F8696B}0.0 &
  \cellcolor[HTML]{FA9874}-0.4 &
  \cellcolor[HTML]{F8696B}0.0 &
  \cellcolor[HTML]{F98770}-0.4 &
  \cellcolor[HTML]{FCB77A}-0.6 &
  \cellcolor[HTML]{F8696B}0.0 &
  \cellcolor[HTML]{FDCF7E}-1.6 &
  \cellcolor[HTML]{FBA576}-0.5 &
  \cellcolor[HTML]{F98A71}-7.2 &
  \cellcolor[HTML]{F8696B}0.0 &
  \cellcolor[HTML]{F8696B}-10.5 &
  \cellcolor[HTML]{FEEB84}0.0 &
  \cellcolor[HTML]{F8696B}0.0 &
  \cellcolor[HTML]{FBAC77}-1.2 &
  \cellcolor[HTML]{F8696B}0.0 &
  \cellcolor[HTML]{E4E483}-0.8 &
  \cellcolor[HTML]{D6E082}5.0 &
  \cellcolor[HTML]{F8696B}0.0 &
  \cellcolor[HTML]{EBE683}-3.1 &
  \cellcolor[HTML]{F8696B}-0.1 &
  \cellcolor[HTML]{FEE482}-3.5 &
  \cellcolor[HTML]{FDD780}-4.1 &
  \cellcolor[HTML]{FDD07E}-0.5 &
  \cellcolor[HTML]{FA9874}0.2 \\
LLaVA Description &
  23 K &
  3.2 &
  \cellcolor[HTML]{F8696B}-0.7 &
  \cellcolor[HTML]{F8696B}-0.5 &
  \cellcolor[HTML]{F8696B}-1.4 &
  \cellcolor[HTML]{F8696B}0.0 &
  \cellcolor[HTML]{FBA977}-3.4 &
  \cellcolor[HTML]{F8696B}0.0 &
  \cellcolor[HTML]{FA9874}-0.3 &
  \cellcolor[HTML]{F8696B}0.0 &
  \cellcolor[HTML]{F8696B}-1.6 &
  \cellcolor[HTML]{F8736D}-1.6 &
  \cellcolor[HTML]{F8696B}0.0 &
  \cellcolor[HTML]{FDC67D}-2.3 &
  \cellcolor[HTML]{F9806F}-1.6 &
  \cellcolor[HTML]{FDD780}-4.7 &
  \cellcolor[HTML]{F8696B}0.0 &
  \cellcolor[HTML]{FCB77A}-1.4 &
  \cellcolor[HTML]{FBB078}-3.2 &
  \cellcolor[HTML]{F8696B}0.0 &
  \cellcolor[HTML]{F8736C}-4.9 &
  \cellcolor[HTML]{F8696B}0.0 &
  \cellcolor[HTML]{FCB77A}-6.9 &
  \cellcolor[HTML]{FBA476}1.4 &
  \cellcolor[HTML]{F8696B}0.0 &
  \cellcolor[HTML]{ABD380}3.1 &
  \cellcolor[HTML]{F8696B}-0.1 &
  \cellcolor[HTML]{FEE983}-1.5 &
  \cellcolor[HTML]{FBAF78}-6.8 &
  \cellcolor[HTML]{FDD67F}-0.1 &
  \cellcolor[HTML]{F98971}-0.1 \\
OK-VQA QG &
  9 K &
  3.2 &
  \cellcolor[HTML]{B2D580}7.2 &
  \cellcolor[HTML]{BCD881}6.9 &
  \cellcolor[HTML]{FCEA84}2.6 &
  \cellcolor[HTML]{F8696B}0.0 &
  \cellcolor[HTML]{F8696B}-9.6 &
  \cellcolor[HTML]{F8696B}0.0 &
  \cellcolor[HTML]{FCC47C}1.4 &
  \cellcolor[HTML]{F8696B}0.0 &
  \cellcolor[HTML]{FA9874}0.2 &
  \cellcolor[HTML]{F98871}-1.3 &
  \cellcolor[HTML]{F8696B}0.0 &
  \cellcolor[HTML]{F8696B}-10.0 &
  \cellcolor[HTML]{FBAD78}-0.3 &
  \cellcolor[HTML]{F8746D}-7.9 &
  \cellcolor[HTML]{F8696B}0.0 &
  \cellcolor[HTML]{FDD47F}2.0 &
  \cellcolor[HTML]{FFEB84}-0.1 &
  \cellcolor[HTML]{F86F6C}0.1 &
  \cellcolor[HTML]{FA9C74}-2.2 &
  \cellcolor[HTML]{F87A6E}0.2 &
  \cellcolor[HTML]{F8696B}-12.7 &
  \cellcolor[HTML]{F1E784}3.9 &
  \cellcolor[HTML]{F97E6F}0.3 &
  \cellcolor[HTML]{FDC97D}-10.6 &
  \cellcolor[HTML]{F7E984}0.6 &
  \cellcolor[HTML]{FA9272}-43.0 &
  \cellcolor[HTML]{FCB77A}-6.2 &
  \cellcolor[HTML]{E5E483}2.9 &
  \cellcolor[HTML]{FA9373}0.1 \\
LLaVA Reasoning &
  77 K &
  3.2 &
  \cellcolor[HTML]{F86E6C}-0.5 &
  \cellcolor[HTML]{F86F6C}-0.2 &
  \cellcolor[HTML]{F8736D}-1.1 &
  \cellcolor[HTML]{F8696B}0.0 &
  \cellcolor[HTML]{FBB178}-2.7 &
  \cellcolor[HTML]{F8696B}0.0 &
  \cellcolor[HTML]{F8696B}-2.2 &
  \cellcolor[HTML]{F8696B}0.0 &
  \cellcolor[HTML]{F8706C}-1.3 &
  \cellcolor[HTML]{F8696B}-1.7 &
  \cellcolor[HTML]{F8696B}0.0 &
  \cellcolor[HTML]{FDC97D}-2.1 &
  \cellcolor[HTML]{FA9974}-0.9 &
  \cellcolor[HTML]{D9E082}-0.6 &
  \cellcolor[HTML]{F8696B}0.0 &
  \cellcolor[HTML]{FDC97D}0.7 &
  \cellcolor[HTML]{FAEA84}0.3 &
  \cellcolor[HTML]{F8696B}0.0 &
  \cellcolor[HTML]{F8696B}-5.5 &
  \cellcolor[HTML]{F8696B}0.0 &
  \cellcolor[HTML]{FDEB84}-3.0 &
  \cellcolor[HTML]{FBAA77}1.5 &
  \cellcolor[HTML]{F8696B}0.0 &
  \cellcolor[HTML]{FFEB84}-5.1 &
  \cellcolor[HTML]{F8696B}-0.1 &
  \cellcolor[HTML]{FEDB80}-8.1 &
  \cellcolor[HTML]{FDCF7E}-4.6 &
  \cellcolor[HTML]{FCBF7B}-1.7 &
  \cellcolor[HTML]{F3E884}2.7
           \\
			\bottomrule
		\end{tabular}%
	}
	\caption
	{Normalized transfer learning performance of mPLUG-Owl. Higher values indicate better transferability. The rows are sorted in descending order of average performance. We multiply the values by a factor of 10 to aid visualization. The highest performance in each column is 10. 
  QG denotes question generation, MC denotes multiple-choice and G denotes open-ended generation. The color scale is normalized along each column. The colors represent values in descending order: green, yellow, orange and red. 
	}
	\label{eval:affinity_matrix_norm_mplugowl_landscape}
\end{table}
\end{landscape}

\section{Mean Cosine Similarity of Target Tasks}
\label{Appendix_target_task_cos_sim_svd}
In this section, we rank the average cosine similarity among the target tasks. We first compute pairwise cosine similarity using the SVD features of target tasks. For each target task, we take the average of all pairs that it is involved in. Finally, we rank all target tasks in a descending order.

\begin{table}[!htp]
\centering
\small
\setlength\tabcolsep{5pt}
\begin{tabular}{@{}lc@{}}
\toprule
Target Task        & Cosine Similarity \\
\midrule
OK-VQA (MC)  & 0.54       \\
VQAv2 (G)    & 0.54       \\
A-OKVQA (MC) & 0.54       \\
GQA (G)      & 0.53       \\
OK-VQA (G)   & 0.53       \\
A-OKVQA (G)  & 0.53       \\
TextVQA (MC) & 0.52       \\
VQAv2 (MC)   & 0.51       \\
TextVQA (G)  & 0.50       \\
CLEVR (G)    & 0.50       \\
ChartQA (G)  & 0.48       \\
OCR-VQA (G)  & 0.45       \\
ScienceQA (MC)   & 0.45       \\
GQA (MC)     & 0.44       \\
Hateful Memes (MC)          & 0.42       \\
VSR (MC)         & 0.41       \\
NY Ranking (MC)      & 0.37       \\
IconQA (MC)      & 0.35       \\
OCR-VQA (MC) & 0.32       \\
TextCaps (G)    & 0.29       \\
RAVEN-FAIR (MC)  & 0.27       \\
Flickr30k (G)   & 0.25       \\
COCO Caption (G) & 0.22       \\
ChartQA (MC) & 0.15       \\
MORE (G)        & 0.15       \\
CLEVR (MC)   & 0.14       \\
\vlbenchmark        & -0.06      \\
NY Explanation (G)   & -0.09      \\
OpenCQA (G)      & -0.27     \\
\bottomrule
\end{tabular}%
\caption
{
Mean cosine similarity, computed from the SVD features, for each target task. The tasks are ranking by descending similarity. 
}
\label{tab:target_task_cos_sim_svd}
\end{table}

\section{Hierarchical Clustering of SVD Similarity}
\label{Appendix_hierarchical_clustering}
\begin{figure}[!thp]
	\centering
	\includegraphics[width=\columnwidth]{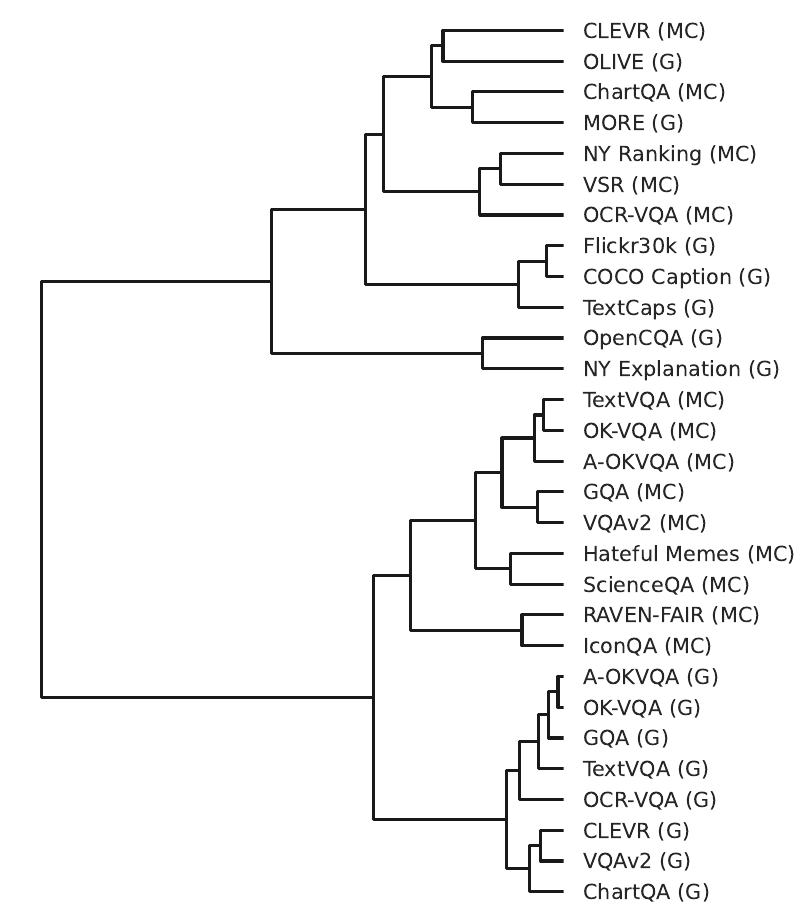}

  \caption
  	{
  	 Hierarchical clustering of target tasks.
	  } 
\label{fig:hcluster}
 \end{figure}

In this section, we perform hierarchical clustering on the SVD similarity features of target tasks using the Ward's linkage criterion which minimizes the total intra-cluster variance.
In Figure~\ref{fig:hcluster}, we show that hierarchical clustering forms meaningful clusters. For example, captioning tasks are clustered together. 
Generative and multiple-choice evaluated target tasks are grouped into different groups. This cluster supports the generative vs multiple-choice evaluation factor from factor analysis.
However, the clusters are not as comprehensive as common factors extracted by factor analysis. For example, hierarchical clustering does not elucidate factors such as reading vs reasoning, and spatial reasoning.

\section{Factor Analysis Details}
Here we show all the factor loadings of the six factors from the residual matrix $\bar{A}$. Communality quantifies the proportion of variance in each target task that is accounted for by the identified factors. A low communality value indicates that a task differs significantly from others in the mix.

\label{Appendix_factor_analysis_details}

\begin{table*}[!htbp]
\centering
\resizebox{1\textwidth}{!}{%
\begin{tabular}{cccccccc}
\toprule 
\textbf{Target   Tasks} & \textbf{Factor 1} & \textbf{Factor 2} & \textbf{Factor 3} & \textbf{Factor 4} & \textbf{Factor 5} & \textbf{Factor 6} & \textbf{Communality} \\
\midrule 
Flickr30k &
  \cellcolor[HTML]{D0CECE}0.97 &
  -0.02 &
  0.00 &
  0.07 &
  0.06 &
  -0.08 &
  0.96 \\
COCO Caption &
  \cellcolor[HTML]{D0CECE}0.93 &
  -0.05 &
  0.00 &
  0.10 &
  -0.02 &
  -0.12 &
  0.90 \\
TextCaps &
  \cellcolor[HTML]{D0CECE}0.83 &
  0.12 &
  -0.20 &
  0.07 &
  0.10 &
  -0.10 &
  0.77 \\
TextVQA (G) &
  -0.19 &
  \cellcolor[HTML]{D0CECE}0.87 &
  0.04 &
  -0.10 &
  -0.14 &
  -0.16 &
  0.85 \\
VQAv2 (MC) &
  \cellcolor[HTML]{D0CECE}-0.34 &
  \cellcolor[HTML]{D0CECE}-0.74 &
  \cellcolor[HTML]{D0CECE}-0.34 &
  -0.01 &
  0.24 &
  -0.02 &
  0.83 \\
ChartQA (G) &
  -0.08 &
  \cellcolor[HTML]{D0CECE}0.67 &
  -0.16 &
  \cellcolor[HTML]{D0CECE}0.31 &
  -0.12 &
  -0.23 &
  0.65 \\
OK-VQA (G) &
  -0.24 &
  \cellcolor[HTML]{D0CECE}0.60 &
  \cellcolor[HTML]{D0CECE}0.51 &
  -0.20 &
  0.20 &
  0.15 &
  0.78 \\
GQA (MC) &
  \cellcolor[HTML]{D0CECE}-0.32 &
  \cellcolor[HTML]{D0CECE}-0.55 &
  -0.18 &
  -0.26 &
  -0.02 &
  0.00 &
  0.50 \\
OK-VQA (MC) &
  \cellcolor[HTML]{D0CECE}-0.43 &
  \cellcolor[HTML]{D0CECE}-0.49 &
  -0.30 &
  0.07 &
  0.22 &
  -0.20 &
  0.62 \\
VQAv2 (G) &
  0.08 &
  0.06 &
  \cellcolor[HTML]{D0CECE}0.85 &
  0.23 &
  0.05 &
  -0.25 &
  0.86 \\
GQA (G) &
  -0.22 &
  -0.01 &
  \cellcolor[HTML]{D0CECE}0.75 &
  -0.05 &
  -0.21 &
  0.12 &
  0.66 \\
A-OKVQA (G) &
  -0.28 &
  \cellcolor[HTML]{D0CECE}0.54 &
  \cellcolor[HTML]{D0CECE}0.59 &
  -0.26 &
  0.23 &
  0.17 &
  0.87 \\
TextVQA (MC) &
  \cellcolor[HTML]{D0CECE}-0.38 &
  -0.12 &
  \cellcolor[HTML]{D0CECE}-0.49 &
  0.02 &
  \cellcolor[HTML]{D0CECE}0.36 &
  -0.23 &
  0.58 \\
OCR-VQA (MC) &
  0.20 &
  -0.14 &
  -0.04 &
  \cellcolor[HTML]{D0CECE}0.65 &
  -0.19 &
  -0.27 &
  0.60 \\
ChartQA (MC) &
  -0.14 &
  0.07 &
  -0.02 &
  \cellcolor[HTML]{D0CECE}0.65 &
  0.19 &
  0.29 &
  0.57 \\
RAVEN-FAIR (MC) &
  0.02 &
  -0.01 &
  0.08 &
  \cellcolor[HTML]{D0CECE}-0.40 &
  -0.04 &
  0.17 &
  0.20 \\
ScienceQA (MC) &
  -0.07 &
  0.00 &
  -0.07 &
  \cellcolor[HTML]{D0CECE}-0.39 &
  -0.05 &
  -0.06 &
  0.17 \\
IconQA (MC) &
  -0.01 &
  -0.09 &
  -0.08 &
  \cellcolor[HTML]{D0CECE}-0.34 &
  -0.05 &
  -0.10 &
  0.14 \\
OCR-VQA (G) &
  -0.01 &
  0.11 &
  -0.04 &
  -0.12 &
  \cellcolor[HTML]{D0CECE}-0.66 &
  0.01 &
  0.46 \\
A-OKVQA (MC) &
  -0.21 &
  \cellcolor[HTML]{D0CECE}-0.35 &
  \cellcolor[HTML]{D0CECE}-0.38 &
  -0.18 &
  \cellcolor[HTML]{D0CECE}0.63 &
  -0.07 &
  0.74 \\
MORE (G) &
  0.22 &
  \cellcolor[HTML]{D0CECE}0.47 &
  -0.22 &
  0.21 &
  \cellcolor[HTML]{D0CECE}0.54 &
  -0.03 &
  0.65 \\
OpenCQA (G) &
  0.17 &
  -0.07 &
  -0.09 &
  0.11 &
  \cellcolor[HTML]{D0CECE}0.32 &
  -0.24 &
  0.21 \\
OLIVE (G) &
  -0.05 &
  0.06 &
  0.09 &
  0.10 &
  -0.08 &
  \cellcolor[HTML]{D0CECE}0.61 &
  0.40 \\
CLEVR (G) &
  -0.17 &
  0.20 &
  0.16 &
  \cellcolor[HTML]{D0CECE}-0.44 &
  \cellcolor[HTML]{D0CECE}-0.34 &
  \cellcolor[HTML]{D0CECE}0.59 &
  0.74 \\
CLEVR (MC) &
  -0.18 &
  -0.13 &
  -0.05 &
  -0.07 &
  0.01 &
  \cellcolor[HTML]{D0CECE}0.55 &
  0.36 \\
VSR (MC) &
  0.15 &
  -0.26 &
  -0.10 &
  0.10 &
  -0.06 &
  \cellcolor[HTML]{D0CECE}0.50 &
  0.37 \\
NY Explanation (G) &
  0.13 &
  -0.03 &
  -0.04 &
  0.26 &
  0.21 &
  -0.10 &
  0.14 \\
NY Ranking (MC) &
  -0.24 &
  -0.30 &
  0.13 &
  0.08 &
  -0.23 &
  0.04 &
  0.22 \\
Hateful Memes (MC) &
  0.05 &
  -0.09 &
  -0.16 &
  -0.14 &
  -0.24 &
  0.05 &
  0.12 \\
  \bottomrule 
\end{tabular}%
}
  \caption
  	{
  	Results of EFA on the residuals $\bar{A}$. Cut-off for factor loadings = 0.3.
	  } 
  \label{fig:Appendix_fa_residuals}
\end{table*}

 \begin{table*}[!htbp]
 \centering
\resizebox{0.6\textwidth}{!}{%
\begin{tabular}{ccccc}
\toprule
\textbf{Target   Tasks} & \textbf{Factor 1} & \textbf{Factor 2} & \textbf{Factor 3} & \textbf{Communality} \\
\midrule
OK-VQA (G)  & \cellcolor[HTML]{D0CECE}0.78 & 0.43                         & 0.44                         & 1.00 \\
A-OKVQA (G) & \cellcolor[HTML]{D0CECE}0.74 & 0.44                         & 0.49                         & 0.98 \\
ChartQA (G) & 0.59                         & \cellcolor[HTML]{D0CECE}0.68 & 0.31                         & 0.91 \\
TextVQA (G) & \cellcolor[HTML]{D0CECE}0.63 & \cellcolor[HTML]{D0CECE}0.66 & 0.38                         & 0.97 \\
OCR-VQA (G) & 0.30                         & \cellcolor[HTML]{D0CECE}0.65 & 0.46                         & 0.73 \\
GQA (G)     & 0.51                         & 0.46                         & \cellcolor[HTML]{D0CECE}0.73 & 1.00 \\
VQAv2 (G)   & \cellcolor[HTML]{D0CECE}0.60 & 0.46                         & \cellcolor[HTML]{D0CECE}0.60 & 0.93 \\
\bottomrule
\end{tabular}%
}
  \caption
  	{
  	Results of EFA on generative VQAs. Cut-off for factor loadings = 0.6.
	  } 
 \label{fig:Appendix_fa_gen_tasks}

\end{table*}

\begin{table*}[!htbp]
\centering
\resizebox{0.6\textwidth}{!}{%
\begin{tabular}{ccccc}
\toprule
\textbf{Target   Tasks} & \textbf{Factor 1} & \textbf{Factor 2} & \textbf{Factor 3} & \textbf{Communality} \\
\midrule
OCR-VQA (MC) & \cellcolor[HTML]{D0CECE}0.81 & 0.31                         & 0.28                         & 0.82 \\
ChartQA (MC) & \cellcolor[HTML]{D0CECE}0.72 & 0.38                         & 0.21                         & 0.70 \\
A-OKVQA (MC) & 0.51                         & \cellcolor[HTML]{D0CECE}0.69 & 0.44                         & 0.93 \\
TextVQA (MC) & 0.53                         & \cellcolor[HTML]{D0CECE}0.69 & 0.39                         & 0.90 \\
OK-VQA (MC)  & 0.59                         & \cellcolor[HTML]{D0CECE}0.64 & 0.44                         & 0.95 \\
GQA (MC)     & 0.23                         & 0.28                         & \cellcolor[HTML]{D0CECE}0.93 & 1.00 \\
VQAv2 (MC)   & 0.50                         & 0.55                         & \cellcolor[HTML]{D0CECE}0.64 & 0.96 \\
\bottomrule
\end{tabular}%
}
  \caption
  	{
  	Results of EFA on multiple-choice VQAS $\bar{A}$. Cut-off for factor loadings = 0.6.
	  } 
  \label{fig:Appendix_fa_mc_tasks}
\end{table*}

\end{document}